\pdfoutput=1

\documentclass[10pt]{siamltex}
\usepackage[]{amsmath,amssymb,epsfig}

\usepackage[top=18mm, bottom=18mm, left=24mm, right=24mm]{geometry}

\newcommand\norm[1]{\left\lVert#1\right\rVert}

\usepackage{epstopdf}

\usepackage{comment}
\usepackage{graphicx}
\usepackage[space]{grffile}

\usepackage{placeins}

\newcommand{\mb}[1]{\mbox{\boldmath$#1$}}

\usepackage{subfigure}
\usepackage{amsmath}
\usepackage{amssymb}
\usepackage{graphicx}
\usepackage{comment}
\usepackage{array}
\usepackage{algorithmic}
\usepackage{algorithm}
\usepackage{url}

\newcolumntype{L}[1]{>{\raggedright\let\newline\\\arraybackslash\hspace{0pt}}m{#1}}
\newcolumntype{C}[1]{>{\centering\let\newline\\\arraybackslash\hspace{0pt}}m{#1}}
\newcolumntype{R}[1]{>{\raggedleft\let\newline\\\arraybackslash\hspace{0pt}}m{#1}}

\usepackage{setspace}
\usepackage{multicol}
\usepackage{multirow}
 \usepackage{hyperref}

\hypersetup{
    colorlinks=true,
    citecolor=red,
    linkcolor=blue,
    urlcolor=blue
  }

\newcounter{ale}

\newenvironment{liste}{\begin{itemize}}{\end{itemize}}
\newcommand{\aliste}{\begin{liste} \setcounter{ale}{1}}
\newcommand{\zliste}{\end{liste}}

\begin{document}


\title{Sync-Rank: Robust Ranking, Constrained Ranking and Rank Aggregation via Eigenvector and SDP Synchronization }
\author{ Mihai~Cucuringu  \thanks{Department of Mathematics, UCLA, 520 Portola Plaza, Mathematical Sciences Building 6363, Los Angeles, CA 90095-1555, email: mihai@math.ucla.edu}}
\maketitle


\begin{center}
     \today
\end{center}


\begin{abstract}
We consider the classic problem of establishing a statistical ranking of a set of $n$ items given a set of inconsistent and incomplete pairwise comparisons between such items. Instantiations of this problem occur in  numerous applications in data analysis (e.g., ranking teams in sports data), computer vision, and machine learning. We formulate the above problem of ranking with incomplete noisy information as an instance of the \textit{group synchronization} problem over the group SO(2) of planar rotations, whose usefulness has been demonstrated in numerous applications in recent years 
in computer vision and graphics, sensor network localization and structural biology. 
Its least squares solution can be approximated by either a spectral or a  semidefinite programming (SDP) relaxation, followed by a rounding procedure. 
We show extensive numerical simulations on both synthetic and real-world data sets (Premier League soccer games, a Halo 2 game tournament and NCAA College Basketball games), which show that our proposed method compares favorably to other ranking methods from the recent literature. 
%
Existing theoretical guarantees on the group synchronization problem  
imply lower bounds on the largest amount of noise permissible in the data while still achieving  exact recovery of the ground truth ranking.
%
We propose a similar synchronization-based algorithm for the rank-aggregation problem, which  integrates in a globally consistent ranking many pairwise rank-offsets or partial rankings, given by different rating systems on the same set of items, an approach which yields significantly more accurate results than other 
aggregation methods, including \textit{Rank-Centrality}, a recent state-of-the-art algorithm. 
Furthermore, we discuss the problem of semi-supervised ranking when there is available information on the ground truth rank  of a subset of players, and propose an algorithm based on SDP  which is able to recover the ranking of the remaining players, subject to such hard constraints.  
Finally, synchronization-based ranking, combined with a spectral technique for the  densest subgraph problem,  
makes it possible to extract locally-consistent partial rankings, in other words, to identify the rank of a  small subset of players whose pairwise rank comparisons are less noisy than the rest of the data, which other methods are not able to identify. 
We discuss a number of related open questions and variations of the ranking problem in other 
settings, which we defer for future investigation.
\end{abstract}


\section{Introduction}   \label{sec:Intro}

We consider the problem of ranking a set of $n$ players, given  ordinal or cardinal pairwise comparisons on their rank offsets. In most practical applications, such available information is usually incomplete, especially in the setting where $n$ is large, and the available data is very noisy, meaning that a large fraction of the pairwise measurements are both incorrect and inconsistent with respect to the existence of an underlying total ordering. In such scenarios, one can at most hope to recover a total (or partial) ordering that is as consistent as possible with the available noisy measurements. For instance, at sports tournaments where all pairs of players meet, in most circumstances the outcomes contain cycles (A beats B, B beats C, and C beats A), and one seeks to recover a ranking that minimizes the number of upsets, where an upset is a pair of players for which the higher ranked player is beaten by the lower ranked player.

Due to the shear size of nowadays data sets, the set of available comparisons between the items is very sparse, with much, or even most, of the data being incomplete, thus rendering the ranking problem considerably harder. 
In addition, the available measurements are not uniformly distributed around the network, a fact which can significantly affect the ranking procedure. Similarly, the noise in the data may not be distributed uniformly throughout the network, with part of the network containing pairwise measurements that are a lot less noisy than the rest of the network, which provides an opportunity to recover partial ranking which are locally consistent with the given data. We investigate this possibility in Appendix \ref{sec:PartialRank}, and show how our proposed method can be combined with recent spectral algorithms for detecting planted cliques or dense subgraphs in a graph.
Furthermore, in many scenarios, the available data is governed by an underlying (static or dynamic) complex network, whose structural properties can play a crucial role in the accuracy of the ranking process if exploited accordingly, as in the case of recent work on time-aware ranking in dynamic citation networks \cite{RankingDynamicNetworks}.

The analysis of many modern large-scale data sets implicitly requires various forms of ranking to allow for the identification of the most important entries, for efficient computation of search and sort operations, or for extraction of main features.
Instances of such problems are abundant in various disciplines, especially in modern internet-related applications such as the famous search engine provided by Google \cite{langville2011google,Pageetal98}, eBay's feedback-based reputation mechanism \cite{xiong2003reputation}, Amazon's Mechanical Turk (MTurk) system for crowdsourcing which enables individuals and businesses to coordinate the use of human labor to perform various tasks \cite{raykar2011ranking,horton2010labor}, the popular movie recommendation system provided by Netflix  \cite{cremonesi2010performance}, the Cite-Seer network of citations  \cite{giles1998citeseer}, or for ranking of college football teams \cite{harville1977use}.

Another setting which can be reduced to the ranking problem, comes up in  the area of exchange economic systems, where an item $i$ can be exchanged for an item $j$ at a given rate; for example, 1 unit of item $i$ is worth $a_{ij} > 0$ units of item $j$. Such information can be collected in the form of a (possibly incomplete)  exchange matrix $A=(a_{ij})$, which is a reciprocal matrix since $ a_{ij} = 1/a_{ij}$ for  non-zero entries $a_{ij}$.      Such reciprocal matrices have been studied since the 1970s with the work of Saaty \cite{SaatyReciprocal}  in the context of paired preference aggregation systems, and more recently by Ma for asset pricing in foreign exchange markets \cite{MaReciprocal}. In this setup, the goal is to compute a universal (non-negative)  value $\pi_i$ associated to each item $i$, such that $ a_{ij} = \frac{\pi_i}{\pi_j}$,  which can easily be seen as equivalent to the pairwise ranking problem via the logarithmic map $ C_{ij} = \log a_{ij} $, whenever $a_{ij} \neq 0 $. In other words, the available pairwise measurement $C_{ij}$ is a, perhaps noisy, measurement of the offset $ \log \pi_i - \log \pi_j$. Note that in arbitrage free economic systems, the  triangular condition $ C_{ij} + C_{jk} + C_{ki} = 0 $ is always satisfied.


Traditional ranking methods, most often coming from the social choice theory literature, 
 have proven less efficient in dealing with nowadays data, for several reasons. Most of this  literature
has been developed with ordinal comparisons in mind, while much of the current data deals with cardinal (numerical) scores for the pairwise comparisons. However, it is also true that in certain applications such as movie or music rating systems, it is somewhat more natural to express preferences in relative terms (e.g., movie $X$ is better than movie $Y$) rather than in absolute terms (e.g., $X$ should be ranked $3^{rd}$ and $Y$ $9^{th}$, or $X$ is better than $Y$ by 6 \textit{units}). In other applications, however, such as sports, the outcome of a match is often a score, for example in soccer, we know what is the goal difference via which team $X$ beat team $Y$. 
Along the same lines, in many instances one is often interested not only in recovering the ordering of the items, but also in associating a score to each of the items themselves, which reflects the level of accuracy or the intensity of the proposed preference ordering. For example, the TrueSkill ranking algorithm developed by Microsoft Research assigns scores to online gamers based on the outcome of games players between pairs of players. Every time a player competes in a new game, the ranking engine updates his estimated score and the associated level of confidence, and in doing so, it learns the underlying inherent skill parameters each player is assumed to have.

There exists a very rich literature on ranking, which dates back as early as the 1940s with the seminal work of Kendall and Smith \cite{KendallSmith1940}, who were interested in recovering the ranking of a set of players from pairwise comparisons reflecting a total ordering. 
Perhaps the most popular ranking algorithm to date is the famous PageRank \cite{Pageetal98}, used by Google to rank web pages in increasing order of their relevance, based on the hyperlink structure of the Web graph. On a related note, Kleinberg's HITS algorithm \cite{KleinberHITS} is another website ranking algorithm in the spirit of PageRank, based on identifying good \textit{authorities} and \textit{hubs} for a given topic queried by the user, and which assigns two numbers to a web page: an authority and a hub weight, which are defined recursively. A higher authority weight occurs if the page is pointed to by pages with high hub weights. And similarly, a higher hub weight occurs if the page points to many pages that have high authority weights.

In another line of work \cite{BravermanMossel}, Braverman and Mossel proposed an algorithm which outputs an ordering based on $O(n \log n)$ pairwise comparisons on adaptively selected pairs. Their model assumes that there exists an underlying true ranking of the players, and the available data comes in the form of noisy comparison results, in which the true ordering of a queried pair is revealed with probability $\frac{1}{2} + \gamma $, for some parameter $ \gamma $ which does not depend on the pair of players that compete against each other. However, such a noise model is somewhat unrealistic in certain instances like chess matches or other sporting events, in the sense that the noisy outcome of the comparison does not depend on the strength of the opponents that are competing (i.e., on their underlying skill level) \cite{Elo1978}.


A very common approach in the  rank aggregation literature is to treat the input rankings as data generated from a probabilistic model, and then learn the Maximum Likelihood Estimator (MLE) of the input data, an idea which has been explored in depth in both the machine learning and computational social choice theory communities include the the Bradley-Terry-Luce (BTL) model \cite{BradleyTerry1952}, the Plackett-Luce (PL) model \cite{Luce1959individual}, the Mallows-Condorcet model \cite{Mallows1957,Condorcet}, and Random Utility Model (RUM) \cite{RUMmodel}.
The $BTL$ model has found numerous applications in recent years, including   pricing in the airline industry \cite{talluri2006theory}, or analysis of professional basketball results \cite{koehler1982application}. 
Much of the related research within the machine learning community has focused  on the development of computationally  efficient algorithms to estimate parameters for some of the above popular models, a line of work commonly referred to as  \textit{learning to rank} \cite{LiuLearningToRank}. We refer the reader to the recent book of Liu \cite{bookLiuLearningToRank},  
for a comprehensive review of the main approaches to learning to rank, and how one may leverage tools from the machine learning community to improve on and propose new ranking models. Another such example is the earlier work of Freund et al. \cite{FreundSchapireSinger}, who proposed Rank-Boost, an efficient algorithm for combining preferences based on the boosting approach from machine learning.

Soufiani et al. propose a class of efficient Generalized Method-of-Moments algorithm for computing parameters of the Plackett-Luce model, by breaking the full rankings into pairwise comparisons, and then computing the parameters that satisfy a set of generalized moment conditions \cite{AzariSoufiani_nips13_b}.  Of independent interest is the approach of breaking full rankings into pairwise comparisons, since the input to the synchronization-based approach proposed in this paper consists of pairwise comparisons. This technique of \textit{rank breaking} was explored in more depth by a subset of the same set of authors \cite{AzariSoufiani_icml14}, with a focus on the consistency of their proposed breaking methods for a variety of models, together with fast algorithms for estimating the parameters.

Other related work includes \cite{JamiesonNowak2011}, whose authors propose to adaptively select pairwise comparisons, an approach which, under certain assumptions, recovers the underlying ranking with much fewer measurements when compared to the more naive approach of choosing at random. 
Kenyon-Mathieu and Schudy  \cite{KenyonMathieu} propose a polynomial time approximation scheme (PTAS) for the minimum feedback arc set problem on tournaments\footnote{When all pairwise comparisons between a set of $n$ players are available, the data can be conveniently represented as a directed complete graph, referred to as \textit{tournament graphs} in the theoretical computer science literature. Such scenarios are very common in practice, especially in sports, where in a \textit{round robin} tournament every two players meet, and the direction of each edge encodes the outcome of the match, or in a more general settings, a preference relation between a set of items.}, an NP-hard problem which arises at tournaments where all pairs of players meet and one seeks to recover a ranking that minimizes the number of upsets.
Very computationally efficient methods based on simple scoring methodologies that come with certain guaranties exist since the work of Huber in the 1960s \cite{HuberRowSum1963} (based on a simple row-sum procedure), and very recently Wauthier et al.  \cite{RankingWauthierJordan} in the context of ranking from a random sample of binary comparisons, who can also account for whether one seeks an 
approximately uniform quality across the ranking, or more accuracy near the top of the ranking than the bottom.



The idea of angular embedding, which we exploit in this paper, is not new, and aside from recent work by Singer in the context of the angular synchronization problem \cite{sync}, has also been explored by Yu \cite{StellaAE}, who observes that embedding in the angular space is significantly more robust to outliers when compared to 
embedding in the linear space. In the later case, the traditional least squares formulation, or its $l_1$ norm formulation, cannot match the performance of the angular embedding approach, thus suggesting that the key to overcoming outliers comes not with imposing additional constraints on the solution, but by adaptively penalizing the inconsistencies between the measurements themselves. Yu's  proposed spectral method returns very satisfactory results in terms of robustness to noise when applied to an image reconstruction problem. 
In recent work, Braxton et al. \cite{osting2012statistical}, propose an $l_1$-norm formulation for the statistical ranking problem, for the case of cardinal data, and provides an alternative to previous recent work utilizing an $l_2$-norm formulation, as in \cite{hirani2010least} and 
\cite{HodgeRanking}.

In very recent work \cite{RankCentrality}, Negahban et al. propose an iterative algorithm for the rank aggregation problem of integrating ranking information from multiple ranking systems, by estimating  scores for the items from the stationary distribution of a certain random walk on the graph of items, where each edge encodes the outcome of pairwise comparisons. We summarize their approach in Section \ref{secsec:RankCent}, and compare against it in the setting of the rank aggregation problem. However, for the case of a single rating system, we propose and compare to a variant of their algorithm. 
 Their work addresses several shortfalls of earlier work \cite{AmmarShah} by a subset of the same authors, who view the available comparison data as partial samples from an unknown distribution over permutations, and reduce ranking to performing inference on this distribution, but in doing so, assume that the comparisons between all pairs of items are available, a less realistic constraint in most practical applications. 

In other also very recent work \cite{serialRank}, which we briefly summarize in Section  \ref{secsec:SER}, Fogel et al. propose a ranking algorithm given noisy incomplete pairwise comparisons by making an expliciti connection to another very popular ordering problem, namely \textit{seriation} \cite{AtkinsSeriation}, where one is given a similarity matrix between a set of items and assumes that a total order exists and aims to order the items along a chain such that the similarity between the items decreases with their distance along this chain.
Furthermore, they demonstrate in \cite{AspremontRelaxPermutation}  the applicability of the same seriation paradigm to the setup of semi-supervised ranking, where additional structural constraints are imposed on the solution. 


\textbf{Contribution of our paper.} The contribution of our present work can be summarized as follows. 
\begin{itemize}
\item We make an explicit connection between ranking and the angular synchronization problem, and use existing spectral and SDP relaxations for the latter problem to compute robust global rankings.

\item We perform a very extensive set of numerical simulations comparing our proposed method with existing state-of-the-art algorithms from the ranking literature, across a variety of synthetic  measurement graphs and noise models, both for numerical (\textit{cardinal}) and binary (\textit{ordinal}) pairwise comparisons between the players. In addition, we compare the algorithms on three real data sets: the outcome of soccer games in the  English Premier League, a Microsoft tournament for the computer game Halo 2, and NCAA College Basketball games.
Overall, we compare (in most instances, favorably) to the two recently proposed state-of-the-art algorithms,  \textit{Serial-Rank} \cite{serialRank}, and \textit{Rank-Centrality} \cite{RankCentrality}, aside from the more traditional Least-Squares method. 
\item Furthermore, we propose and compare to a very simple ranking method based on Singular Value Decomposition, which may be on independent interest as its performance (which we currently investigate theoretically in a separate ongoing work) is comparable to that of a recent state-of-the-art method.
\item We propose a method for ranking in the semi-supervised setting where a subset of the players have a prescribed rank to be enforced as a hard constraint.
\item We also adjust the synchronization approach to the setting of the rank aggregation problem of integrating ranking information from multiple rating systems that provide independent, incomplete and inconsistent pairwise comparisons for the same set of players, with the goal of producing a single global ranking.
\item Finally, we show that by combining Sync-Rank with recent algorithms for the \text{planted clique} and \text{densest subgraph problem}, we are able to  identify \textit{planted locally-consistent} partial rankings, which other methods are not able to extract.
\end{itemize}

The advantage of the synchronization-based ranking algorithm (Sync-Rank)  stems from the fact that it is a computationally simple, non-iterative algorithm that is model independent and relies exclusively on the available data, which may come as either pairwise ordinal or cardinal comparisons.
Existing theoretical guarantees from the recent literature on the group synchronization problem \cite{sync,littleGrothendieck,tightSDPMLEsync,afonso} 
trivially translate to lower bounds for the largest amount of noise permissible in the measurements that would still allow for an exact or almost exact recovery of the underlying ground truth ranking.  
We point point out that a perfect recovery of the angles in the angular synchronization problem is not a necessary condition for a perfect recovery of the underlying ground truth ranking, since it  suffices that only the relative ordering of the angles is preserved.

The remainder of this paper is organized as follows. 
Section \ref{sec:OtherMethods} summarizes related methods against which we compare, with a focus on the recent Serial-Rank algorithm.
Section \ref{sec:GroupAngSync} is a review of the angular synchronization problem and existing results from the literature. 
Section \ref{sec:Sync-Rank} describes the Sync-Rank algorithm for  ranking via eigenvector and SDP-based synchronization. 
Section \ref{sec:numexp} provides an extensive numerical comparison of Sync-Rank with other methods from the literature.
%
Section \ref{sec:RankAggregation} considers the rank aggregation, which we solve efficiently via the same spectral and SDP relaxations of the angular synchronization problem.
In Section \ref{sec:constRanking} we consider the constrained ranking problem and propose to solve it via a modified SDP-based synchronization algorithm.
Section  \ref{sec:varOpen} summarizes several variations and open problems related to ranking, while   
Section \ref{sec:summaryDisc} is a  summary and discussion. 
Appendix \ref{sec:PartialRank} proposes an algorithm for extracting locally-consistent partial rankings from comparison data.
Appendix \ref{sec:appSER} summarizes the recently proposed Serial-Rank algorithm. 
Finally, in Appendix   \ref{sec:appEngland}  we provide additional numerical results for the English Premier League soccer  data set. 


\section{Related Methods}  \label{sec:OtherMethods}
In this section, we briefly summarize the Serial-Rank algorithm recently introduced in \cite{serialRank}, which performs spectral ranking via seriation and was shown to compare favorably to other classical ranking methods, some of which we discussed in the Introductory section. 
In addition, we summarize the very recent Rank-Centrality algorithm proposed by Negahban et al. \cite{RankCentrality}, which we used for the rank aggregation problem discussed in Section \ref{sec:RankAggregation}, and also propose a modification of it for the setting of a single rating system, making it amenable to both cardinal and ordinal comparisons. 
Finally, we consider two other approaches for obtaining a global ranking based on Singular Value Decomposition (SVD) and the popular method of Least Squares (LS).

\subsection{Serial Rank and Generalize Linear Models}
\label{secsec:SER}
In very recent work \cite{serialRank}, Fogel et al. propose a seriation algorithm for ranking a set of players given noisy incomplete pairwise comparisons between the players. The gist of their approach is to assign similar rankings to players that compare similarly with all other players.  They do so by constructing a similarity matrix from the available pairwise comparisons, relying on existing seriation methods to reorder the similarity matrix and thus recover the final rankings. 
The authors make an explicit connection between the ranking problem and another related classical ordering problem, namely \textit{seriation}, where one is given a similarity matrix between a set of $n$ items and assumes that the items have an underlying ordering on the line such that the similarity between items decreases with their distance. In other words, the more similar two items are, the closer they should be in the proposed solution. 
By and large, the goal of the seriation problem is to recover the underlying linear ordering based on unsorted, inconsistent and incomplete pairwise similarity information. We briefly summarize their approach in Appendix \ref{sec:appSER}.

\subsection{Ranking via Singular Value Decomposition}  \label{secsec:SVD}
An additional ranking method we propose, and compare against, is based on the traditional  Singular Value Decomposition (SVD) method. The applicability of the SVD-Rank approach stems from the observation that, in the case of cardinal measurements (\ref{cardComp}), the noiseless matrix of rank offsets $C = (C_{ij})_{1\leq i,j \leq n}$, 
is a skew-symmetric matrix of even rank 2 since
\begin{equation}
	R = r \boldsymbol{e}^T - \boldsymbol{e} r^T
\end{equation}
where $\boldsymbol{e}$ denotes the all-ones column vector. In the noisy case, $C$ is a random perturbation of a rank-2 matrix. We consider the top two singular vectors of $C$, order their entries by their size, extract the resulting rankings, and choose between the first and second singular vector based on whichever one minimizes the number of upsets. Note that since the singular vectors are obtained via a global sign, we (again) choose the ordering which minimizes the number of upsets. Though a rather naive approach, SVD-Rank returns, under the multiplicative uniform noise model, results that are comparable to those of very recent algorithms such as Serial-Rank \cite{serialRank} and Rank-Centrality \cite{RankCentrality}.
A previous application of SVD to ranking has been explored in Gleich and Zhukov \cite{Gleich04svdbased}, for studying relational data as well as developing a  method for interactive refinement of the search results.
To the best of our knowledge, we are not aware of other work that considers SVD-based ranking for the setting considered in this paper.
An interesting research direction, which we are pursuing in ongoing work, is to analyze the  performance of SVD-Rank using tools from the 
random matrix theory literature on rank-2 deformations of random matrices \cite{Benaych_georges_thesingular}.


\subsubsection{Rank-2 Decomposition in the ERO model} \label{secsec:SVD_ERO}
Note that for the \textbf{Erd\H{o}s-R\'{e}nyi Outliers} ERO($n,p,\eta$) model given by (\ref{ERoutliers}), the following decomposition could render the SVD-Rank method amenable to a theoretical analysis. Note that the expected value of the entries of $C$ is given by
\begin{equation}
	\mathbb{E} C_{ij} = (r_i - r_j) (1-\eta) \alpha,
\end{equation}
in other words, $ \mathbb{E} C $ is a rank-2 skew-symmetric  matrix
\begin{equation}
	\mathbb{E} C = (1-\eta) \alpha (  r \boldsymbol{e}^T - \boldsymbol{e} r^T)
\end{equation}
Next, one may decompose the given data matrix $C$ as
\begin{equation}
	C = \mathbb{E} C + R
	\label{decompC_EC_R}
\end{equation}
where $R = C - \mathbb{E} C$ is a random skew-symmetric matrix whose elements have zero mean and are given by
\begin{equation}
R_{ij} = \left\{
 \begin{array}{rll}
 (r_i - r_j)  [ 1 - (1-\eta) \alpha ] &  \text{for a correct edge,} & \text{ w. prob. } (1-\eta) \alpha \\
 q - (r_i - r_j) (1-\eta) \alpha   &  \text{for an incorrect edge,} & \text{ w. prob. } \eta \alpha \text{ and } q \sim \text{Uniform}[-(n-1), n-1] 	\\
	- (r_i - r_j) (1-\eta) \alpha      &  \text{for a missing edge},   & \text{ w. prob. } 1-\alpha,  	\\
     \end{array}
   \right.
\label{ERoutliers_Rij}
\end{equation}
whenever $i \neq j$, which renders  the given data matrix $C$ decomposable into a low-rank (rank-2) perturbation of a random skew-symmetric matrix. The case $\alpha=1$ which corresponds to the complete graph $G=K_n$ simplifies (\ref{ERoutliers_Rij}), and is perhaps a first step towards a  theoretical investigation.

\iftrue
\subsubsection{Rank-2 Decomposition in the MUN model} \label{secsec:SVD_MUN}
A similar decomposition holds for the other noise model we have considered, \textbf{Multiplicative Uniform Noise}, MUN($n,p,\eta$), given by (\ref{MUN_model}). As above,  
\begin{eqnarray}
   & \mathbb{E} C_{ij} = r_i - r_j
\label{ECij_MUN}
\end{eqnarray}
and a similar decomposition $ C = \mathbb{E} C + R $ as in (\ref{decompC_EC_R}) holds, where the zero mean entries of the random matrix $R$ give by
\begin{eqnarray}
   R_{ij} = (r_i - r_j) \epsilon
\label{R_ij_MUN}
\end{eqnarray}
with $ \epsilon \sim [-\eta, \eta]$. Note that, as opposed to  (\ref{ERoutliers_Rij}),  the entries are no longer independent. To limit the dependency between the entries of the random matrix, one may further assume that there are no comparisons whenever $r_i - r_j $ is large enough, i.e., between players who are far apart in the rankings, an assumption which may seem natural in certain settings such as chess competitions, where it is less common that a highly skilled chess master plays against a much weaker player. It would be interesting to investigate whether this additional assumption could make the SVD approach amenable to a theoretical analysis in light of recent results from the random matrix theory literature by Anderson and Zeitouni \cite{AndersonZeitouni},  which relax the independence condition and consider finite-range dependent random matrices that allow for dependency between the entries which are ''nearby" in the matrix.

\subsection{Ranking via Least Squares}   \label{secsec:LS}
We also compare our proposed ranking method with the more traditional least-squares approach. Assuming the number of edges in $G$ is given by $m=|E(G)|$, we denote by $B$ the edge-vertex incidence matrix of size $m \times n$ whose entries are given by
\begin{equation}
B_{ij} = \left\{
 \begin{array}{rll}
  1   &   \text{ if } (i,j) \in     E(G),    & \text{ and  } i > j  	\\
 -1   &   \text{ if } (i,j) \in     E(G),    & \text{ and  } i <  j  	\\
  0  &   \text{ if } (i,j) \notin  E(G)  &
  \end{array}
   \right.
\label{incidenceMtxL}
\end{equation}
and by $w$ the vector of length $m \times 1$ which contains the pairwise rank measurements  $w(e) = C_{ij}$, for all edges $e=(i,j) \in E(G)$. We obtain the least-squares solution to the ranking problem by solving the following minimization problem
\begin{equation}
	\underset{ x \in \mathbb{R}^n }{\text{ minimize } } \;\; || B x - w ||_2^2
\label{rank_LS}
\end{equation} 
We point out here the work of Hirani et al \cite{hirani2010least}, who show that the problem of least-squares ranking on graphs has far-reaching rich connections with various other research areas, including spectral graph theory and multilevel methods for graph Laplacian systems, Hodge decomposition theory and random clique complexes in topology. 





\subsection{The Rank-Centrality algorithm}  \label{secsec:RankCent}

In recent work \cite{RankCentrality}, Negahban et al. propose an iterative algorithm for the rank aggregation problem by estimating  scores for the items from the stationary distribution of a certain random walk on the graph of items, where edges encode the outcome of pairwise comparisons. The authors propose this approach in the context of the rank aggregation problem, which, given as input a collection of sets of pairwise comparisons over $n$ players or partial rankings (where each such set is provided by an independent rating system, or member of a jury of size $k$) the goal is to provide a global ranking that is as consistent as possible with the given measurements of all $k$ ranking systems.

At each iteration of the random walk, the probability of transitioning from vertex $i$ to vertex $j$ is directly proportional to how often player $j$ beat player $i$ across all the matches the two players confronted, and is zero if the two players have never played a game before. In other words, the random walk has a higher chance of transitioning to a more skillful neighbors, and thus the frequency of visiting a particular node, which reflects the rank or the skill level of the corresponding players, is thus encoded in the stationary distribution of the associated Markov Chain. Such an interpretation of the stationary distribution of a Markov chain can be traced back to early work on the topic of \textit{network centrality} from the network science literature. Network centrality-based tools have been designed to measure which nodes of the graph (or other network structures) are most important \cite{booknewman,WassermanSocialBook}, some of which having a natural interpretation in terms of information flow within a network \cite{EstradaCookCLX}. One of the most popular applications of network centrality is the  PageRank algorithm \cite{Pageetal98} for computing the relative importance of a web page on the web graph.  More recently, dynamic centrality measures have been proposed for the analysis of temporal network data in neuroscience, for studying the functional activity in the human brain using functional magnetic resonance imaging \cite{BrainMasonJCN}.
%

In the context of the popular BTL model, the authors of \cite{RankCentrality} propose the following approach for computing the Markov matrix, which we adjust to render it applicable to both ordinal and cardinal measurements, in the case of a single rating system. Note that in Section \ref{sec:RankAggregation} where we discuss the rank aggregation problem in the context of multiple rating systems, we rely on the initial \textit{Rank-Centrality}  algorithm introduced in \cite{RankCentrality}.


For a pair of items $i$ and $j$, let $Y_{ij}^{(l)}$ be equal to 1 if player $j$ beats  player $i$, and 0 otherwise, during the $l^{th}$ match between the two players, with $l=1,\ldots,k$.  The BTL model assumes that $ \mathbb{P}(Y_{ij}^{(l)} = 1) = \frac{w_j}{w_i + w_j}$, where  $w$ represent  the underlying vector of positive real weights associated to each player. The approach in \cite{RankCentrality} starts by estimating the fraction of times players $j$ has defeated player $i$, which is denoted by 
\begin{equation}
a_{ij} = \frac{1}{k} \sum_{l=1}^{k} Y_{ij}^{l},
\label{aij_RC}
\end{equation}
as long as players $i$ and $j$ competed in at least one match, and $0$ otherwise. Next, consider the symmetric matrix 
\begin{equation}
A_{ij} = \frac{a_{ij}}{ a_{ij} + a_{ji} },
\label{Aij_RC}
\end{equation}
which converges to $ \frac{w_j}{w_i + w_j}$, as $k \to \infty$.
To define a valid transition probability matrix, the authors of \cite{RankCentrality} scale all the edge weights by  $1/d_{max}$ and consider the resulting random walk 
\begin{equation}
P_{ij} = \left\{
     \begin{array}{rl}
 \frac{1}{d_{max}} A_{ij} & \;\; \text{ if } i \neq j \\
1 -  \frac{1}{d_{max}}  \sum_{k\neq i}  A_{ik} & \;\; \text{ if }  i=j, \\
     \end{array}
   \right.
\label{P_RC}
\end{equation}
where $ d_{max}$ denotes the maximum out-degree of a node, thus making sure that each row sums to 1. The stationary distribution $\pi$ is the top left eigenvector of $P$, and its entries denote the final numerical scores associated to each node, which, upon sorting, induce a ranking of the $n$ players.

To render the above approach applicable\footnote{Otherwise, in the ordinal case, $A_{ij}$ is either $0$ or $1$} in the case when $k=1$ of a single rating system (for both cardinal and ordinal measurement), but also when $k>1$ for the case of cardinal measurements, we propose the following alternatives to designing the winning probabilities $a_{ij}$ given by (\ref{aij_RC}), and inherently the final winning probability matrix $A$ in (\ref{Aij_RC}). Once we have an estimate for $A$ (given by (\ref{myA_ordinal_RC}) for ordinal data, respectively by  (\ref{myA_cardinal_RC}) for cardinal data), we proceed with building the transition probability $P$ as in (\ref{P_RC}) and consider its stationary distribution.
Note that we also make these two new methods proposed below applicable to the setting of multiple rating systems $ k  > 1$, by simply averaging out the resulting winning probabilities $A_{ij}^{(l)}, l=1,\ldots,k$, given by each rating system via (\ref{myA_ordinal_RC}) and (\ref{myA_cardinal_RC}), across all rating systems
\begin{equation}
A_{ij} =  \frac{1}{k} \sum_{l=1}^{k} A_{ij}^{(l)},
\label{juryVersionMyA}
\end{equation}
and then consider the transition probability matrix $P$ as in (\ref{P_RC}) and its stationary distribution.

\subsubsection{Adjusted Rank-Centrality for ordinal measurements}
\label{sec:sub_adjRC_ordinal}
To handle the case of ordinal measurements, we propose a hybrid approach that combines Serial-Rank and Rank-Centrality, and yields more accurate results than the former one, as it can be seen in Figure \ref{fig:Meth6_n200_ord}.
We proceed by computing the $S^{match}$ matrix as in the Serial-Rank algorithm (given by (\ref{SijMatch}) and (\ref{SMatch})  in  Appendix \ref{sec:appSER}) that counts the number of matching  comparisons between $i$ and $j$ with other third reference items $k$. The intuition behind this similarity measure is that players that beat the same players and are beaten by the same players should have a similar ranking in the final solution. Note that $S_{ij} \leq n $ for any pair of players\footnote{This is since $C_{ii}$ is defined to be $1$, otherwise it would be true that $S_{ij} \leq n-2 $.}, and thus $ \frac{S_{ij}}{n} \in [0,1]$

Note that, whenever $S_{ij}$ is very large, say $ S_{ij} \approx n $, meaning the two players are very similar, then the quantity 
$ 1- \frac{S_{ij}}{n}$ is small and close to zero, and thus a good proxy for the difference in the winning probabilities $A_{ij}$ and $A_{ji}$ defined in (\ref{Aij_RC}). In other words, if two players are very similar, it is unlikely that, had they played a lot of matches against each other, one player will defeat the other in most of the matches. On the other hand, if two players are very dissimilar, and thus  $S_{ij}$ is close to zero and $ 1- \frac{S_{ij}}{n}$ is close to one, then it must be that, had the two players met in many matches, one would have defeated the other in a significant fraction of the  games. With these observations in mind,  and in the spirit of (\ref{Aij_RC}), we design the matrix $A$ of winning probabilities such that 
\begin{equation}
\left\{
     \begin{array}{r}
     A_{ij} + A_{ji} = 1 \\
     | A_{ij} - A_{ji} | = \frac{S_{ij}}{n}. \\
     \end{array}
	\right.
\label{System2by2RC}
\end{equation}
for a pair of players that met in a game. We lean the balance in favor of the player who won in the (single) direct match, and assign to him the larger winning probability. Keeping in mind that $A_{ij}$ should be a proxy for the fraction of times player $j$ defeated player $i$ (thus whenever $C_{ij}>0$ it must be that $A_{ij} > A_{ji}$), the above system of equations (\ref{System2by2RC}) yields 
\begin{equation}
A_{ij} = \left\{
     \begin{array}{rl}
1 - \frac{1}{2} \frac{S_{ij}}{n}, & \;\; \text{ if } C_{ij} > 0 \\
   \frac{1}{2} \frac{S_{ij}}{n}, & \;\; \text{ if }C_{ij} < 0. \\
0   , & \;\; \text{ if } C_{ij} = 0. \\
     \end{array}
   \right.
\label{myA_ordinal_RC}
\end{equation}
We remark that, in the case of  outliers given by the ERO noise model (\ref{ERoutliers}), our above proposed version of Rank-Centrality (denoted as RC), when used in the setting of multiple rating systems, performs much better than the original Rank-Centrality algorithm (denoted as RCO), as shown in the bottom plot of Figure \ref{fig:ErrorsJuryERO}.

\subsubsection{Adjusted Rank-Centrality for cardinal measurements}
\label{sec:sub_adjRC_cardinal}

For the case of cardinal measurements, we propose a similar matrix $A$ of winning probabilities, and incorporate the magnitude of the score into the entries of $A$. The intuition behind defining the winning probability is given by the following reasoning. Whenever $C_{ij}$ takes the largest possible (absolute) value (i.e., assume $ C_{ij} = (n-1) $, thus $j$ defeats $i$ by far), we define the winning probability that player $j$ defeats player $i$ to be largest possible, i.e., $A_{ij} = 1$, and in general, the larger the magnitude of $C_{ij}$, the larger $A_{ij}$ should be. 
On the other hand, whenever $C_{ij}$ has the smallest possible (absolute) value (i.e., assume $ C_{ij} = 1 $), then the wining probability should be as small as possible, i.e., close to $\frac{1}{2}$. With these two observations in mind, we define the winning probability matrix as 
\begin{equation}
A_{ij} = \left\{
     \begin{array}{rl}
\frac{1}{2}  + \frac{1}{2} \frac{C_{ij}}{n-1}, & \;\; \text{ if } C_{ij} > 0 \\
\frac{1}{2} - \frac{1}{2} \frac{C_{ij}}{n-1}, & \;\; \text{ if }C_{ij} < 0. \\
0, & \;\; \text{ if } C_{ij} = 0. \\
     \end{array}
   \right.
\label{myA_cardinal_RC}
\end{equation}

\section{The Group Synchronization Problem}  \label{sec:GroupAngSync}

Finding group elements from noisy measurements of their ratios is known as the \textit{group synchronization} problem. For example, the synchronization problem over the special orthogonal group $SO(d)$ consists of estimating a set of $n$ unknown $d\times d$  matrices $R_1,\ldots,R_n \in SO(d)$ from noisy measurements of a subset of their pairwise ratios $R_i R_j^{-1}$
\begin{align}
	&  \underset{R_1,\ldots,R_n \in SO(d)}{\text{minimize}}  \sum_{(i,j) \in E}^{ } w_{ij} \|  R_i^{-1} R_j - R_{ij} \|_{F}^{2}  ,   
\label{genSyncMinimization}
\end{align}
where $||\cdot||$ denotes the Frobenius norm, and $w_{ij}$ are non-negative weights representing the confidence in the noisy pairwise measurements $R_{ij}$. Spectral and semidefinite programming relaxations for solving an instance of the above synchronization problem were originally introduced and analyzed by Singer \cite{sync} in the context of angular synchronization, over the group SO(2) of planar rotations, where one is asked to estimate $n$ unknown angles 
\begin{equation}
	\theta_1,\ldots,\theta_n \in [0,2\pi),
\end{equation}
given $m$ noisy measurements $\Theta_{ij}$ of their offsets 
\begin{equation}
	\Theta_{ij} = \theta_i - \theta_j \mod 2\pi.
	\label{eqTheta_ij}
\end{equation}
The difficulty of the problem is amplified on one hand by the amount of noise in the offset measurements, and on the other hand by the fact that $m \ll {n \choose 2}$, i.e., only a very small subset of all possible pairwise offsets are measured. In general, one may consider other groups $\mathcal{G}$ (such as SO($d$), O($d$)) for which there are available noisy measurements $g_{ij}$ of ratios between the group elements
$ g_{ij} = g_i g_j^{-1}, g_i, g_j \in \mathcal{G}.$ 
The set $E$ of pairs $(i,j)$ for which a ratio of group elements is available can be realized as the edge set of a graph $G=(V,E)$, $|V|=n, |E|=m$, with vertices corresponding to the group elements $g_1,\ldots,g_n$, and edges to the available pairwise measurements $ g_{ij} = g_i g_j^{-1}$. 
%
%
%
%

In \cite{sync}, Singer analyzed the following noise model, where each edge in the measurement graph $G$ is present with probability $p$, and each available measurement is either correct with probably $1-\eta$ or a random measurement with probability $\eta$. For such a noise model with outliers, the available measurement matrix $\Theta$ is given by the following mixture
\begin{equation}
\Theta_{ij} = \left\{
 \begin{array}{rll}
 \theta_i - \theta_j & \;\; \text{ for a correct edge,} & \text{ with probability } p (1-\eta) \\
 \sim Uniform(S^1)  & \;\; \text{ for an incorrect edge,} & \text{ with probability }  p \eta  \\
		     0      & \;\; \text{ for a missing edge},   & \text{ with probability } 1-p.  	\\
     \end{array}
   \right.
\label{AmitNoiseTheta}
\end{equation}
Using tools from random matrix theory, in particular rank-1 deformations of large random matrices \cite{FeralPeche}, Singer showed in \cite{sync} that for the complete graph (thus $p=1$),  the spectral relaxation for the angular synchronization problem given by (\ref{FinalRelaxAmitMaxi}) and summarized in the next Section \ref{secsec:syncEIG}, 
undergoes a phase transition phenomenon, with the top eigenvector of the Hermitian matrix $H$ in (\ref{mapToCircle}) 
exhibiting above random correlations with the underlying ground truth solution as soon as 
\begin{equation}
	1 - \eta > \frac{1}{\sqrt{n}}.
\end{equation}
In other words, even for very small values of $1 - \eta$ (thus a large noise level), the eigenvector synchronization method summarized in Section  \ref{secsec:syncEIG} is able to successfully recover the ground truth angles if there are enough pairwise measurements available, i.e., whenever $n (1-\eta)^2$ is large enough.
For the general case of Erd\H{o}s-R\'{e}nyi graphs, the same phenomenon is encountered as soon as 
$  1 - \eta >  \sqrt{ \frac{n^5}{8m^3}}$, where $m$ denotes the number of edges in the measurement graph $G$.

\subsection{Spectral Relaxation} \label{secsec:syncEIG}
Following the approach introduced in \cite{sync}, we build the $n \times n$ sparse Hermitian matrix $H = (H_{ij})$ whose elements are either $0$ or points on the unit circle in the complex plane
\begin{equation}
 {H}_{ij} = \begin{cases}
 e^{\imath \theta_{ij}} & \text{if } (i,j) \in E^P \\
  0 & \text{if } (i,j) \notin E^P.
\end{cases}
\label{mapToCircle}
\end{equation}
In an attempt to preserve the angle offsets as best as possible, Singer considers the following maximization problem
\begin{equation}
\underset{  \theta_1,\ldots,\theta_n \in [0,2\pi)   }{\text{maximize}}  \sum_{i,j=1}^{n} e^{-\iota \theta_i} H_{ij} e^{\iota \theta_j}
\label{AmitMaxim}
\end{equation}
which gets incremented by $+1$ whenever an assignment of angles $\theta_i$ and $\theta_j$ perfectly satisfies the given edge constraint $  \Theta_{ij} = \theta_i - \theta_j \mod 2\pi$ (i.e., for a \textit{good} edge), while the contribution of an incorrect assignment (i.e., of a \textit{bad} edge) will be uniformly distributed on the unit circle in the complex plane. 
Note that (\ref{AmitMaxim}) is equivalent to the formulation in 
(\ref{genSyncMinimization}) by exploiting properties of the Frobenius norm, and relying on the fact that it is possible to represent group elements for the special case of SO(2) as complex-valued numbers. 
%
%
Since the non-convex optimization problem in (\ref{AmitMaxim}) is difficult to solve computationally, Singer introduced the following spectral relaxation 
\begin{equation}
\underset{  z_1,\ldots,z_n \in \mathbb{C};  \;\;\; \sum_{i=1}^n |z_i|^2 = n  }{\text{maximize}} \;\; \sum_{i,j=1}^{n}  \bar{z_i} H_{ij} z_j
\label{RelaxAmitMaxi}
\end{equation}
by replacing the individual constraints $z_i = e^{\iota \theta_i}$ having unit magnitude by the much weaker single constraint $\sum_{i=1}^n |z_i|^2 = n $. Next, we recognize the resulting maximization problem in (\ref{RelaxAmitMaxi}) as the maximization of a quadratic form whose solution is known to be given by the top eigenvector of the Hermitian matrix $H$, 
which has an orthonormal basis over $\mathbb{C}^n$, with real eigenvalues $\lambda_1 \geq \lambda_2 \geq \ldots  \geq \lambda_n$ and corresponding  eigenvectors $v_1, v_2, \ldots, v_n$. In other words, the spectral relaxation of the non-convex optimization problem in (\ref{AmitMaxim}) is given by 
\begin{equation}
\underset{  || z || ^2 = n }{\text{maximize}} \; z^* H z
\label{FinalRelaxAmitMaxi}
\end{equation}
which can be solved via a simple eigenvector computation, by setting $z = v_1$, where $v_1$ is the top eigenvector of $H$, satisfying 
$H v_1 = \lambda_1 v_1$, with $ || v_1 ||^2= n$, corresponding to the largest eigenvalue $ \lambda_1$. Before extracting the final estimated angles, we consider the following normalization of $H$ using the diagonal matrix $D$, whose diagonal elements are given by
$D_{ii} = \sum_{j=1}^N |H_{ij}|$, and define
\begin{equation}
\mathcal{H} = D^{-1} H,
\label{Rnormalization}
\end{equation}
which is similar to the Hermitian matrix $D^{-1/2} H D^{-1/2}$ through
        $$\mathcal{H} = D^{-1/2} (D^{-1/2}  H D^{-1/2}) D^{1/2}. $$
Thus, $\mathcal{H}$ has $n$ real eigenvalues $\lambda_1^\mathcal{H}  > \lambda_2^\mathcal{H}  \geq \cdots \geq \lambda_n^\mathcal{H} $ with corresponding $n$ orthogonal (complex valued)  eigenvectors $v_1^\mathcal{H} ,\ldots,v_n^\mathcal{H} $, satisfying $\mathcal{H} v_i^\mathcal{H}  = \lambda_i^{\mathcal{H}} v_i^\mathcal{H} $. Finally, we define the estimated rotation angles $\hat{\theta}_1,...,\hat{\theta}_n$ 
using the top eigenvector $v_1^{\mathcal{H}}$ via  
\begin{equation}
\label{est-r}
e^{\iota \hat{\theta}_i} = \frac{v_1^{\mathcal{H}}(i)}{|v_1^{\mathcal{H}}(i)|} , \;\;\;\; i=1,2,\ldots, n
\end{equation}
Note that the estimation of the rotation angles $\theta_1,\ldots,\theta_n$ is up to an additive phase since $e^{i\alpha}v_1^{\mathcal{H}}$ is also an eigenvector of $\mathcal{H}$ for any $\alpha \in \mathbb{R}$.
We point out that the only difference between the above approach and the angular synchronization algorithm in \cite{sync} is the normalization (\ref{Rnormalization}) of the matrix prior to the computation of the top eigenvector, considered in our previous work \cite{asap2d}, and formalized
in \cite{AmitVDM} via the notion of \textit{graph connection Laplacian} $\mathcal{L} = D - H$ (or its normalized version), for which it can be shown that the bottom $d$  eigenvectors can be used to recover the unknown elements of SO($d$), after a certain rounding procedure.

\subsection{Semidefinite Programming Relaxation} \label{secsec:syncSDP}

A second relaxation proposed in \cite{sync} as an alternative to the spectral relaxation, is via the following semidefinite programming formulation. In an attempt to preserve the angle offsets as best as possible, one may consider the following maximization problem 
\begin{equation}
\sum_{i,j=1}^{n} e^{-\iota \theta_i} H_{ij} e^{\iota \theta_j} = \text{trace}(H \Upsilon),
\end{equation}
where $ \Upsilon $ is the (unknown) $n \times n$ Hermitian matrix of rank-1 given by 
\begin{equation}
\Upsilon_{ij} = e^{\iota (\theta_i-\theta_j)}
\label{rankOneCons}
\end{equation}
with ones in the diagonal $\Upsilon_{ii}, \forall i=1,2,\ldots,n$.
Note that, with the exception of the rank-1 constraint, 
all the remaining constraints are convex and altogether define the following SDP relaxation for the angular synchronization problem 
\begin{equation}
	\begin{aligned}
	& \underset{\Upsilon \in \mathbb{C}^{n \times n}}{\text{maximize}}
	& & trace( H \Upsilon) \\
	& \text{subject to}
	& & \Upsilon_{ii} = 1 & i=1,\ldots,n \\
		& & &   \Upsilon \succeq 0,
	\end{aligned}
 \label{SDP_program_SYNC}
\end{equation}
which can be solved via standard methods from the convex optimization literature \cite{Vandenberghe94sdp}.
We remark that, from a computational perspective, solving such SDP problems is computationally feasible only for relative small-sized problem (typically with  several thousand unknowns, up to about $n=10,000$), though there exist distributed methods for solving such convex optimization problems, such as the popular Alternating Direction Method of Multipliers (ADMM) \cite{Boyd_ADMM} which can handle large-scale problems arising nowadays in statistics and machine learning \cite{ADMM_Sparse_Low_Rank}.

As pointed out in \cite{sync}, this program is very similar to the well-known   Goemans-Williamson SDP relaxation for the famous MAX-CUT problem of finding the maximum cut in a weighted graph, the only difference being the fact that here we optimize over the cone of complex-valued Hermitian positive semidefinite matrices, not just real symmetric matrices.
Since the recovered solution is not necessarily of rank-1, the estimator is obtained from the best rank-1 approximation of the optimal solution matrix $\Theta$ via a Cholesky decomposition. We  plot in Figure \ref{fig:rkSDP_Meth6_n200_num} the recovered ranks of the SDP relaxation for the ranking problem, and point out the interesting phenomenon that, even for noisy data, under favorable noise regimes, the SDP program is still able to find a rank-1 solution. The tightness of this relaxation has been explained only recently in the work of Bandeira et al. \cite{bandeira2014tightness}.
The advantage the SDP relaxation brings is that it explicitly imposes the unit magnitude constraint for $e^{\iota \theta_i}$, which we cannot otherwise enforce in the spectral relaxation solved via the eigenvector method.

\section{Sync-Rank: Ranking via Synchronization}
\label{sec:Sync-Rank}

We now consider the application of the angular synchronization framework \cite{sync} to the ranking problem. The underlying idea has also been considered in \cite{StellaAE} in the context of image denoising, who suggested, similar to \cite{sync}, to perform the denoising step in the angular embedding space as opposed to the linear space, and observed increased robustness against sparse outliers in the measurements.

\begin{figure}[h!]
\begin{center}
\includegraphics[width=0.35\textwidth]{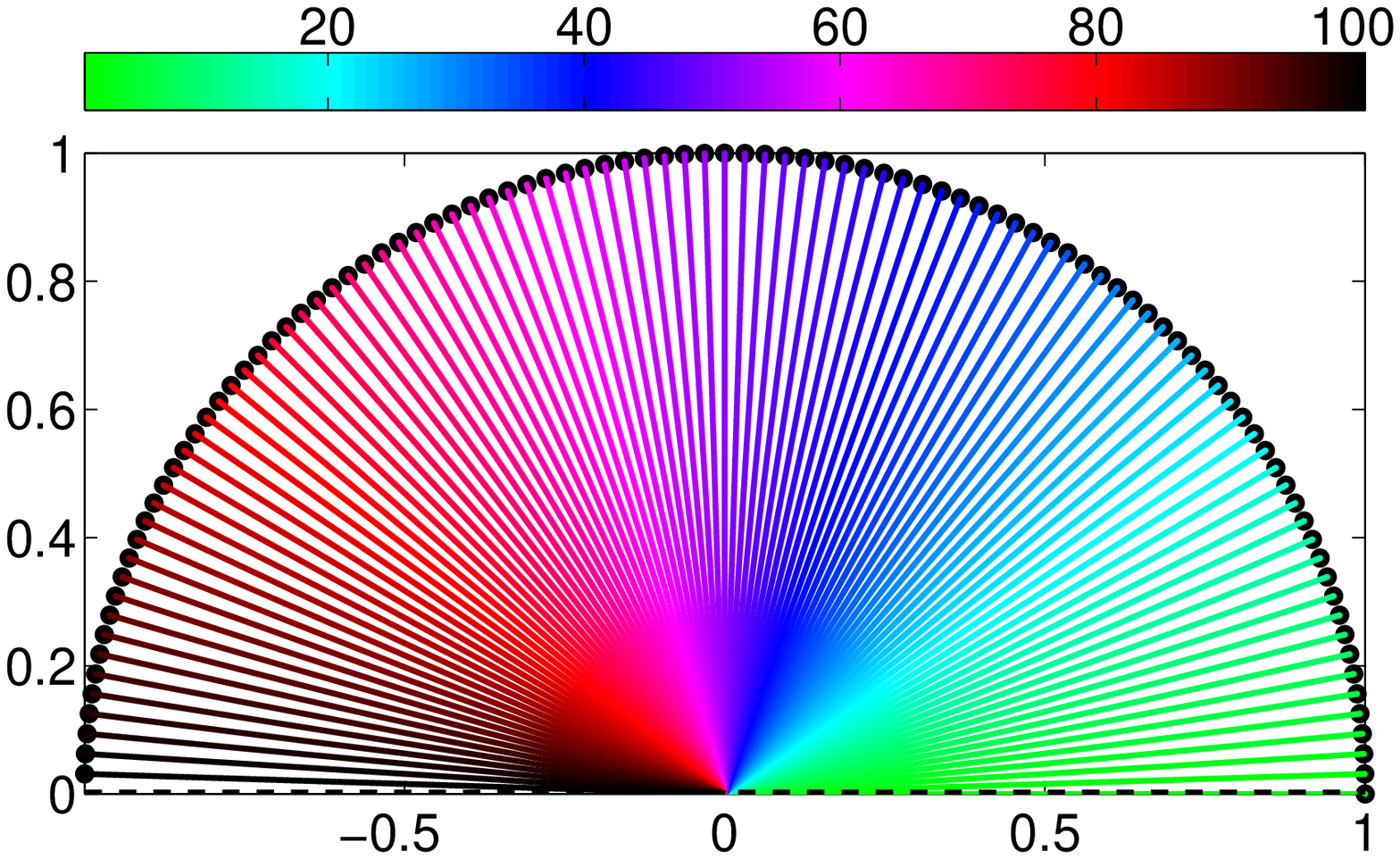}
\includegraphics[width=0.25\textwidth]{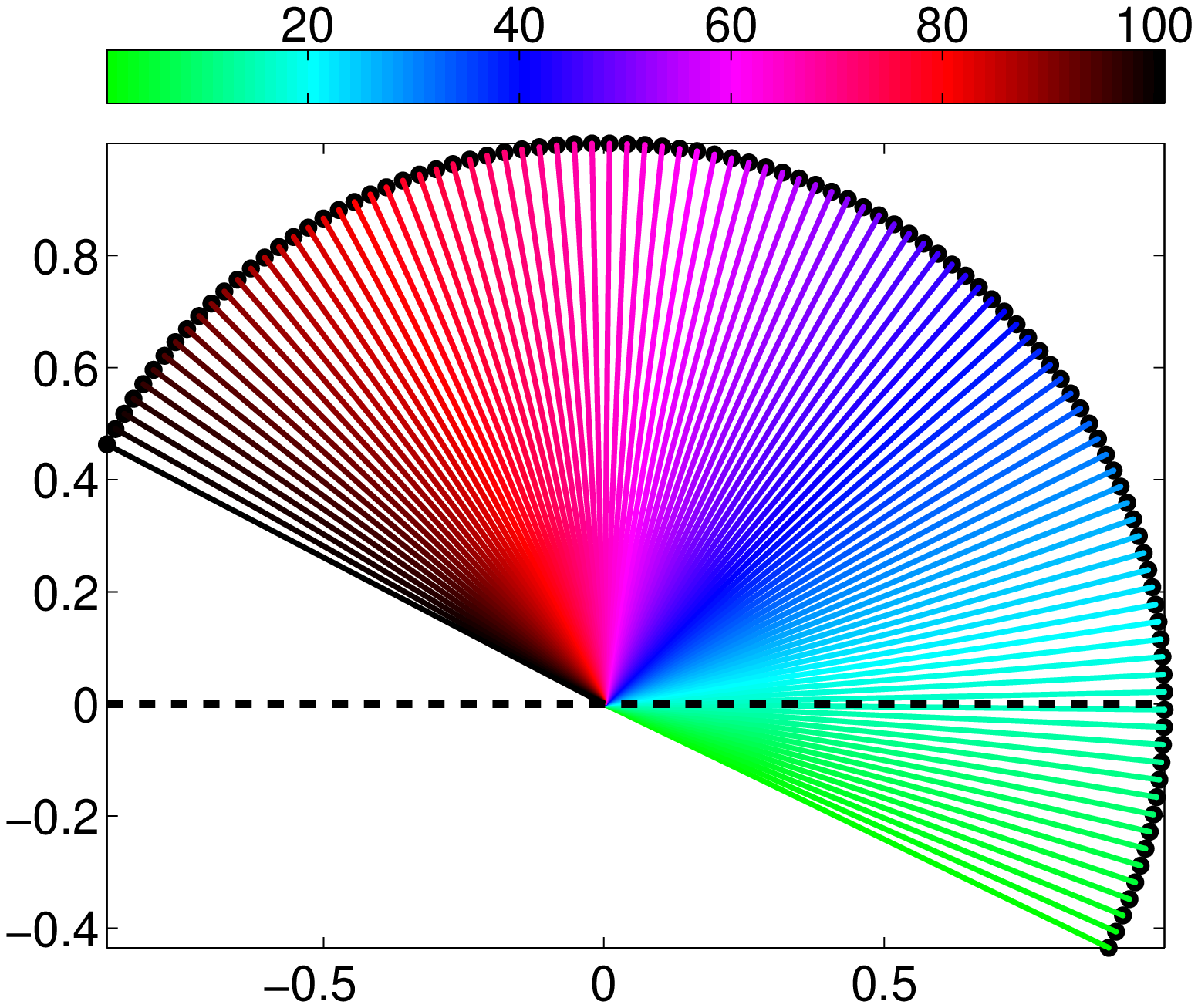}
\end{center}
\vspace{-3mm}
\caption{ (a) Equidistant mapping of the rankings $1,2,\ldots,n$ around half a circle, for $n=100$, where the rank of the $i^{th}$ player is $i$. (b) The recovered solution at some random rotation, motivating the step which computes the best circular permutation of the recovered rankings, which minimizes the number of upsets with respect to the initially given pairwise   measurements.}
\label{fig:wrapHalfCircleEx}
\end{figure}

Denote the true ranking of the $n$ players to be $ r_1 < r_2 < \ldots < r_n$, and assume without loss of generality that $r_i = i$, i.e., the rank of the $i^{th}$ player is $i$. In the ideal case, the ranks can be imagined to lie on a one-dimensional line, sorted from $1$ to to $n$, with the pairwise rank comparisons given, in the noiseless case, by $C_{ij} = r_i - r_j$ (for cardinal measurements) or $C_{ij} = \mbox{sign}(r_i - r_j)$ (for ordinal measurements). In the angular embedding space, we consider the ranks of the players mapped to the unit circle, say fixing $r_1$ to have a zero angle with the $x$-axis, and the last player $r_n$ corresponding to angle equal to $\pi$. In other words, we imagine the $n$ players wrapped around a fraction of the circle, interpret the available rank-offset measurements as angle-offsets in the angular space, and thus arrive at the setup of the angular synchronization problem detailed in Section \ref{sec:GroupAngSync}.

We also remark that the modulus used to wrap the players around the circle plays an important role in the recovery process. If we choose to map the players across the entire circle, this would cause ambiguity at the end points, since the very highly ranked players will be positioned very close (or perhaps even mixed) with the very poorly ranked players. To avoid the confusion, we simply choose to map the $n$ players to the upper half of the unit circle $[0,\pi]$.

In the ideal setting, the angles obtained via synchronization would be as shown in the left plot of Figure \ref{fig:wrapHalfCircleEx}, from where one can easily infer the ranking of the players by traversing the upper half circle in an anti-clockwise direction. However, since the solution to the angular synchronization problem is computed up to a global shift  (see, for example, the right plot of Figure \ref{fig:wrapHalfCircleEx}), an additional post-processing step is required to accurately extract the underlying ordering of the players that best matches the observed data. To this end, we simply compute the best circular permutation of the initial  rankings obtained from synchronization, that minimizes the number of upsets in the given data. We illustrate this step with an actual noisy instance of Sync-Rank in Figure \ref{fig:SyncExCircular_n100_025}, where plot (a) shows the rankings induced by the initial angles recovered from synchronization, while plot (b) shows the final ranking solution, obtained after shifting by the best circular permutation.

Denote by $\text{\boldmath$s$} = [s_1, s_2, \ldots, s_n]$ the ordering induced by the angles recovered from angular synchronization, when sorting the angles from smallest to largest, where $s_i$ denotes the label of the player ranked on the $i^{th}$ position. For example, $s_1$ denotes the player with the smallest corresponding angle $\theta_{x_1}$.
To measure the accuracy of each candidate circular permutation $\sigma$, we first compute the  pairwise rank offsets associated to the induced ranking, via
\begin{equation}
P_{\sigma}( \text{\boldmath$s$} ) =  \left( \sigma( \text{\boldmath$s$} ) \otimes \mb{1}^T -  \mb{1}  \otimes  \sigma( \text{\boldmath$s$} )^T  \right) \circ A
\label{distTwoRanks}
\end{equation}
where $\otimes$ denotes the outer product of two vectors $ x \otimes y = x y^T$, 
$\circ$ denotes the Hadamard product of two matrices (entrywise product), and $A$ is the adjacency matrix of the graph $G$.
In other words, for an edge $ (u,v) \in E(G)$, it holds that 
$ \left( P_{\sigma}( \text{\boldmath$s$} ) \right)_{uv} = \sigma(s)_u - \sigma(s)_v$, i.e., the resulting rank offset after applying the cyclic shift \footnote{The circular or cyclic shift $\sigma$ is given by $ \sigma(i) = (i+1) \text{ mod } n$. The result of repeatedly applying circular shifts to a given $n$-tuple $[x_1, \ldots,x_n]$ are often denoted circular shifts of the tuple.}. Next, we choose the circular permutation which minimizes the $l_1$ norm\footnote{This is just the $l_1$ norm of the vectorized form of the matrix 
$ \norm{X}_1 =   \sum_{i=1}^{n} \sum_{j=1}^{n} | X_{ij}| $} of the  residual matrix
\begin{equation}
\sigma =  \underset{\sigma_1,\ldots,\sigma_n }{\text{arg min} }  \;\;\;
 \frac{1}{2} \norm{  \mbox{sign}( P_{\sigma_i}( \text{\boldmath$s$} )) -  \mbox{sign}(C) }_1  
\label{argMinSign}
\end{equation}
which counts the total number of upsets.
Note that an alternative to the above error  is given by 
\begin{equation}
\sigma =  \underset{\sigma_1,\ldots,\sigma_n }{\text{arg min} }  \;\;\;
\norm{ P_{\sigma_i}( \text{\boldmath$s$} ) - C }_1  
\label{argMinNoSign}
\end{equation}
which takes into account the actual magnitudes of the offsets, not just their sign.

\begin{figure}[h!]
\begin{center}
\includegraphics[width=0.3\textwidth]{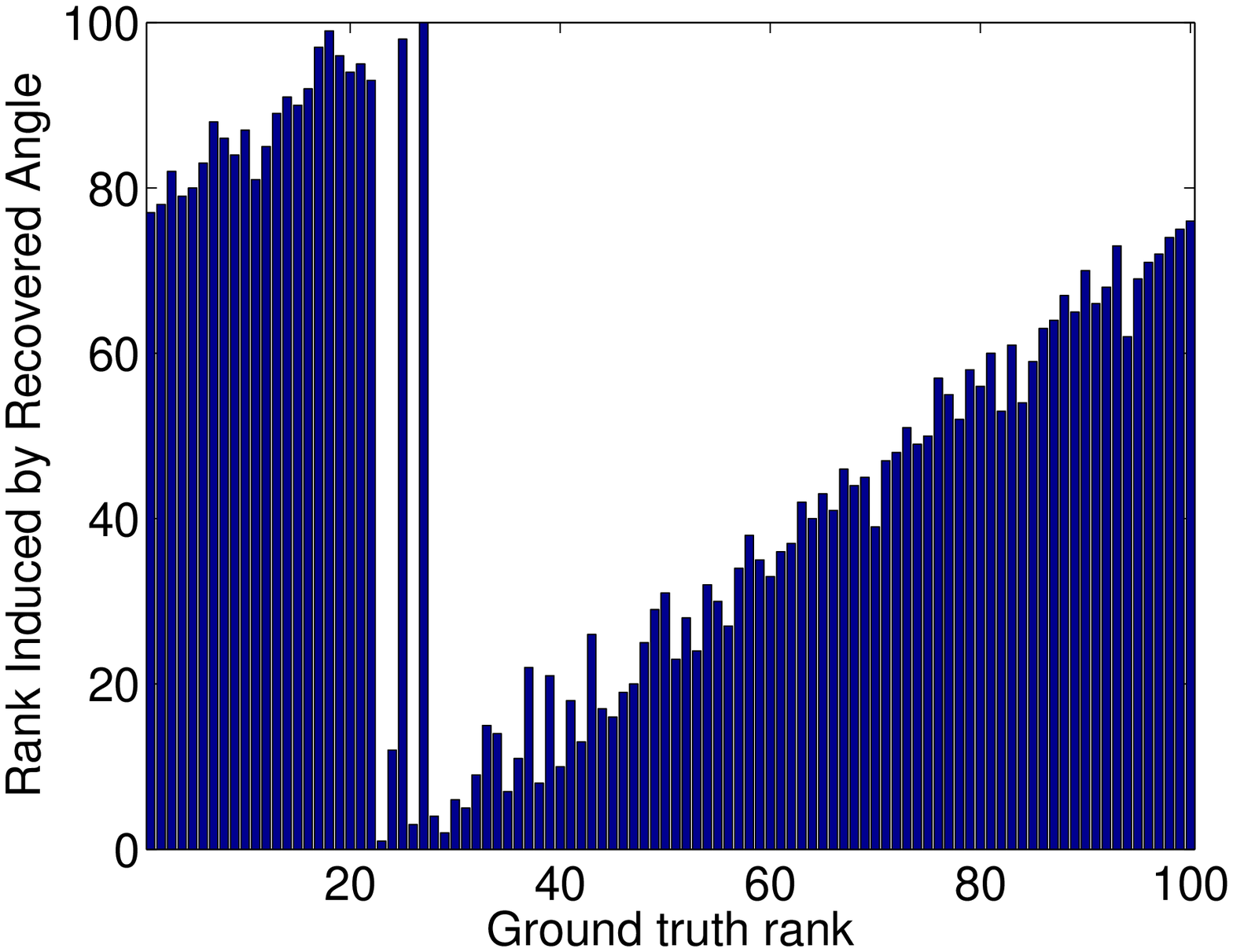}
\includegraphics[width=0.3\textwidth]{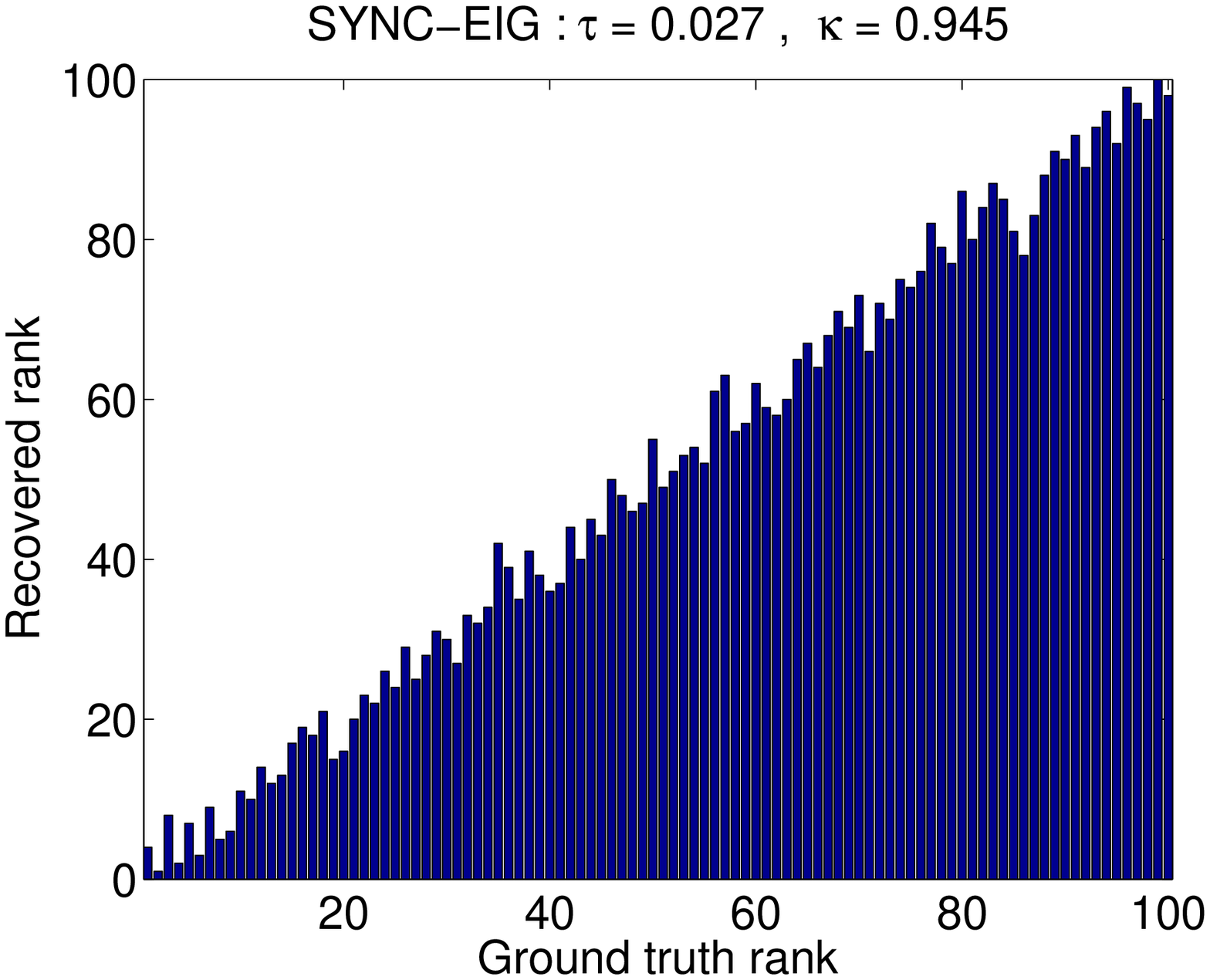}
\end{center}
\vspace{-2mm}
\caption{An illustration of the steps for recovering the ranking from the eigenvector synchronization approach, for an Erd\H{o}s-R\'{e}nyi measurement graph $G(n=100,p=0.5)$ with cardinal comparisons, and outliers chosen uniformly at random with probability $\eta = 0.25$ as in the ERO model (\ref{ERoutliers}).   
Left: the ranking induced by the initially obtained $\theta$ angles, estimated via angular synchronization. 
Right: the ranking obtained after shifting the set of angles by the best circular permutation which minimizes the number of upsets (i.e., inverted rankings) in the available measurements;  $\tau$ (respectively, $\kappa$) denote the Kendall distance (respectively, correlation) between the recovered and the ground truth ranking.}
\label{fig:SyncExCircular_n100_025}
\end{figure}

We summarize  the above steps of the Sync-Rank approach 
in Algorithm \ref{Algo:listSync}. Note that, throughout the paper, we denote by SYNC (or SYNC-EIG) the version of synchronization-based ranking which relies on the spectral relaxation of the synchronization problem, and by SYNC-SDP the one obtained via the SDP relaxation.
\begin{algorithm}[h!]
\begin{algorithmic}[1]
\REQUIRE $G=(V,E)$ the graph of pairwise comparisons and $C$ the $n \times n $ matrix of pairwise comparisons (rank offsets), such that whenever $ij \in E(G)$ we have available a (perhaps noisy) comparison between players $i$ and $j$, either a cardinal comparison ($C_{ij} \in [-(n-1), (n-1)]$) or an ordinal comparison $C_{ij} = \pm 1$.
\STATE Map all rank offsets $C_{ij}$ to an angle $\Theta_{ij} \in [0, 2 \pi  \delta )$ with $\delta \in [0,1)$, using the transformation 
\begin{equation}
	C_{ij}  \mapsto \Theta_{ij} :=  2 \pi  \delta  \frac{ C_{ij}}{n-1} 
	\label{transfToCircle}
\end{equation}
We choose $\delta = \frac{1}{2}$, and hence $\Theta_{ij} :=  \pi  \frac{ C_{ij}}{n-1} $.
\STATE Build the $n \times n$  Hermitian matrix $H$  
with $ {H}_{ij} = e^{\imath \theta_{ij}}  \text{, if } (i,j) \in E $, and $ {H}_{ij}=0$ otherwise, as in (\ref{mapToCircle}).
\STATE Solve the angular synchronization problem  via either its spectral  (\ref{FinalRelaxAmitMaxi}) or SDP (\ref{SDP_program_SYNC}) relaxation, and denote the recovered solution by 
$ \hat{r}_i =  e^{\imath \hat{\theta}_i} = \frac{v_1^R(i)}{|v_1^R(i)|} , \;\;\;\; i=1,2,\ldots, n$, where $v_1$ denotes the recovered eigenvector
\STATE Extract the corresponding set of angles $\hat{\theta}_1,\ldots,\hat{\theta}_n \in [0,2\pi)$ from $\hat{r}_1, \ldots,\hat{r}_n$.
\STATE Order the set of angles $\hat{\theta}_1,\ldots,\hat{\theta}_n$ in increasing order, and denote the induced ordering  by $ \text{\boldmath$s$} = s_1, \ldots, s_n$.
\STATE Compute the best circular permutation $\sigma$ of the above ordering \text{\boldmath$s$} that minimizes the resulting number of upsets  with respect to the initial rank comparisons given by $C$
\begin{equation}
\sigma =  \underset{\sigma_1,\ldots,\sigma_n }{\text{arg min} }  \;\;\;
 || \text{sign} (P_{\sigma_i}( \text{\boldmath$s$} )  ) -  \text{sign}(C) ||_1  
\end{equation}
with  $P$ defined as in (\ref{distTwoRanks}).
\STATE  Output as a final solution the ranking induced by the circular permutation $\sigma$.
\end{algorithmic}
\caption{ Summary of the Synchronization-Ranking (Sync-Rank) Algorithm }
\label{Algo:listSync}
\end{algorithm}
Figure \ref{fig:BarpsMethodComp} is a comparison of the rankings obtained by the different methods: SVD, LS, SER, SER-GLM, RC, and SYNC-EIG, for an instance of the ranking problem given by the Erd\H{o}s-R\'{e}nyi measurement graph 
$G(n=100,p=0.5)$ with cardinal comparisons, and outliers chosen uniformly at random with probability $\eta = \{ 0, 0.35, 0.75\} $, according to the ERO($n,p,\eta$) noise model given by (\ref{ERoutliers}).
Note that at low levels of noise, all methods yield satisfactory results, but as the noise level increases, only Sync-Rank is able to recover a somewhat accurate solution, and significantly outperforms the results obtained by any of the other methods in terms of the number of flips (i.e., Kendall distance) with respect to the original ranking.

\begin{figure}[h!]
\begin{center}
\subfigure[ $\eta = 0$ ]
{\includegraphics[width=0.18\textwidth]{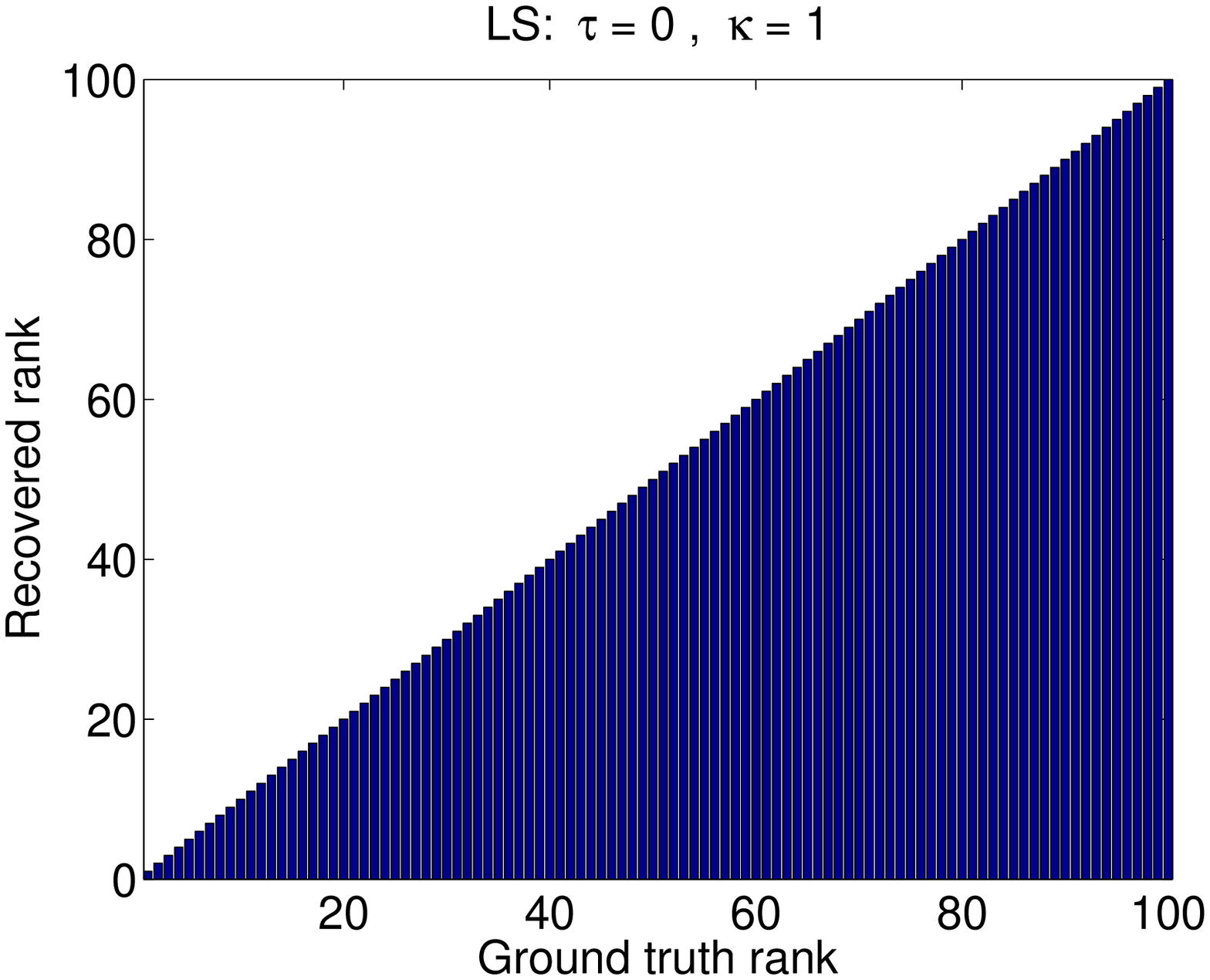}
\includegraphics[width=0.18\textwidth]{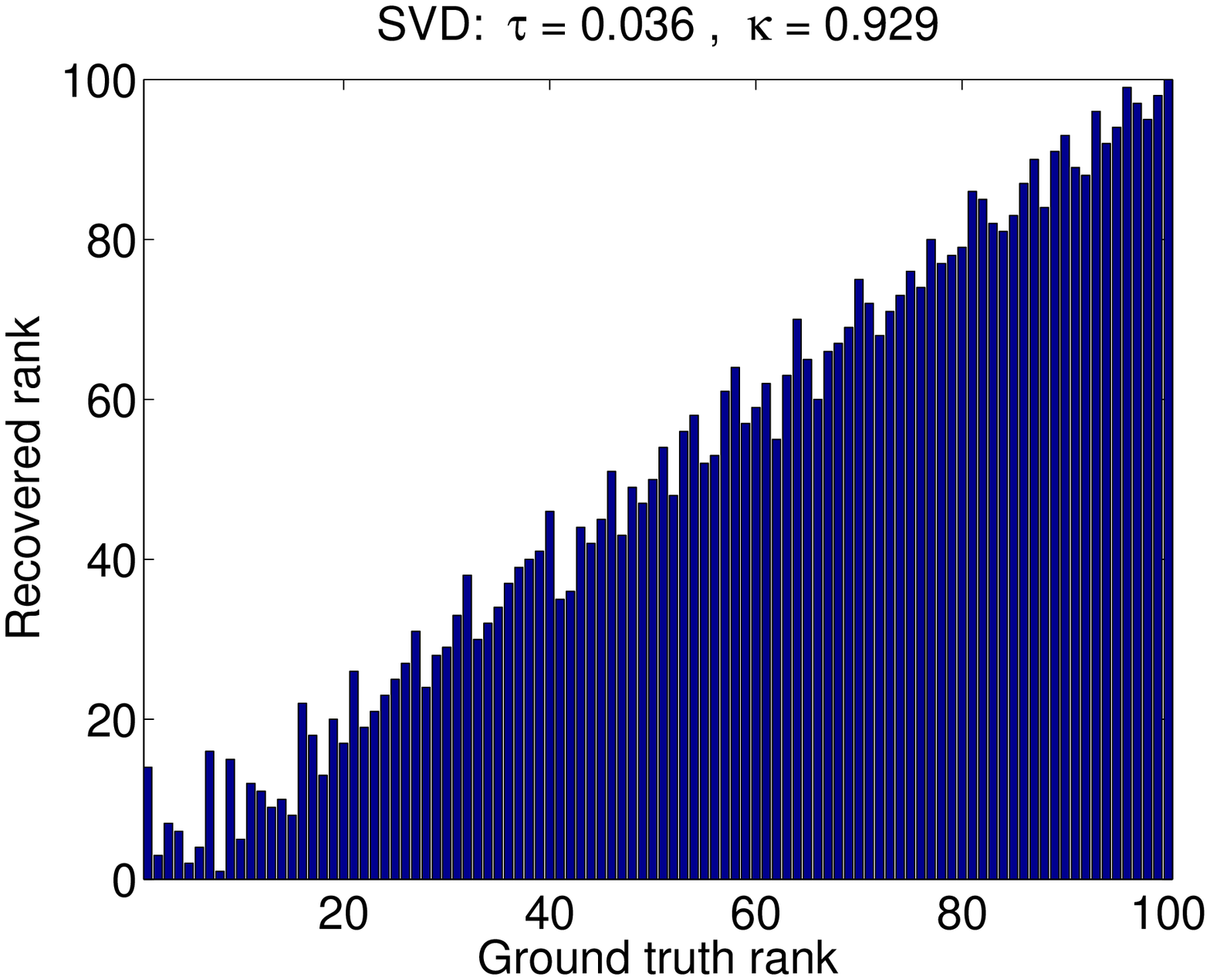}
\includegraphics[width=0.18\textwidth]{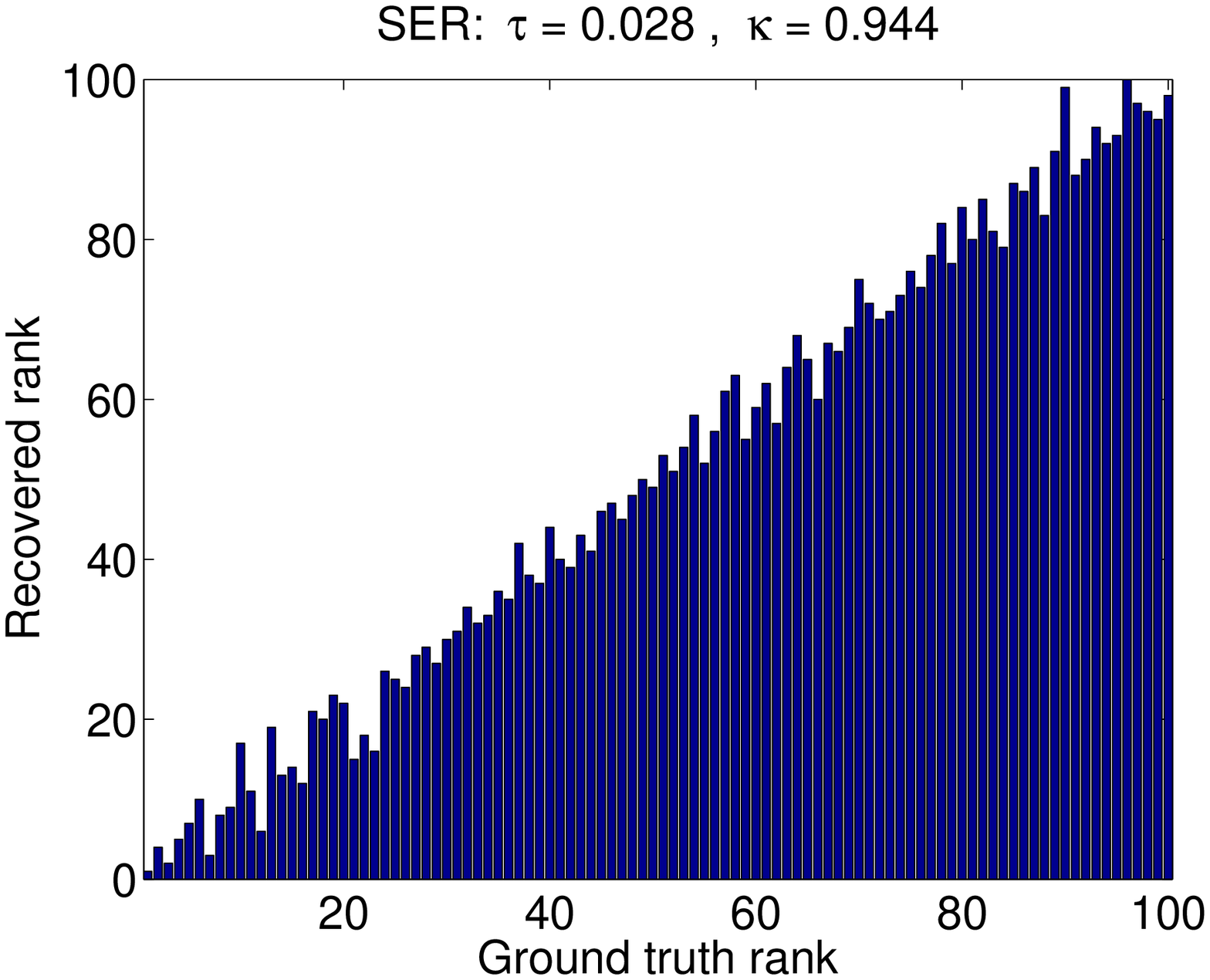}
\includegraphics[width=0.18\textwidth]{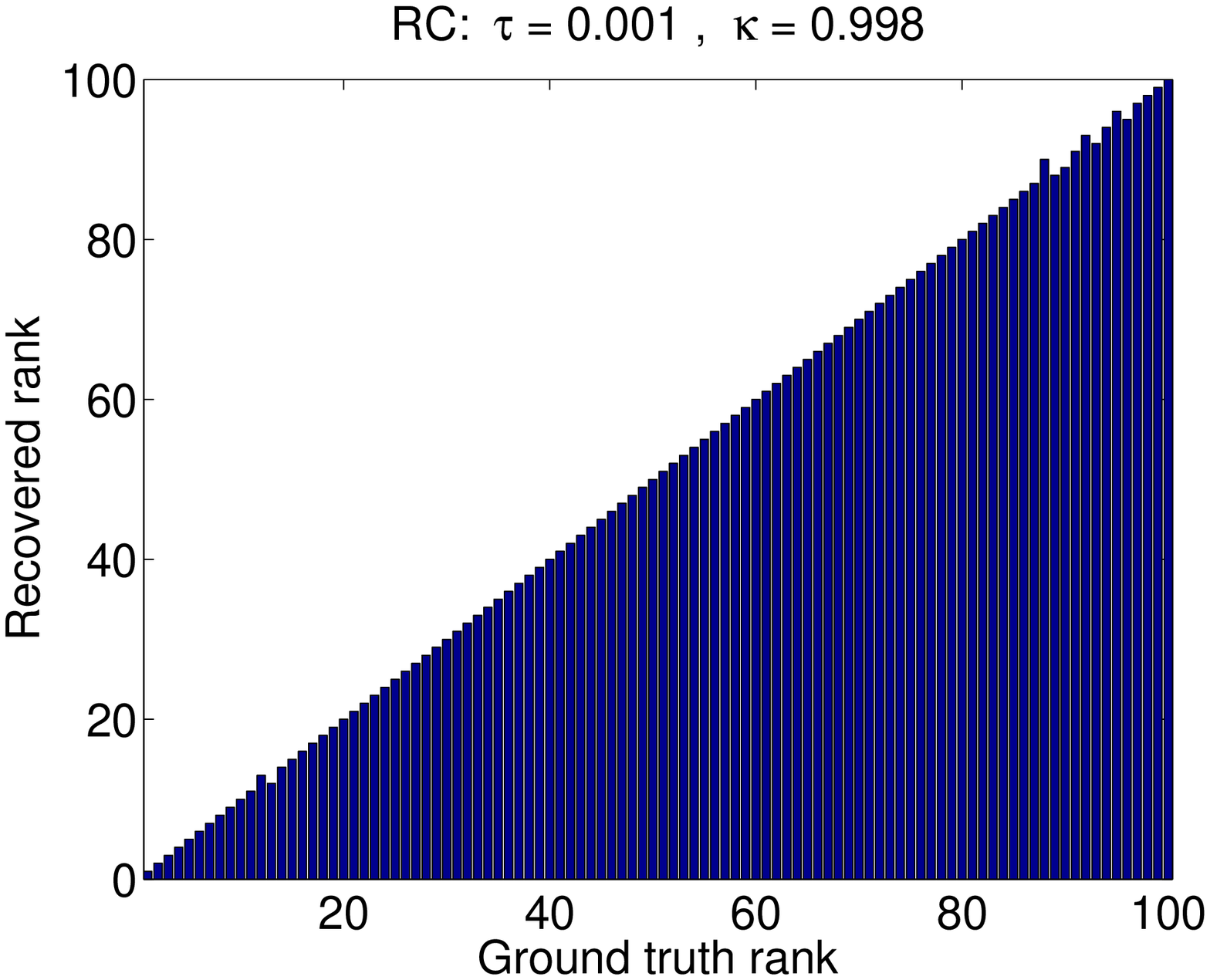}
\includegraphics[width=0.18\textwidth]{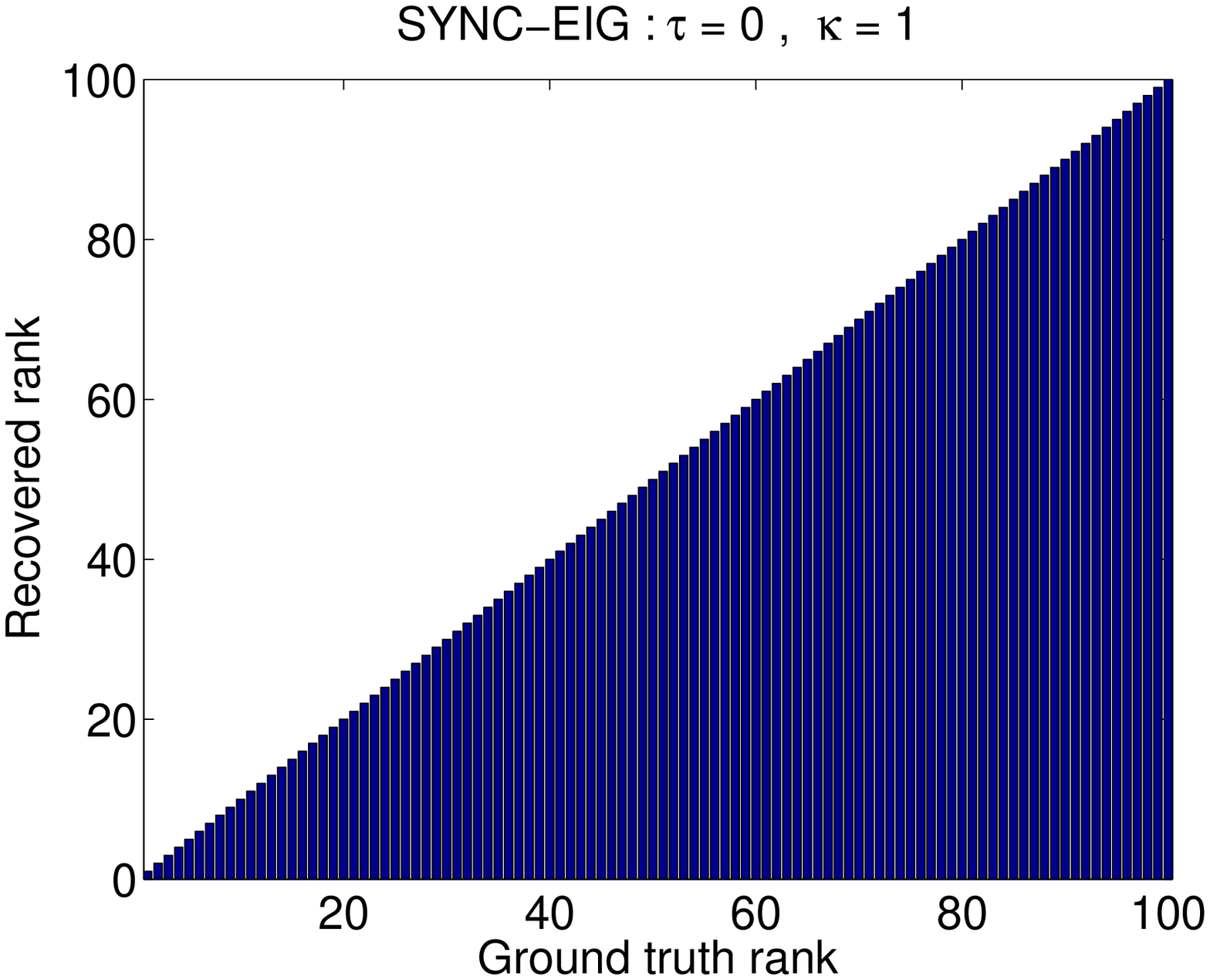}
}
\vspace{-3mm}
\subfigure[ $\eta = 0.35$ ]
{\includegraphics[width=0.18\textwidth]{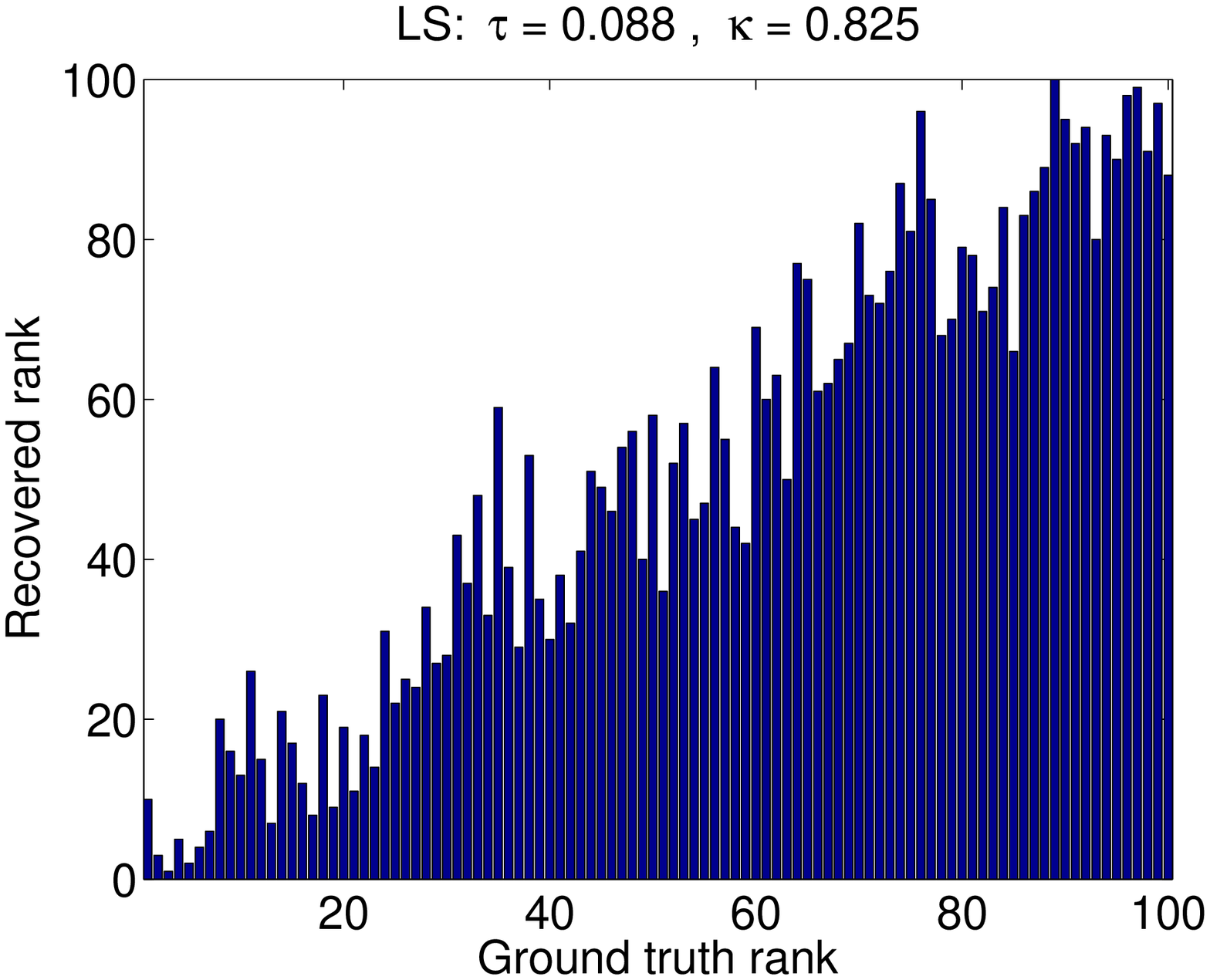}
\includegraphics[width=0.18\textwidth]{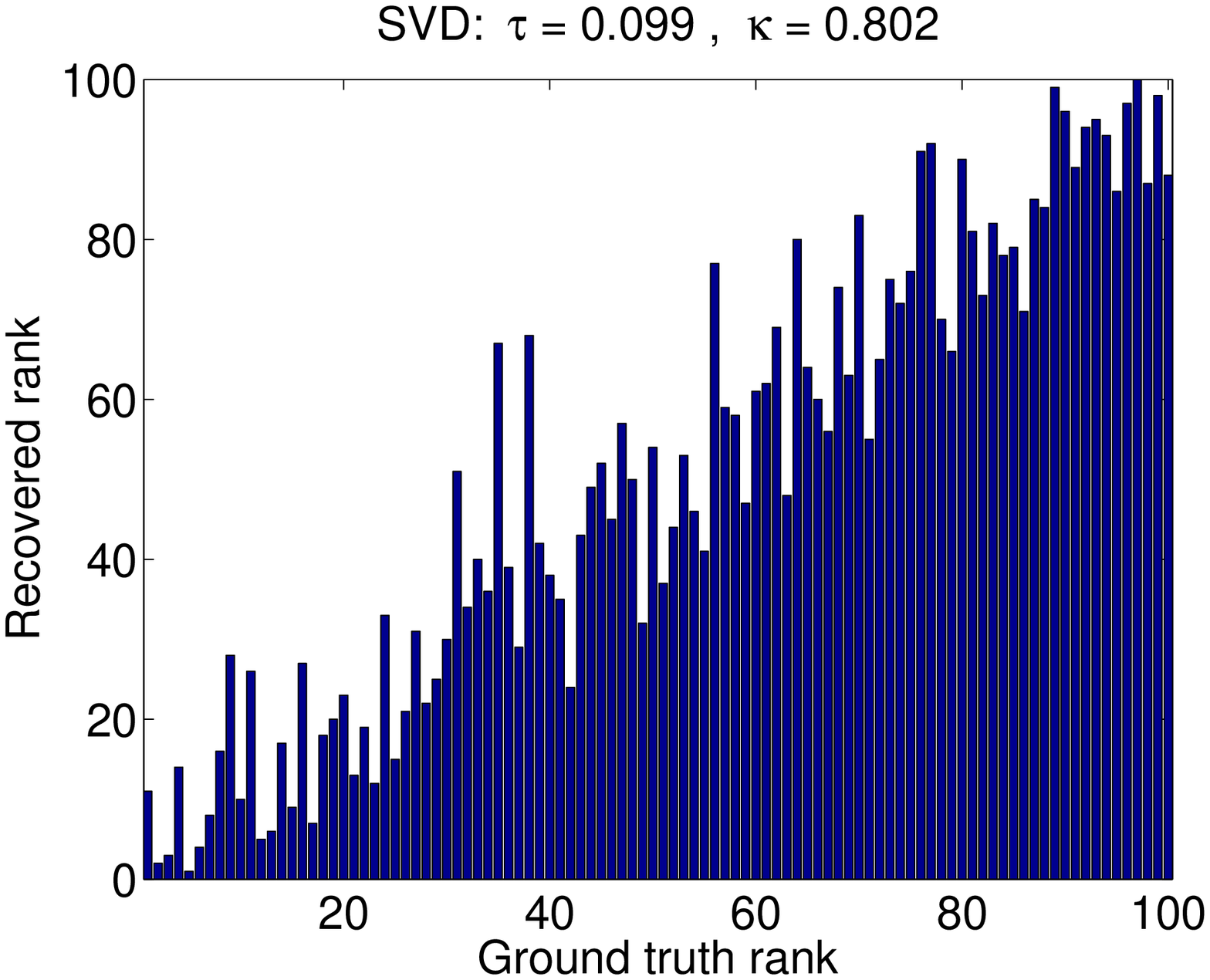}
\includegraphics[width=0.18\textwidth]{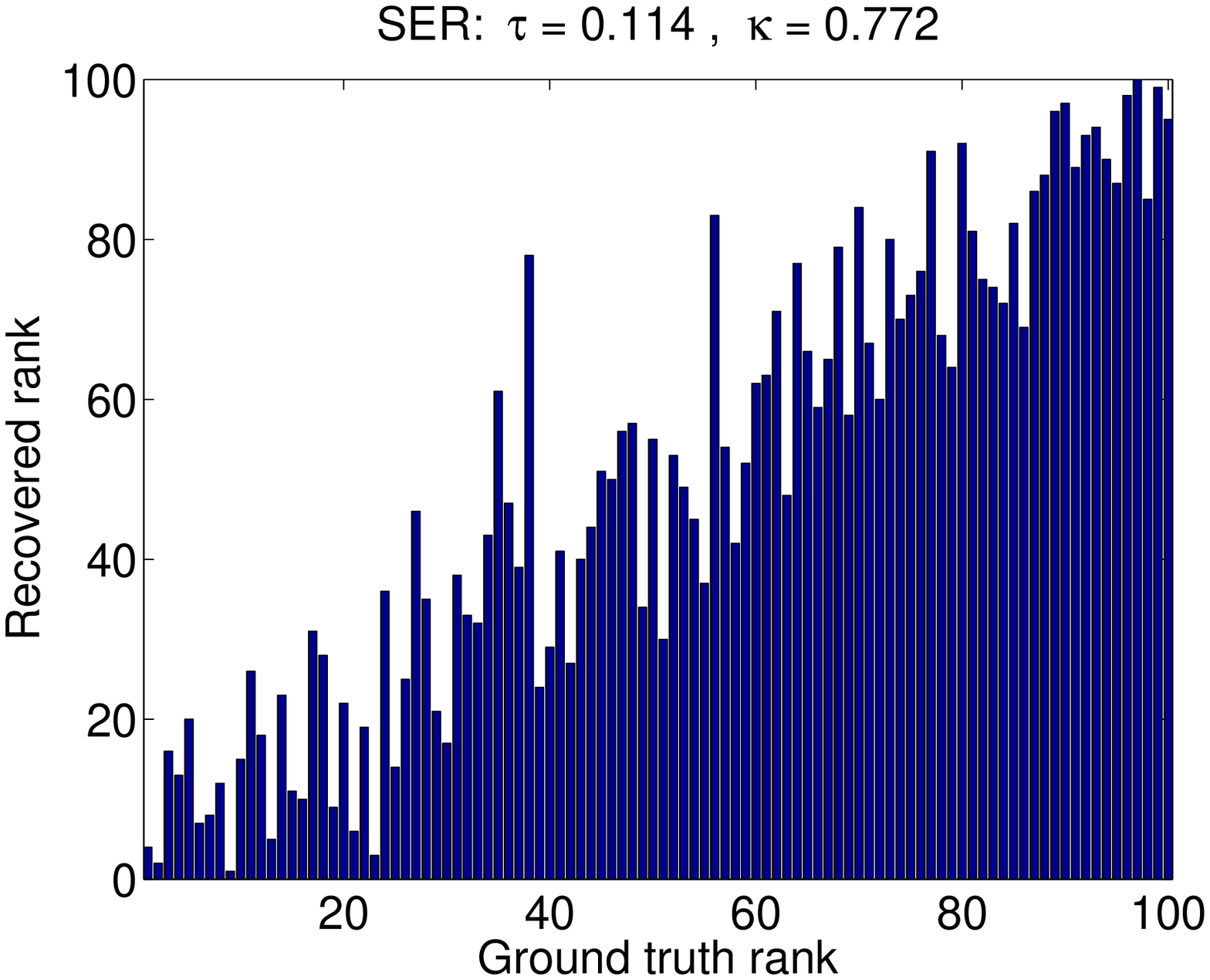}
\includegraphics[width=0.18\textwidth]{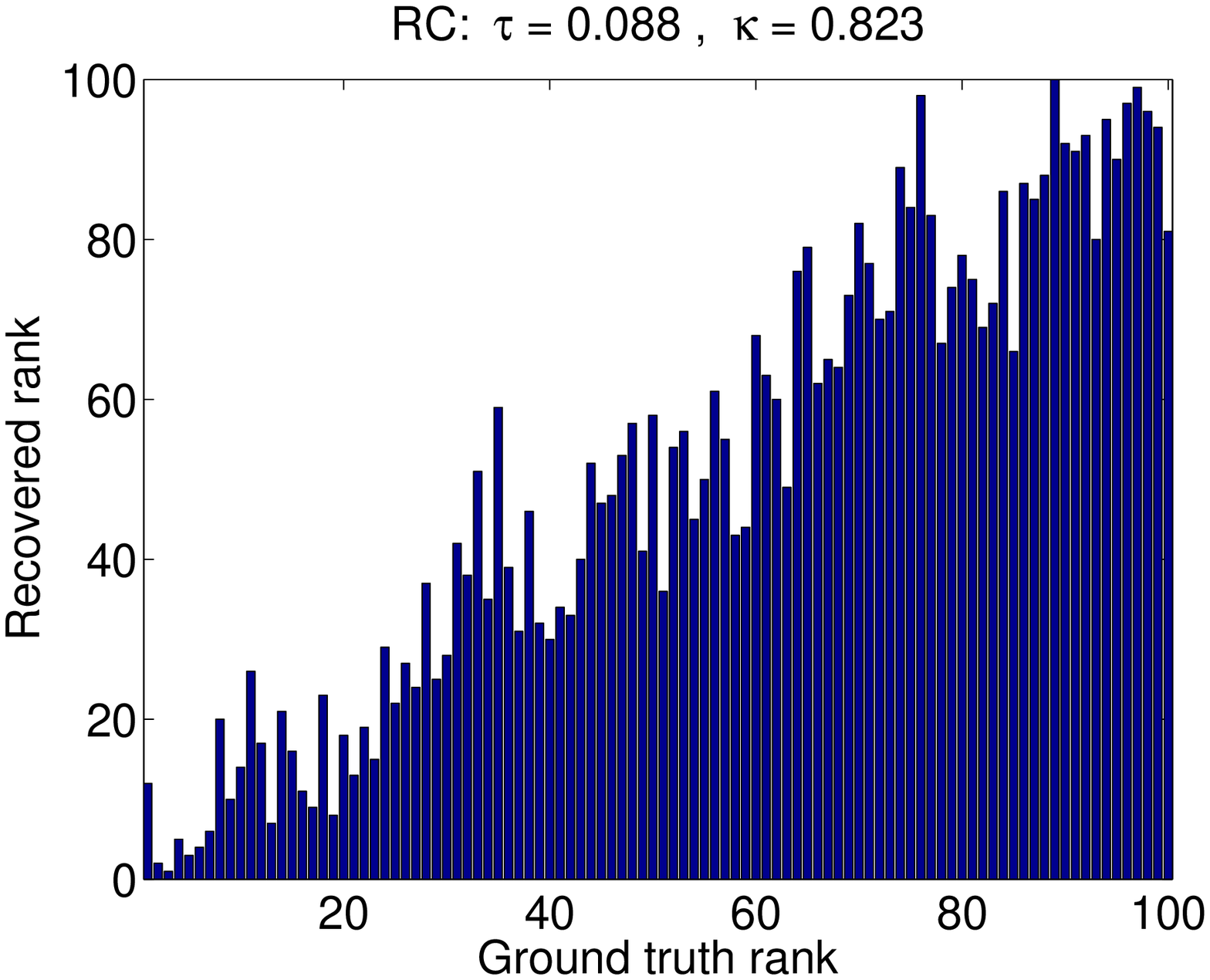}
\includegraphics[width=0.18\textwidth]{PLOTS/ExBars/FinalRankSync_eig_sim0_OutliersER_n100_p0p5_num_eta0p35.eps}
}
\vspace{-3mm}
\subfigure[ $\eta = 0.75$ ]
{\includegraphics[width=0.18\textwidth]{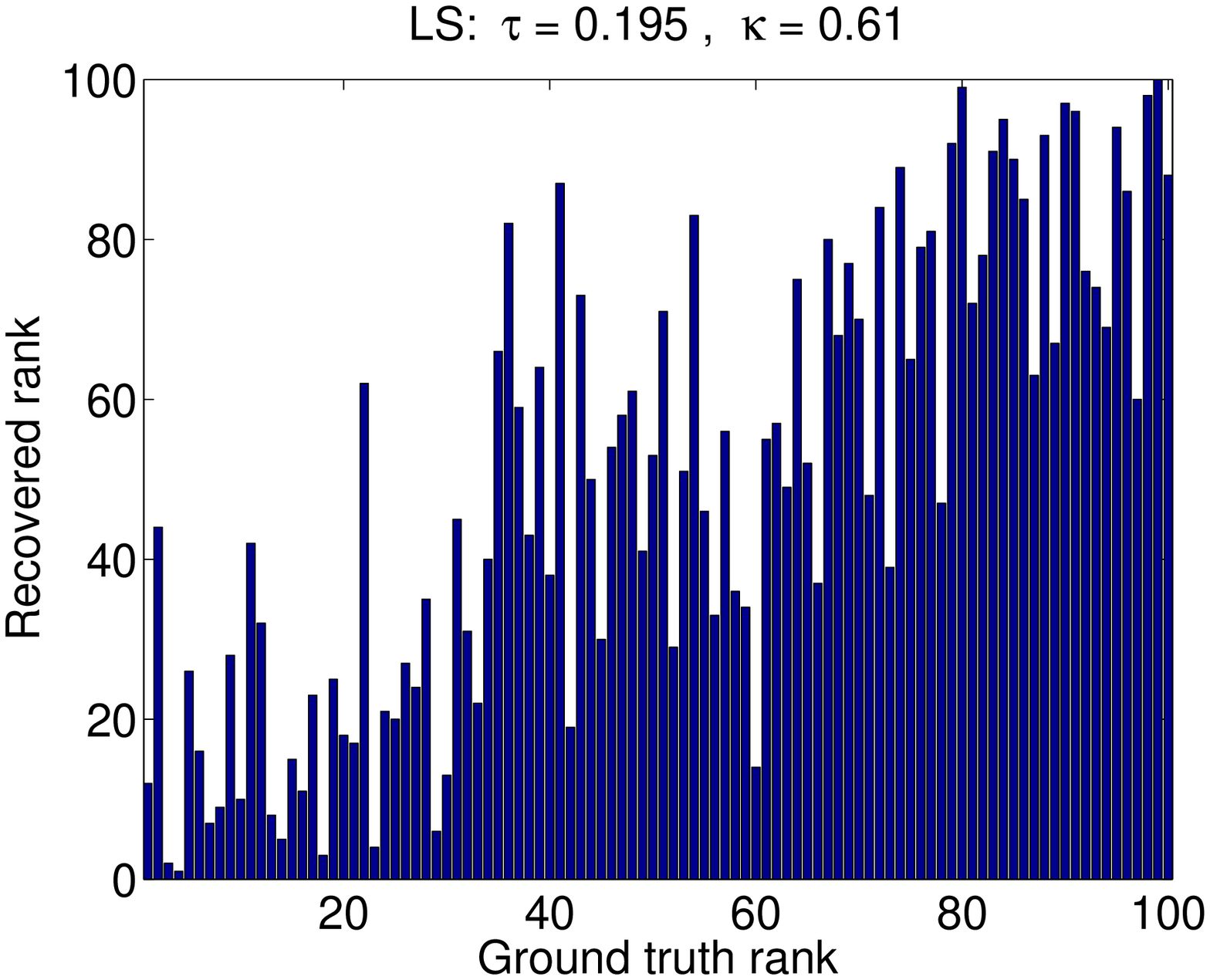}
\includegraphics[width=0.18\textwidth]{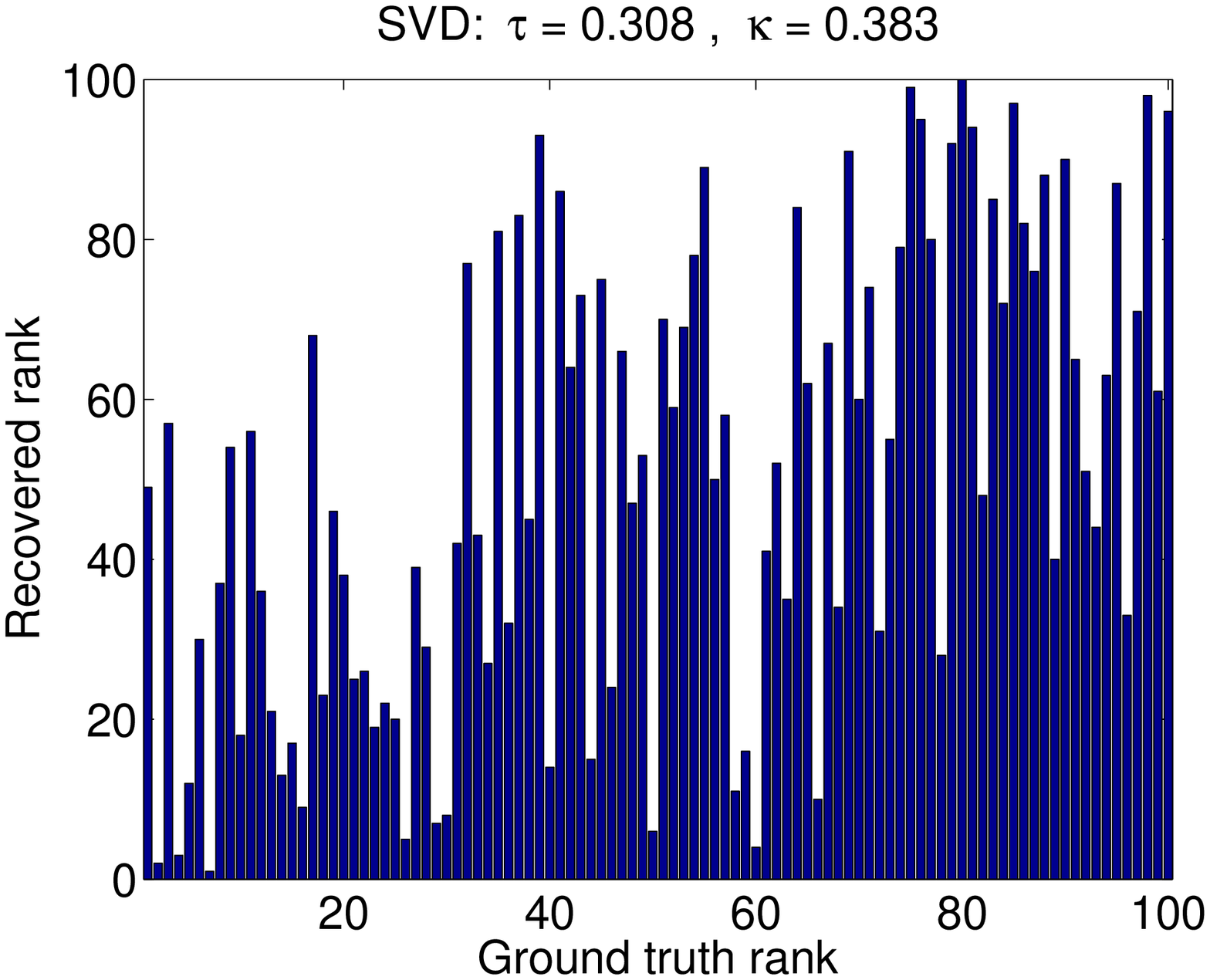}
\includegraphics[width=0.18\textwidth]{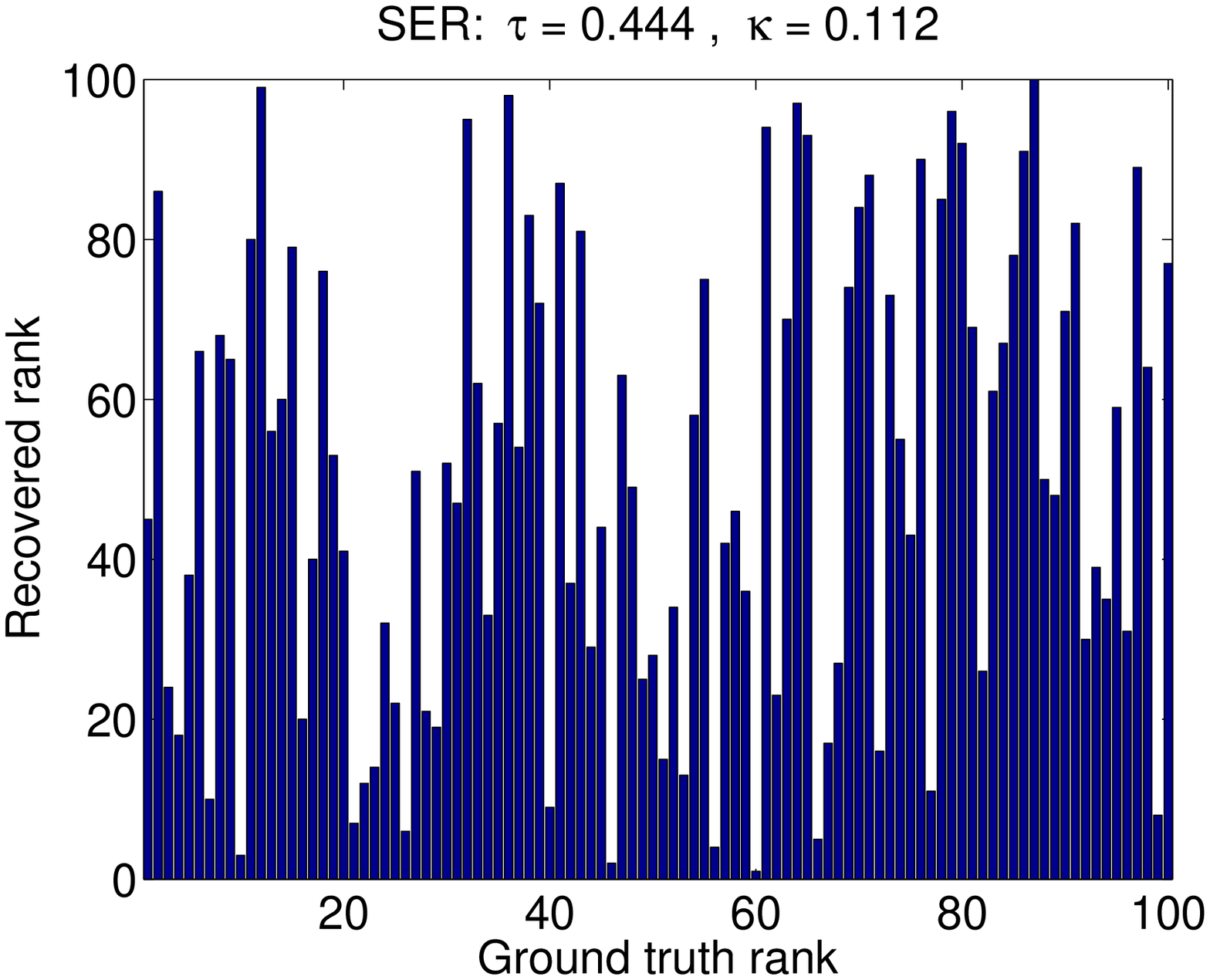}
\includegraphics[width=0.18\textwidth]{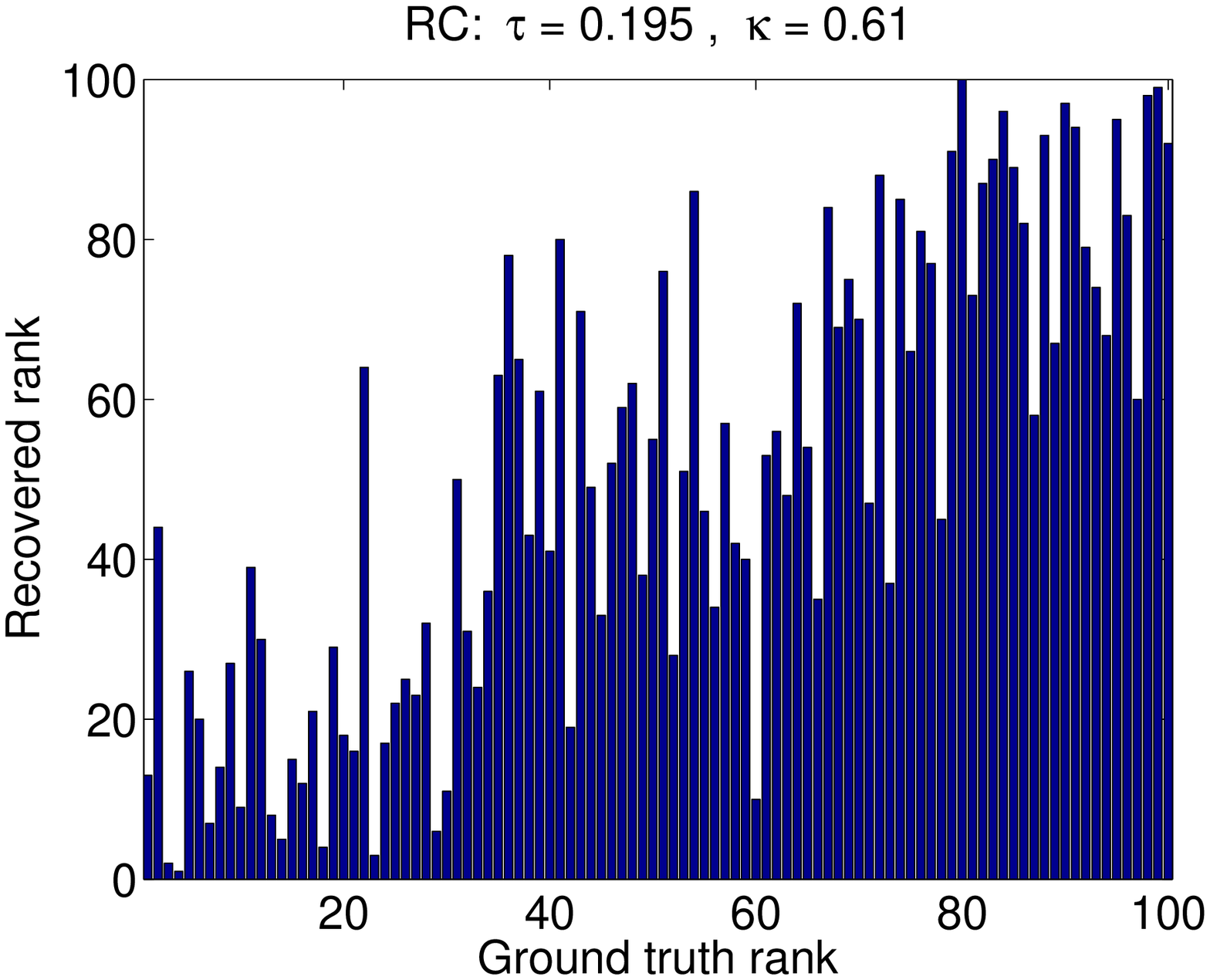}
\includegraphics[width=0.18\textwidth]{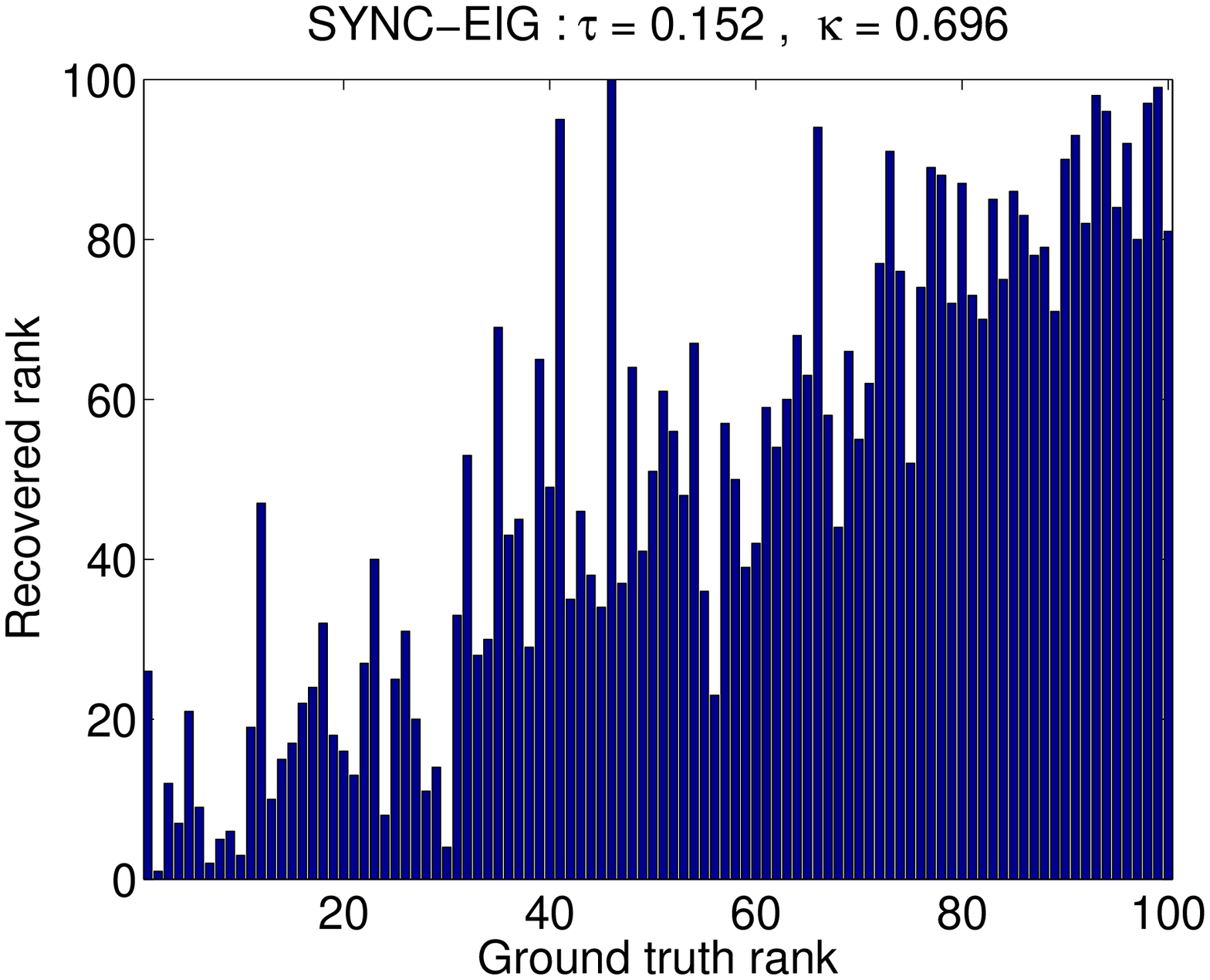}
}
\end{center}
\caption{A comparison of the rankings obtained by the different methods: LS, SVD, SER, RC and SYNC-EIG (eigenvector synchronization), for the graph ensemble $G(n=100,p=0.5)$ with cardinal comparisons, and outliers chosen uniformly at random with probability $\eta = \{ 0, 0.35, 0.75\} $, according to the ERO($n, \eta$) noise model given by (\ref{ERoutliers}). We omit the results obtained via SER-GLM and SYNC-SDP, since they were very similar to SER, respectively SYNC-EIG, for this particular instance.}
\label{fig:BarpsMethodComp}
\end{figure}

%




\subsection{Synchronization Ranking for Ordinal Comparisons}
\label{secsec:syncWeighted}
When the pairwise comparisons are ordinal, and thus $C_{ij}= \pm 1, \forall ij \in E$, all the angle offsets in the synchronization problem will have constant magnitude, which is perhaps undesirable. To this end, we report 
on a scoring method for the synchronization problem, for which the magnitude of the resulting entries (used as input for synchronization) give a more accurate description of the rank-offset between a pair of players. For an ordered pair $(i,j)$, we define the \textit{Superiority Score} of $i$ with respect to $j$ as follows, in a manner somewhat analogous to the  $S_{ij}^{match}$ score (\ref{SijMatch}) used by the Serial-Rank algorithm. Let
\begin{equation}
	W_{ij} = \mathcal{L}(i) \cap \mathcal{H}(j) = \{ k \; | \; C_{ik} = -1 \text{ and } C_{jk} = 1  \}
\end{equation}
where $ \mathcal{L}(x) $ denotes the set of nodes with rank lower than $x$, 
and $ \mathcal{H}(x) $ denotes the set of nodes with rank higher than $x$. Based on this, the final score (rank offset) used as input for the Sync-Rank algorithm is given by 
\begin{equation}
S_{ij} = \left\{
	\begin{array}{rl}
   	- ( W_{ij} - W_{ji} )  & \text{ if }	W_{ij} \geq W_{ji} 	\\
   	  W_{ji} - W_{ij} & \text{ otherwise }	\\
     \end{array}
   \right.
\label{SijDefSuperiority}  %
\end{equation}
To preserve skew-symmetry, we set $S_{ji}  = - S_{ij} $. The philosophy behind this measure is as follows. In the end, we want $S_{ij}$  to reflect the true rank-offset between two nodes.  
One can think of $ W_{ij} $ as the number of witnesses favorable to $i$ (\textit{supporters} of $i$), which are players ranked lower than $i$ but higher than $j$.
Similarly,  $ W_{ji} $ is the number of witnesses favorable to $j$ (\textit{supporters} of $j$), which are players ranked lower than $j$ but higher than $i$. Then, the final rank offset is given by the difference of the two values $ W_{ij}$ and  $W_{ji}$ (though one could also perhaps consider the maximum). The sign adjustment is just so that we have $ S_{ij} $ being a proxy for $r_i - r_j$ (if $r_i$ is high-rank and thus $r_i$ is a small number, and $r_j$ is low-rank and thus $r_j$ is a large number, then $r_i - r_j$ should be strongly negative).
We solve the resulting synchronization problem via the eigenvector method, and denote the approach by SYNC-SUP. 
While this approach yields rather poor results (compared to the initial synchronization method) on the synthetic data sets, its performance is comparable to that of the other methods when applied to real data sets, and quite  surprisingly, it performs twice as good as any of the other methods when applied to the NCAA basketball data set. We attribute this rather intriguing fact to the observation (based on a preliminary investigation) that for this particular data set, the measurement graphs for each season are somewhat close to having a one dimensional structure, with teams grouped in leagues according to their strengths, and it is less likely for a highly ranked team to play a match against a very weak team. It would be interesting to investigate the performance of the all algorithms on a measurement graph that is a disc graph with nodes (i.e., players) lying on a one-dimensional line, and there is an edge (i.e., match) between two nodes if and only if they are at most $r$ units apart.



\FloatBarrier
\section{Noise models and experimental results}   \label{sec:numexp}

We compare the performance of the Sync-Rank algorithm with that of other methods across a variety of measurement graphs, varying parameters such as the number of nodes, edge density, level of noise and underlying noise model. We detail below the two noise models we have experimented with, and then proceed with the performance outcomes of extensive numerical simulations. 

\subsection{ Measurement  and Noise Models}
In most real world scenarios, noise is inevitable and imposes additional significant challenges, aside from those due to sparsity of the measurement graph.  
To each player $i$ (corresponding to a node of $G$), we associate a corresponding unique positive integer weight $r_i$. For simplicity, one may choose to think of $r_i$ as taking values in  $ \{1,2,\ldots,n\}$. In other words, we assume there is an underlying ground truth ranking of the $n$ players, with the most skillful player having rank $1$, and the least skillful one having rank $n$.
We denote by 
\begin{equation}
\boldsymbol{Cardinal \;\; measurement:} \;\;\;\;   O_{ij} =  r_i - r_j  
\label{cardComp}
\end{equation}
the ground truth \textbf{cardinal} rank-offset of a pair of players. Thus, in the case of cardinal comparisons, the available measurements  $C_{ij}$ are noisy versions of 
the ground truth $O_{ij}$ entries. On the other hand, an \textbf{ordinal} measurement is a pairwise comparison between two players which reveals only who the higher-ranked player is, or in the case of items, the preference relationship between the two items, i.e., 
 $C_{ij}=1$ is item $j$ is preferred to item $i$, and $C_{ij} = -1$ otherwise, thus without revealing the intensity of the preference relationship
 \begin{equation}
\boldsymbol{Ordinal \;\; measurement:} \;\;\;\;   O_{ij} =  \mbox{sign}(r_i - r_j).
\label{ordComp}
\end{equation}
This setup is commonly encountered in classical social choice theory, under the name of \textit{Kemeny} model, where $O_{ij}$, and takes value $1$ if player $j$ is ranked higher than player $i$, and $-1$ otherwise.
%
Given (perhaps noisy)  versions of the pairwise cardinal or ordinal measurements $O_{ij}$ given by (\ref{cardComp}) or  (\ref{ordComp}), the goal is to recover an ordering (ranking) of all players that is as consistent as possible with the given data. 
We compare our proposed ranking methods with those summarized in Section \ref{sec:OtherMethods}, on measurement graphs of varying edge density, under two different noise models, and for both ordinal and cardinal comparisons.

To test for robustness against incomplete measurements, we use a measurement graph $G$ of size $n$ given by the popular Erd\H{o}s-R\'{e}nyi model $G(n,p)$, where edges between the players are present independently with probability $p$. In our experiments, we consider three different values of $p=\{0.2, 0.5, 1\}$, the latter case corresponding to the complete graph  $K_n$ on $n$ nodes, which corresponds to having a ranking comparison between all possible pairs of players. 
To test the robustness against noise, we consider the following two noise models detailed below. We remark that, for both models, noise is added such that the resulting measurement matrix remains skew-symmetric.  

\subsubsection{Multiplicative Uniform Noise (MUN) Model}
In the \textbf{Multiplicative Uniform Noise} model, which we denote by MUN($n, p, \eta$), noise is multiplicative and uniform, meaning that, for cardinal measurements, instead of the true rank-offset measurement $ O_{ij} =  r_i - r_j $, we actually measure 
\begin{equation}
C_{ij} =  O_{ij} ( 1 + \epsilon ), \;\; \text{ where } \;\; \epsilon \sim [-\eta,\eta].
\end{equation}
An equivalent formulation 
is that to each true rank-offset measurement we add random noise $\epsilon_{ij}$ (depending on $ O_{ij}$)  uniformly distributed in $[-\eta O_{ij}, \eta O_{ij}]$, i.e.,
\begin{equation}
C_{ij} = 
\left\{
\begin{array}{rlll}
r_i - r_j + \epsilon_{ij} &  \epsilon_{ij} \sim Discrete \;  Unif.([-\eta (r_i - r_j),\eta (r_i - r_j)]) & \text{for an existing edge} &    \text{w. p. } p \\
0 &  &  \text{for a missing edge}   & \text{w. p. } 1-p.\\
\end{array}
\right.
\label{MUN_model}
\end{equation}
Note that we cap the erroneous measurements at $n-1$ in absolute value. Thus, whenever $ C_{ij} > n-1 $ we set it to $n-1$, and whenever $ C_{ij} < -(n-1) $ we set it to $-(n-1)$, since the furthest away two players can be is $n-1$ positions. 
The percentage noise added is $100 \eta$, (e.g., $\eta=0.1$ corresponds to $10\%$ noise).
Thus, if $\eta = 50\%$, and the ground truth rank offset $O_{ij} =  r_i - r_j = 10 $, then the  available measurement $C_{ij}$ is a random number in $[5,15]$.

\subsubsection{Erd\H{o}s-R\'{e}nyi Outliers (ERO) Model}
The second noise model we experiment with, is an \textbf{Erd\H{o}s-R\'{e}nyi Outliers} model, abbreviated  by ERO($n,p, \eta$), where the available measurements are given by the following mixture model
\begin{equation}
C_{ij} = \left\{
 \begin{array}{rll}
 r_i - r_j & \;\; \text{ for a correct edge,} & \text{ with probability } (1-\eta) p \\
 \sim Discrete \; Unif. [-(n-1), n-1]  & \;\; \text{ for an incorrect edge,} & \text{ with probability } \eta p	\\
		     0      & \;\; \text{ for a missing edge},   & \text{ with probability } 1-p,  	\\
     \end{array}
   \right.
\label{ERoutliers}
\end{equation}
Note that this noise model is similar to the one considered by Singer \cite{sync}, in which angle offsets are either perfectly correct with probability $1-\eta$, and randomly distributed on the unit circle, with
probability $\eta$.

We expect that in most practical applications, the first  noise model (MUN) is perhaps more realistic, where most rank offsets are being perturbed by noise (to a smaller or larger extent, proportionally to their magnitude), as opposed to having a combination of perfectly correct rank-offsets and randomly chosen rank-offsets.


\subsection{Numerical comparison synthetic data sets}  \label{secsec:NumerExpSynth}

This section compares the spectral and SDP synchronization-based ranking with the other methods briefly summarized in Sec. \ref{sec:OtherMethods}, and also in Table  \ref{tab:methodAbbrev}. More specifically, we compare against the 
Serial-Rank algorithm, based on both $S^{match}$ (\ref{SijMatch}) and $S^{glm}$ (\ref{def:SimGLM}) summarized in  Sec. \ref{secsec:SER} and Appendix \ref{sec:appSER}, the SVD-based approach in Sec. \ref{secsec:SVD}, the method of Least Squares in Sec. \ref{secsec:LS}, and finally the Rank-Centrality algorithm in Sec. \ref{secsec:RankCent}.

\begin{table}[tpb]
\begin{minipage}[b]{0.99\linewidth}
\begin{center}
\begin{tabular}{|c|l|l|}
\hline
 Acronym & Name & Section \\
\hline
SVD 		&  SVD  Ranking &  Sec. \ref{secsec:SVD}  \\
LS 		& Least Squares Ranking & Sec. \ref{secsec:LS} \\
SER 		& Serial-Ranking &  Sec. \ref{secsec:SER}  \\
SER-GLM 		& Serial-Ranking in the GLM model &  Sec. \ref{secsec:SER} \\
RC 			& Rank-Centrality &  Sec. \ref{secsec:RankCent}  \\
SYNC & Synchronization-Ranking via the spectral relaxation &  Sec. \ref{sec:Sync-Rank} \\
SYNC-SUP		& Synchronization-Ranking based on the Superiority Score (spectral relaxation) &  Sec. \ref{secsec:syncWeighted} \\
SYNC-SDP 	& Synchronization-Ranking via the SDP relaxation & Sec. \ref{secsec:syncSDP}  \\
\hline
\end{tabular}
\end{center}
\end{minipage} 
\caption{Names of the algorithms we compare, their acronyms, and respective Sections.}
\label{tab:methodAbbrev}
\end{table}

We measure the accuracy of the recovered solutions, using the popular Kendall distance, i.e., the number of pairs of candidates that are ranked in different order (flips), in the two permutations, the original one and the recovered one.  Given two rankings $\pi_1$ and $\pi_2$, their Kendall distance is defined as 
\begin{equation}
 \kappa(\pi_1,\pi_2) =  \frac{ | \{ (i,j): i<j, [ \pi_1(i)<\pi_1(j) \; \wedge  \; \pi_2(i) > \pi_2(j)]   \; \vee \; [ \pi_1(i) > \pi_1(j) \; \wedge \; \pi_2(i) < \pi_2(j) ] \} | }{ {n \choose 2 }} = \frac{ nr.  flips}{ {n \choose 2 }}
\end{equation}
We compute the Kendall distance on a logarithmic scale ($\log_2$), and  remark that in the error curves, for small levels of noise, the  error levels for certain methods are missing due to them being equal to 0.
%
%
%
In Figure \ref{fig:Meth6_n200_num} we compare the methods for the case of cardinal pairwise measurements, while in  Figure \ref{fig:Meth6_n200_ord} we do so for ordinal measurements.
The SYNC and SYNC-SDP methods are far superior to the other methods in the case of cardinal measurements, and similar to the other methods for ordinal measurements.
In Figures \ref{fig:rkSDP_Meth6_n200_num} we plot the recovered rank of the SDP program  (\ref{SDP_program_SYNC}). Note that for favorable levels of noise, the SDP solution is indeed of rank 1 even if we did not specifically  enforce this constraint, a phenomenon explained only recently in the work of Bandeira et al.  \cite{bandeira2014tightness}, who investigated the tightness of this SDP relaxation. 



\begin{figure}[h!]
\begin{center}
\subfigure[$p=0.2$, cardinal, MUN ]{\includegraphics[width=0.2530\textwidth]{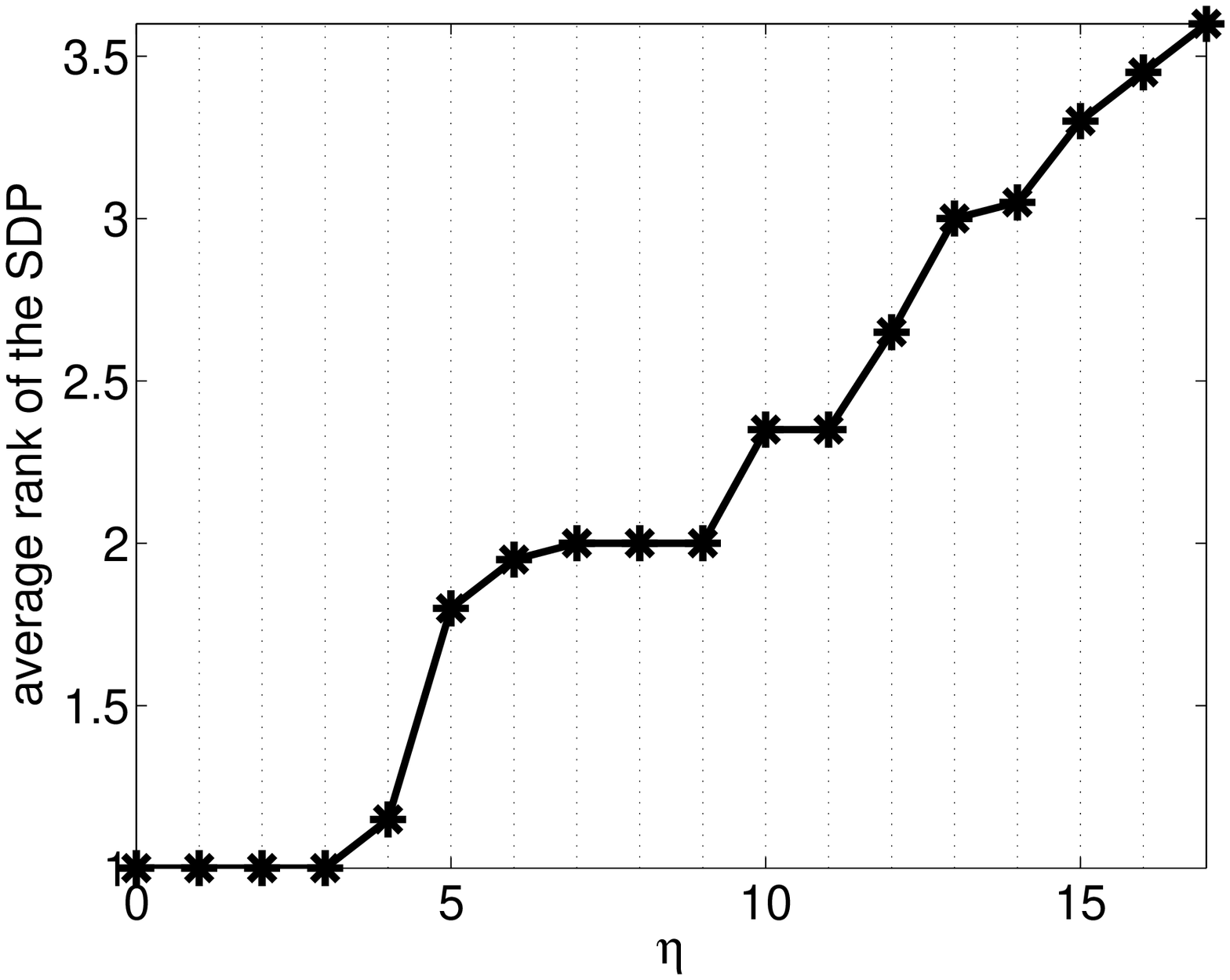}}
\subfigure[$p=0.5$, cardinal, MUN]{\includegraphics[width=0.2530\textwidth]{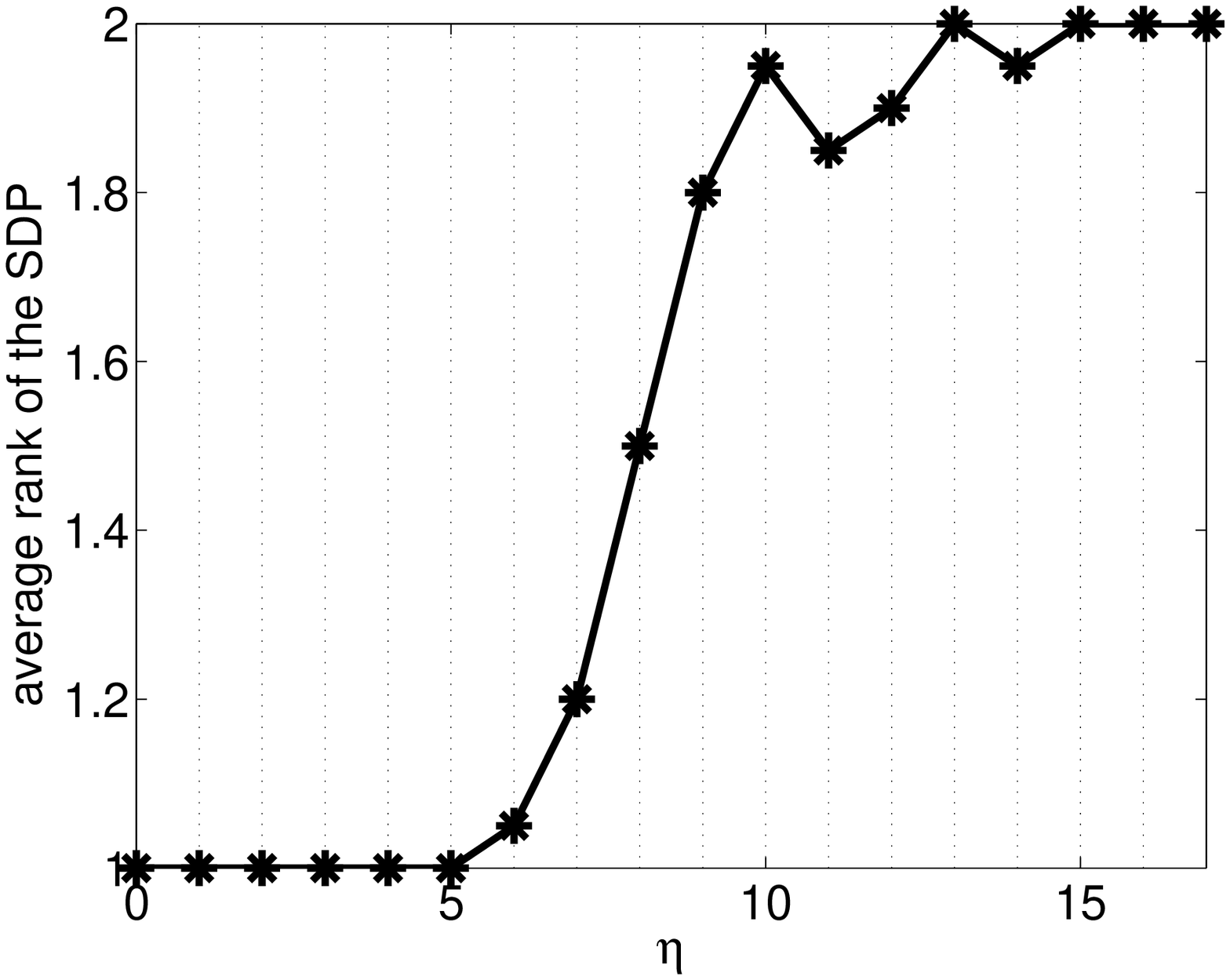}}
\subfigure[$p=1$, cardinal, MUN ]{\includegraphics[width=0.2530\textwidth]{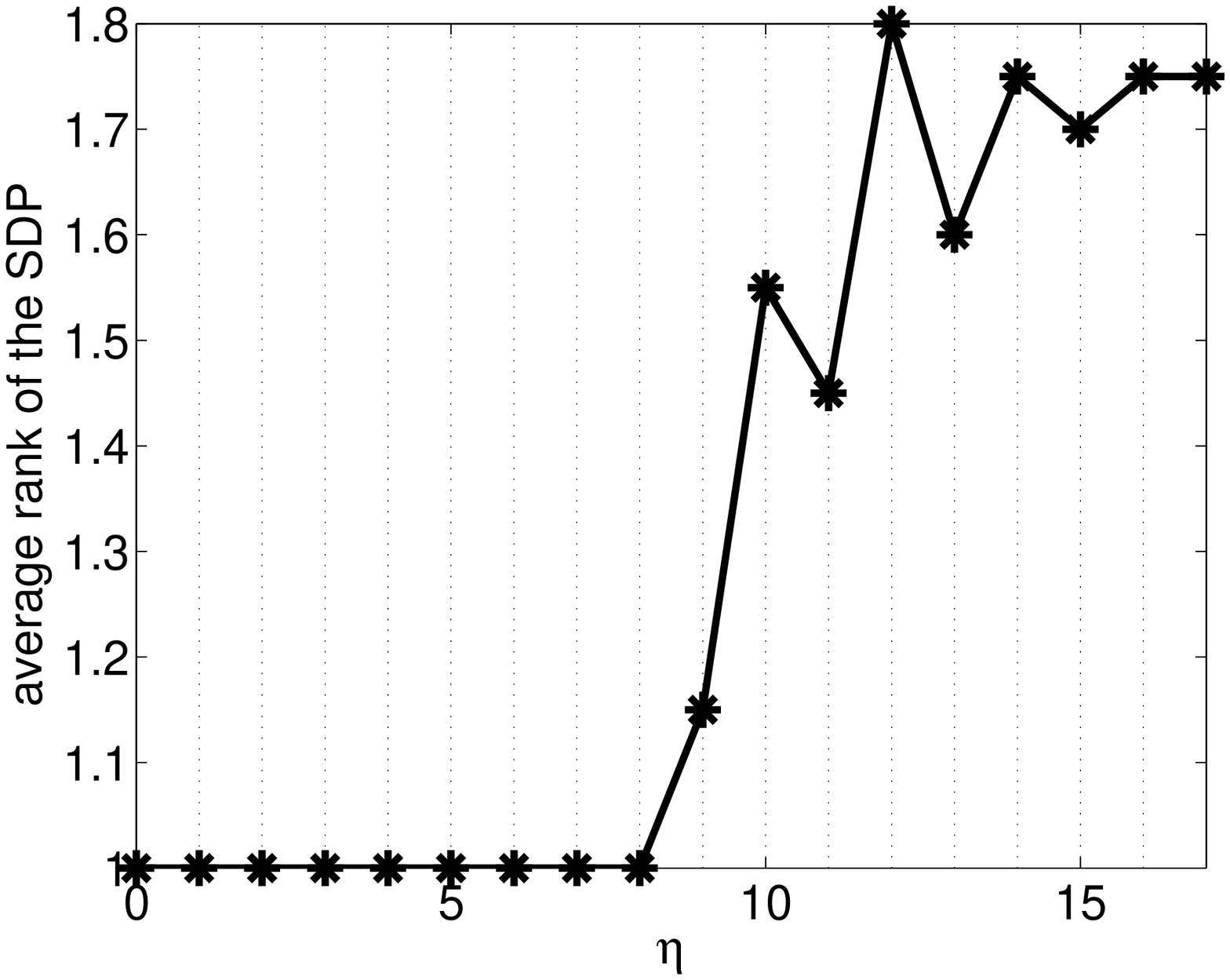}}

\subfigure[ $p=0.2$, cardinal, ERO ]{\includegraphics[width=0.253025\textwidth]{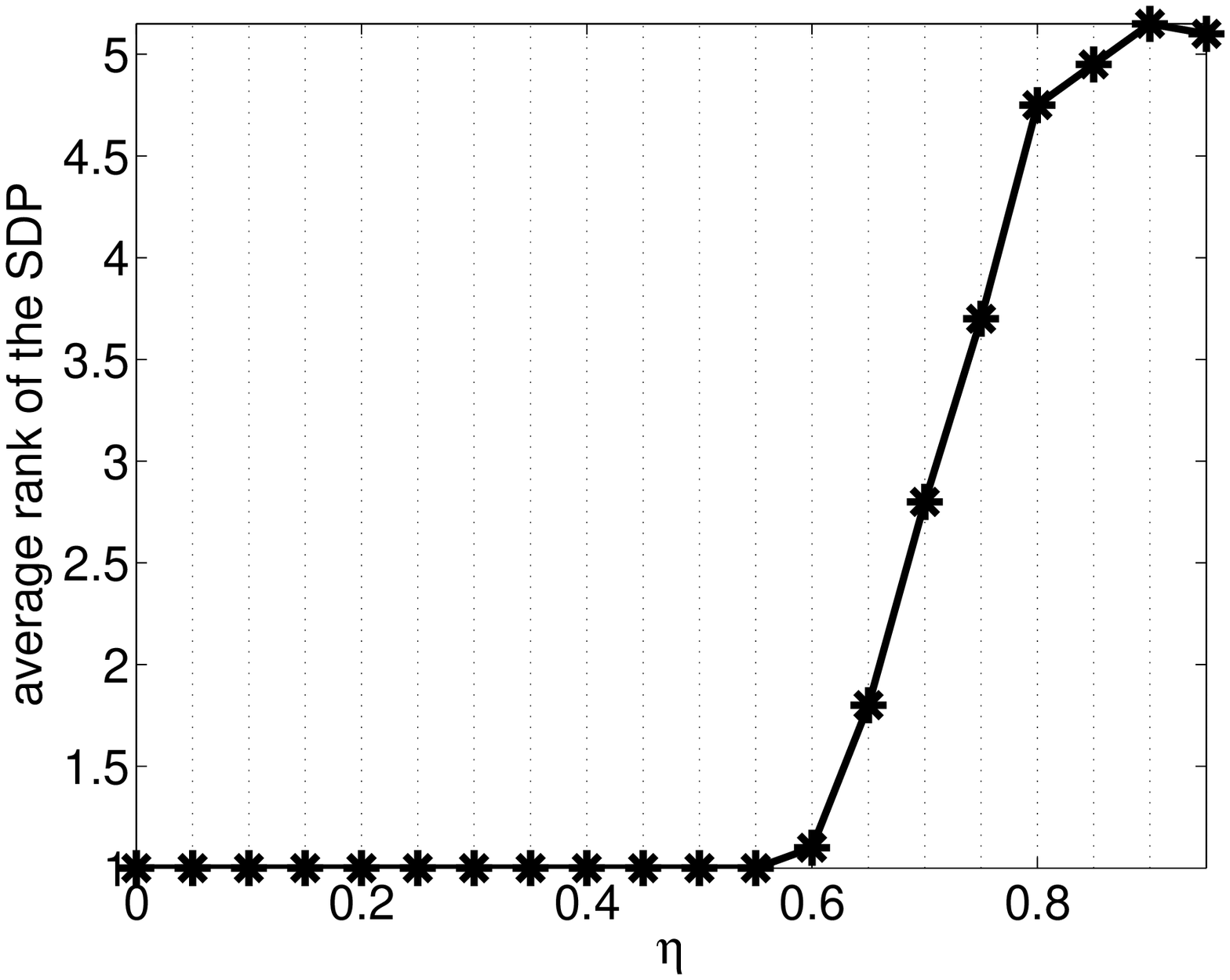}}
\subfigure[ $p=0.5$, cardinal, ERO ]{\includegraphics[width=0.253025\textwidth]{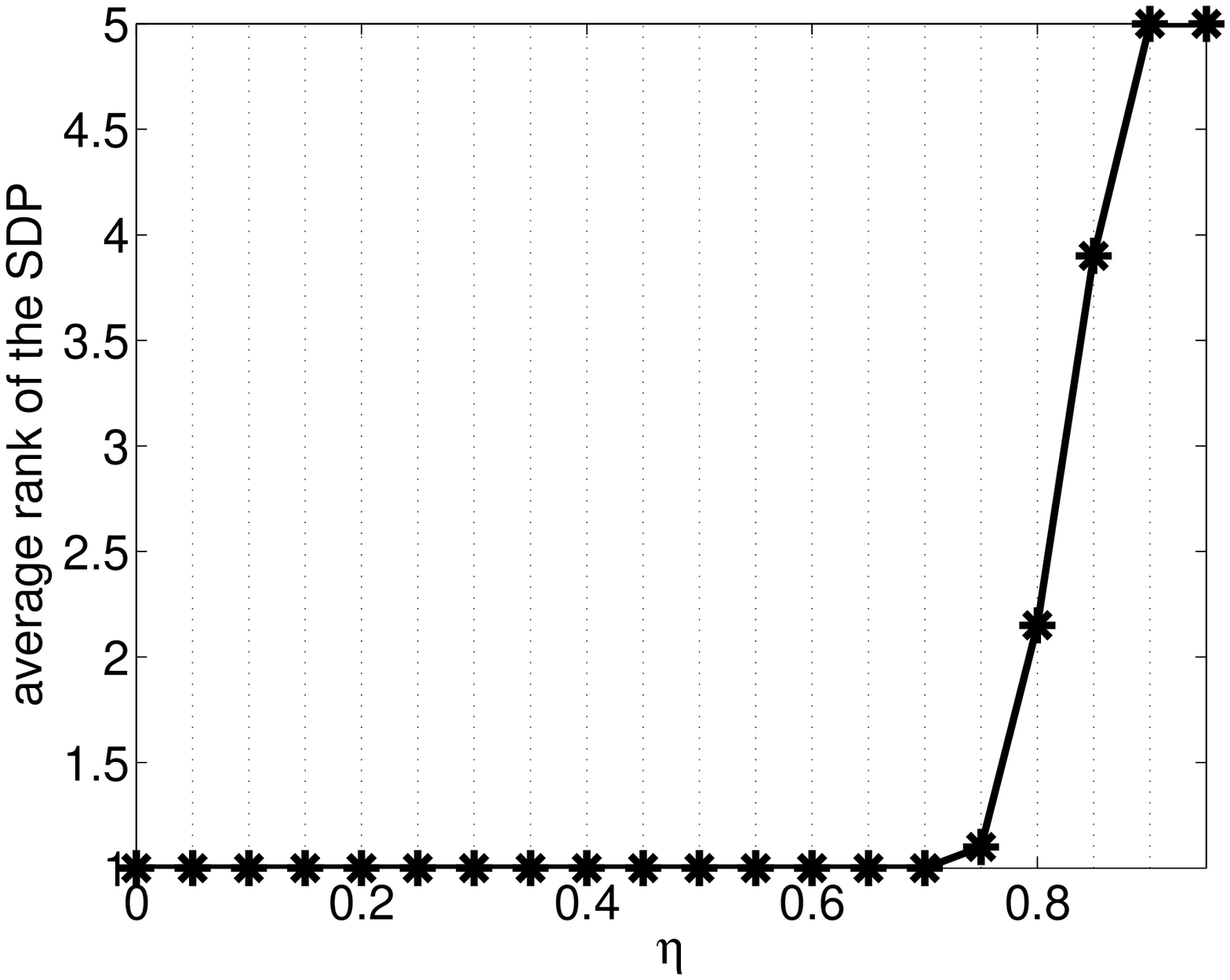}}
\subfigure[ $p=1$, cardinal, ERO  ]{\includegraphics[width=0.253025\textwidth]{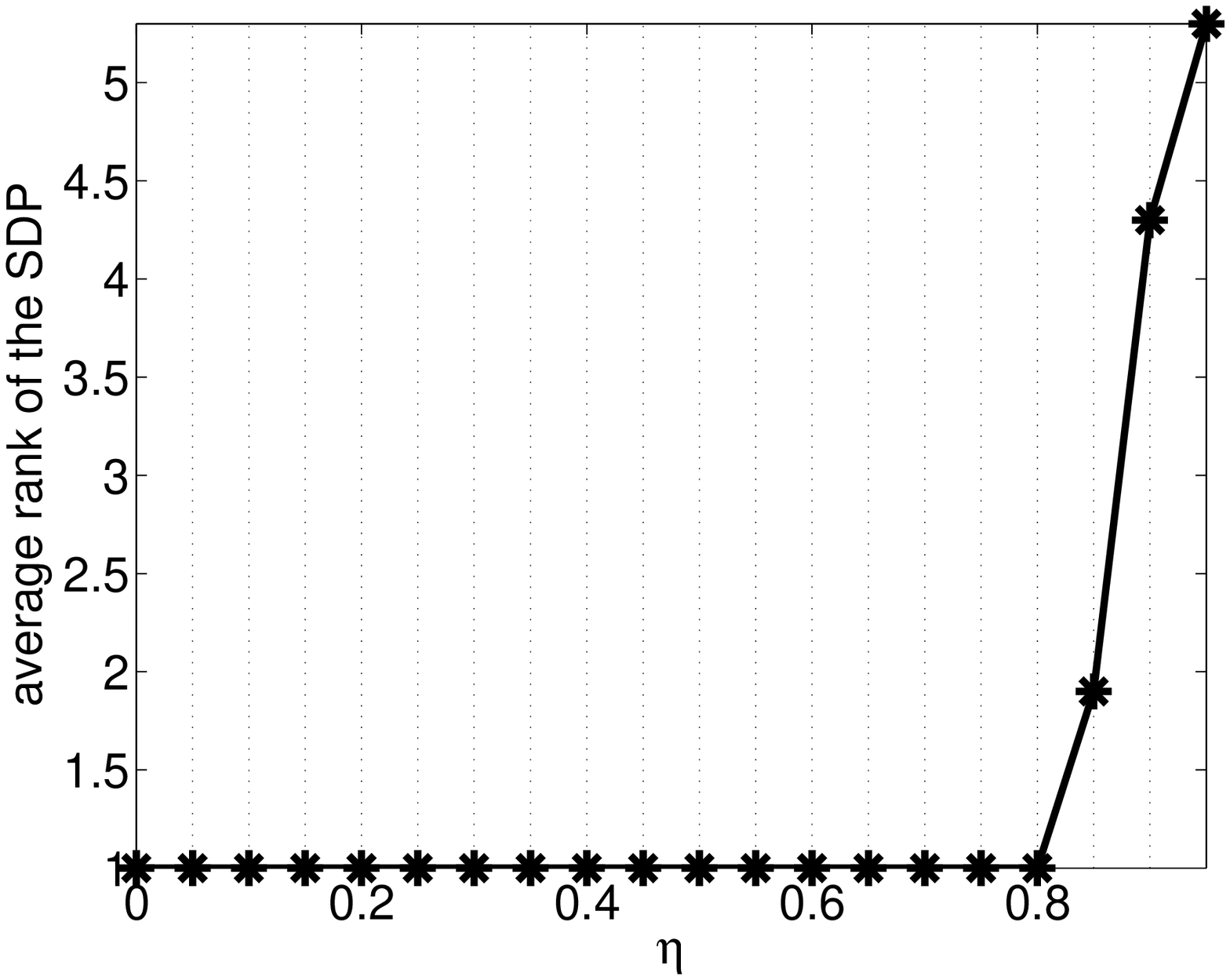}}
\end{center} 
\caption{The rank of the recovered solution from the SDP program (\ref{SDP_program_SYNC}), for the case of cardinal comparisons, as the vary the amount of noise. Top: Multiplicative Uniform Noise (MUN($n=200,p,\eta$)). Bottom: Erd\H{o}s-R\'{e}nyi Outliers (ERO($n=200,p, \eta$)).  We average the results over 20 experiments.}
\label{fig:rkSDP_Meth6_n200_num}
\end{figure}

\begin{figure}[h!]
\begin{center}
\subfigure[$p=0.2$, cardinal, MUN ]{\includegraphics[width=0.2930\textwidth]{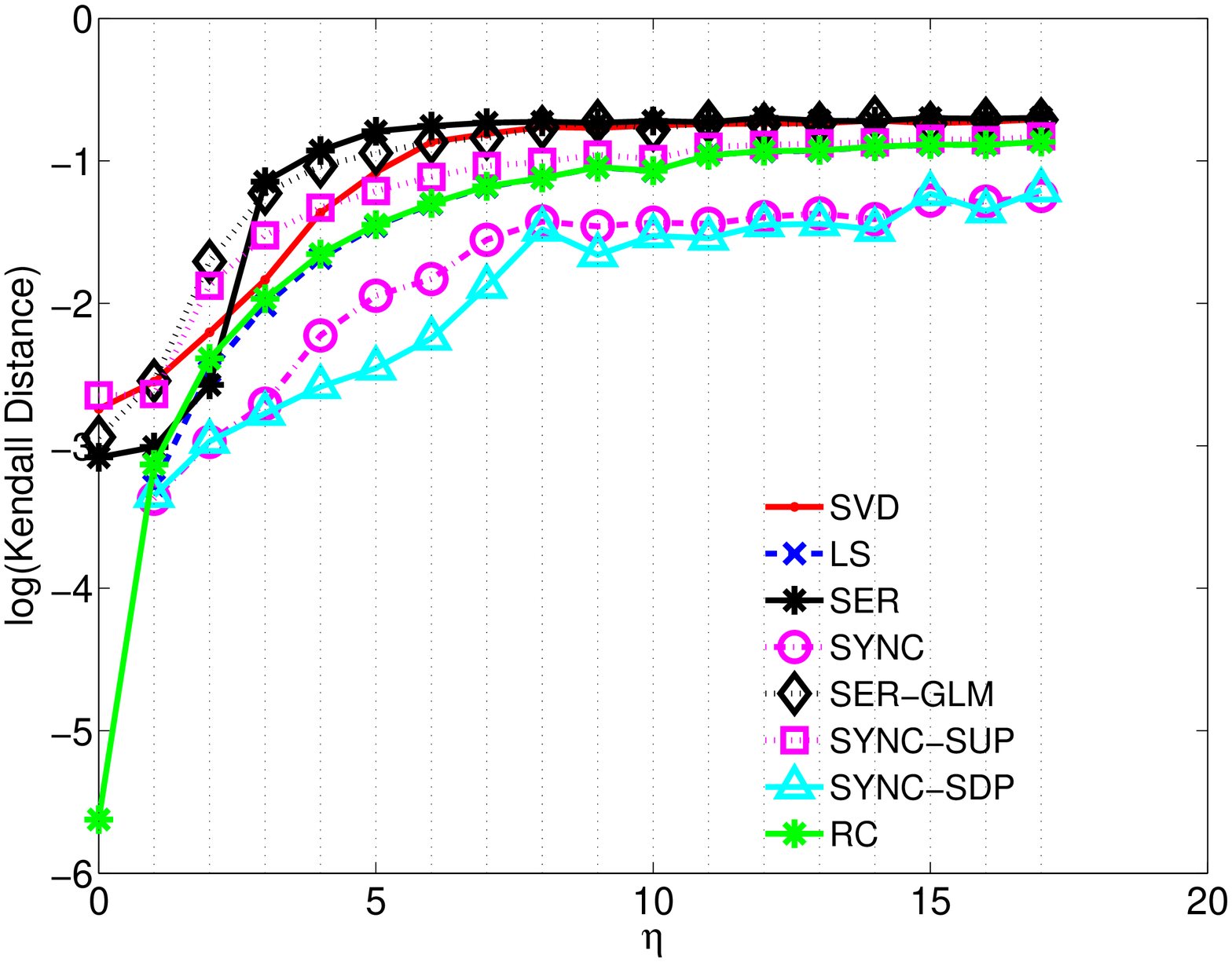}}
\subfigure[$p=0.5$, cardinal, MUN]{\includegraphics[width=0.2930\textwidth]{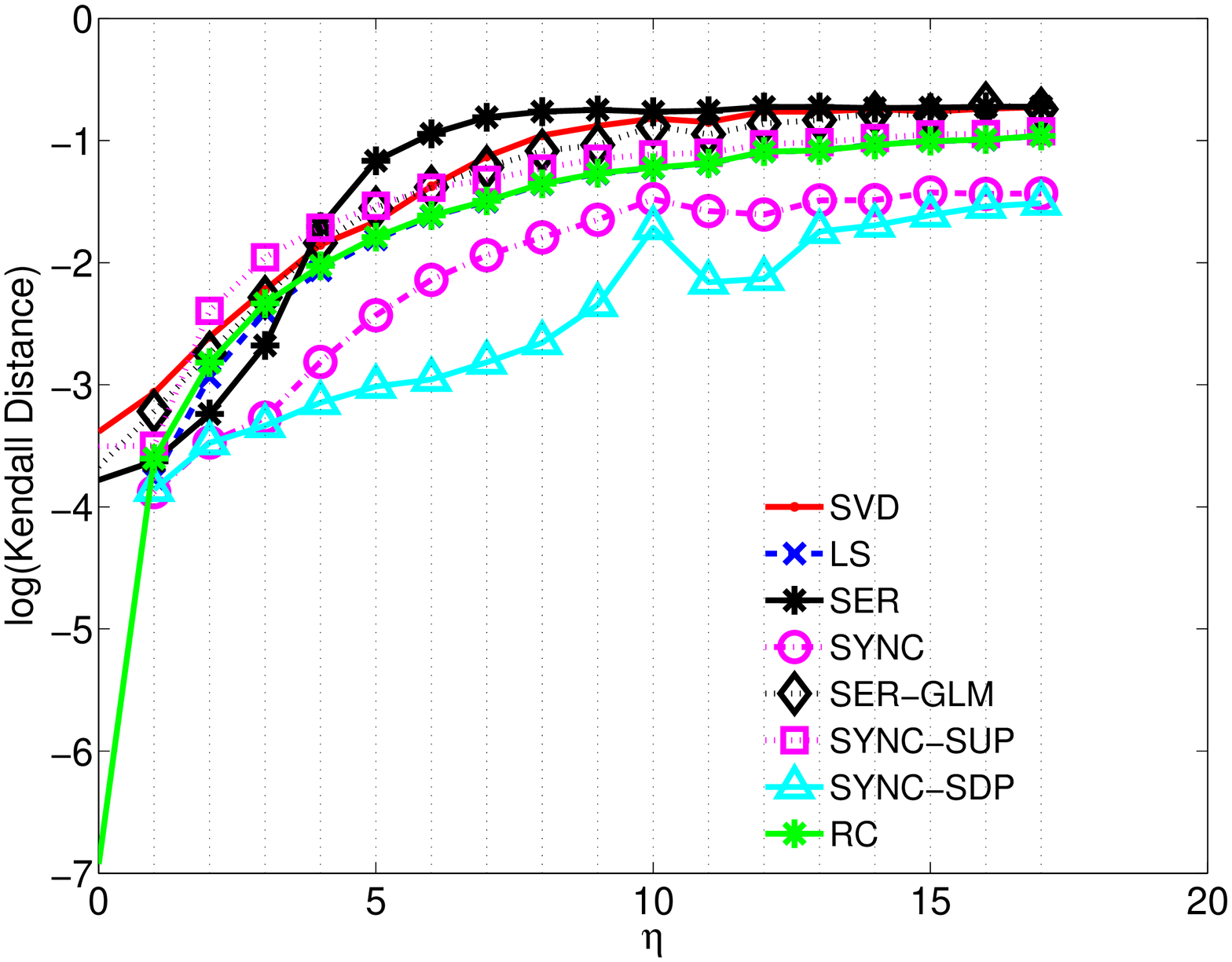}}
\subfigure[$p=1$, cardinal, MUN ]{\includegraphics[width=0.2930\textwidth]{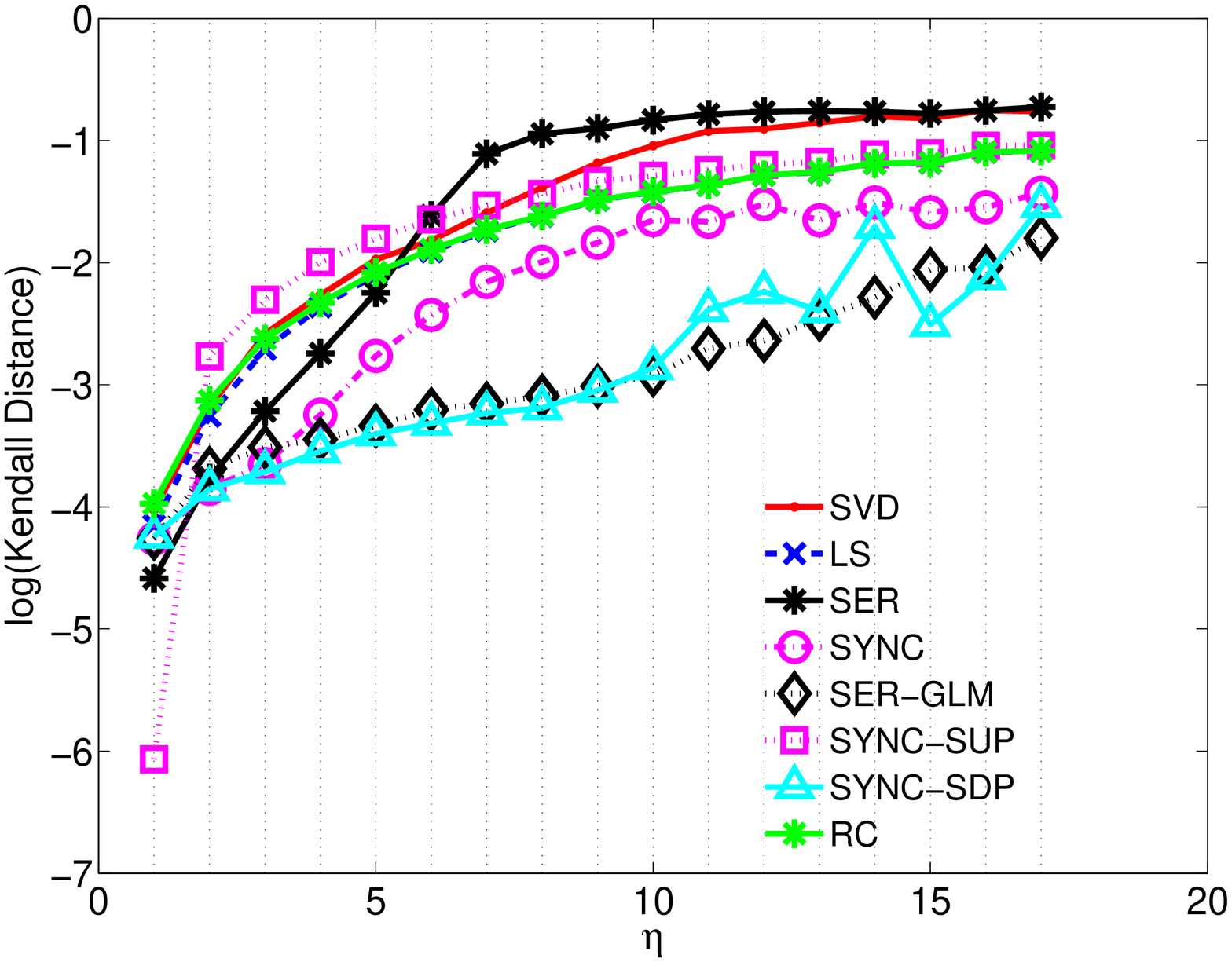}}
\subfigure[ $p=0.2$, cardinal, ERO ]{\includegraphics[width=0.2930\textwidth]{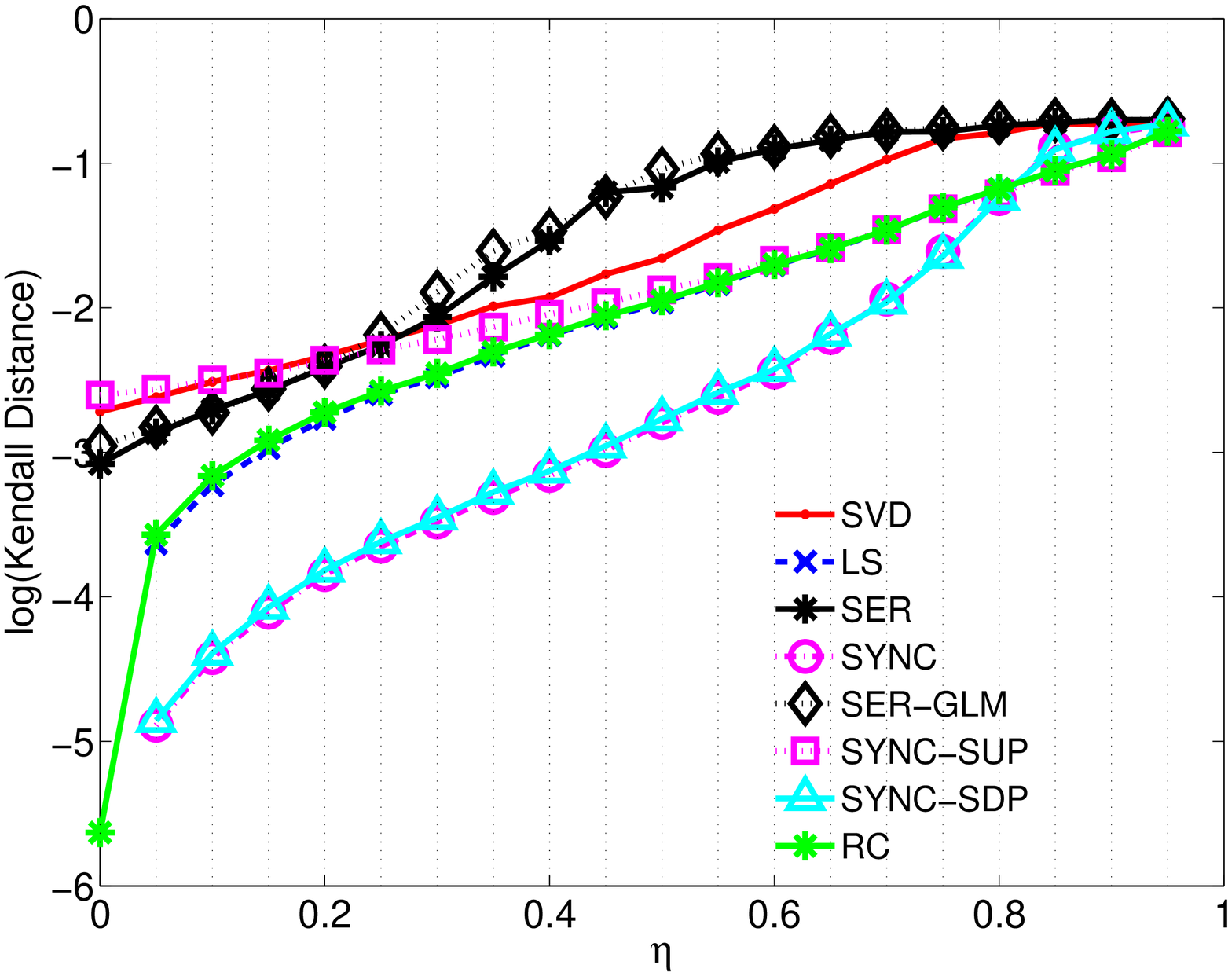}}
\subfigure[ $p=0.5$, cardinal, ERO ]{\includegraphics[width=0.2930\textwidth]{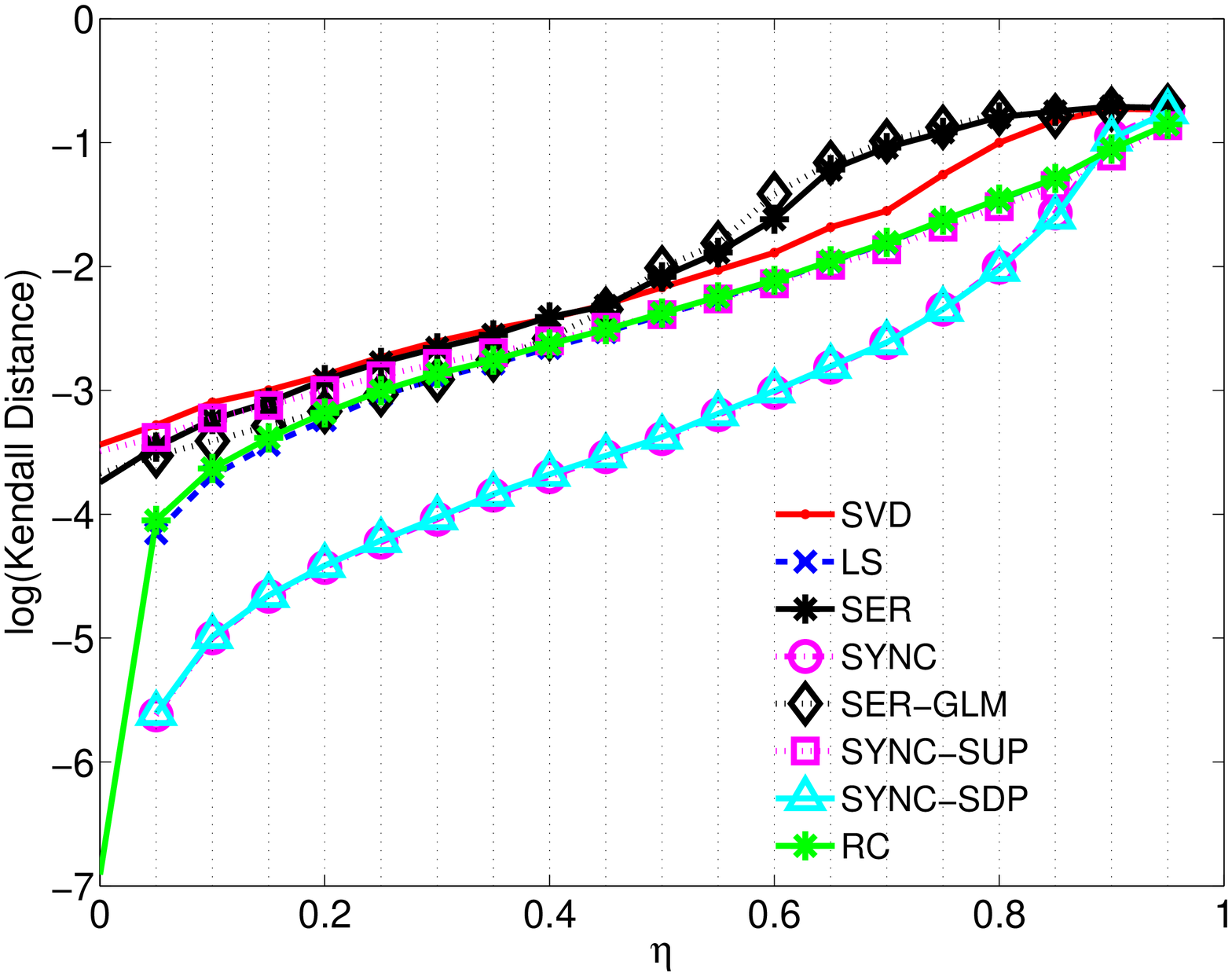}}
\subfigure[ $p=1$, cardinal, ERO  ]{\includegraphics[width=0.2930\textwidth]{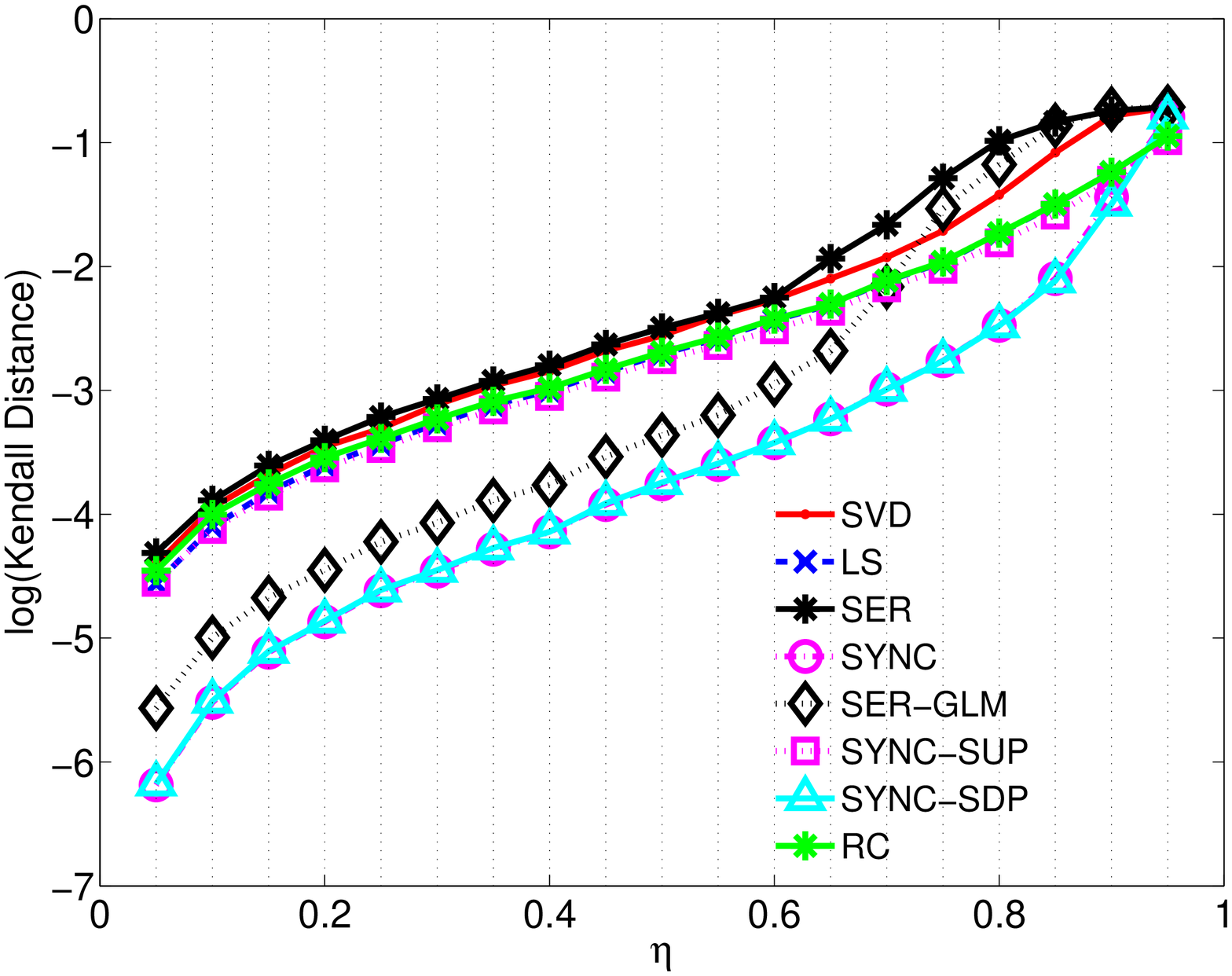}}
\end{center} 
\caption{Comparison of all ranking algorithms in the case of cardinal  comparisons. Top: Multiplicative Uniform Noise (MUN($n=200,p,\eta$)). Bottom: Erd\H{o}s-R\'{e}nyi Outliers (ERO($n=200,p,\eta$)). We average the results over 20 experiments.}
\label{fig:Meth6_n200_num}
\end{figure}

\begin{figure}[h!]
\begin{center}
\subfigure[$p=0.2$, ordinal, MUN ]{\includegraphics[width=0.2930\textwidth]{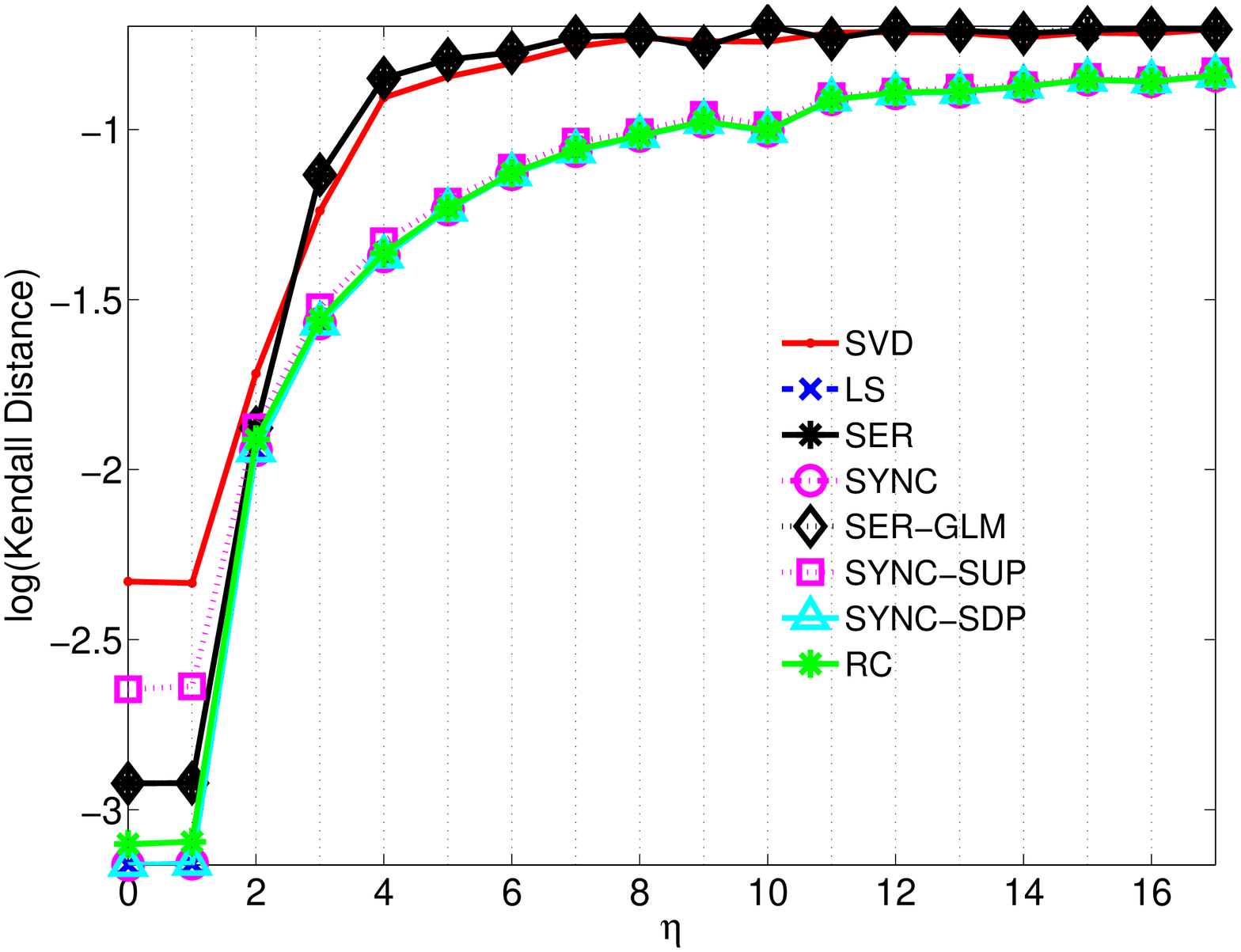}}
\subfigure[$p=0.5$, ordinal, MUN]{\includegraphics[width=0.2930\textwidth]{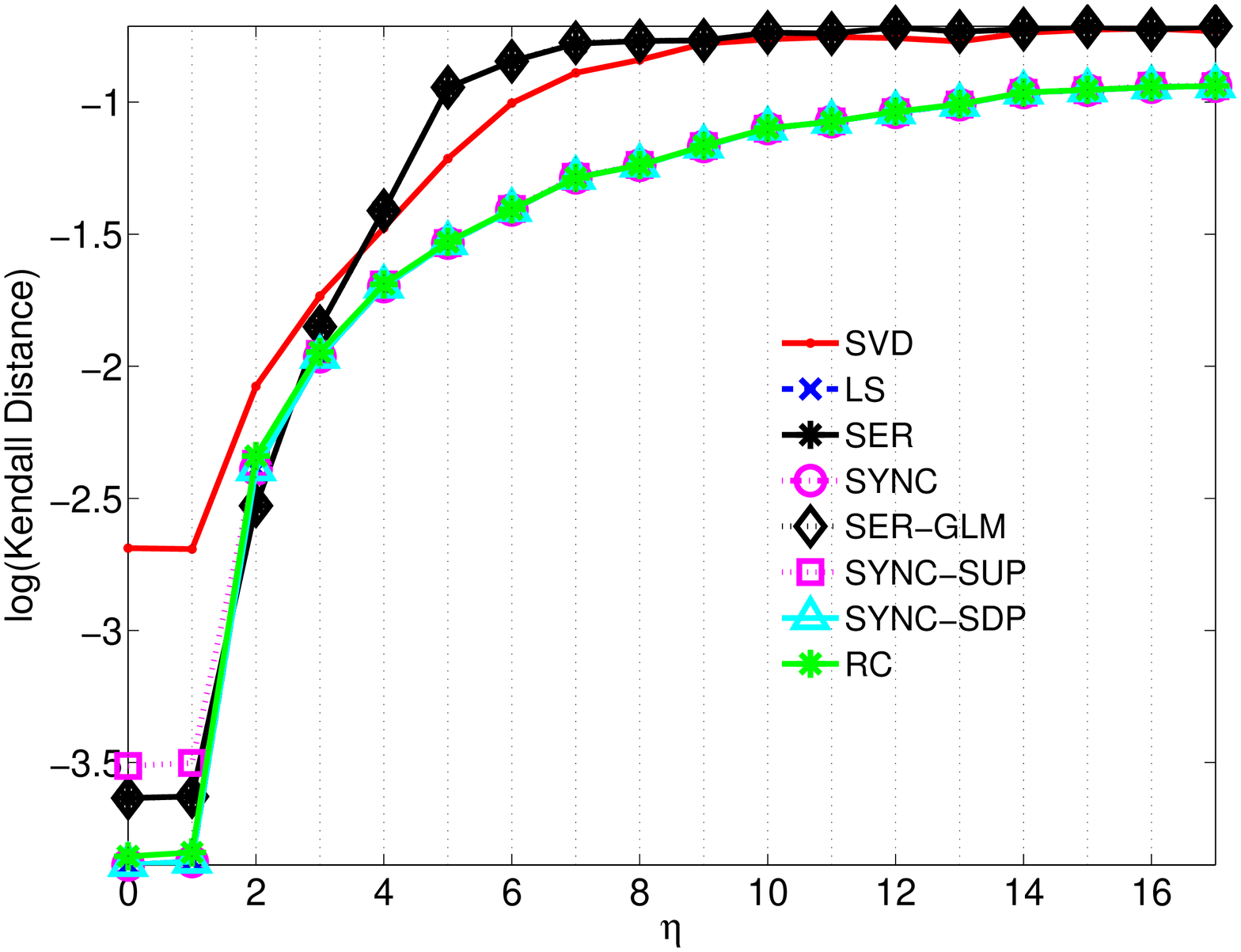}}
\subfigure[$p=1$, ordinal, MUN ]{\includegraphics[width=0.2930\textwidth]{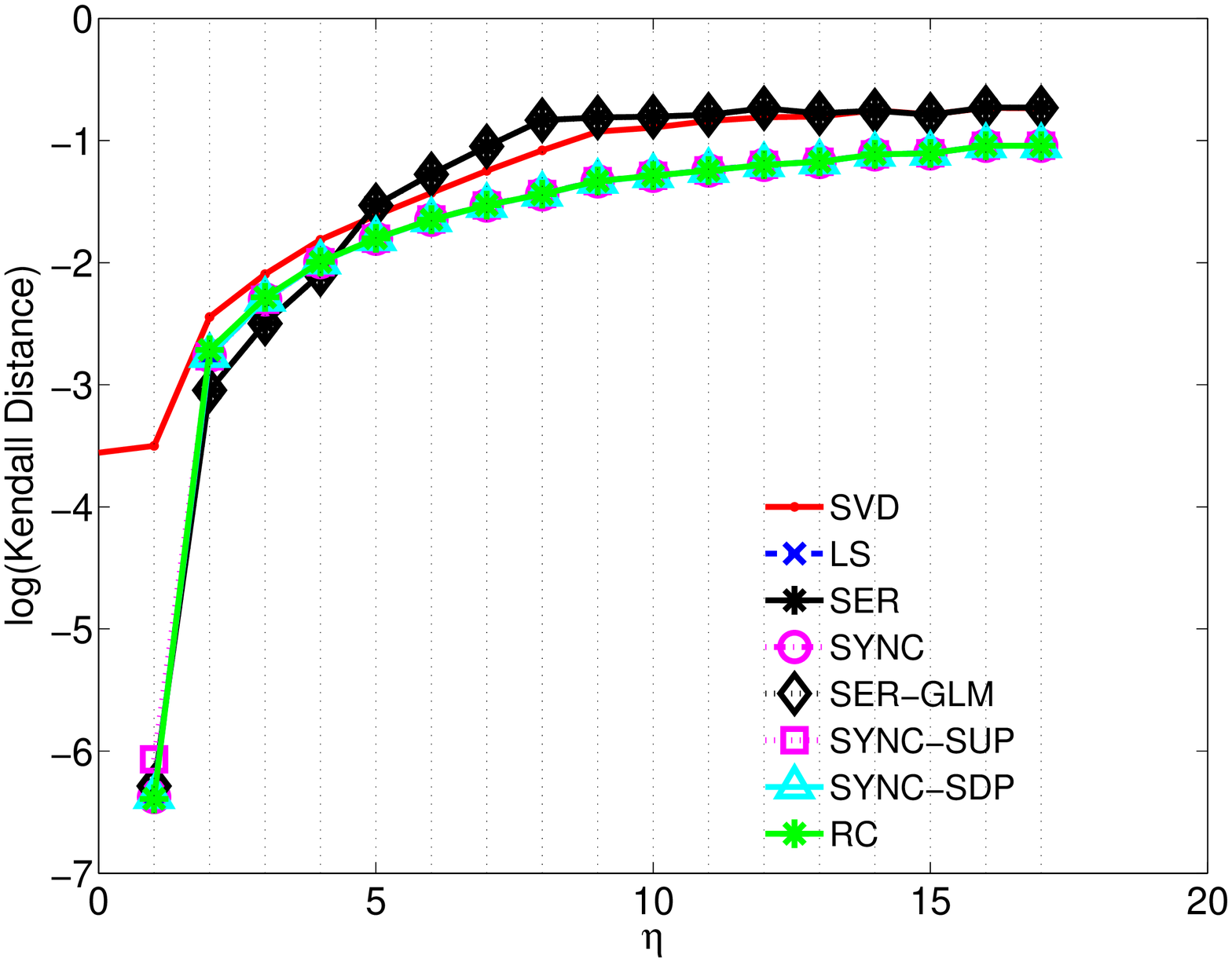}}
\subfigure[ $p=0.2$, ordinal, ERO ]{\includegraphics[width=0.2930\textwidth]{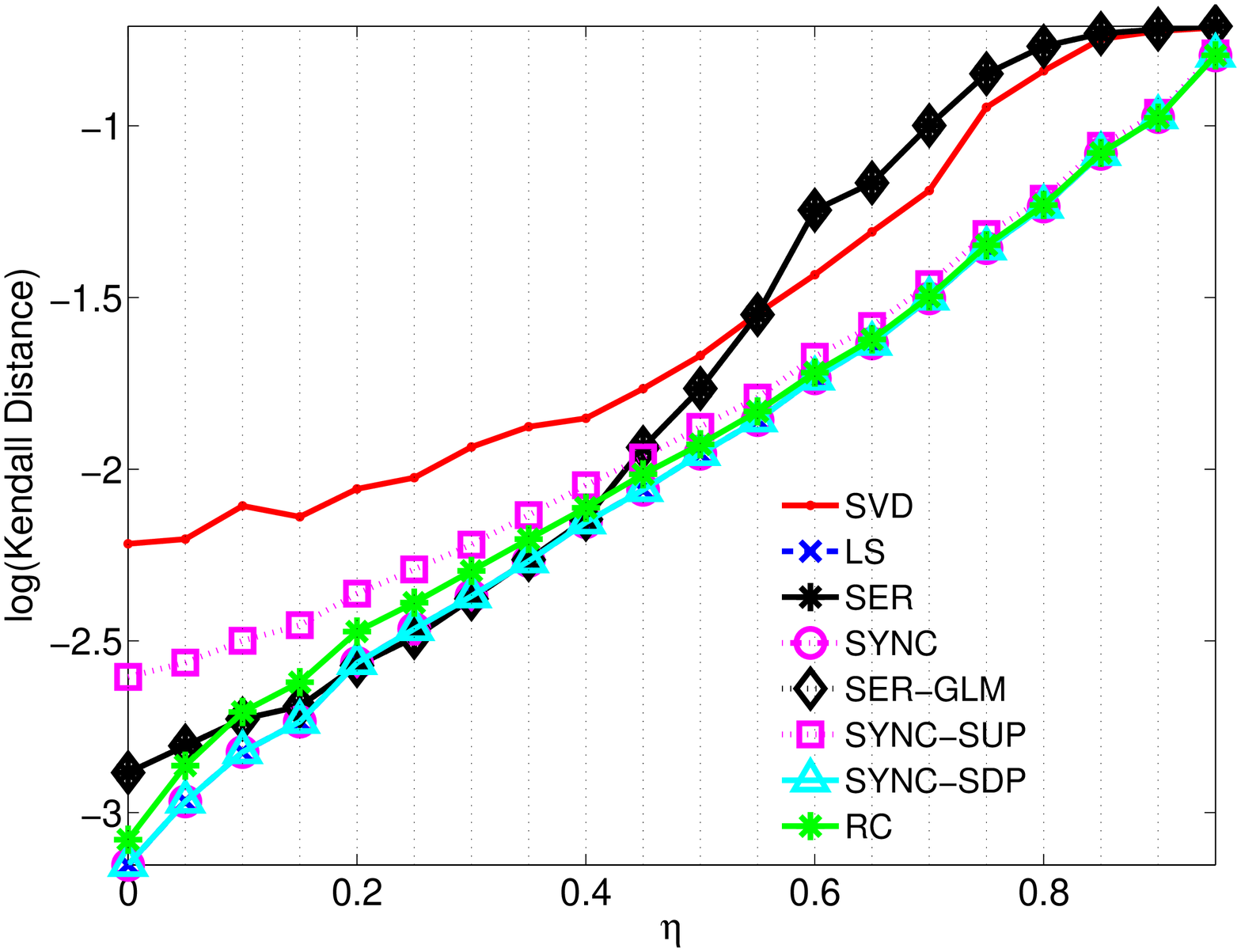}}
\subfigure[ $p=0.5$, ordinal, ERO ]{\includegraphics[width=0.2930\textwidth]{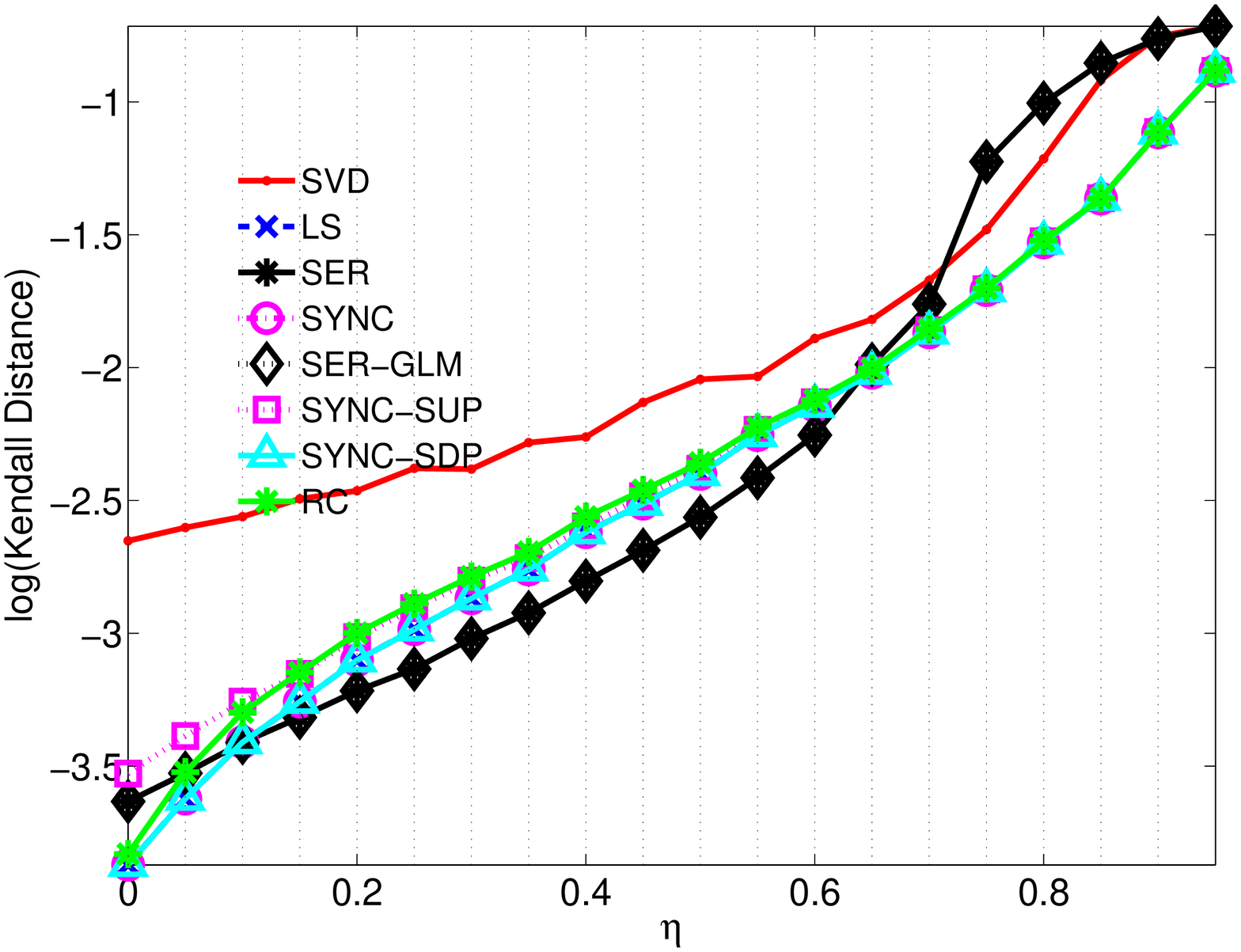}}
\subfigure[ $p=1$, ordinal, ERO  ]{\includegraphics[width=0.2930\textwidth]{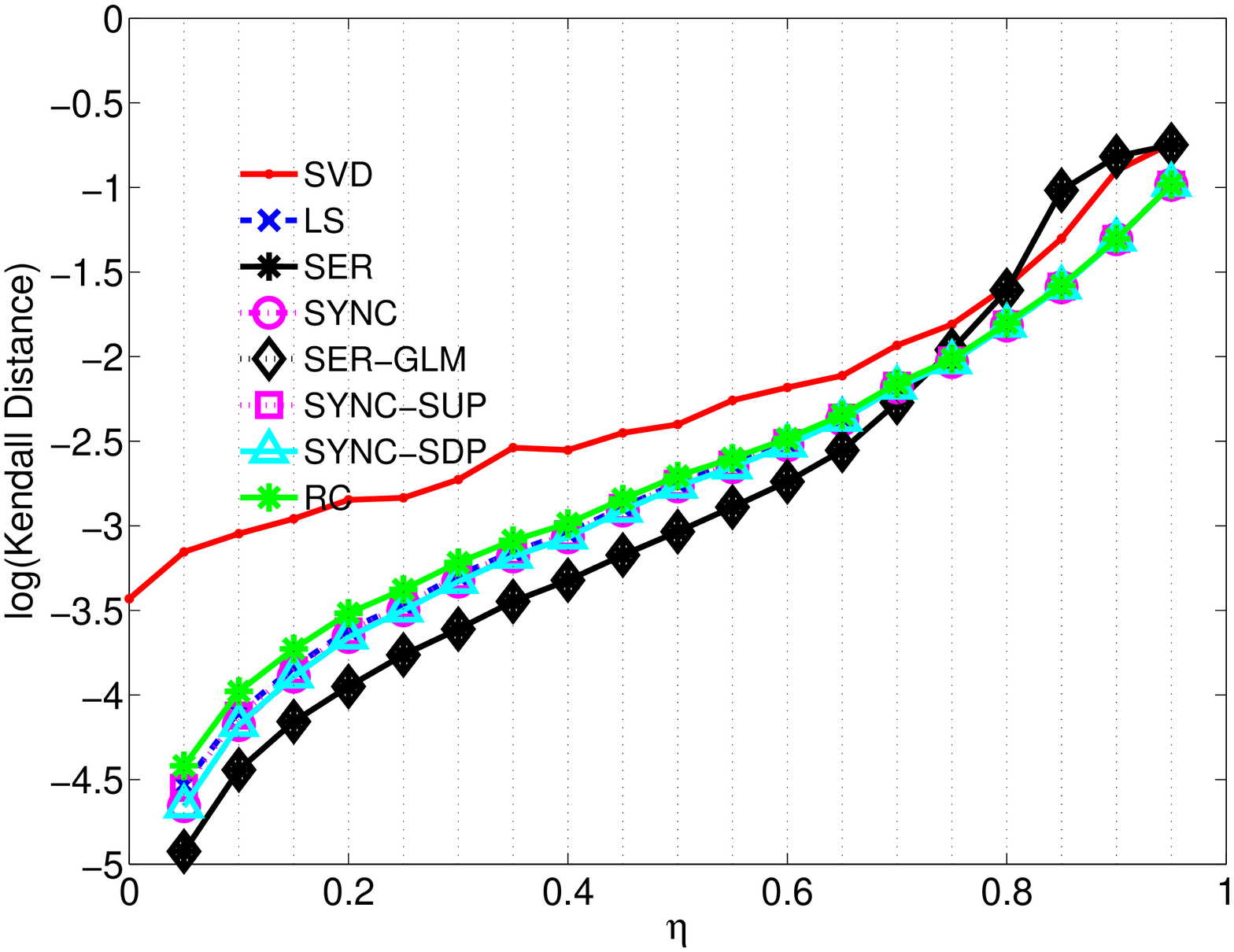}}
\end{center}
\caption{ Comparison of all methods for ranking with ordinal comparisons. Top: Multiplicative Uniform Noise (MUN($n=200,p,\eta$)). Bottom: Erd\H{o}s-R\'{e}nyi Outliers (ERO($n=200,p,\eta$)). We average the results over 20 experiments.}
\label{fig:Meth6_n200_ord}
\end{figure}

\FloatBarrier

\subsection{Numerical Comparison on the English Premier League Data Set}  \label{secsec:NumerExpEnglad}

The first real data set we consider is from the analysis of sports data, in particular, several years of match outcomes from the English Premier League soccer data set. 
We consider the last 3 complete seasons:  2011-2012, 2012-2013, and 2013-2014, both home and away games.  During each championship, any pair of teams meets exactly twice, both home and away.
We pre-process the data in several ways, and propose several methods to extract information from the game outcomes, i.e., of building the pairwise comparison matrix $C$ from the raw data given by $C^{home}$ and $C^{away}$, and report on the numerical results we obtain across all scenarios. The four different criteria we experiment with are as follows
\begin{enumerate}
\item $C^{tpd}$: \textit{Total-Point-Difference}, for each pair of teams, we aggregate the total score for the two matches teams $i$ and $j$ played against each other
\begin{equation}
 C^{tpd}_{ij} = C_{ij}^{home} + C_{ij}^{away}
 \label{Ctpd}
\end{equation}
\item $C^{stpd}$: \textit{Sign-Total-Point-Difference}
 considers the winner after aggregating the above score
\begin{equation}
 C^{stpd}_{ij} = \text{sign}( C^{tpd}_{ij} ) =\text{sign}(C_{ij}^{home} + C_{ij}^{away})
 \label{Cstpd}
\end{equation}
\item $C^{nw}$: \textit{Net Wins}: the number of times a team has beaten the other one. In the soccer data set, this takes value $\{-2,1,0,1,2\}$. In the Halo data set, the number of games a pair of players may play against each other is larger.
\begin{equation}
 C^{nw}_{ij} =  \text{sign}(C_{ij}^{home}) + \text{sign}(C_{ij}^{away})
 \label{Cnw}
\end{equation}
\item $C^{snw}$: \textit{Sign(Net Wins)} ($C^{snw}=\text{sign}(C^{nw}$): only considers the winner in terms of the number of victories.
\begin{equation}
 C^{snw}_{ij} =  \text{sign}( C^{nw}_{ij} )
 \label{Csnw}
\end{equation}
\end{enumerate}
Note that $C^{tpd}$ and $C^{nw}$ lead to cardinal measurements, while $C^{stpd}$ and  
$C^{snw}$ to ordinal ones.
Each user may also interpret each variant of the pre-processing step as a different winning criterium for the game under consideration. We remind the reader that for soccer games, a win is compensated with 3 points, a tie with 1 point, and a loss with 0 points, the final ranking of the teams is determined by the cumulative points a team gathers throughout the season.  
Given a particular criterium, we are interested in finding an ordering that minimizes the number of upsets, or weighted upsets as defined below.

We denote by $\hat{r}_i$ the estimated rank of player $i$ as computed by the method of choice. Recall that lower values of $\hat{r}_i$ correspond to higher ranks (better players or more preferred items). We then construct the induced (possibly incomplete) matrix of induced  pairwise  rank-offsets
\begin{equation} \nonumber
\hat{C}_{ij} = \begin{cases}
 \hat{r}_i - \hat{r}_j  & \text{if } (i,j) \in E(G)\\
  0 & \text{if } (i,j) \notin  E(G),
\end{cases}.
\label{recoveredC}
\end{equation}
and remark that $\hat{C}_{ij}  < 0$ denotes that the rank of player $i$ is higher than the rank of player $j$.
To measure the accuracy of a proposed reconstruction, we rely on the following three metrics. First, we use the popular metric that counts the number of \textit{upsets} (lower is better)
\begin{equation}
	  Q^{(u)}= \sum_{i=1}^{n-1} \sum_{j=i+1}^{n} \mb{1}_{ \{ \text{sign}( C_{ij} \hat{C}_{ij}) = -1 \} } 
\end{equation}
which counts the number of disagreeing ordered comparisons. It contributes with a $+1$ to the summation whenever the ordering in the provided data contradicts the ordering in the induced ranking. Next we define the following two measures of correlation (higher is better) between the given rank comparison data, and the one induced by the recovered solution
\begin{equation}
	Q^{(s)} = \sum_{i=1}^{n-1}   \sum_{j=i+1}^{n} C_{ij} \; \text{sign}(\hat{C}_{ij})
\end{equation}
and the very similar one
\begin{equation}	
	Q^{(w)} = \sum_{i=1}^{n-1}   \sum_{j=i+1}^{n}  C_{ij}   \hat{C}_{ij}
\end{equation}

 \begin{table}[tpb]
\begin{minipage}[b]{0.95\linewidth}
\begin{center}
\begin{tabular}{|c| C{1cm} |C{1cm}|C{1cm}|C{1cm}|C{1cm}|C{1cm}|C{1cm}|C{1cm}|C{1cm}|}
\hline
 Team   & SVD & LS & SER & SYNC &  SER-GLM & SYNC SUP &  SYNC SDP & RC & GT  \\
\hline
Arsenal   & 1  & 4  & 5  & 4  & 4  & 5  & 4  & 4  & 4  \\
 Aston Villa   & 19  & 16  & 18  & 16  & 20  & 19  & 16  & 16  & 15   \\
 Cardiff   & 17  & 20  & 16  & 20  & 15  & 20  & 20  & 20  & 20   \\
 Chelsea   & 8  & 1  & 1  & 1  & 2  & 1  & 1  & 1  & 3   \\
 Crystal Palace   & 14  & 12  & 14  & 12  & 16  & 13  & 12  & 13  & 11   \\
 Everton   & 5  & 5  & 4  & 5  & 5  & 6  & 5  & 5  & 5   \\
 Fulham   & 15  & 17  & 17  & 17  & 17  & 16  & 17  & 17  & 19   \\
 Hull   & 13  & 14  & 13  & 14  & 13  & 14  & 14  & 14  & 16   \\
 Liverpool   & 3  & 3  & 3  & 3  & 3  & 3  & 3  & 3  & 2   \\
 Man City   & 2  & 2  & 2  & 2  & 1  & 2  & 2  & 2  & 1   \\
 Man United   & 4  & 7  & 7  & 7  & 8  & 7  & 7  & 7  & 7   \\
 Newcastle   & 10  & 11  & 10  & 11  & 11  & 10  & 11  & 11  & 10   \\
 Norwich   & 18  & 15  & 19  & 15  & 18  & 18  & 15  & 15  & 18   \\
 Southampton   & 7  & 8  & 8  & 8  & 7  & 8  & 8  & 8  & 8   \\
 Stoke   & 11  & 9  & 9  & 9  & 9  & 9  & 9  & 9  & 9   \\
 Sunderland   & 20  & 18  & 20  & 18  & 19  & 17  & 18  & 18  & 14   \\
 Swansea   & 9  & 10  & 11  & 10  & 10  & 12  & 10  & 10  & 12   \\
 Tottenham   & 6  & 6  & 6  & 6  & 6  & 4  & 6  & 6  & 6   \\
 West Brom   & 16  & 19  & 15  & 19  & 12  & 15  & 19  & 19  & 17   \\
 West Ham   & 12  & 13  & 12  & 13  & 14  & 11  & 13  & 12  & 13   \\ 
\hline
Nr. upsets & 66 & 44 & 44 & 44 & 52 & 48 & 44 & 46 & 54   \\ 
Score/100 &    9.5 &   10.3 &   10.0 &   10.3 &   10.0 &   10.1 &   10.3 &   10.2 &   10.4   \\ 
W-Score/1000 &    9.5 &   10.0 &    9.9 &   10.0 &   10.0 &   10.0 &   10.0 &   10.0 &   10.6   \\ 
Corr w. GT &   0.69 &   0.87 &   0.80 &   0.87 &   0.75 &   0.84 &   0.87 &   0.86 &   1.00   \\ 
\hline
\end{tabular}
\end{center}
\end{minipage} 
\caption{English Premier League Standings 2013-2014, based on the input matrix $C^{nw}$ (\textit{the number of net wins between a pair of teams}) given by (\ref{Cnw}). GT denotes the final official ranking at the end of the season. }
\label{tab:EnglandStandings_nrWonLost}
\end{table}

We show in Table \ref{tab:EnglandStandings_nrWonLost} the rankings obtained by the different methods we have experimented with, for the 2013-2014 Premier League Season, when the input is based on the $C^{nw}$ measurements (\textit{Net Wins}).
The final column, denoted as GT, denotes the final official ranking at the end of the season. 
We sort the teams alphabetically, and show their assigned rank by each of the methods. The bottom row of  Table \ref{tab:EnglandStandings_nrWonLost}  computes the different quality measures defined above, $Q^{(u)}$, $Q^{(s)}$, and $Q^{(w)}$. The very last row in the Table computes the Kendall correlation between the respective ranking and the official final standing GT. We plot similar results in  Appendix  \ref{sec:appEngland}, in Tables   \ref{tab:EnglandStandings_sumGoalDif} ($C^{tpd}$ based on  \textit{Total-Goal-Difference}), 
\ref{tab:EnglandStandings_signsumGoalDif}  ($C^{stpd}$ based on  \textit{Sign-Total-Goal-Difference}), and finally,
\ref{tab:EnglandStandings_signNrWonLost} ($C^{snw}$ based on  \textit{Sign(Net Wins)}).

We remark that, across the different type of inputs considered, LS, SYNC and SYNC-SDP (abbreviated by SDP in the table) correlate best with the official ranking\footnote{We do not believe that correlation with the official standings is a good measure of success, simply because they are based on different rules, i.e., on accumulation of points for each win, tie, or loss. One can think of the four different types of pre-processing criteriums as four different input data sets with pairwise comparisons, and the goal is to propose a solution that best agrees with the input data, whatever that is.}, and in almost all scenarios (with very few exceptions) achieve the highest values for the $Q^{(s)}$ and $Q^{(w)}$ correlation scores. In terms of the number of upsets, the synchronization based methods alternate the first place in terms of quality with SER, depending on the pre-processing step and the input used for the ranking procedure. 
In addition, we show in Figure \ref{fig:EnglandErors} the Q-scores associated to three recent seasons in Premier League: 2011-2012, 2012-2013, and 2013-2014, together with the mean across these three seasons. We show the number of upsets in Figure \ref{fig:EnglandErors}  (a), and the correlations scores in (b) and (c), across the four different possible types of inputs 
$C^{nw}$, 
$C^{snw}$, 
$C^{tpd}$, and 
$C^{stpd}$.




\begin{figure}[h!]
\begin{center}
\subfigure[Nr upsets  $Q^{(u)}$  (lower is better)]{
\includegraphics[width=0.24\textwidth]{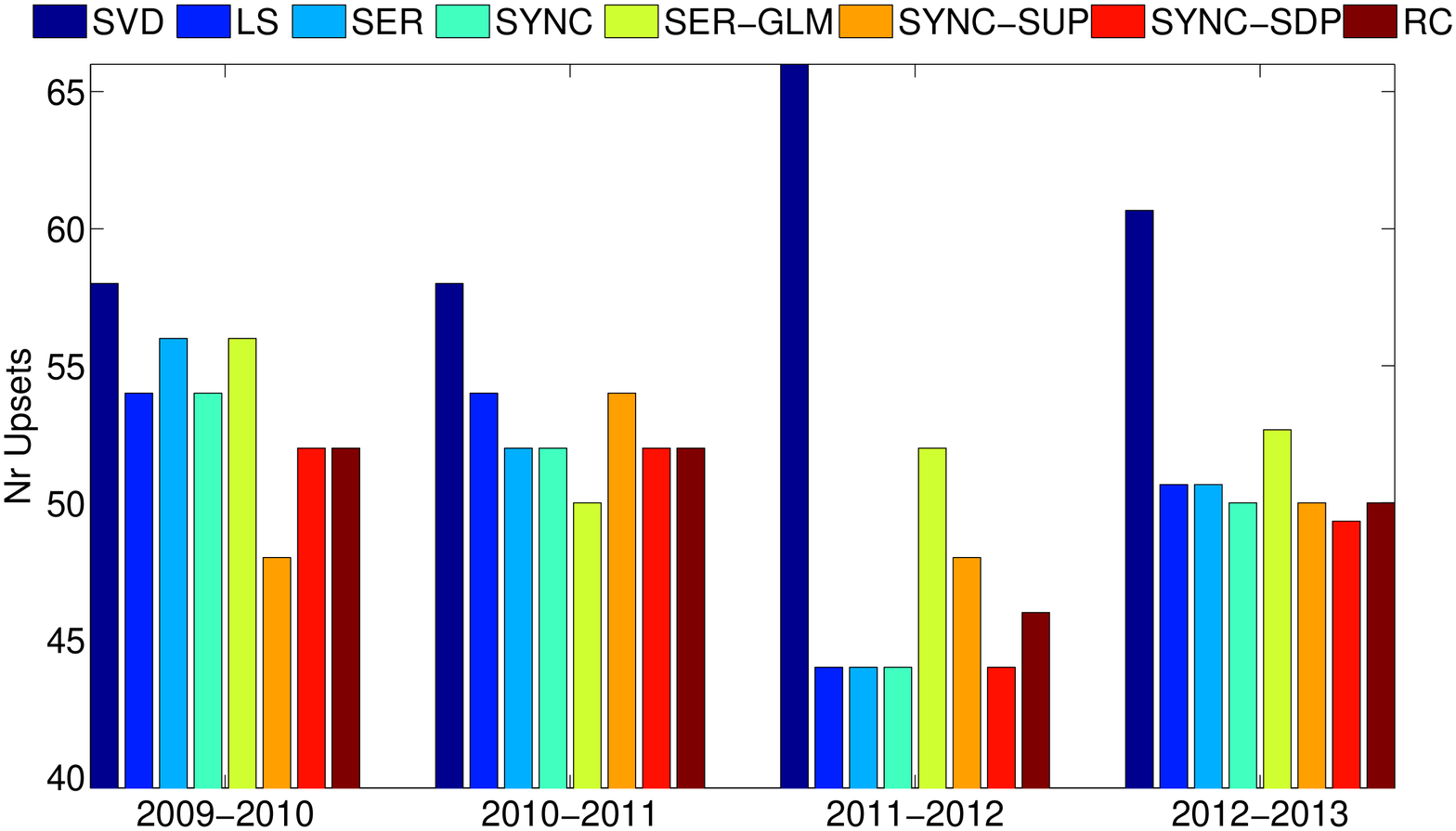}
\includegraphics[width=0.24\textwidth]{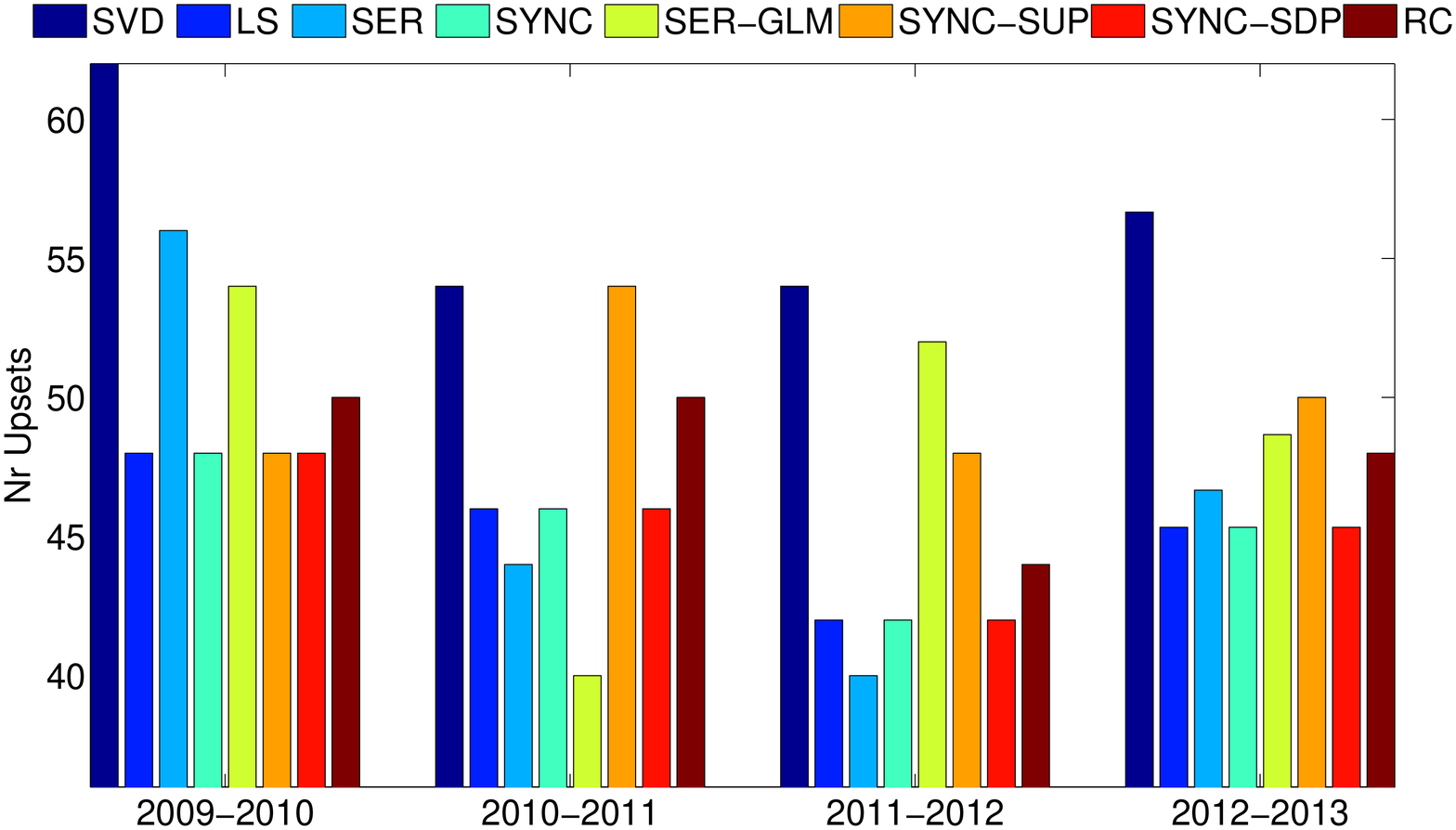}
\includegraphics[width=0.24\textwidth]{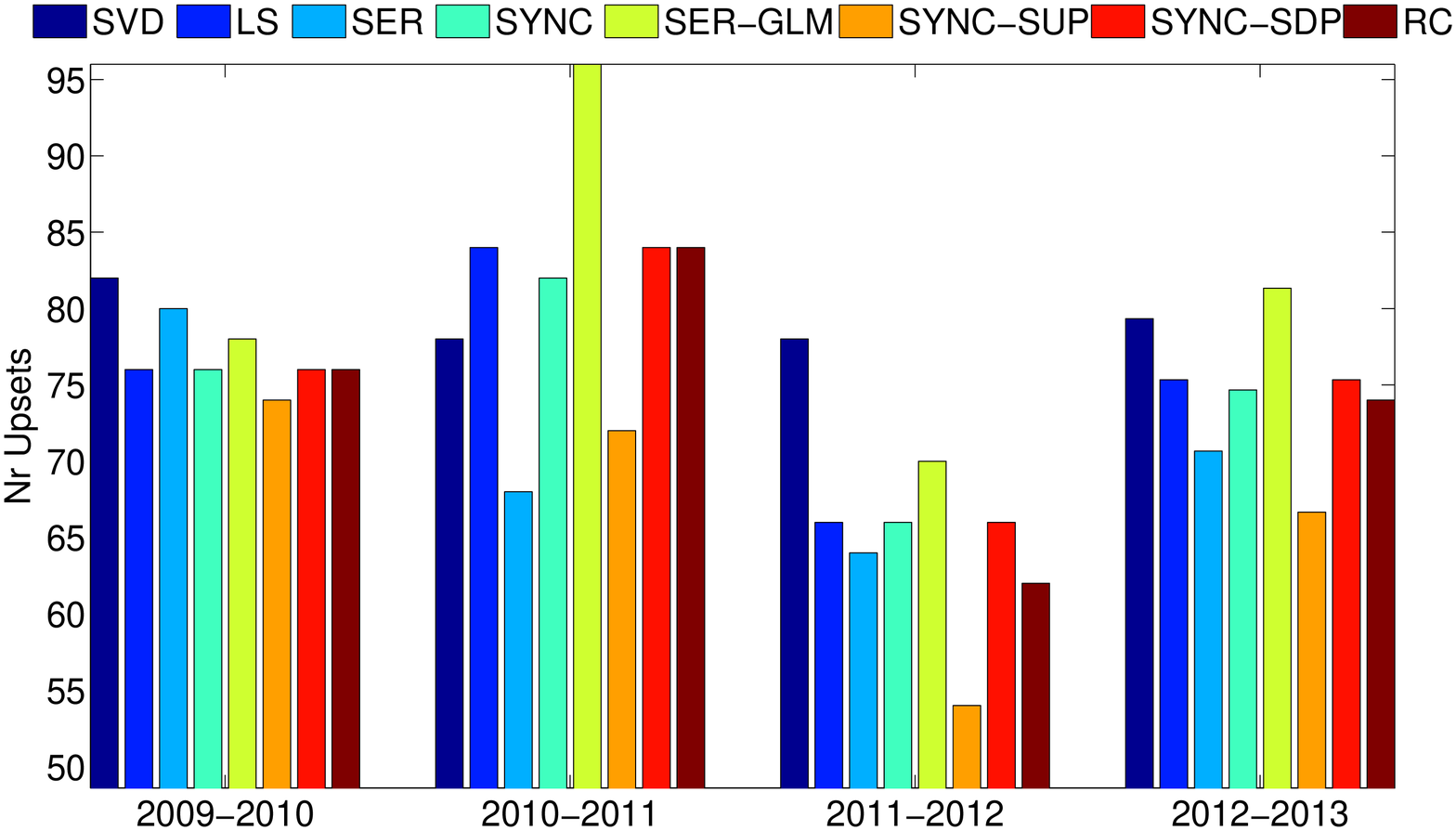}
\includegraphics[width=0.24\textwidth]{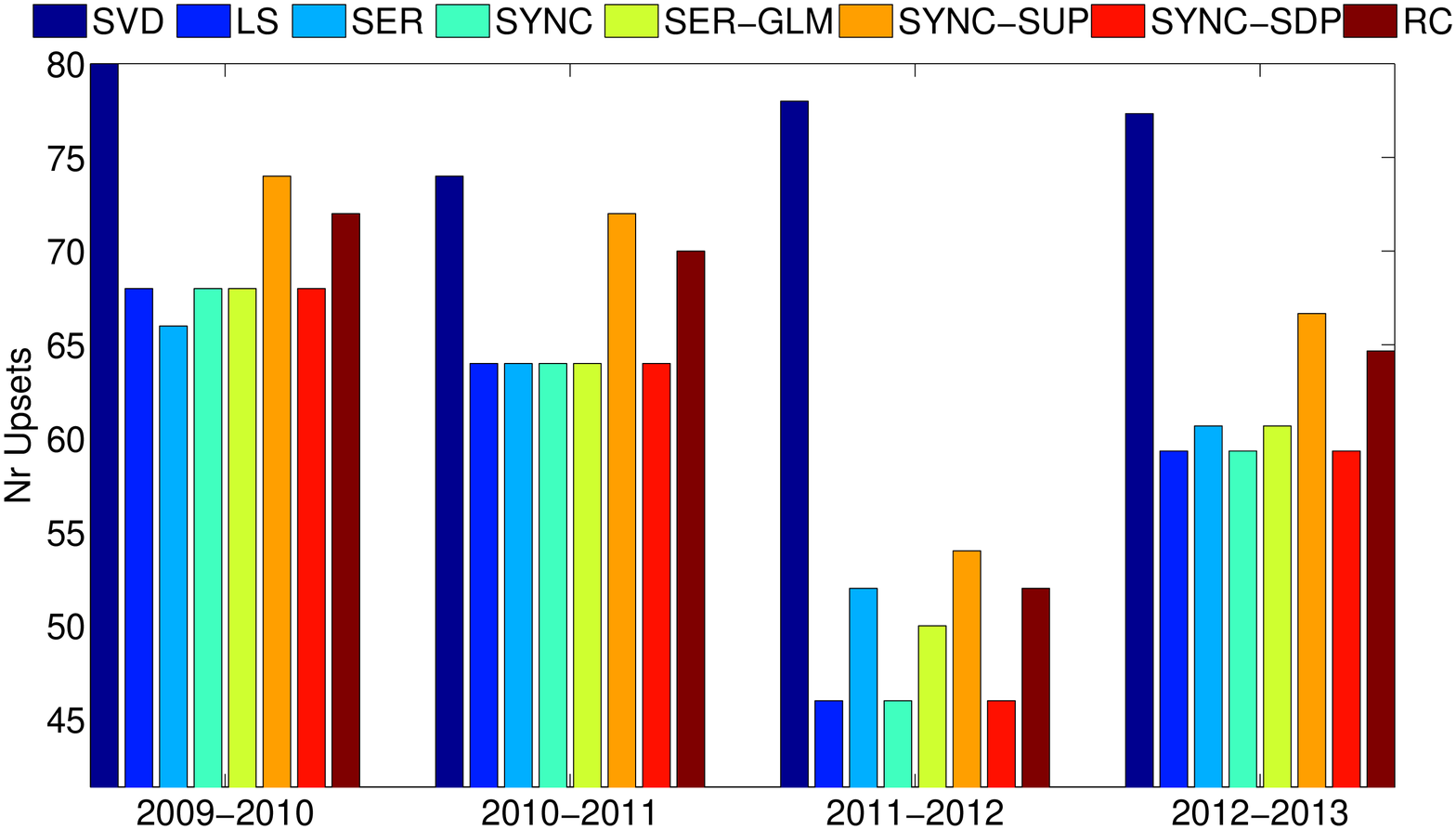}
}
\subfigure[$Q^{(s)}$ correlation score (higher is better)]{
\includegraphics[width=0.24\textwidth]{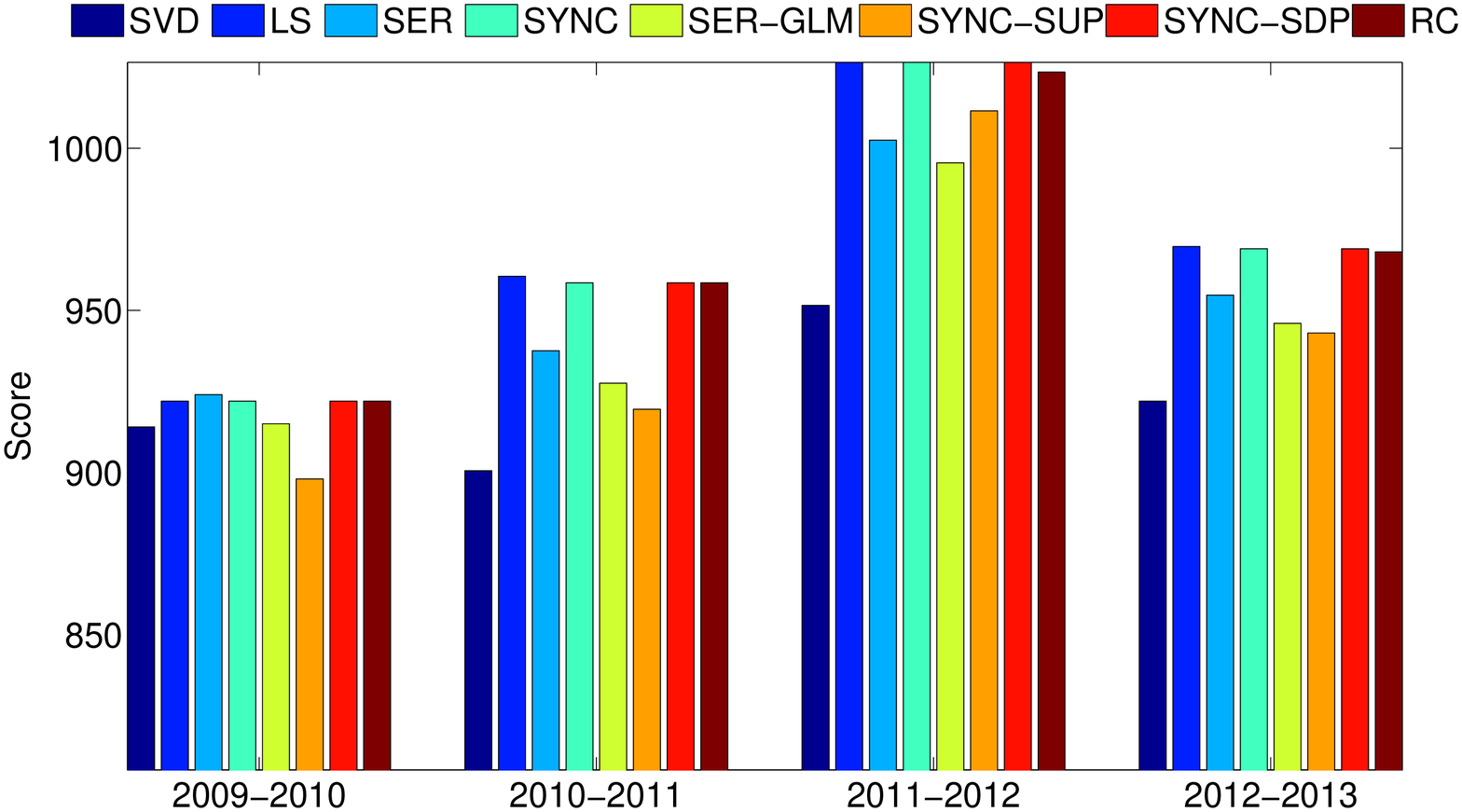}
\includegraphics[width=0.24\textwidth]{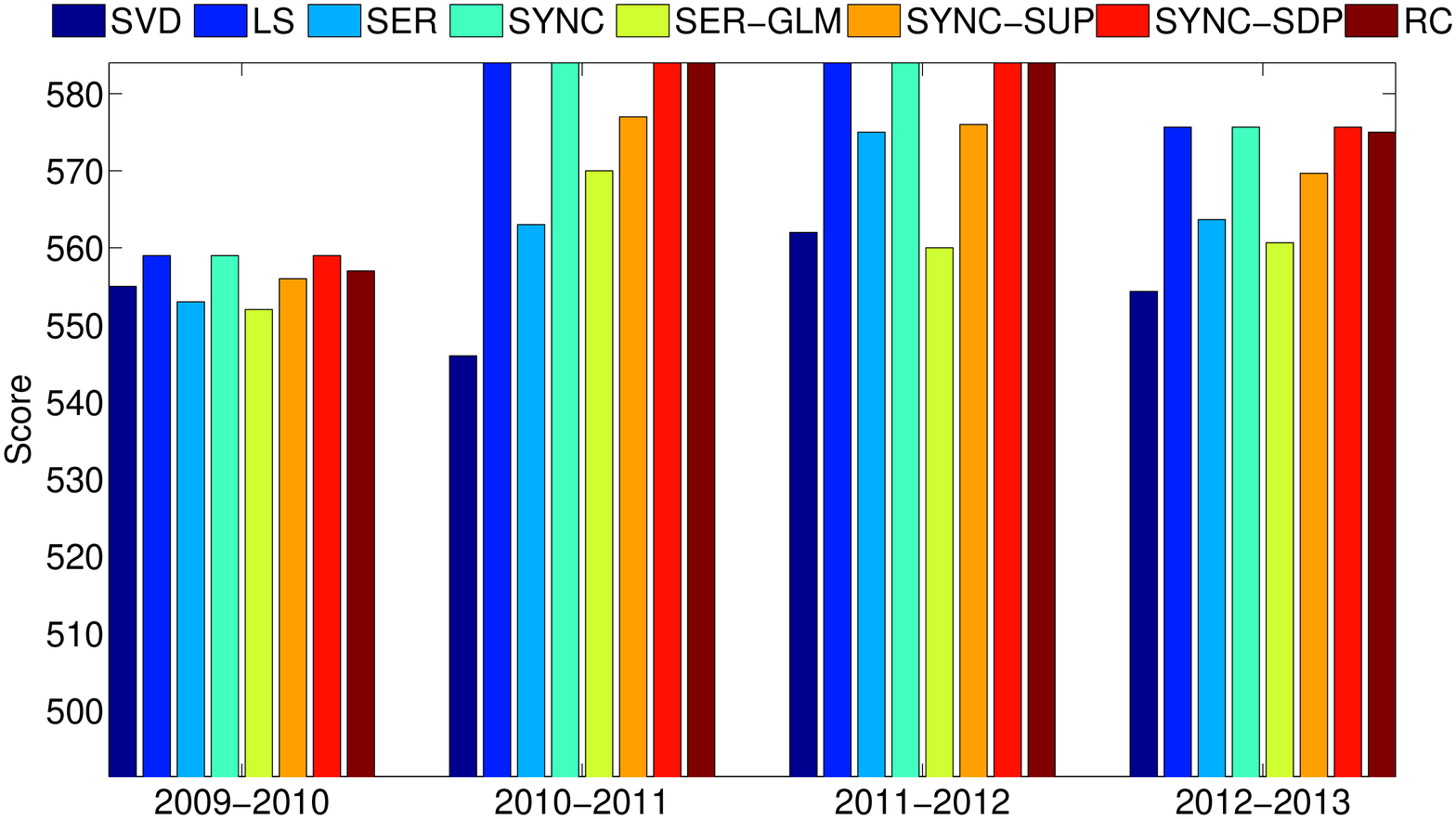}
\includegraphics[width=0.24\textwidth]{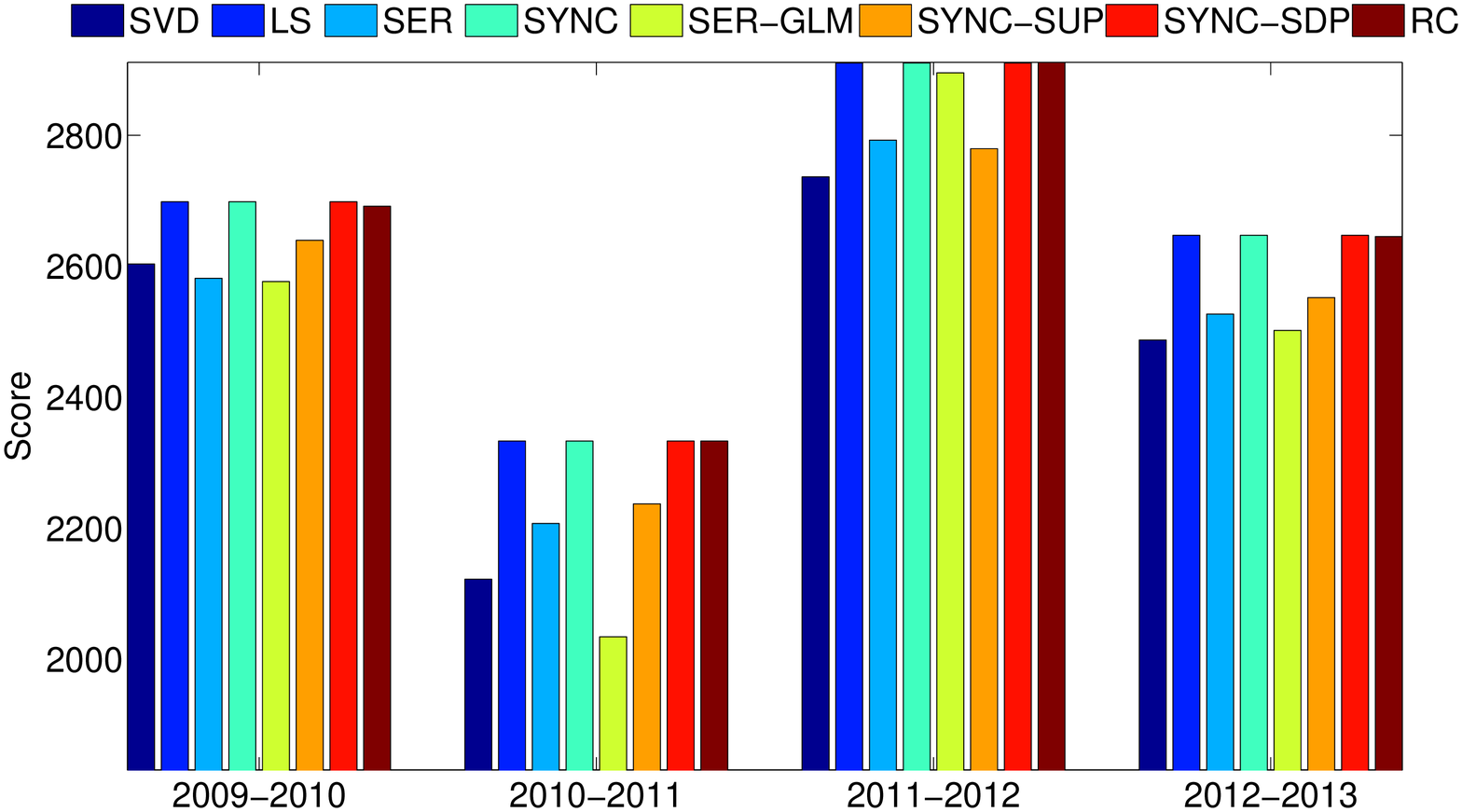}
\includegraphics[width=0.24\textwidth]{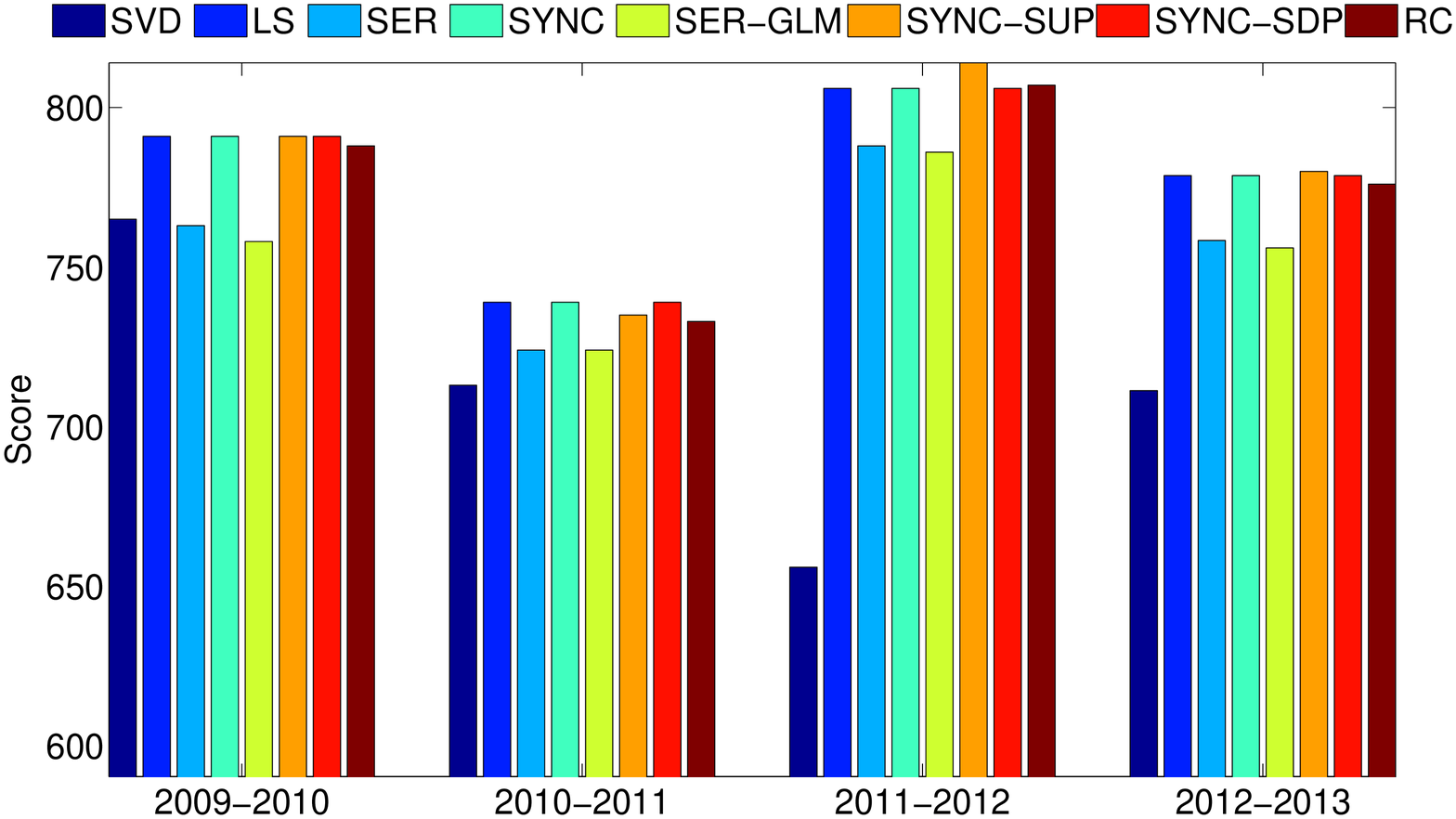}
}
\subfigure[$Q^{(w)}$  correlation Score (higher is better)]{
\includegraphics[width=0.24\textwidth]{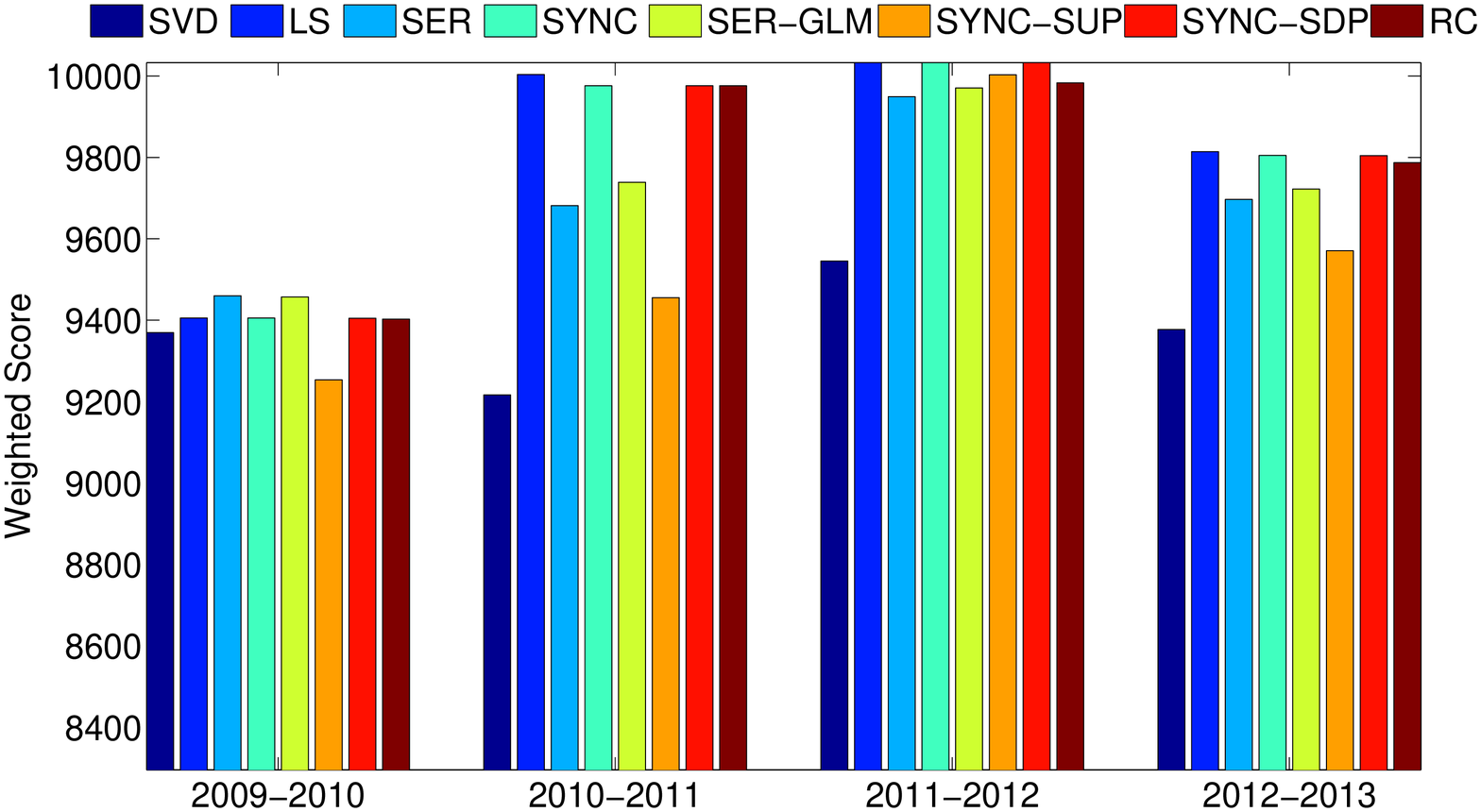}
\includegraphics[width=0.24\textwidth]{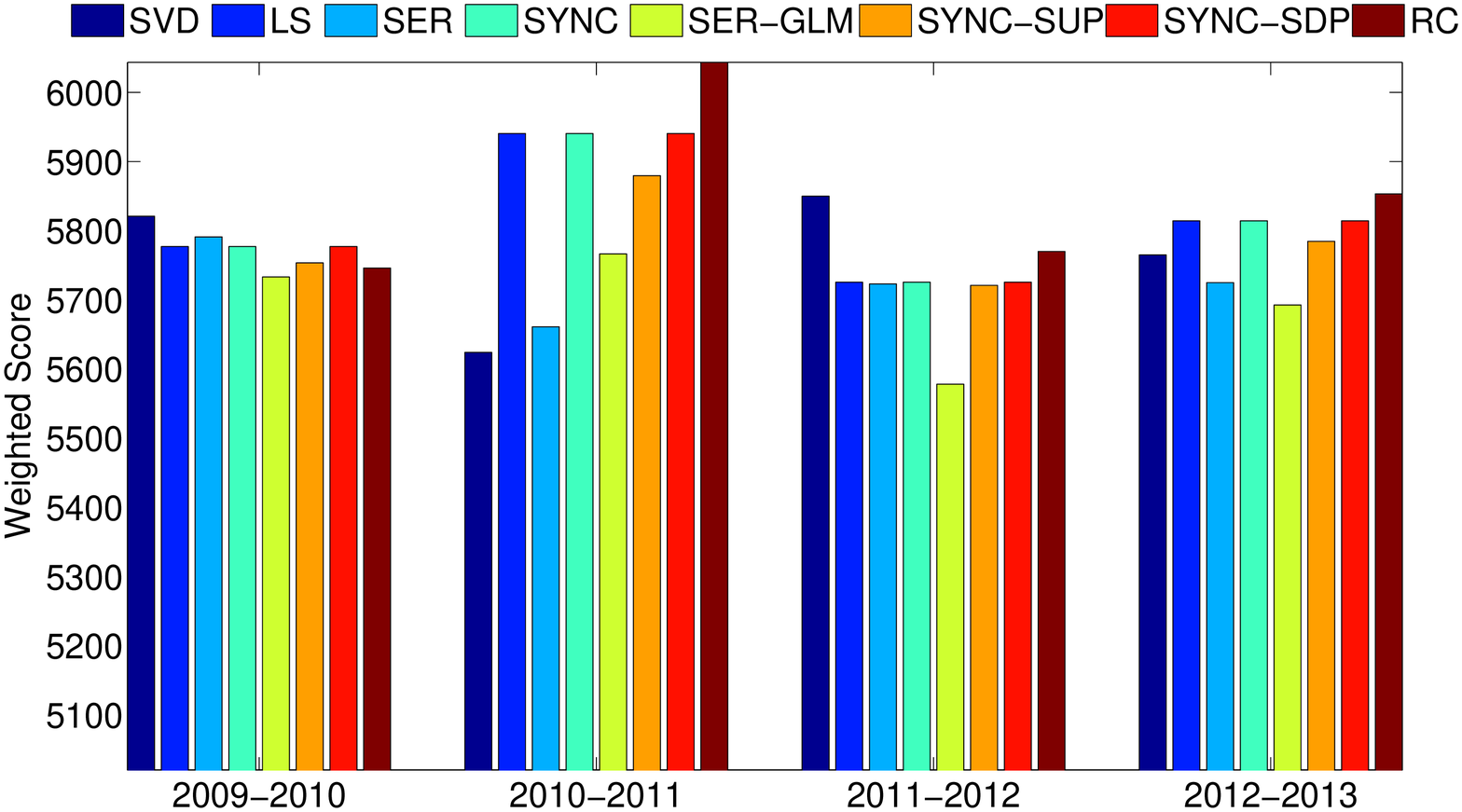}
\includegraphics[width=0.24\textwidth]{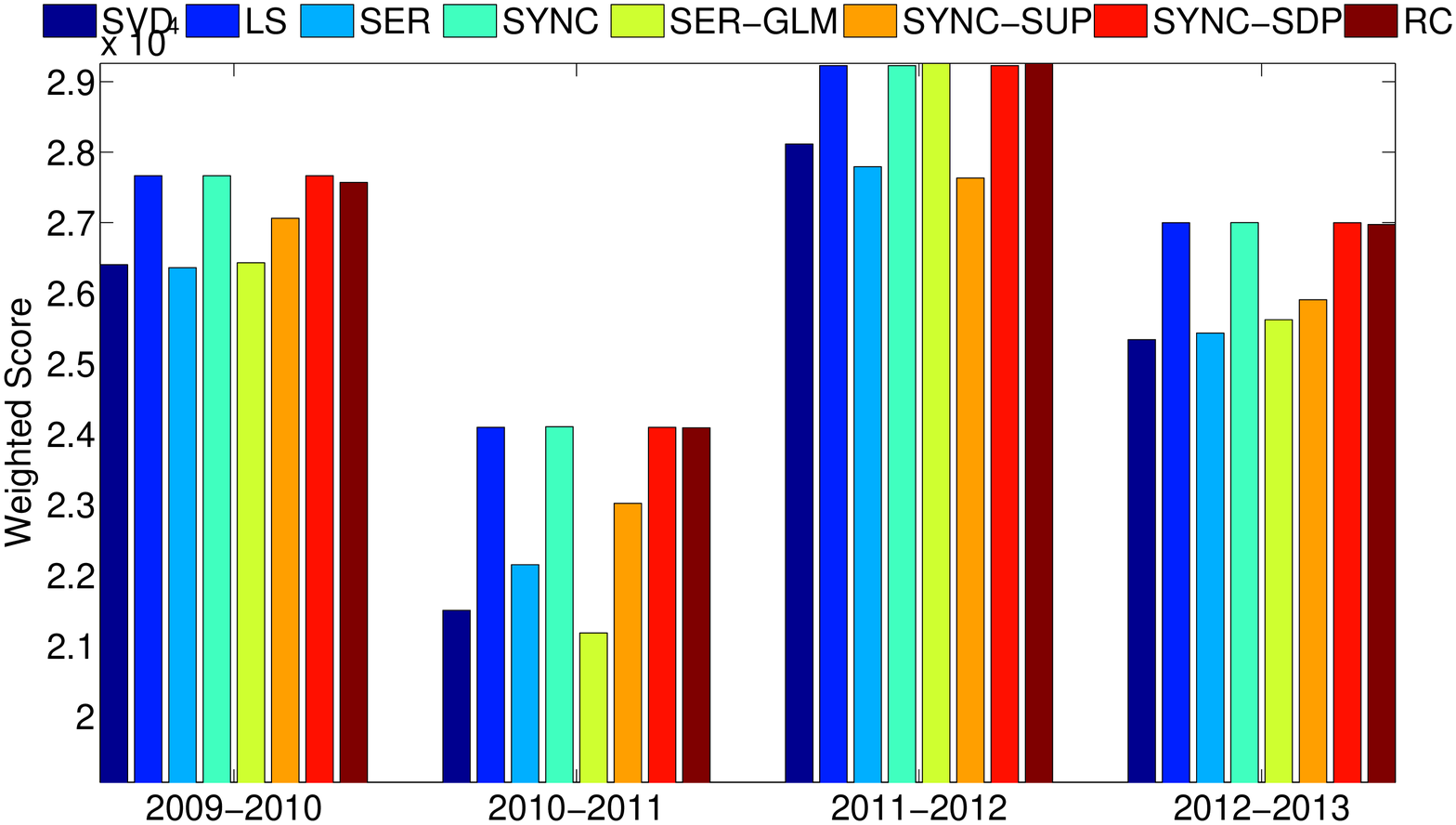}
\includegraphics[width=0.24\textwidth]{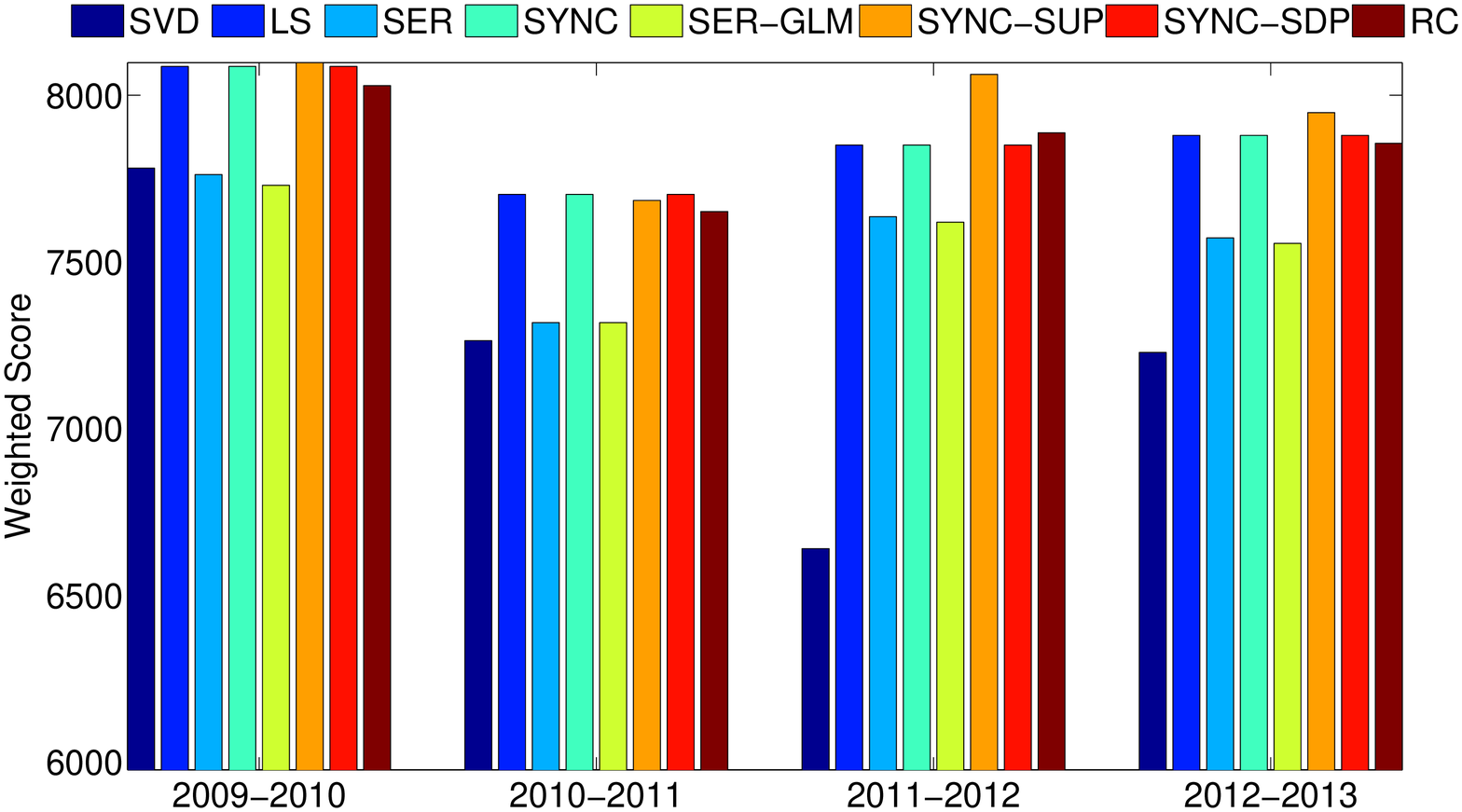}
}
\vspace{-4mm}
\caption{English Premier League 2011-2014. The input data is as follows.
Column 1:   
$C^{nw}$, 
Column 2: 
$C^{snw}$, 
Column 3: 
$C^{tpd}$, and 
Column 4: 
$C^{stpd}$ (as we number the columns left to right).
}
\end{center}
\label{fig:EnglandErors}
\end{figure}

\subsection{Numerical Comparison on the Halo-2 Data Set}  \label{secsec:NumerExpHalo}

In this section, we detail the outcome of numerical experiments performed on a second real data set of game outcomes gathered during the Beta testing period for the Xbox game Halo 2 \footnote{Credits for the use of the Halo 2 Beta Dataset are given to Microsoft Research Ltd. and Bungie.}. 
There are 606 players, who play a total of $6227$ head-to-head games. We remove the low degree nodes, i.e., discard the players who have played less than 3 games. 
The resulting comparison graph has $n=535$ nodes, and $6109$ edges, with average (respectively maximum) degree of the graph being $22.8$ (respectively $125$) and standard deviation $19.6$. 
We show the histogram of resulting degrees in Figure \ref{fig:HaloDegDist}.

\begin{figure}[h!]
\begin{center}
\subfigure[Nr upsets  $Q^{(u)}$  (lower is better)]{
\includegraphics[width=0.24\textwidth]{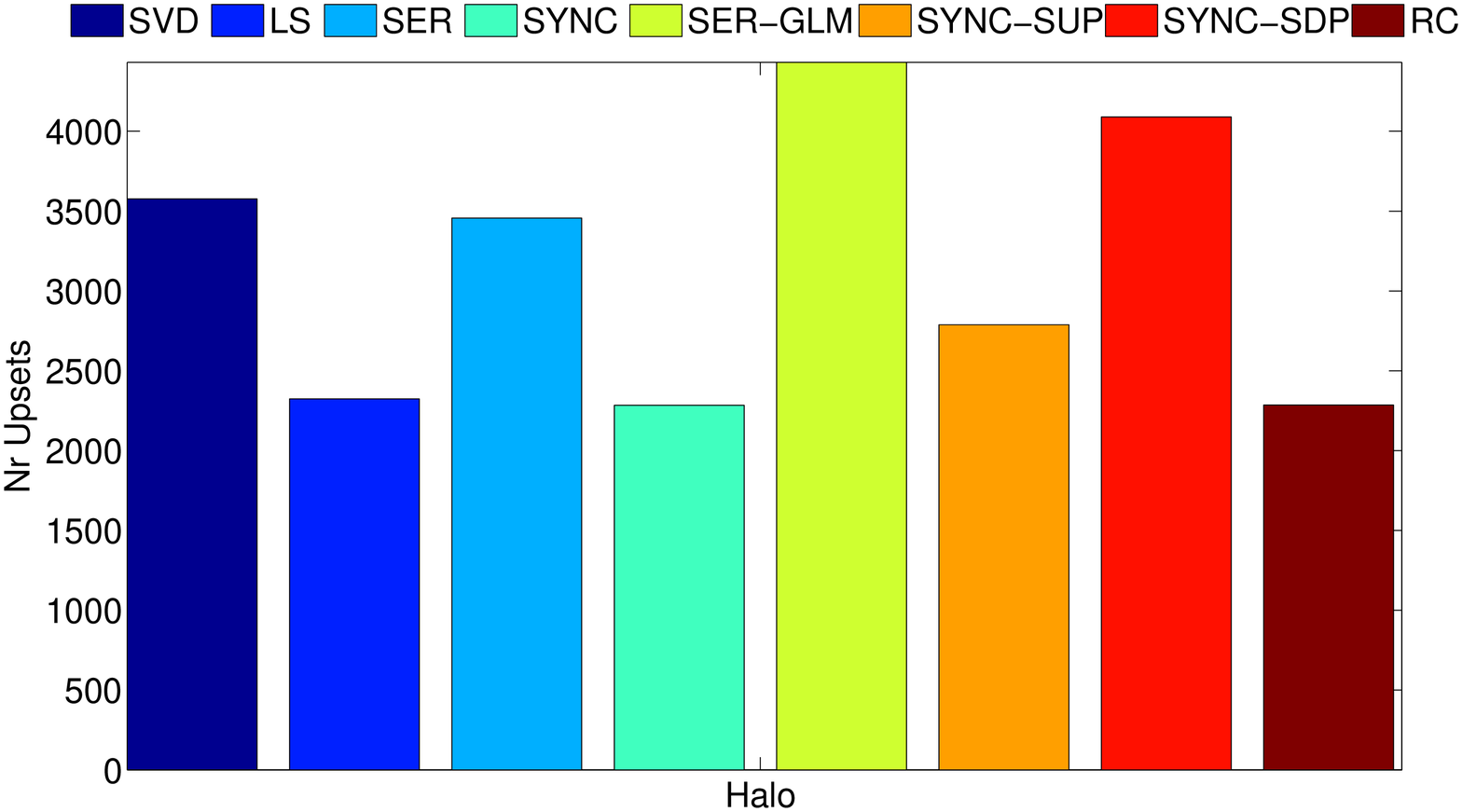}
\includegraphics[width=0.24\textwidth]{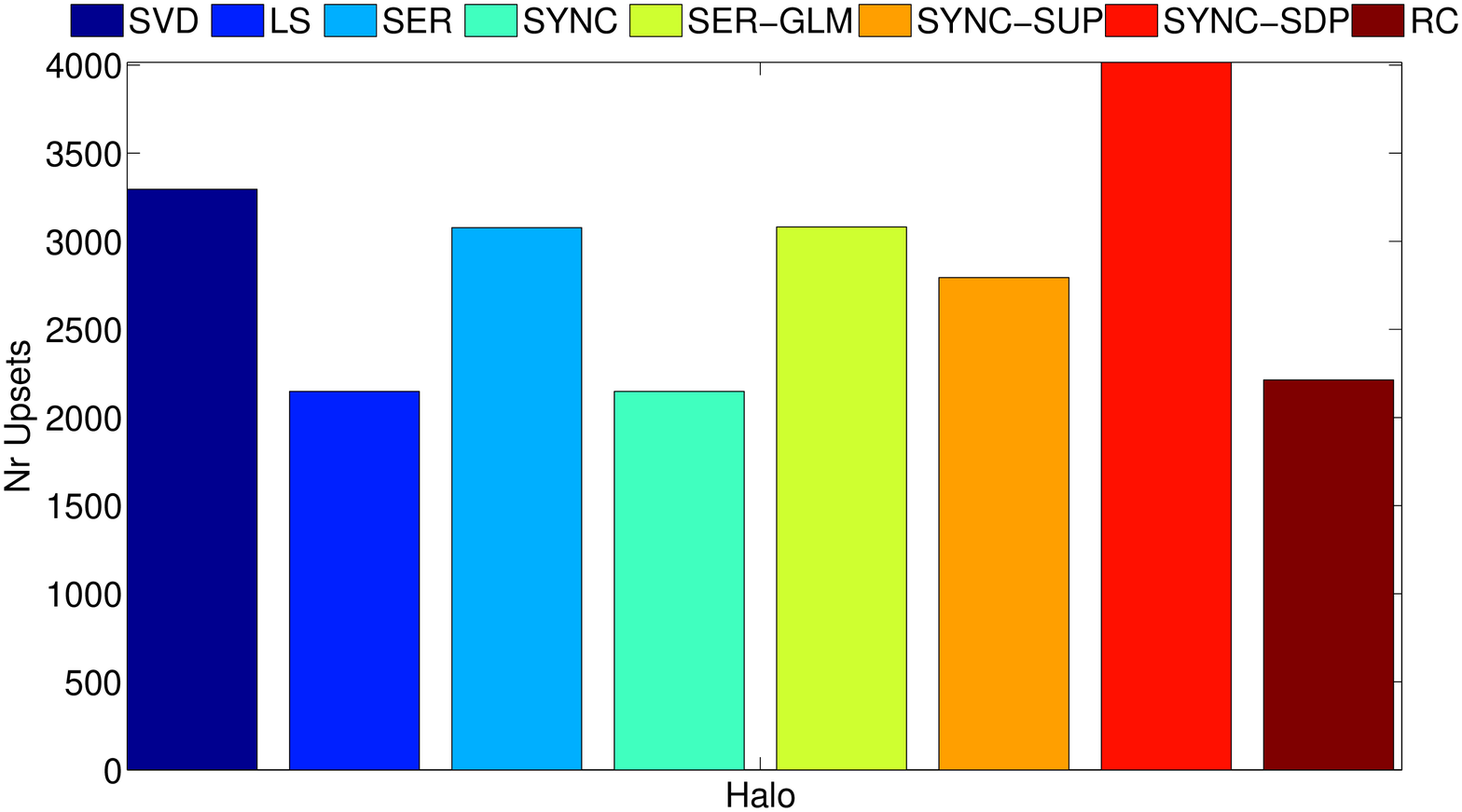}
\includegraphics[width=0.24\textwidth]{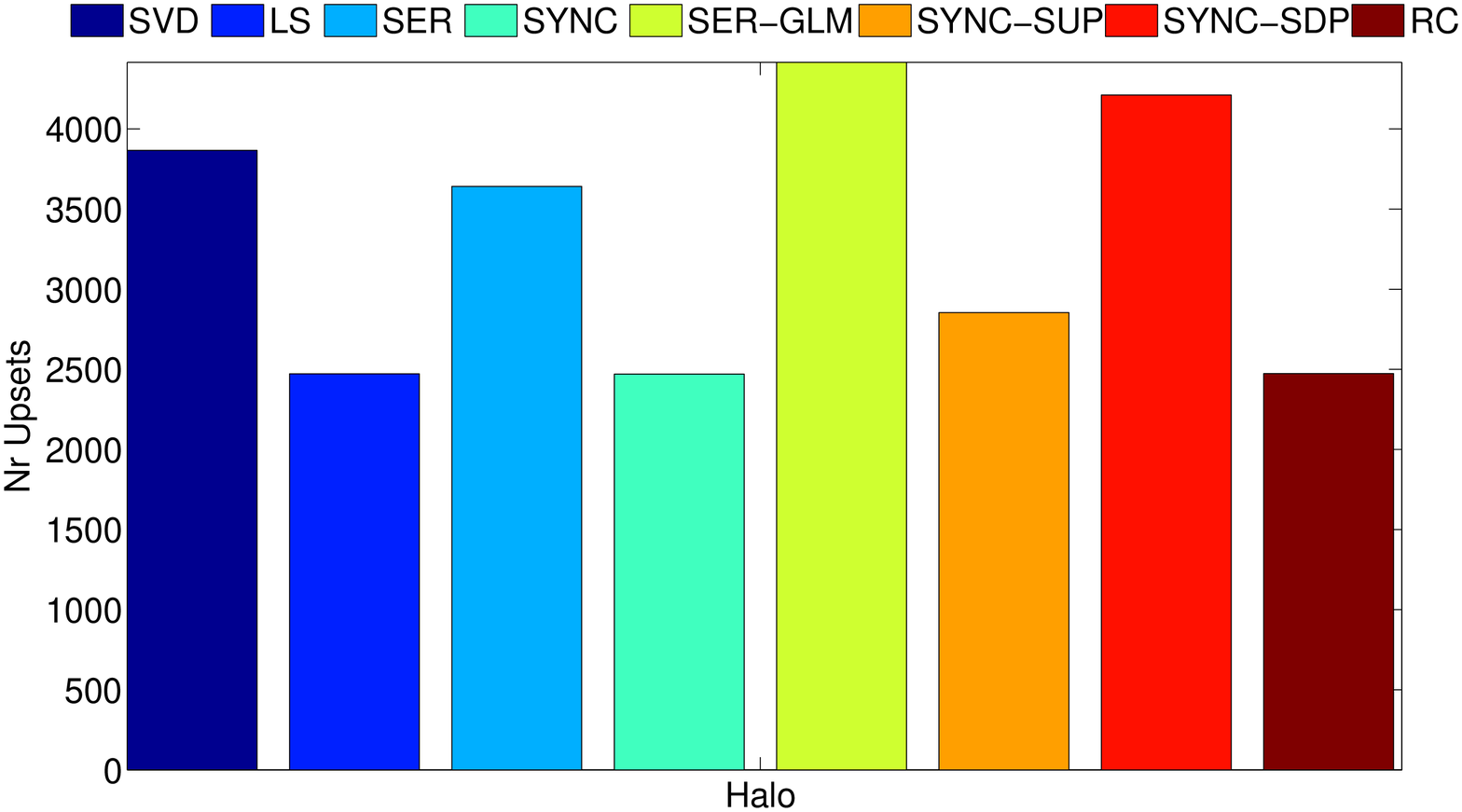}
\includegraphics[width=0.24\textwidth]{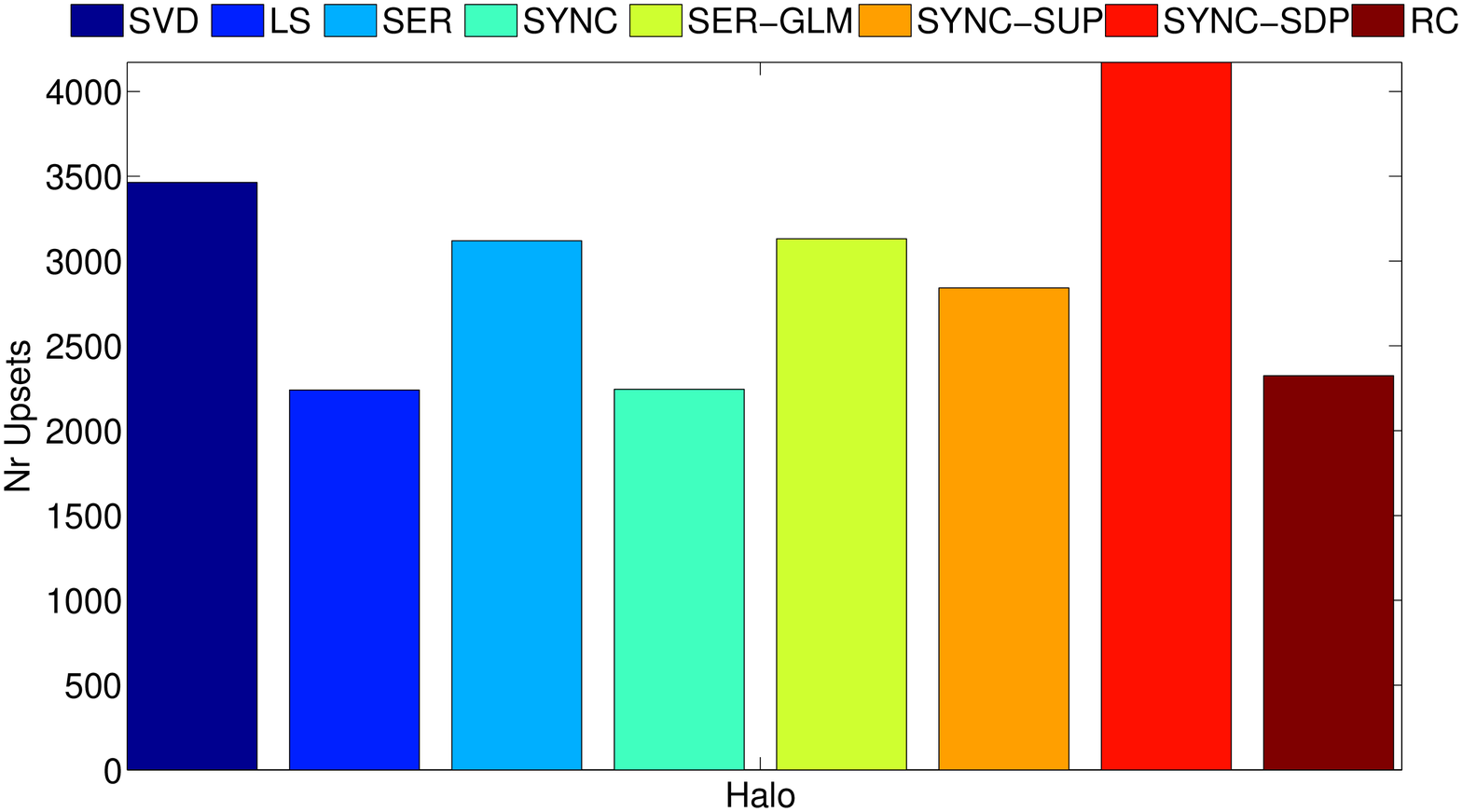}
}
\subfigure[ $Q^{(s)}$ correlation score (higher is better)]{
\includegraphics[width=0.24\textwidth]{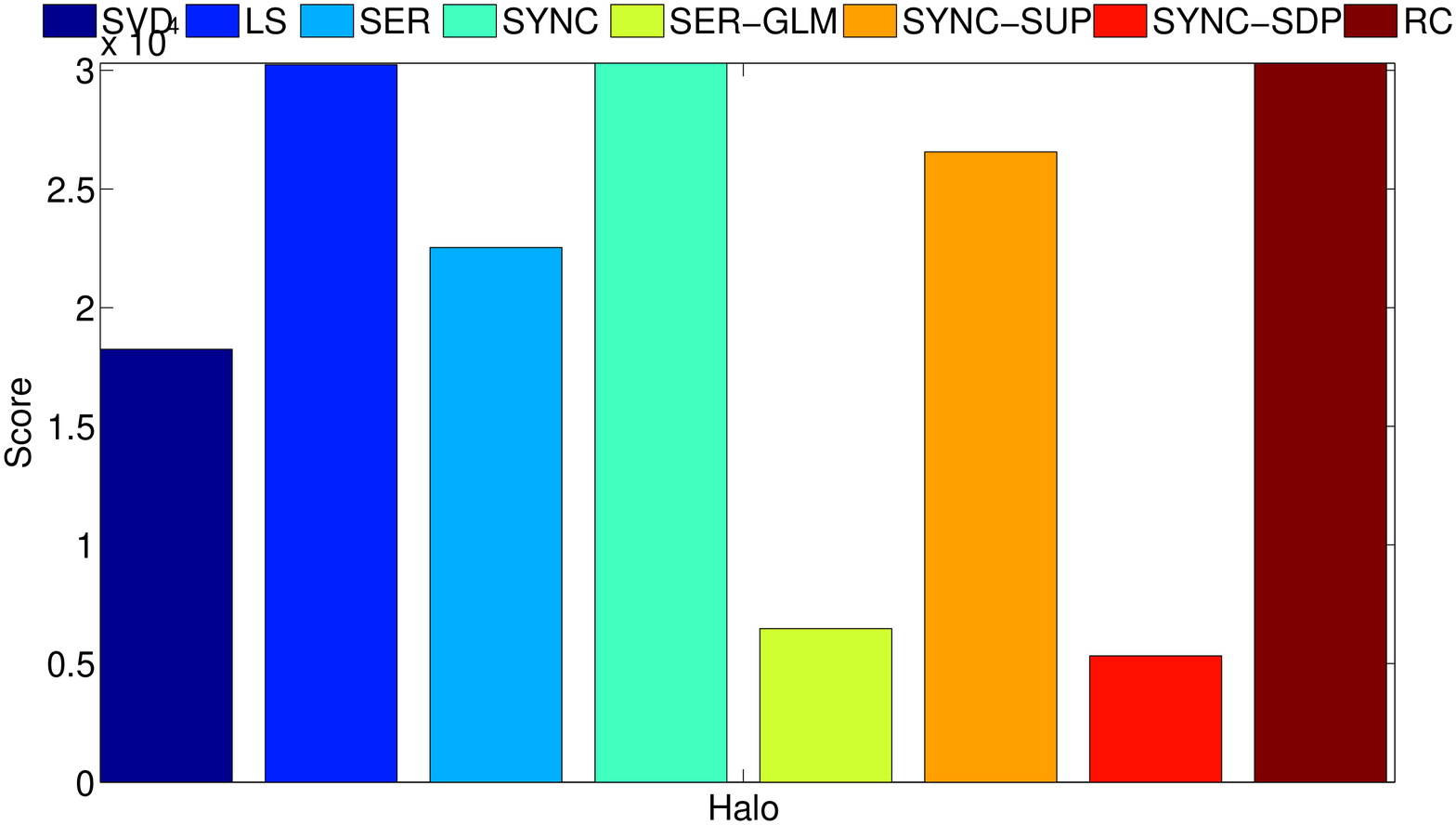}
\includegraphics[width=0.24\textwidth]{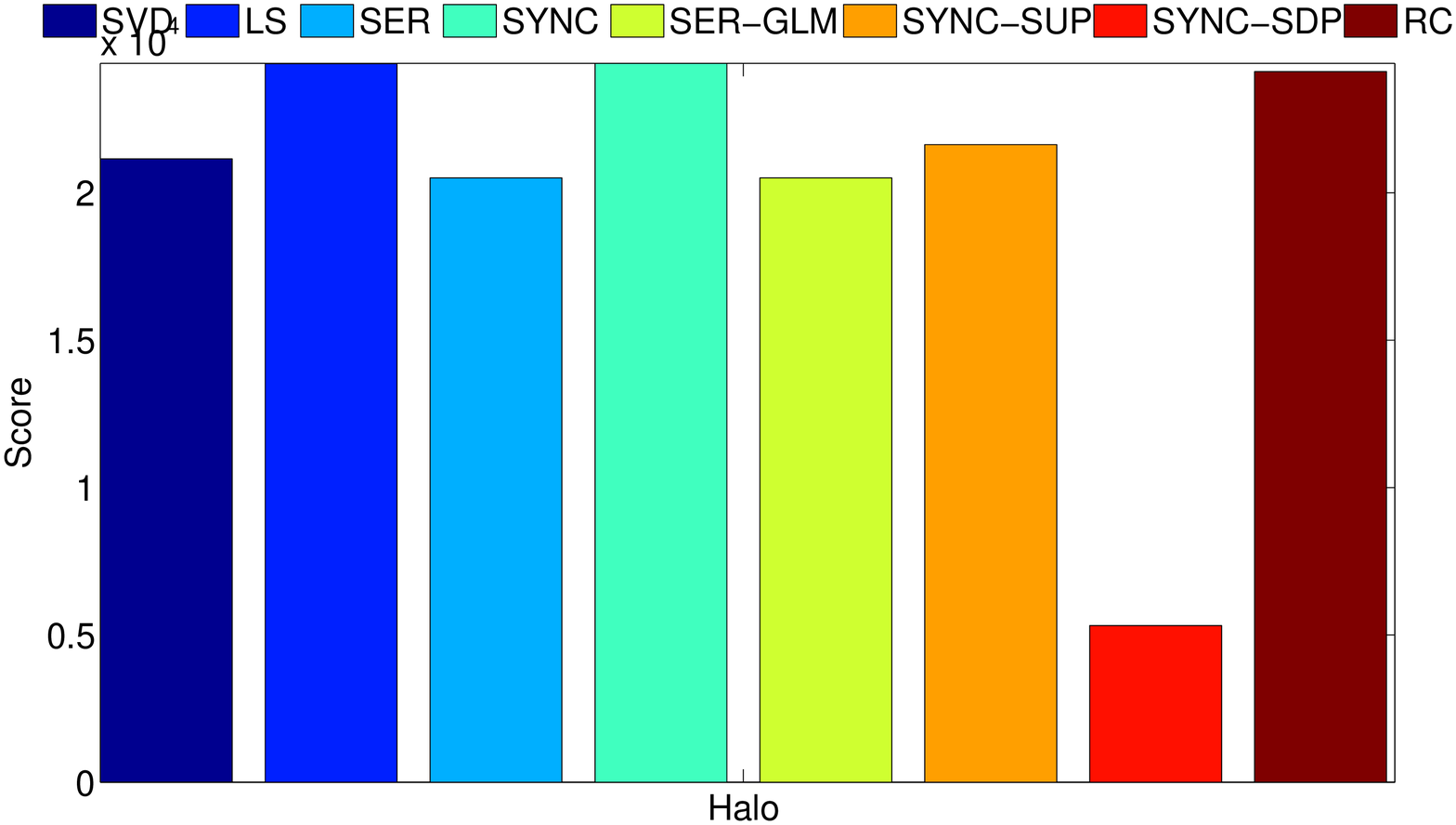}
\includegraphics[width=0.24\textwidth]{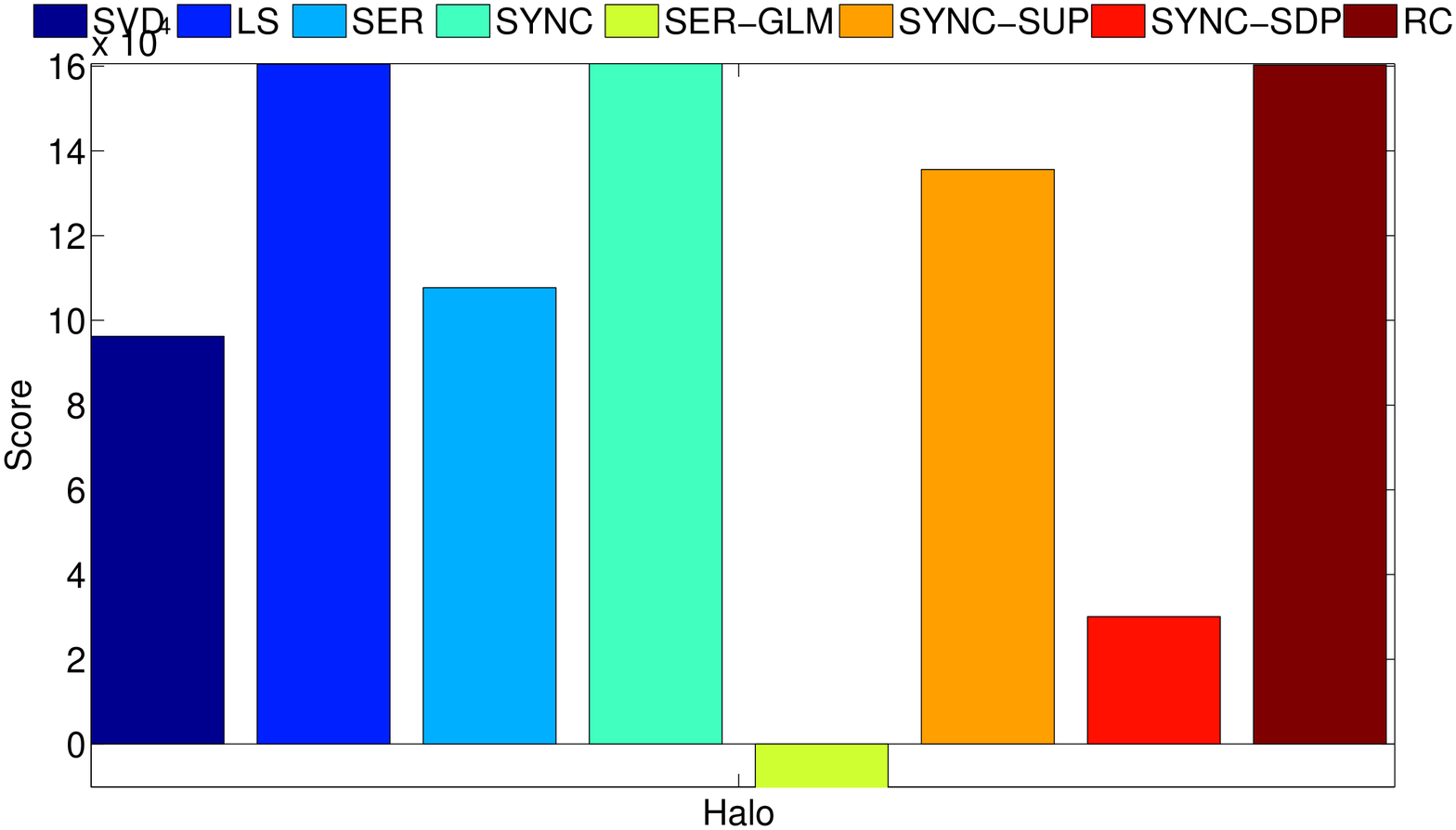}
\includegraphics[width=0.24\textwidth]{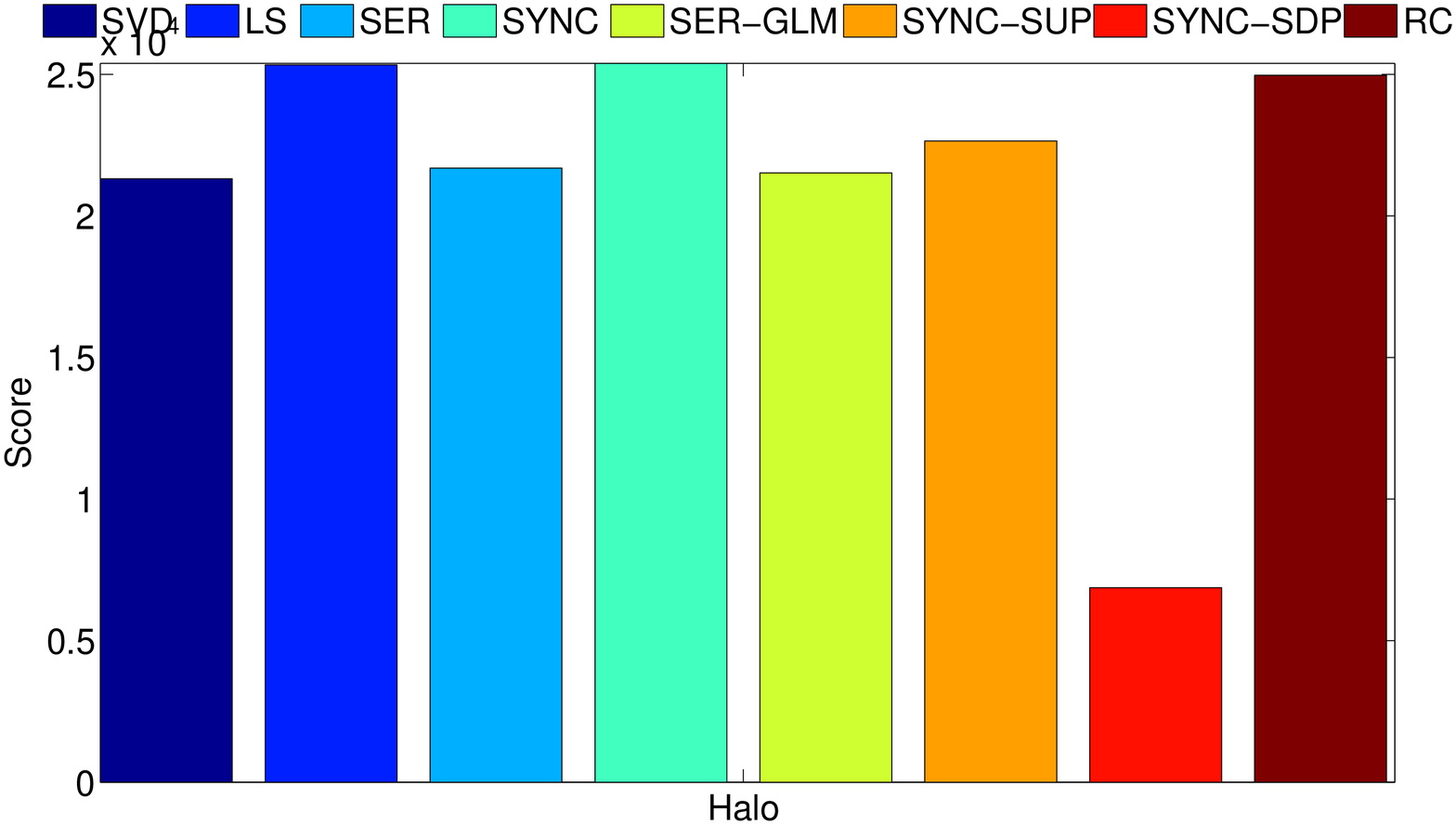}
}
\subfigure[ $Q^{(w)}$  correlation score (higher is better)]{
\includegraphics[width=0.24\textwidth]{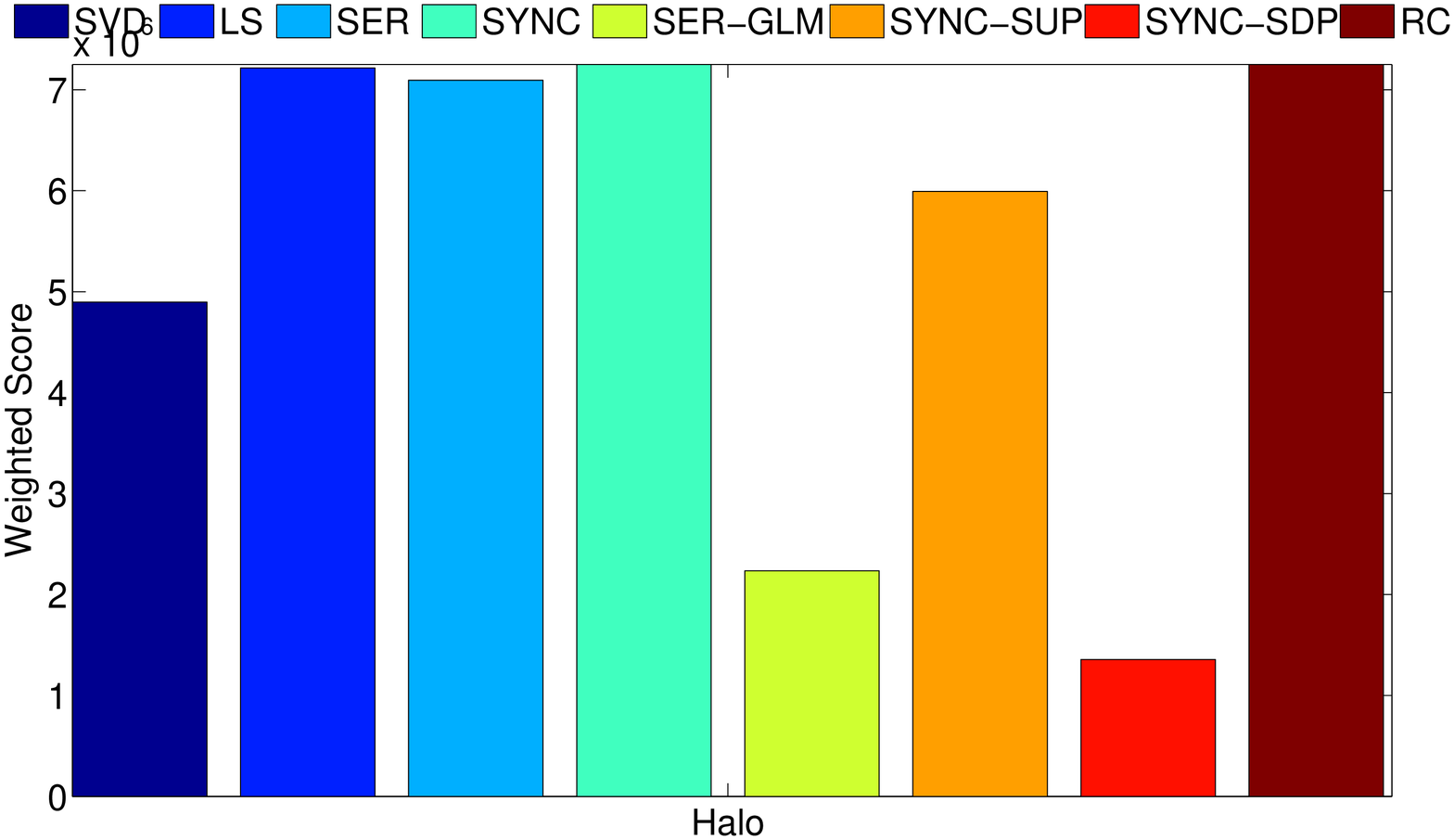}
\includegraphics[width=0.24\textwidth]{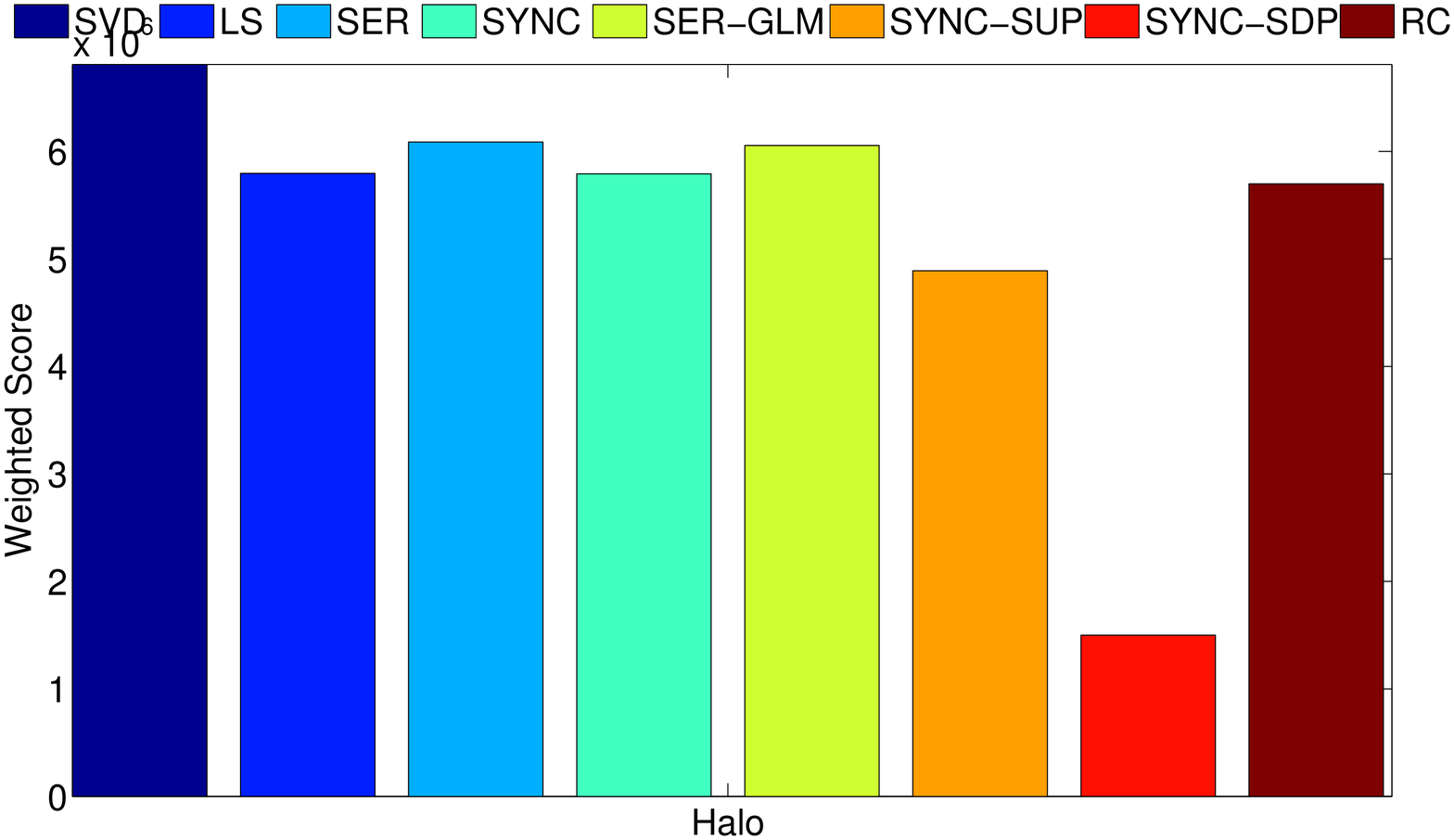}
\includegraphics[width=0.24\textwidth]{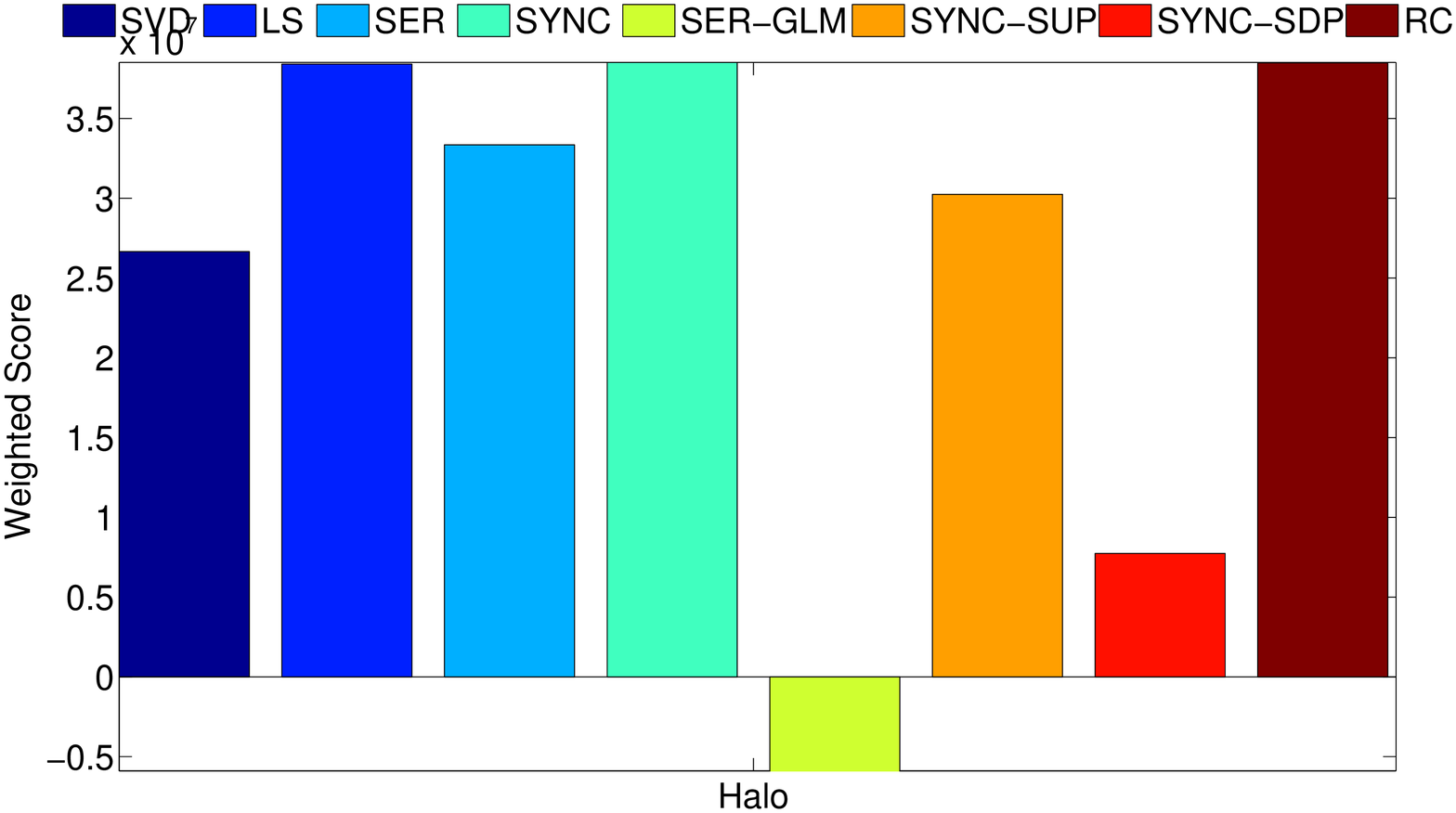}
\includegraphics[width=0.24\textwidth]{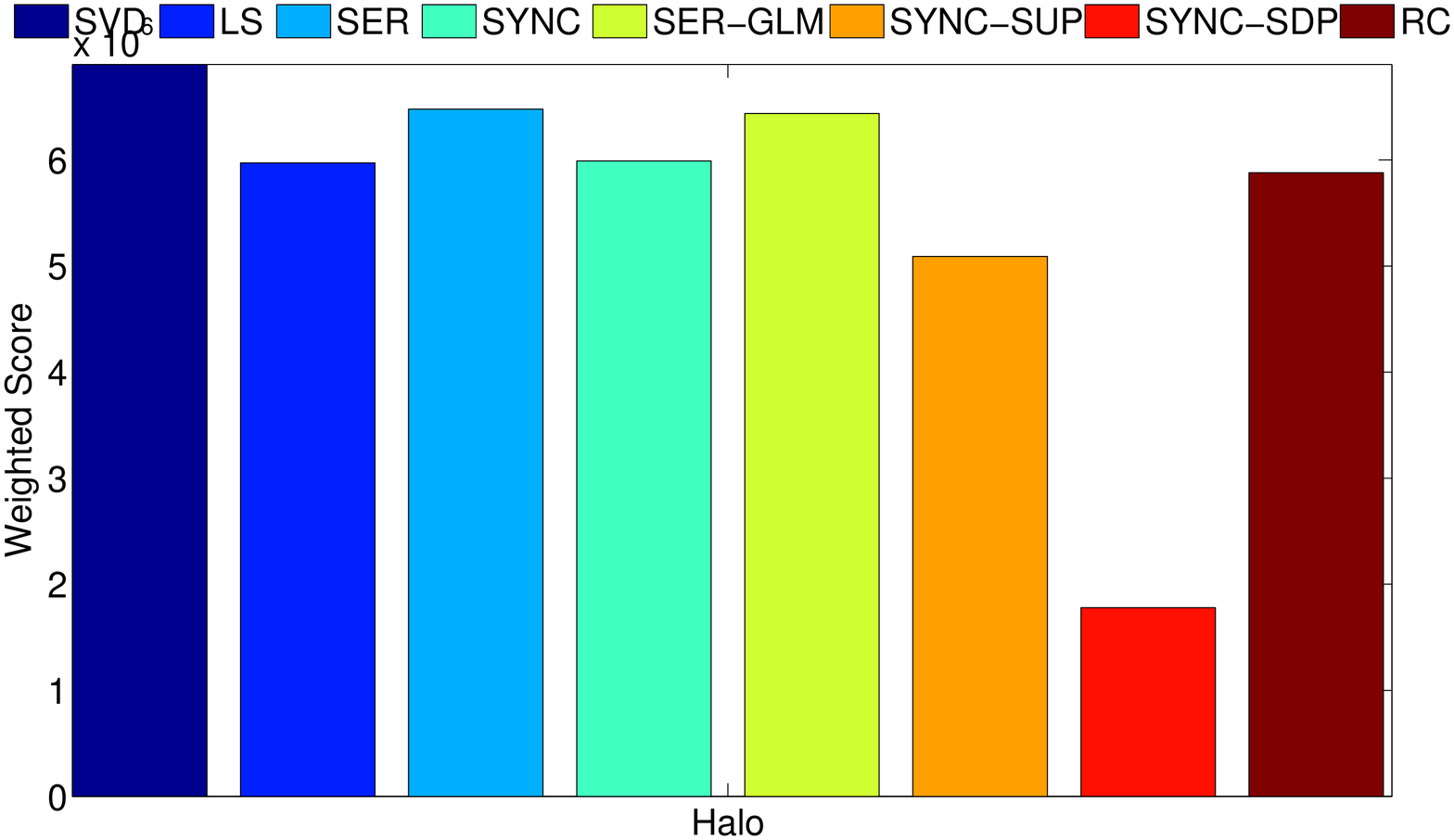}
}
\caption{A comparison of the methods on the Halo data set, based on Head-to-Head matches. 
The input is as follows. 
Column 1:   
$C^{nw}$, 
Column 2: 
$C^{snw}$, 
Column 3: 
$C^{tpd}$, and 
Column 4: 
$C^{stpd}$ (as we number the columns left to right).
}
\end{center}
\label{fig:HaloErors}
\end{figure}

We plot in Figure \ref{fig:HaloErors} the results obtained, in terms of the number of upsets (a), and the two correlation scores in (b) and (c), 
as we compare all the methods considered so far, across the four different possible types of inputs $C^{nw}$,  $C^{snw}$,  $C^{tpd}$, and  $C^{stpd}$ (which index the columns from left to right, in Figure \ref{fig:HaloErors}).  
We remark that LS, SYNC, and RC achieve the lowest number of upsets $Q^{(u)}$, across all possible inputs, and also the best results for the 
$Q^{(s)}$ correlation score. The ranking in terms of performance based on 
the correlation score  $Q^{(w)}$ varies across the different types of inputs, with SVD achieving the best results in two instances.

\subsection{Numerical Comparison on the NCAA College Basketball Data Set} \label{secsec:NumerExpBasket}

Our third and final real data set contains the outcomes of NCAA College Basketball matches during the regular season, for the time interval 1985 - 2014. During the regular season, most often it is the case that a pair of teams play against each other at most once. For example, during the 2014 season, for which 351 teams competed, there were a total of 5362 games, with each team playing on average about 30 games.
We remark that in the earlier years, there were significantly less teams participating in the season, and therefore  games played, which also explains the increasing number of upsets for the more recent years, as shown in Figure  
\ref{fig:BasketErors} (a). For example, during the 1985 season, 282 teams participated,  playing a total of 3737 games. 
In Figure \ref{fig:BasketErors} (a) we compare the performance of all methods across the years, for the case of both cardinal and ordinal measurements, i.e., when we consider the point difference or simply the winner of the game.
Similarly, in Figure  \ref{fig:BasketErors} (b) we compute the average number of upsets across all  years in the interval 1985-2014, for both cardinal and ordinal measurements, i.e., when we record the actual point difference or  only the winner. In the less frequent cases when a pair of teams play against each other more than once, we simply average out the available scores. 
We remark that the SYNC-SUP method, i.e., eigenvector-based synchronization based on  the superiority score given by (\ref{SijDefSuperiority}), significantly outperforms all other methods in terms of the number of upsets. The second best results are obtained (in no particular order) by LS, SYNC, SYNC-SDP and RC, which all yield very similar results, while SVD and SER are clearly the worst performers. We left out from the simulations the SER-GLM method due to its large computational running time.  

\begin{figure}[h!]
\begin{center}
\subfigure[Number of upsets, for each year in the interval 1985 - 2014.]{
\includegraphics[width=0.53\textwidth]{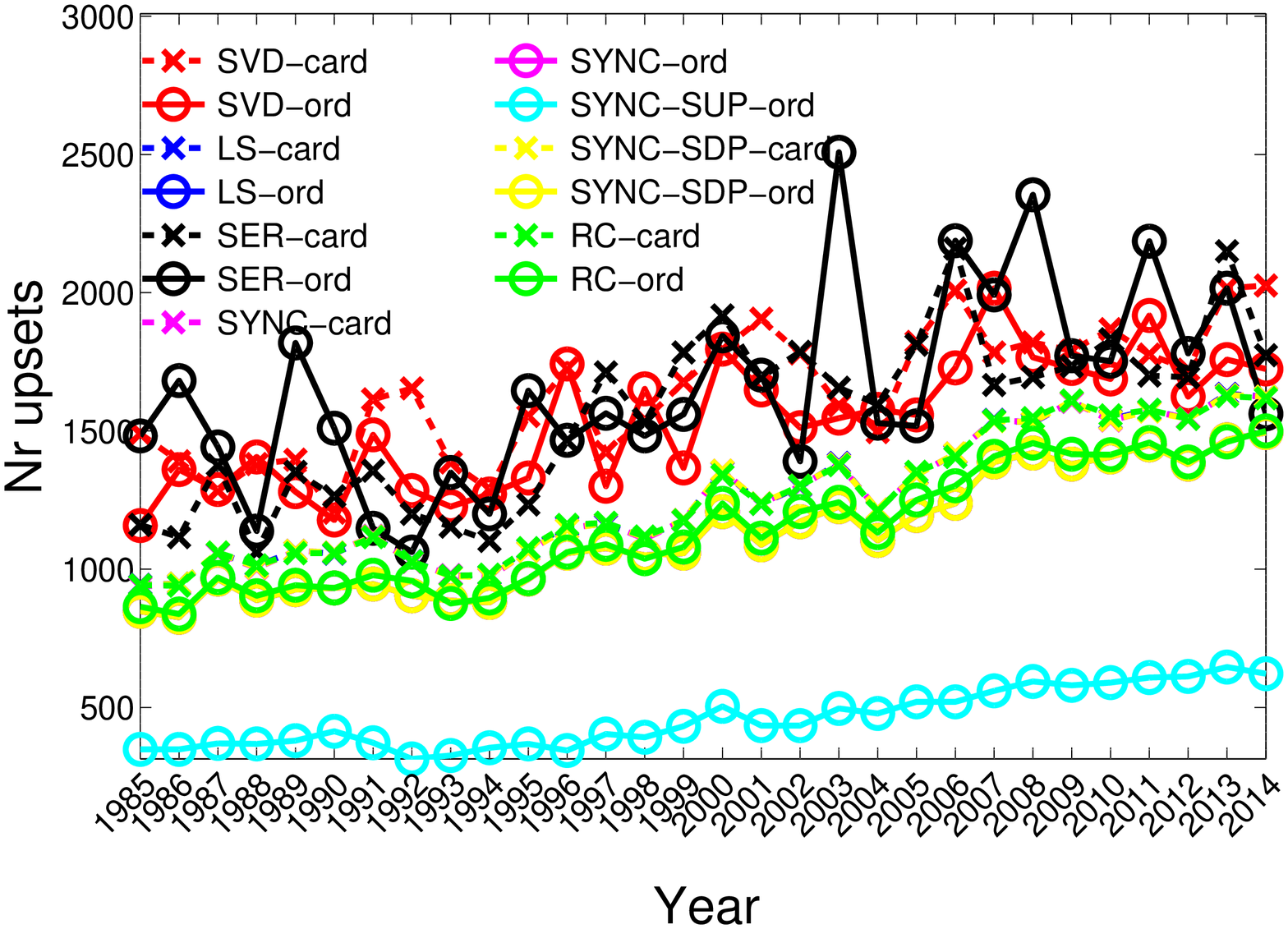}}
\subfigure[Average number of upsets across 1985 - 2014.]{
\includegraphics[width=0.44\textwidth]{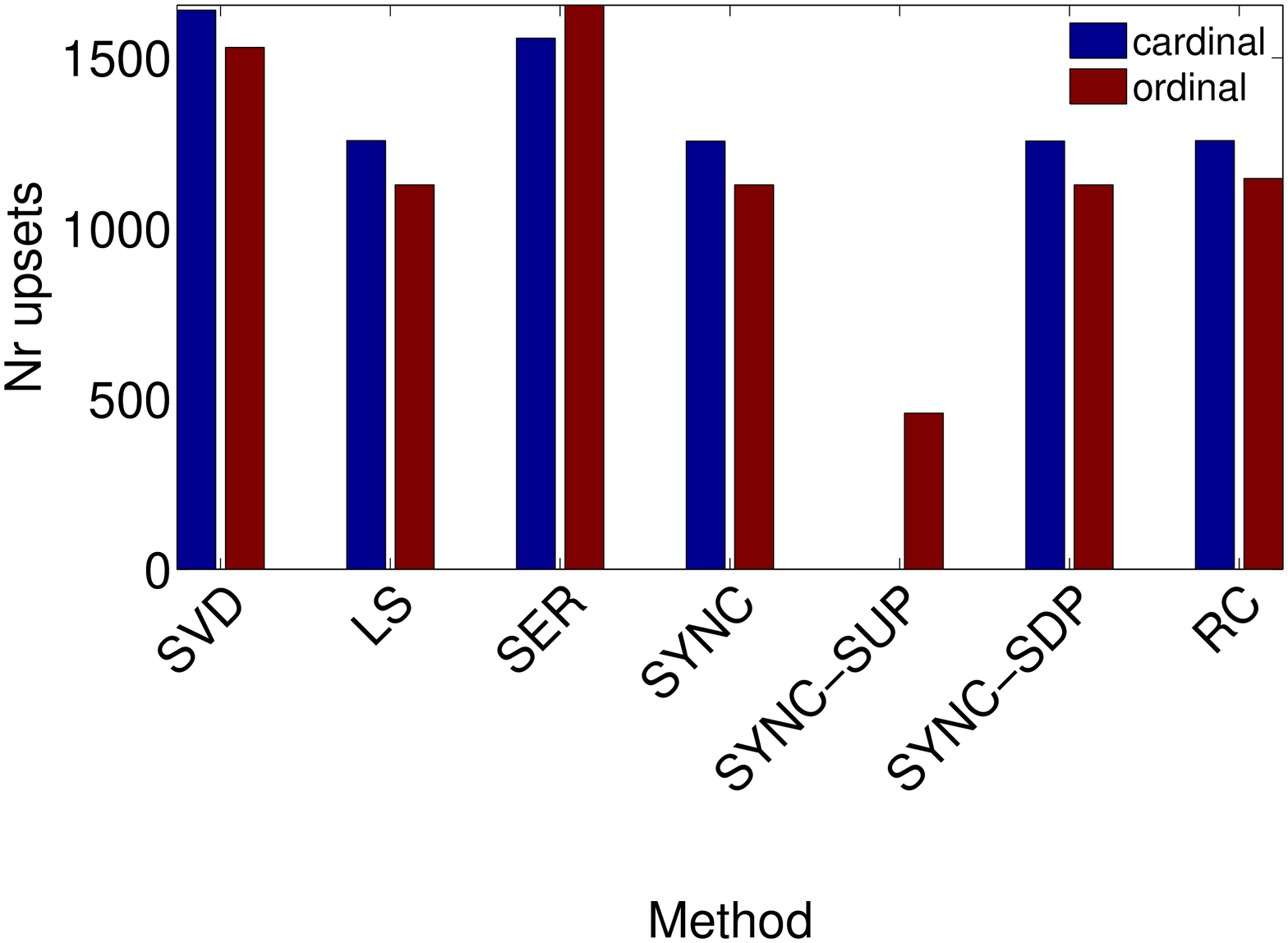}}
\caption{A comparison of the methods on the NCAA College Basketball  Championship data set, over the interval 1985-2014, in terms of the number of upsets, across the years (left) and averaged over the entire period (right). In the left figure, we append the suffix ''\textit{ord}", respectively ''\textit{card}", when the input to the ranking algorithms is given by ordinal measurements that record only the winner of each match, respectively cardinal measurements that rely on the point difference for each match.
}
\end{center}
\label{fig:BasketErors}
\end{figure}

\FloatBarrier

 \clearpage

\section{Rank Aggregation}  \label{sec:RankAggregation}

In many practical instances, it is often the case that multiple voters or rating systems provide incomplete and inconsistent rankings or pairwise comparisons between the same set of players or items. For example, at a contest, two players compete against each other, and $k$ judges decide what the score should be. Or perhaps, as in certain championships,  teams may play against each other multiple times, as is the case in the USA Basketball Championship where teams may play multiple 
matches against each other throughout a season. In such settings, a natural question to ask is, given a list of $k$ (possibly incomplete) matrices of size $n \times n$ corresponding to ordinal or cardinal rank comparisons on a set of $n$ players, how should one aggregate all the available data and produce a single ranking of the $n$ players, that is as consistent as possible with the observed data? A particular instance of this setup occurs when many partial (most likely inconsistent) rankings are provided on a subset of the items; for example, given a partial ranking of size $k$, the resulting comparison matrix will only have  $k \choose 2$ nonzero elements corresponding to all paired comparisons of the $k$ items (chosen out of all $n$ items). The difficulty of the problem is amplified on one hand by the sparsity of the measurements, since each rating system only provides a comparison for a small set of pairs of items, and on the other hand by the amount of inconsistent information in the provided individual preferences. For example, given a set of 3 items A,B,C, the first judge may prefer A to B, the second one B to C, and the third one C to A.



Questions regarding consistency of various voting methods have a long and rich history in the social sciences literature, and date back as far as the 18th century with the work of the French mathematician Marquis de Condorcet \cite{Condorcet}, most famous for his paradox   showing that majority preferences become intransitive when at least three options are available, in other words, it is possible for a certain electorate  to express a preference for candidate A over B, a preference for  B over C, and a preference for C over A, all from the same set of ballots. 
Aside from sports and related competitions, instances of the \textit{rank aggregation} problem appear in a wide variety of learning and social data analysis problems, such as recommendation systems and collaborative-filtering for ranking movies for a user based on the movie rankings provided by other users, 
and information retrieval where one is interested in combining the results of different search engines.

\begin{figure}[h!]
\begin{center}
\includegraphics[width=0.5\textwidth]{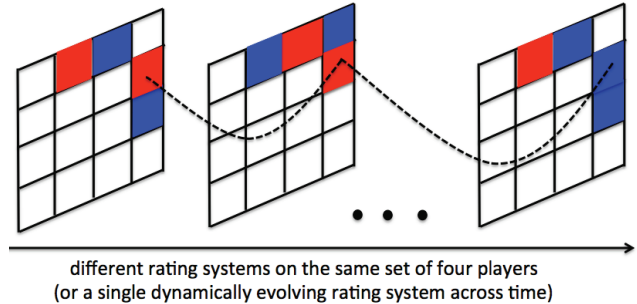}
\caption{ A multilayered network of pairwise rank comparisons. Each \text{slice} corresponds to a (possibly incomplete) pairwise comparison matrix provided by a different rating system. }
\label{fig:many_voters_demo_EPS}
\end{center}
\end{figure}

Mathematically speaking, the rank aggregation problem can be formulated as follows. Assume there exist $k$ judges, each of which makes available a noisy incomplete matrix $C^{(i)}, i=1,\ldots,k$ on the pairwise comparisons between a (perhaps randomly chosen) set of pairs of players, as illustrated for example in Figure \ref{fig:many_voters_demo_EPS}. A dotted line across different slices of the data denotes the fact that the measurements correspond to the same pair of players.
Note that in practice, it is not necessarily the case that the same set of players appears in each comparison matrix $C^{(i)}, i=1,\ldots,k$. However, for simplicity, we do assume this is the case. Furthermore, the pattern of pairwise existing and missing comparisons may be, and usually is, different across the above sequence of comparison matrices. We denote by  $\Theta^{(i)}, i=1,\ldots,k$ the resulting matrices of angle offsets after the mapping (\ref{transfToCircle}), and by  $H^{(i)}, i=1,\ldots,k$, the corresponding matrices after mapping to the complex plane (\ref{mapToCircle}).

\textbf{Naive approach A.} One possible naive approach to this problem would be to solve in parallel, via a method of choice, each of the $C^{(i)}, i=1,\ldots,k$ matrices, thus producing $k$ possible rankings, one for each rating system
\begin{equation}
 \hat{r}_1^{(i)}, \ldots, \hat{r}_n^{(i)}, i=1,\ldots,k.
\end{equation}
To obtain the final rank for each player $ t=1,\ldots,n$, one could then just average the $k$ ranks of player $t$ produced by each of the above $k$ rankings, sort the resulting scores in decreasing order, and consider the induced ranking
\begin{equation}
  \hat{r}_t^{(AVG)}  = \frac{ \hat{r}_t^{(1)} + \ldots, \hat{r}_t^{(k)}  }{k} , \forall t=1,\ldots,n
\label{rankAverage}
\end{equation}
In Figures \ref{fig:juryAllNaiveMeth}, \ref{fig:ErrorsJuryMUN} and \ref{fig:ErrorsJuryERO}, we denote by SVD-PAR, LS-PAR, SER-PAR, SYNC-PAR, SER-GLM-PAR, SYNC-SDP-PAR this approach for rank aggregation which runs, individually on each matrix $C^{(i)}, i=1,\ldots,k$  the various methods considered so far, and finally averages out the obtained rankings across all rankings proposed by a given method. 

\textbf{Naive approach B.}  Alternatively, another, albeit still naive, approach would be to first average out all the available matrices $C^{(i)}, i=1,\ldots,k$ into 
\begin{equation}
 \bar{C} = \frac{ C^{(1)} + C^{(2)} + \ldots + C^{(k)}}{k}
 \label{barCavg}
\end{equation}
and extract a final ranking by whatever preferred method that takes as input matrix $\bar{C}$, which we denote by $\bar{\Theta}$ after mapping to the circle. 
In Figures \ref{fig:juryAllNaiveMeth}, \ref{fig:ErrorsJuryMUN} and \ref{fig:ErrorsJuryERO}, we denote by SVD-AVG, LS-AVG, SER-AVG, SYNC-AVG, SER-GLM-AVG, SYNC-SDP-AVG the approach for rank aggregation which runs the various methods considered so far on the averaged $\bar{C}$ matrix. 

\begin{figure}[h!]
\label{fig:juryAllNaiveMeth}
\begin{center}
\subfigure[ SVD-PAR, $\kappa=0.32 $  ]{\includegraphics[width=0.16\textwidth]{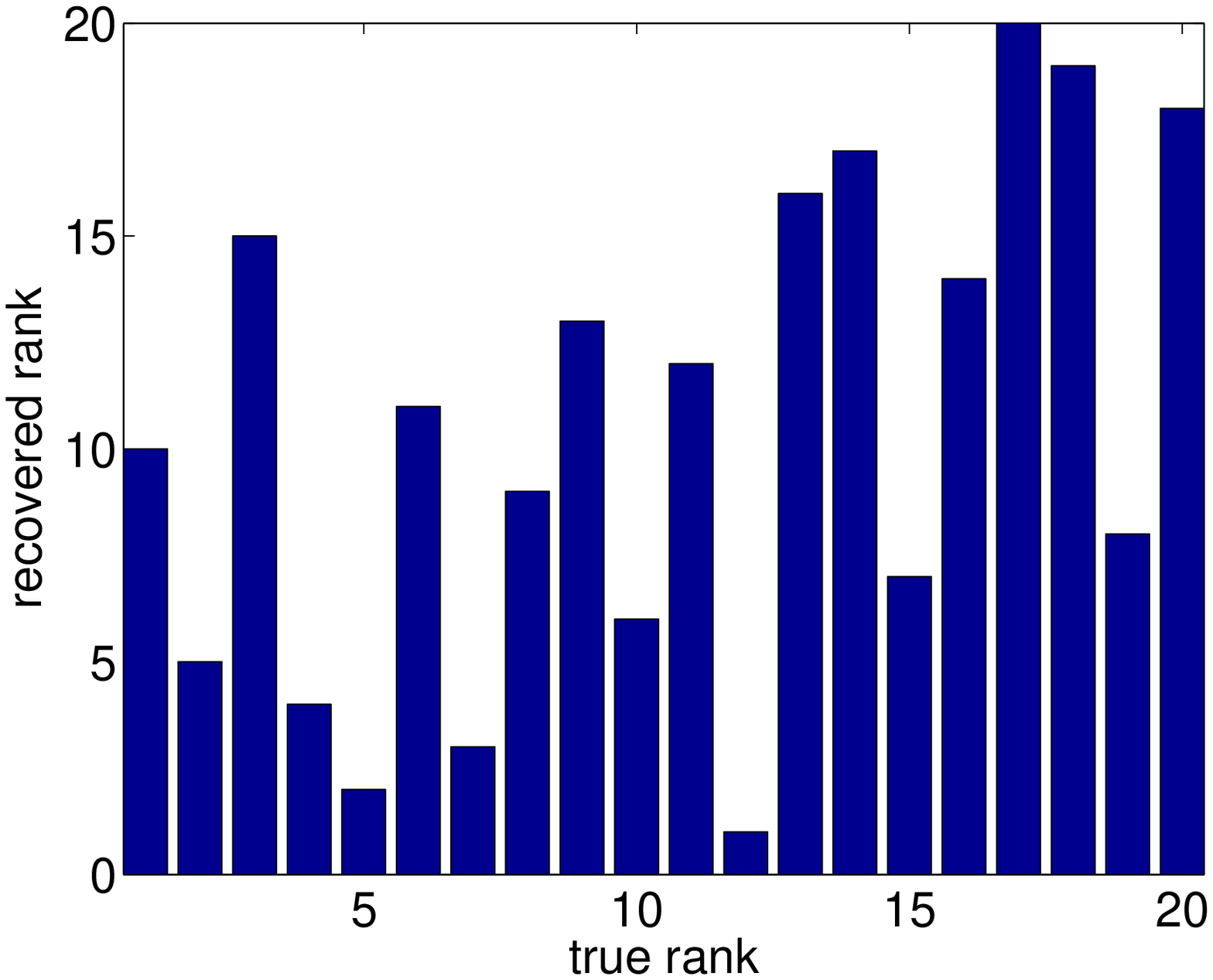}}
\subfigure[ LS-PAR, $\kappa=0.23 $  ]{\includegraphics[width=0.16\textwidth]{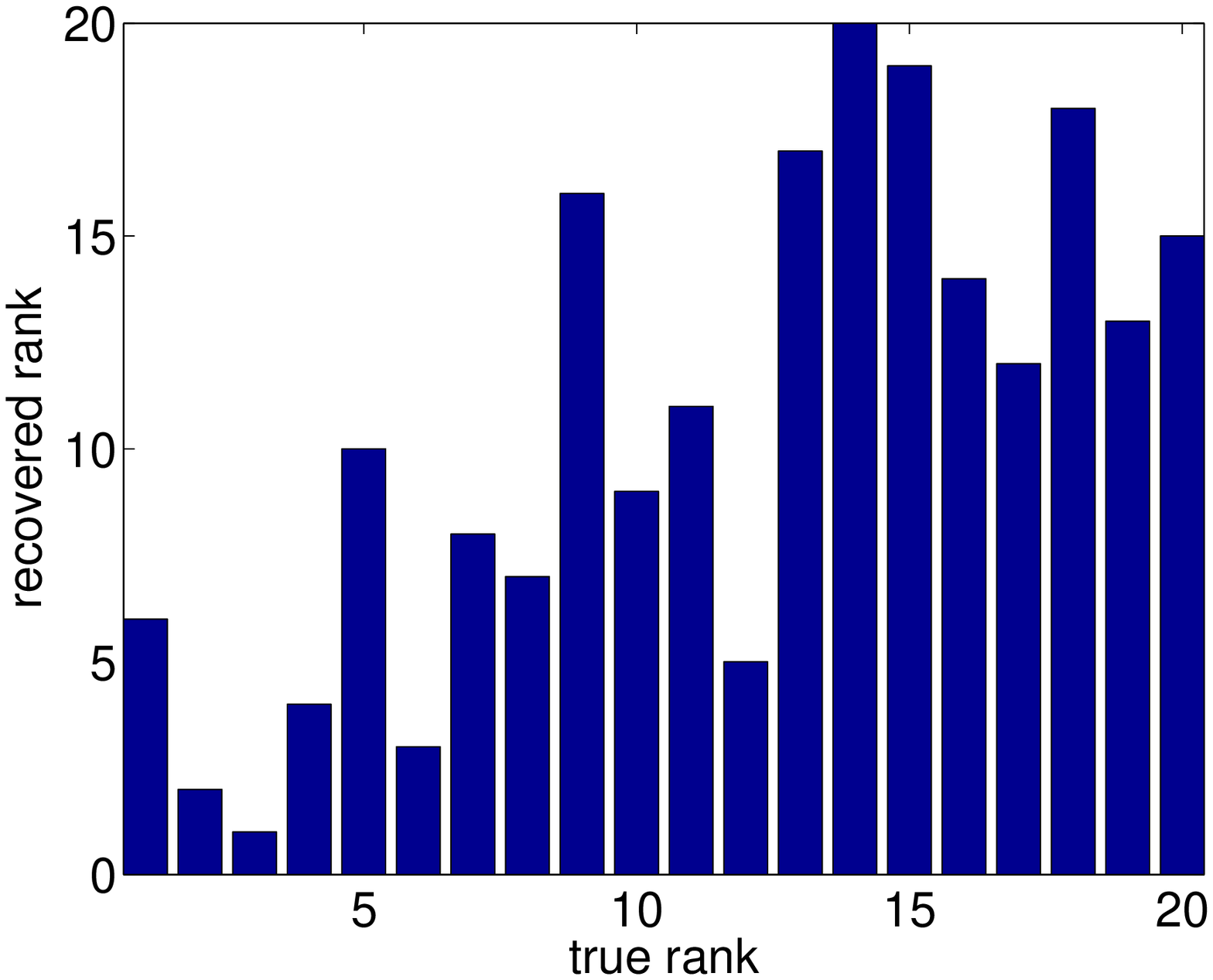}}
\subfigure[ SER-PAR, $\kappa= 0.42$  ]{\includegraphics[width=0.16\textwidth]{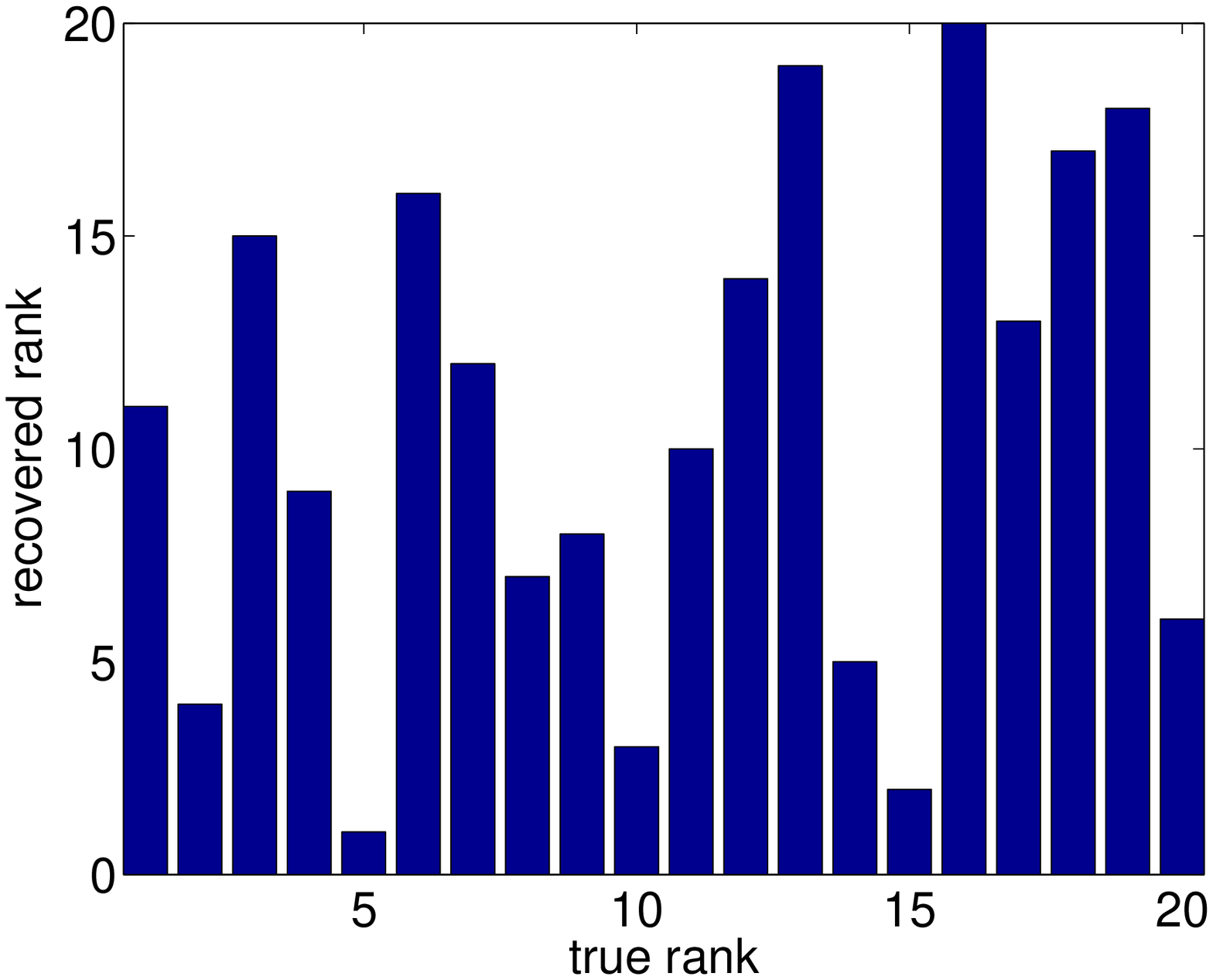}}
\subfigure[ SER-GLM-PAR, $\kappa= 0.22 $  ]{\includegraphics[width=0.16\textwidth]{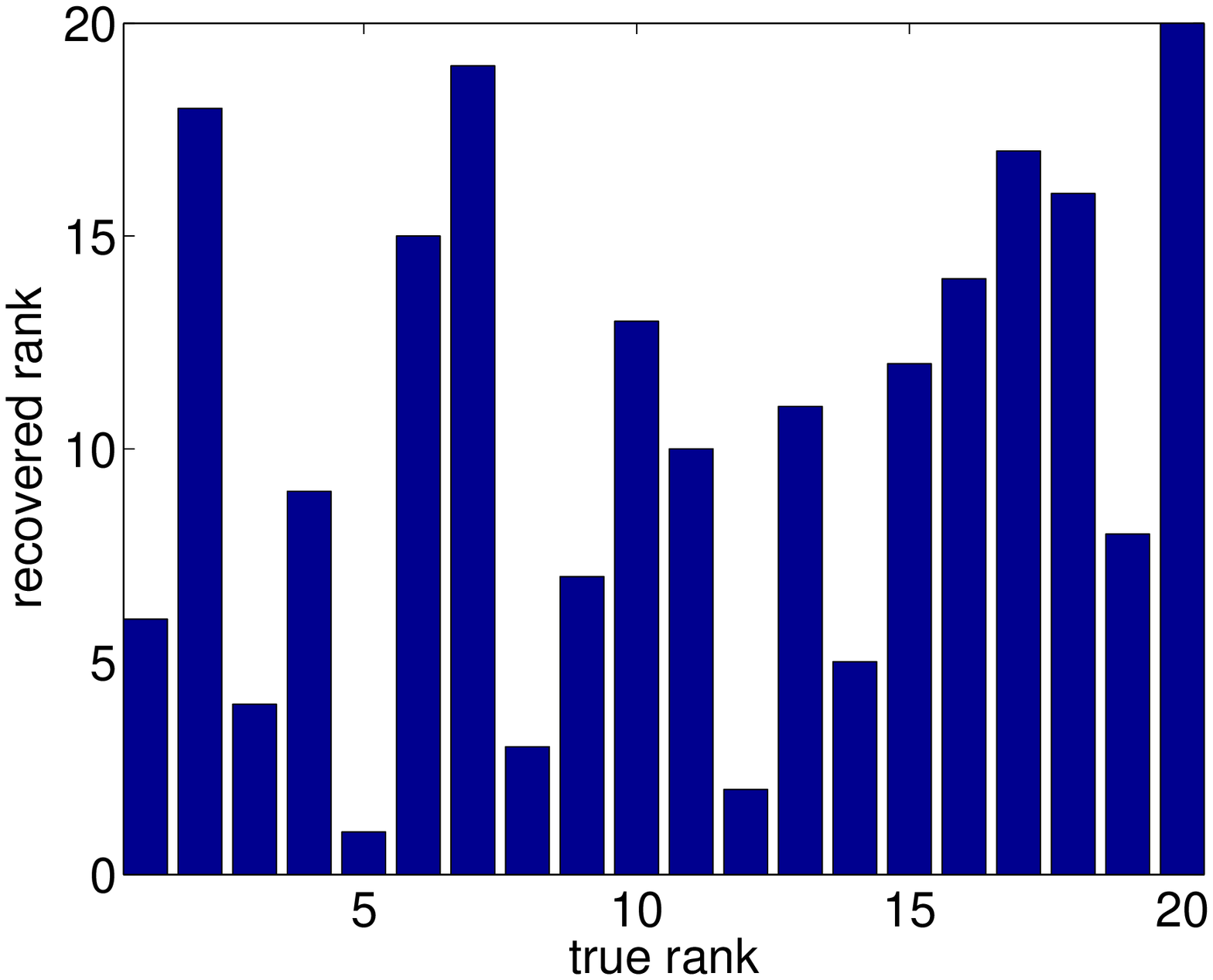}}
\subfigure[ SYNC-EIG-PAR, $\kappa= 0.37$  ]{\includegraphics[width=0.16\textwidth]{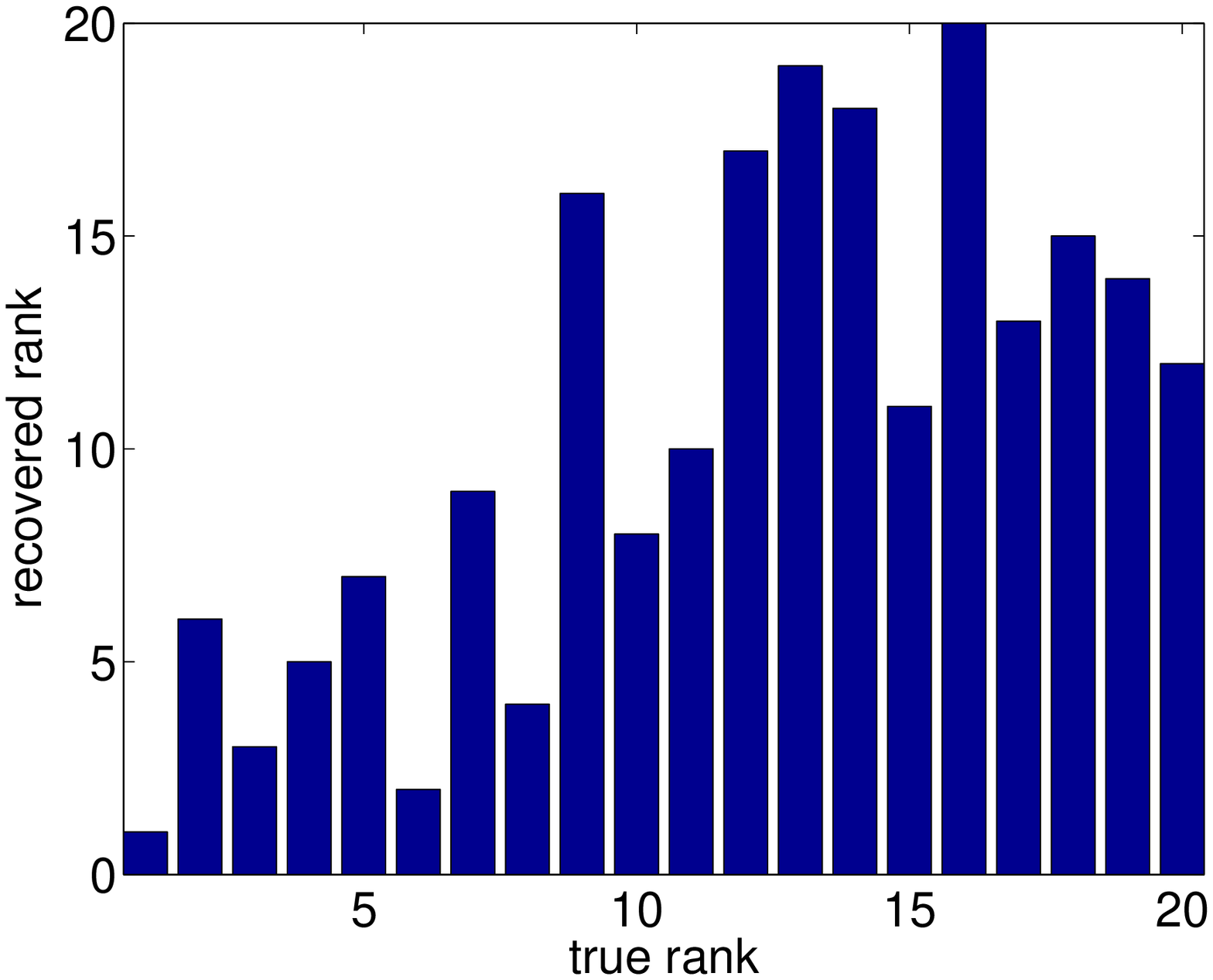}}
\subfigure[ SYNC-SDP-PAR, $\kappa=0.23 $  ]{\includegraphics[width=0.16\textwidth]{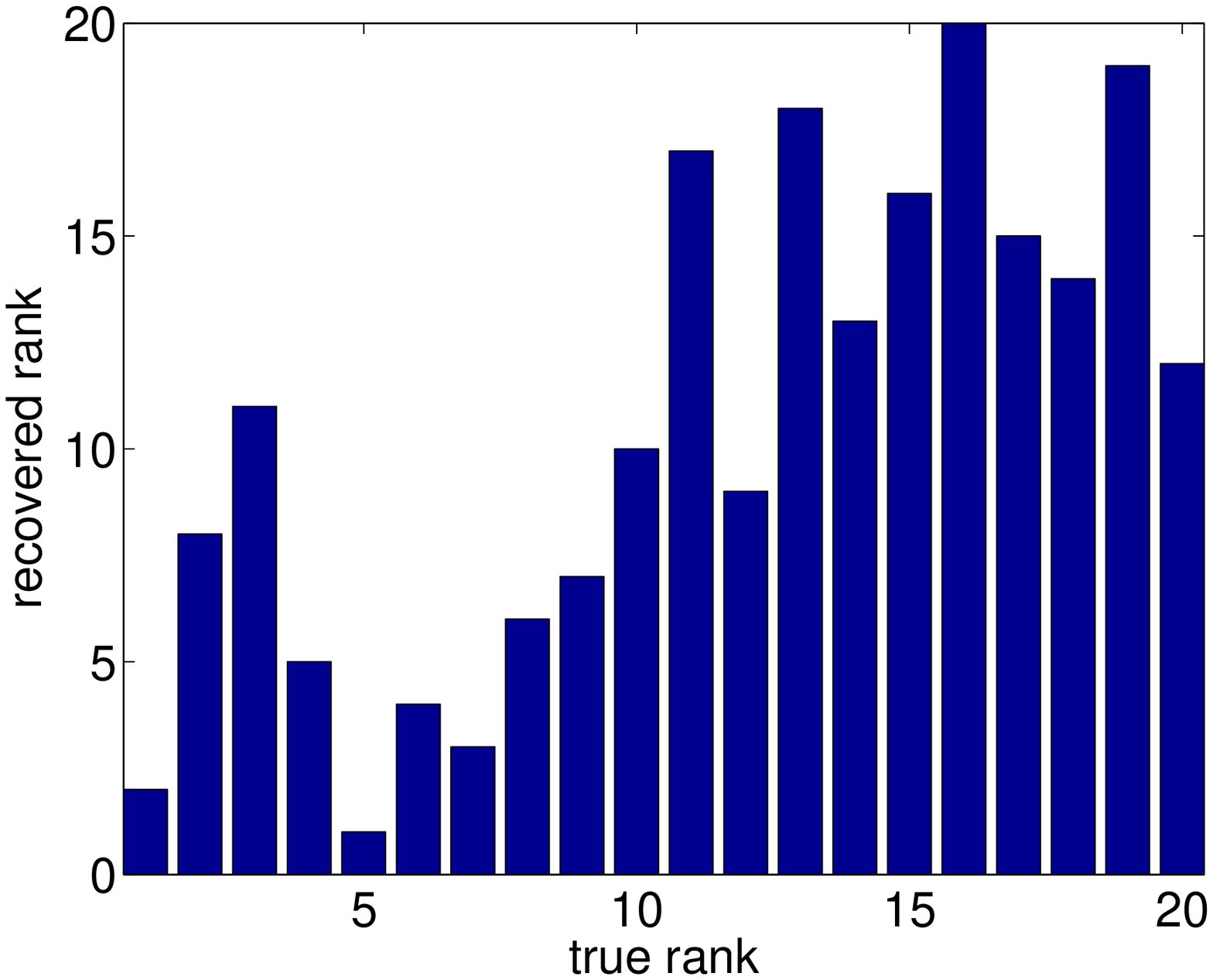}}
\subfigure[ SVD-AVG, $\kappa= 0.17$]{\includegraphics[width=0.16\textwidth]{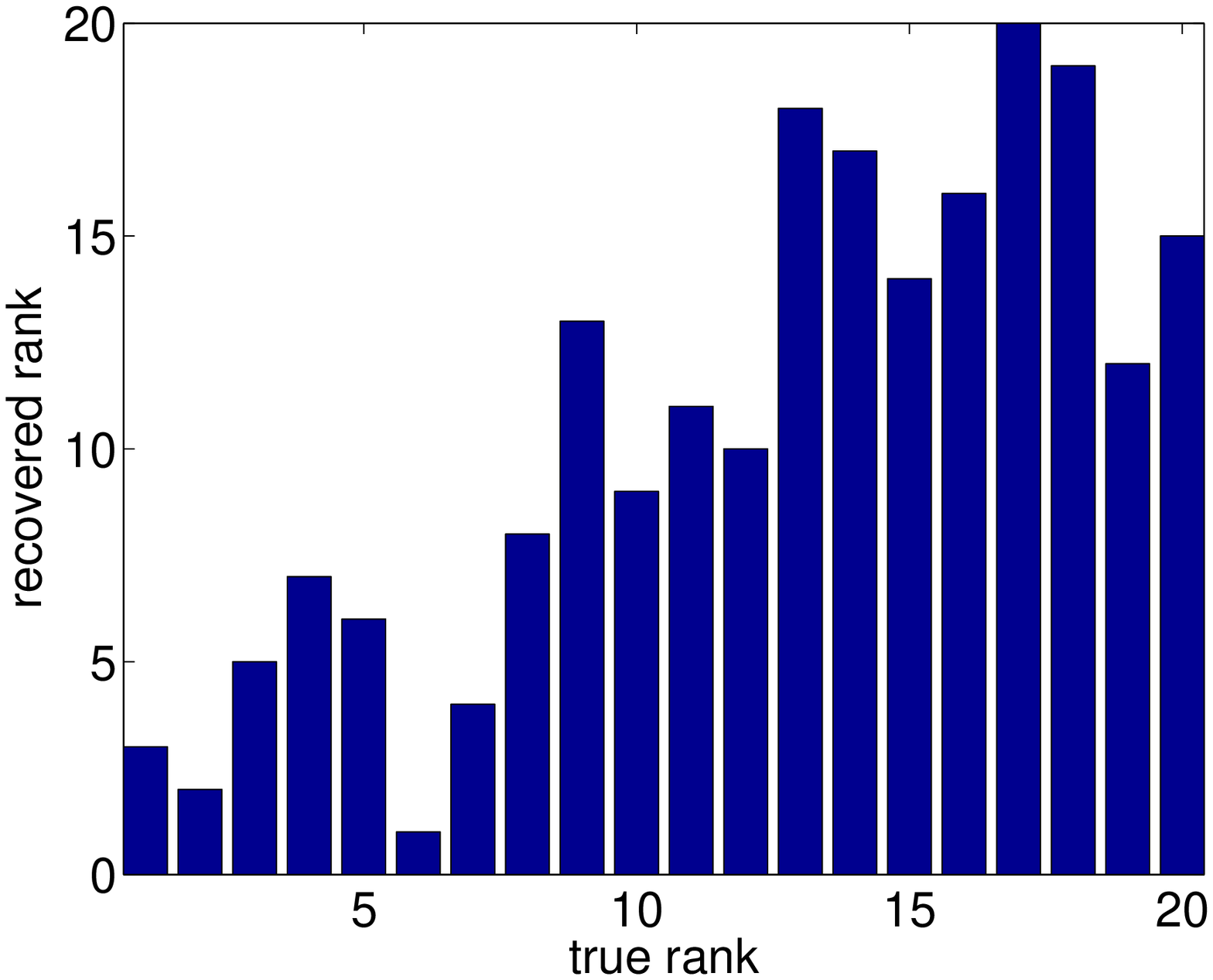}}
\subfigure[ LS-AVG, $\kappa= 0.16$ ]{\includegraphics[width=0.16\textwidth]{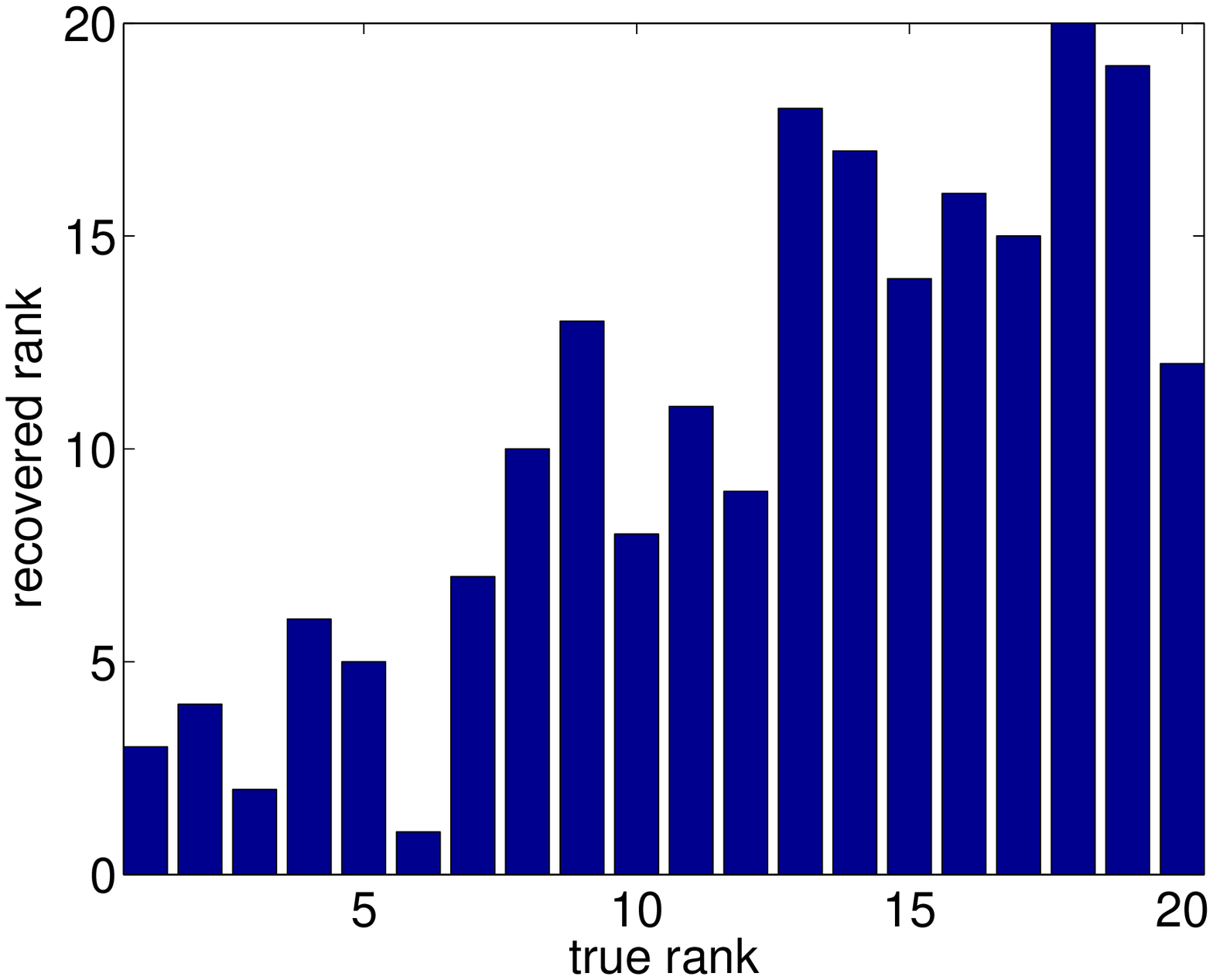}}
\subfigure[ SER-AVG, $\kappa= 0.18$ ]{\includegraphics[width=0.16\textwidth]{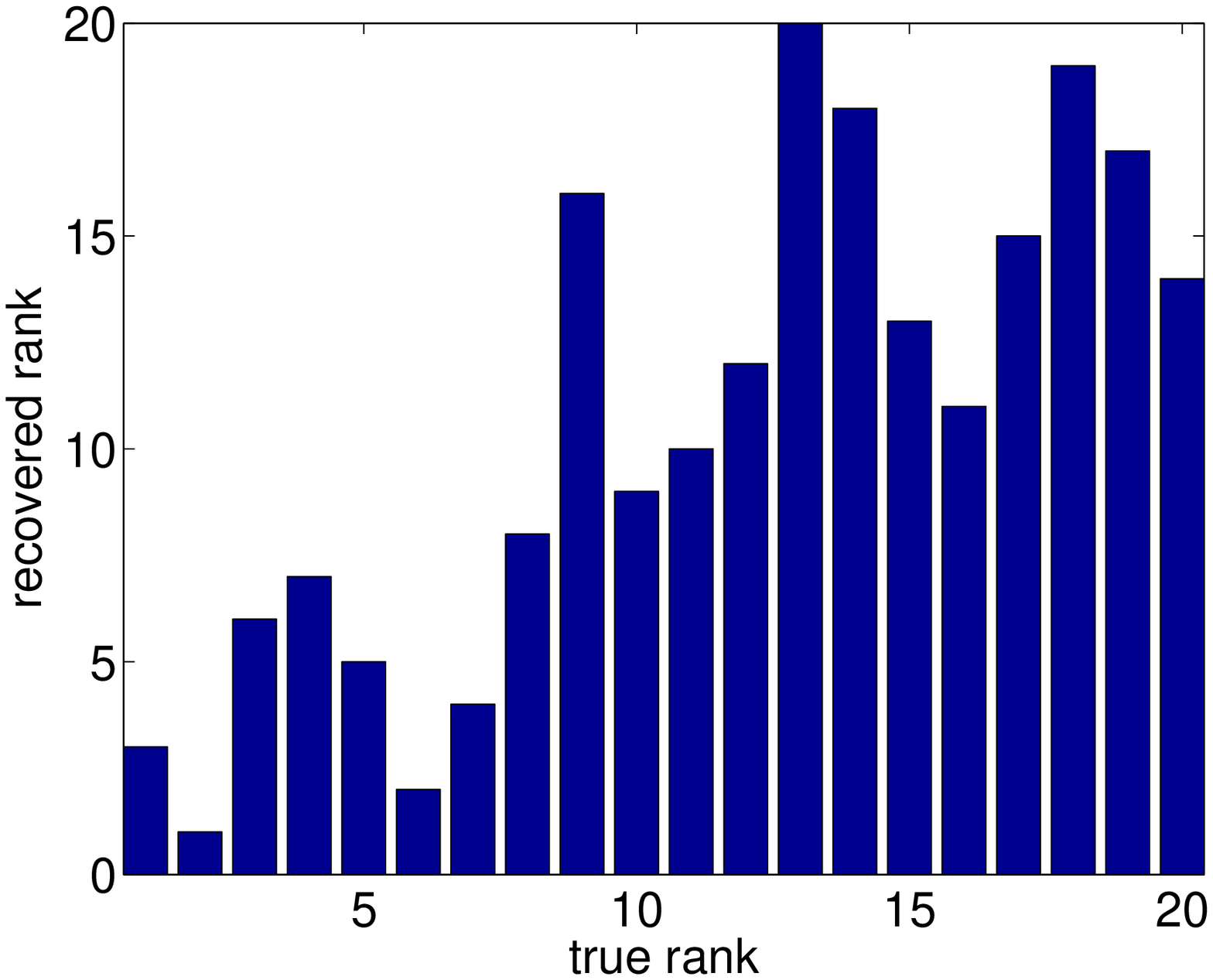}}
\subfigure[ SER-GLM-AVG, $\kappa= 0.31 $ ]{\includegraphics[width=0.16\textwidth]{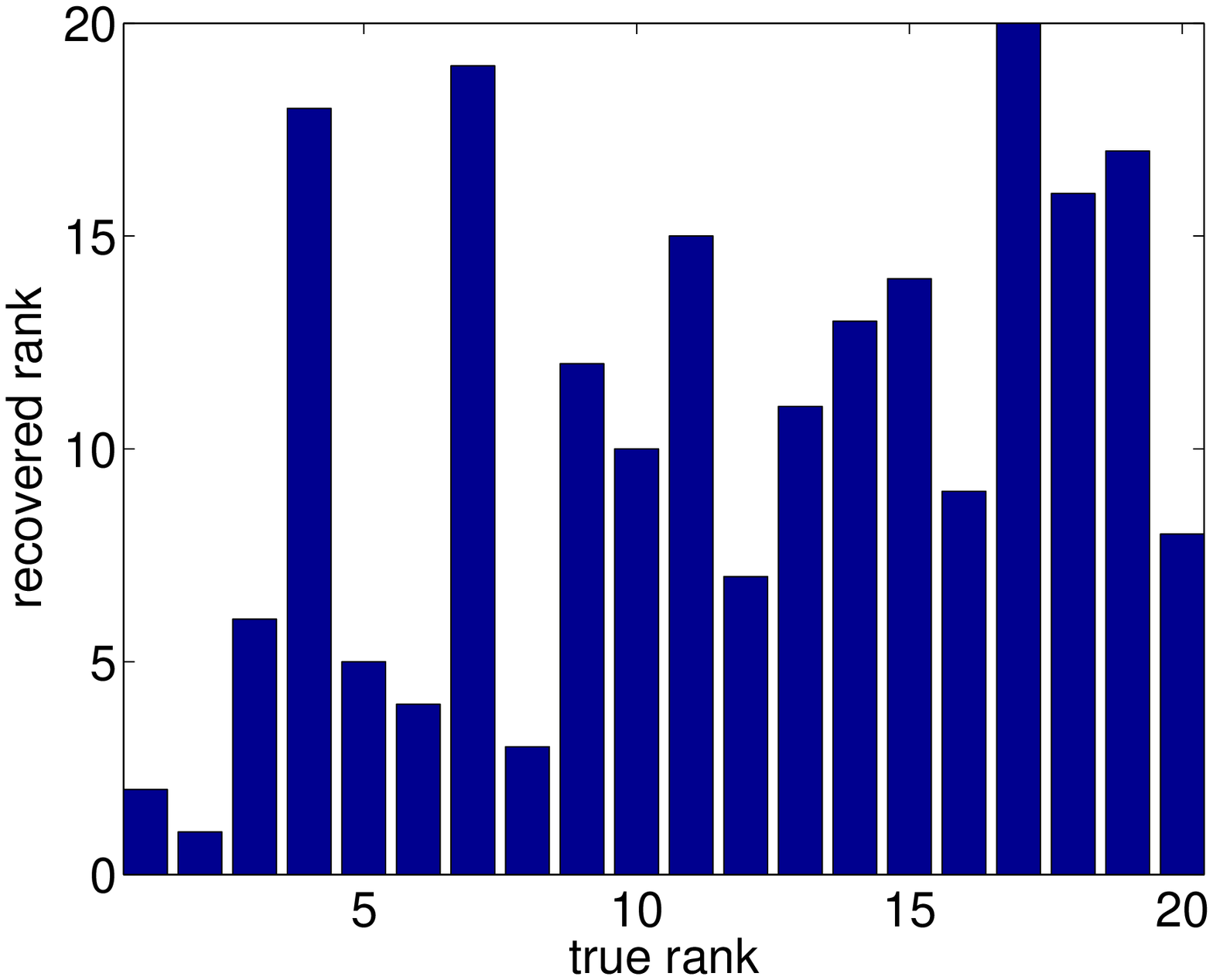}}
\subfigure[ SYNC-EIG-AVG, $\kappa=0.16 $ ]{\includegraphics[width=0.16\textwidth]{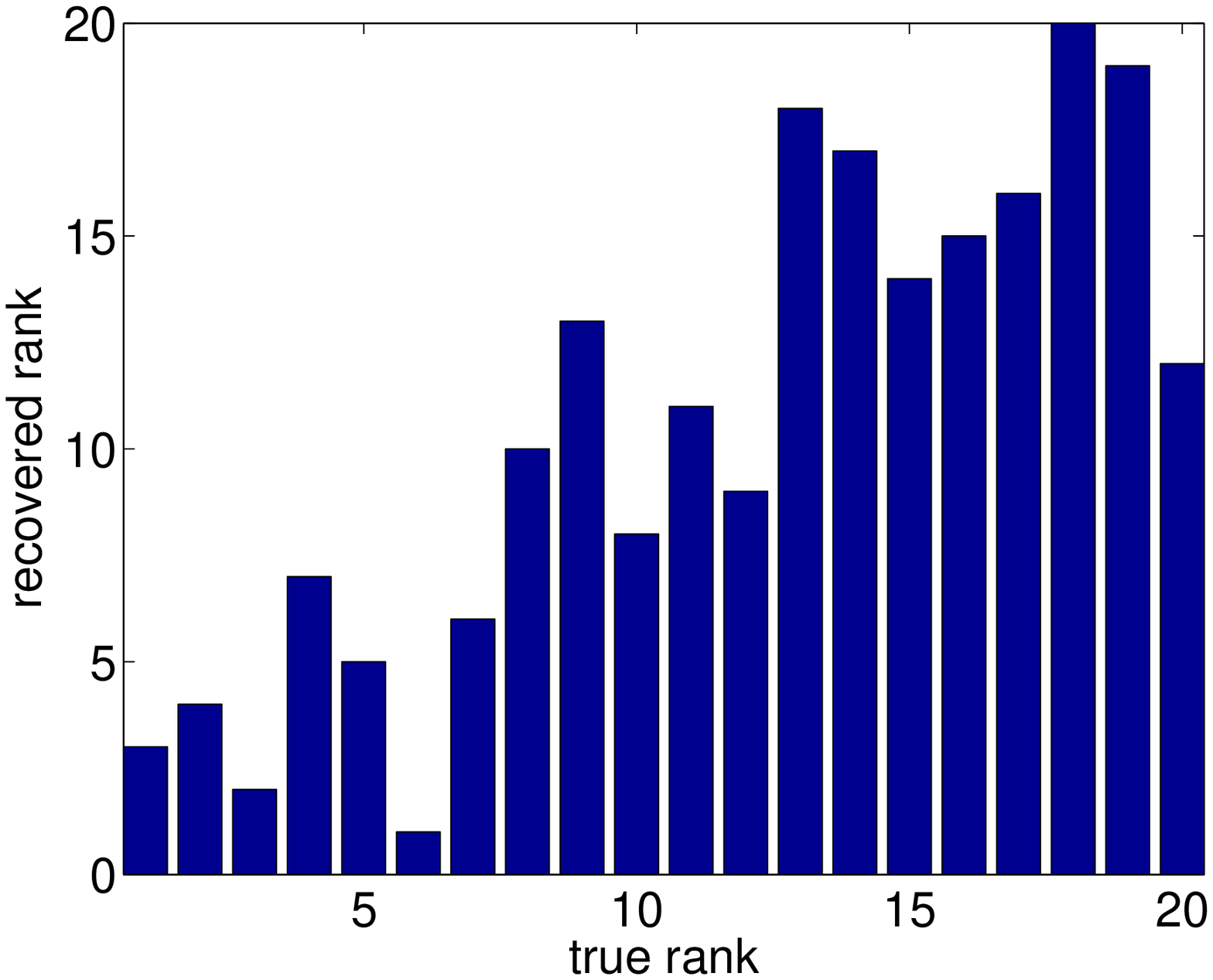}}
\subfigure[ SYNC-SDP-AVG, $\kappa= 0.17$ ]{\includegraphics[width=0.16\textwidth]{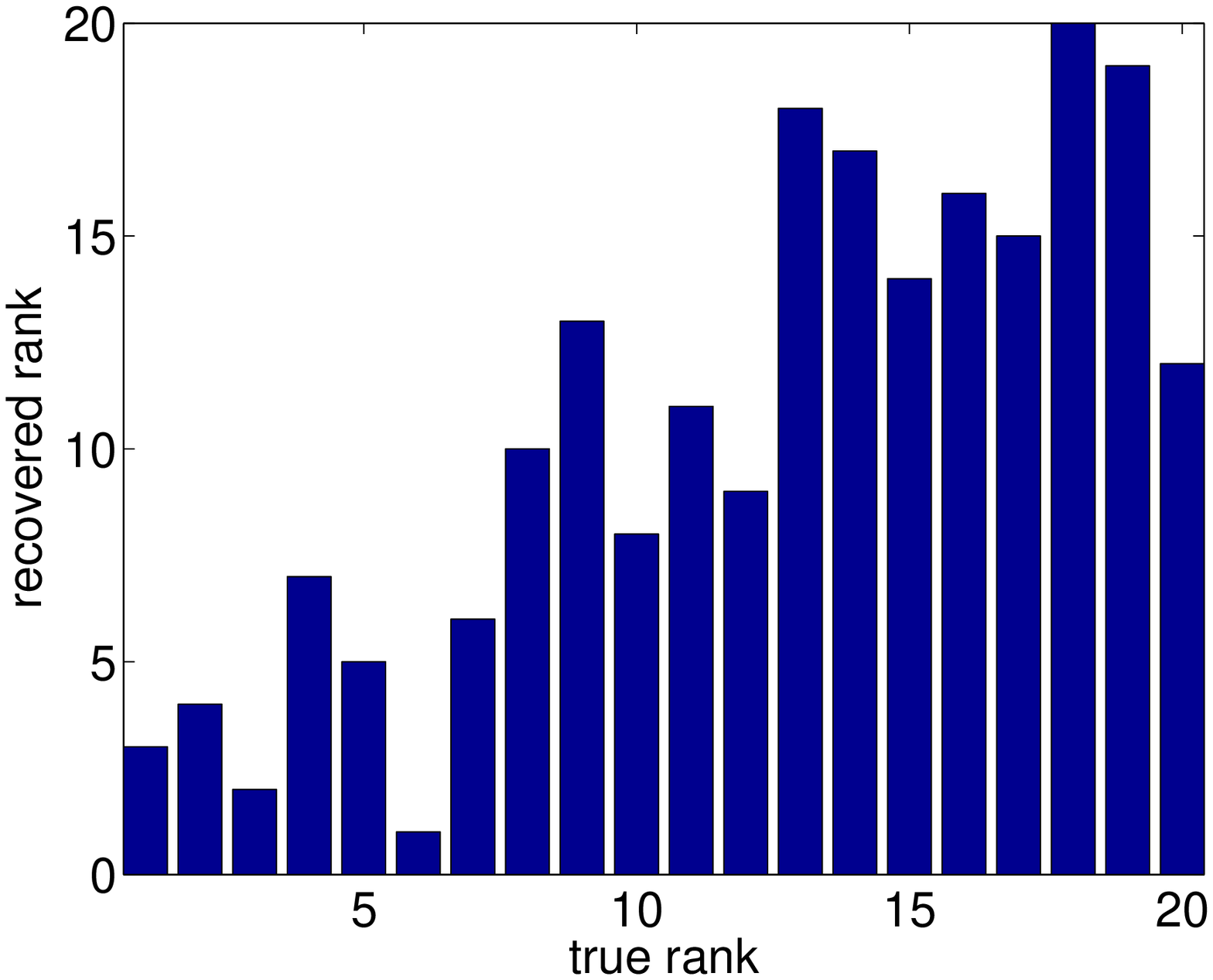}}
\caption{ Output of various rank aggregation methods for multiple jury ranking, for an instance of the rank aggregation problem, with cardinal measurements and outliers given by the ERO($n =20, p=0.2, \eta=0.5$) noise model, and a jury of size $k=5$. 
Top row (Naive Approach A): we run all ranking methods individually for each  matrix $\theta^{(i)}, i=1\ldots,5$, and combine the resulting rankings by a simple average as in (\ref{rankAverage}).
Bottom row (Naive Approach B): we first average the given $k$ measurement matrices into a single measurement matrix $\bar{\Theta}$ as in (\ref{barCavg}), and then we run all ranking methods using $\bar{\Theta}$ as input. 
The Kendal distance is denote by $\kappa$. Note that all proposed solutions are significantly worse than that of the SDP-AGG algorithm shown in Figure \ref{fig:PipelineJuryDemo} (m).}
\end{center}
\end{figure}

\subsection{Rank Aggregation via enlarged SDP-based synchronization}   \label{secsec:rankAggEnlarged}

A less naive approach to the above rank aggregation problem would be to consider the block diagonal matrix of size $N \times N$, with $N = nk$, given by 
\begin{equation}
	 \boldmath{C}_{N \times N} = \text{ \textbf{blkdiag}} \left(  C^{(1)}, C^{(2)}, \ldots, C^{(k)} \right),
\end{equation}
and its counterpart, the Hermitian matrix \text{\boldmath$H$}$_{N \times N}$ obtained after mapping to the circle.  Denote by $\mathcal{A}_i, i=1,\ldots,n$ the set of indices corresponding to player $i$ in \text{\boldmath$C$}. For example, $\mathcal{A}_1  =   \{1, n+1, 2n+1, \ldots, (k-1)n+1 \}$, and in general, $\mathcal{A}_i = \{i, n+i, 2n+i, \ldots, (k-1)n+i \}$. We denote by  \text{\boldmath$G$}  the graph  underlying the pattern of existing entries in  \text{\boldmath$C$}. 
Note that \text{\boldmath$G$} is disconnected and has exactly $k$ connected components.
We denote by \text{\boldmath$Q$} the multipartite graph of constraints, on the same set of nodes $V(\text{\boldmath$G$})$, where an edge connects the same player across different rating systems $C^{(i)}$, thus rendering the union \text{\boldmath$G$}$\cup$\text{\boldmath$Q$}  graph connected. 

The rank aggregation  problem can now be formulated as follows
\begin{equation}
	\begin{aligned}
	& \underset{  r_1, \ldots, r_N; \;\; r_i \in \{1,2,\ldots,n\}  }{\text{minimize}}
	& & \sum_{ij \in E( \text{\boldmath$G$} ) } | r_i - r_j - \text{\boldmath$C$}_{ij} |^2  &  \\
	& \text{subject to}
	&  &   r_i = r_j, \;\; \forall \; i,j \in \mathcal{A}_u,\;\; \forall u=1,\ldots,k.
	\end{aligned}
\label{setup_minimization_rankAgg}
\end{equation}

With the synchronization problem in mind, in the angular embedding space, the above corresponds to 
\begin{equation}
	\begin{aligned}
	& \underset{  \theta_1, \ldots, \theta_N; \;\; \theta_i \in [0,2\pi) }{\text{maximize}}
	& & \sum_{ij \in E( \text{\boldmath$G$} ) }  e^{-\iota \theta_i}  \; \text{\boldmath$H$}_{ij} \; e^{\iota \theta_j}  &  \\
	& \text{subject to}
	&  &   \theta_i = \theta_j, \;\; \forall \; i,j \in \mathcal{A}_u,\;\; \forall u=1,\ldots,k.
	\end{aligned}
\label{sync_minimization_rankAgg}
\end{equation}
whose relaxation is given by 
\begin{equation}
	\begin{aligned}
	& \underset{  z =[z_1, \ldots, z_N]: \; z_i \in  \mathbb{C}, \;\;  || z ||^2 = N }{\text{maximize}}
	& &   z^* \; \text{\boldmath$H$} \; z  &  \\
	& \text{subject to}
	&  &   z_i = z_j, \;\; \forall \; i,j \in \mathcal{A}_u,\;\; \forall u=1,\ldots,k.
	\end{aligned}
\label{syncComplex_minimization_rankAgg} 
\end{equation}
Without the constraints, one could use the eigenvector synchronization method to solve the above relaxation, however the additional constraints that a subset of the elements must correspond to the same unknown element render this approach no longer feasible. 
Note that the case $k=1$ corresponds to the usual synchronization problem when no additional information is available. Unfortunately, since the above constrained quadratic program (\ref{syncComplex_minimization_rankAgg}) can no longer be cast as an eigenvector problem, we consider again the same relaxation via semidefinite programming and simply encode the additional information as hard constraints in the SDP
\begin{equation}
	\begin{aligned}
	& \underset{\Upsilon \in \mathbb{C}^{N \times N}}{\text{maximize}}
	& & Trace( \text{\boldmath$H$}  \Upsilon) \\
	& \text{subject to}
	& & \Upsilon_{ij} = 1 & \;\; \text{ if }  i,j \in \mathcal{A}_u, \;\; u=1,\ldots,k \\
		& & &   \Upsilon \succeq 0,
	\end{aligned}
 \label{SDP_program_rank_agg}
\end{equation}
where the maximization is taken over all semidefinite positive complex-valued matrices  $\Upsilon \succeq 0$ of size $N \times N$. Since $\Upsilon$ is not necessarily a rank-one matrix, the SDP-based estimator is given by the best rank-one approximation to  $\Upsilon$, which can be computed via an eigen-decomposition. Note that after the projection, the enforced constraints may no longer be necessarily perfectly satisfied in the case of very noisy data, which may result in an SDP solution of rank higher than 1. In practice, we consider the top eigenvector $ v_1^{\Upsilon}$ corresponding to the largest eigenvalue of  $\Upsilon$, which, in the noisy setting, is not necessarily piecewise constant. Thus, for each set $\mathcal{A}_i, i=1,\ldots,n$, we average the real and complex part of the entries of $ v_1^{\Upsilon}$ whose support is given by $\mathcal{A}_i$, and denote the resulting vector of size $n$ by $ \bar{v}_1^{\Upsilon} $ whose entries are given by
\begin{equation}
\bar{v}_1^{\Upsilon}(i) = \frac{1}{k} \sum_{t \in \mathcal{A}_i} Real(v_1^{\Upsilon}(t) )  + \iota \frac{1}{k} \sum_{t \in \mathcal{A}_i} Imag( v_1^{\Upsilon}(t) ), \;\;\;\;\;\; i = 1,\ldots,n
\label{avgComplexEigvect}
\end{equation}
Finally, we extract the corresponding angles from $ \bar{v}_1^{\Upsilon} $, and consider the induced ranking of the $n$ players, after accounting for the best circular permutation that minimizes the number of upsets in the given data measurements. 
\begin{figure}[h!]
\begin{center}
\subfigure[ Ground truth ranking]{\includegraphics[width=0.24\textwidth]{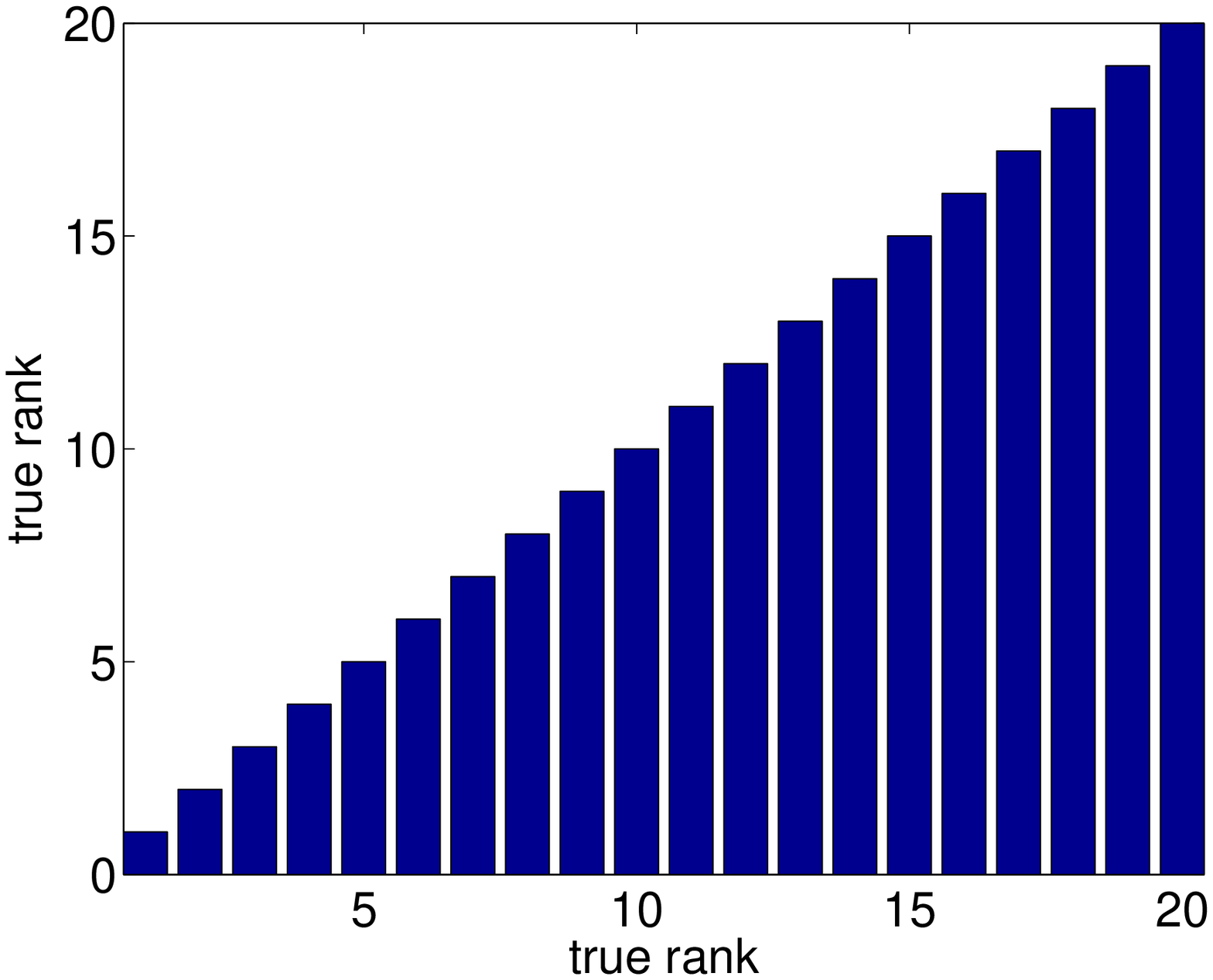}}
\subfigure[ Ground truth rank-offset matrix ]{\includegraphics[width=0.24\textwidth]{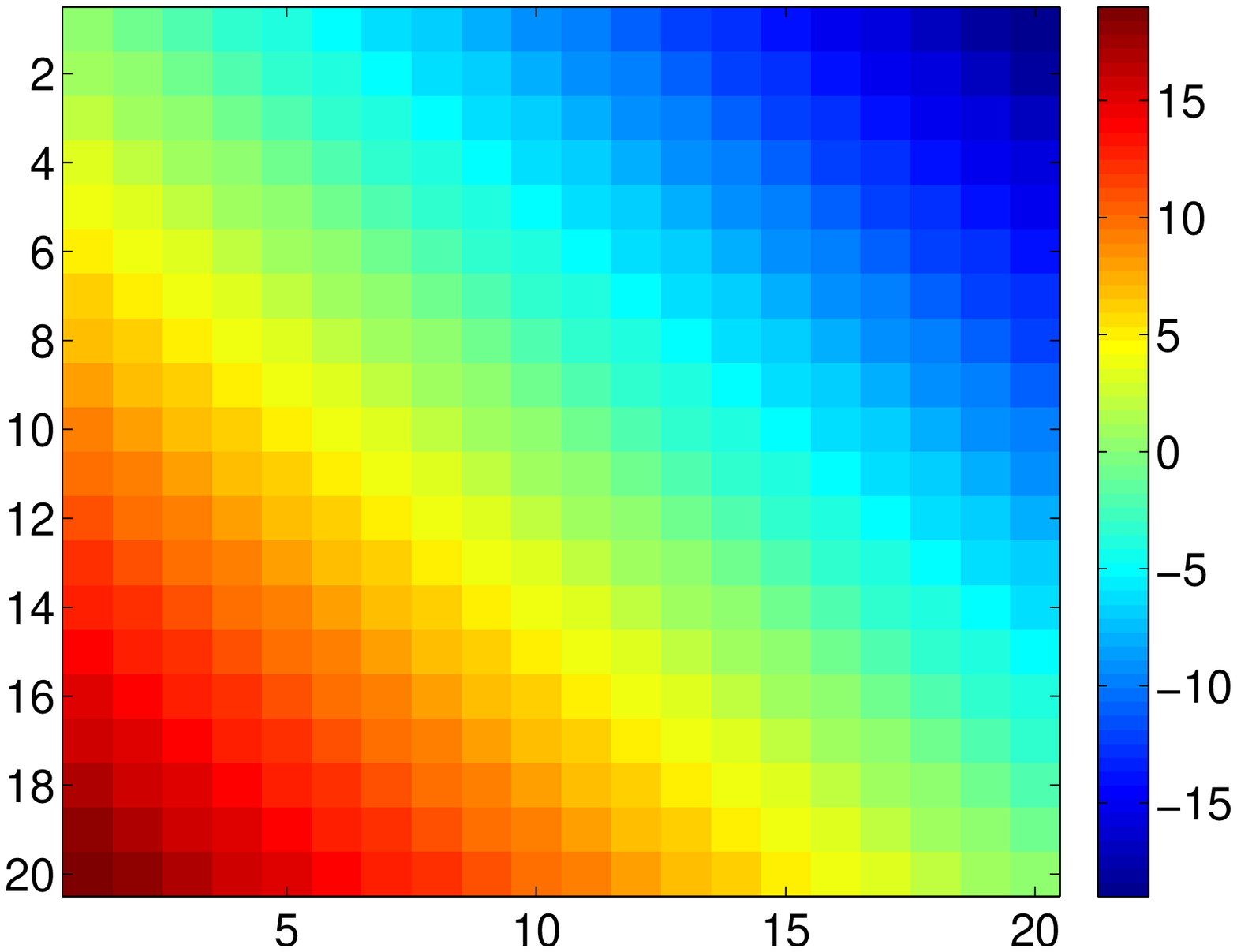}}
\subfigure[ Shuffled ground truth rank-offset matrix ]{\includegraphics[width=0.24\textwidth]{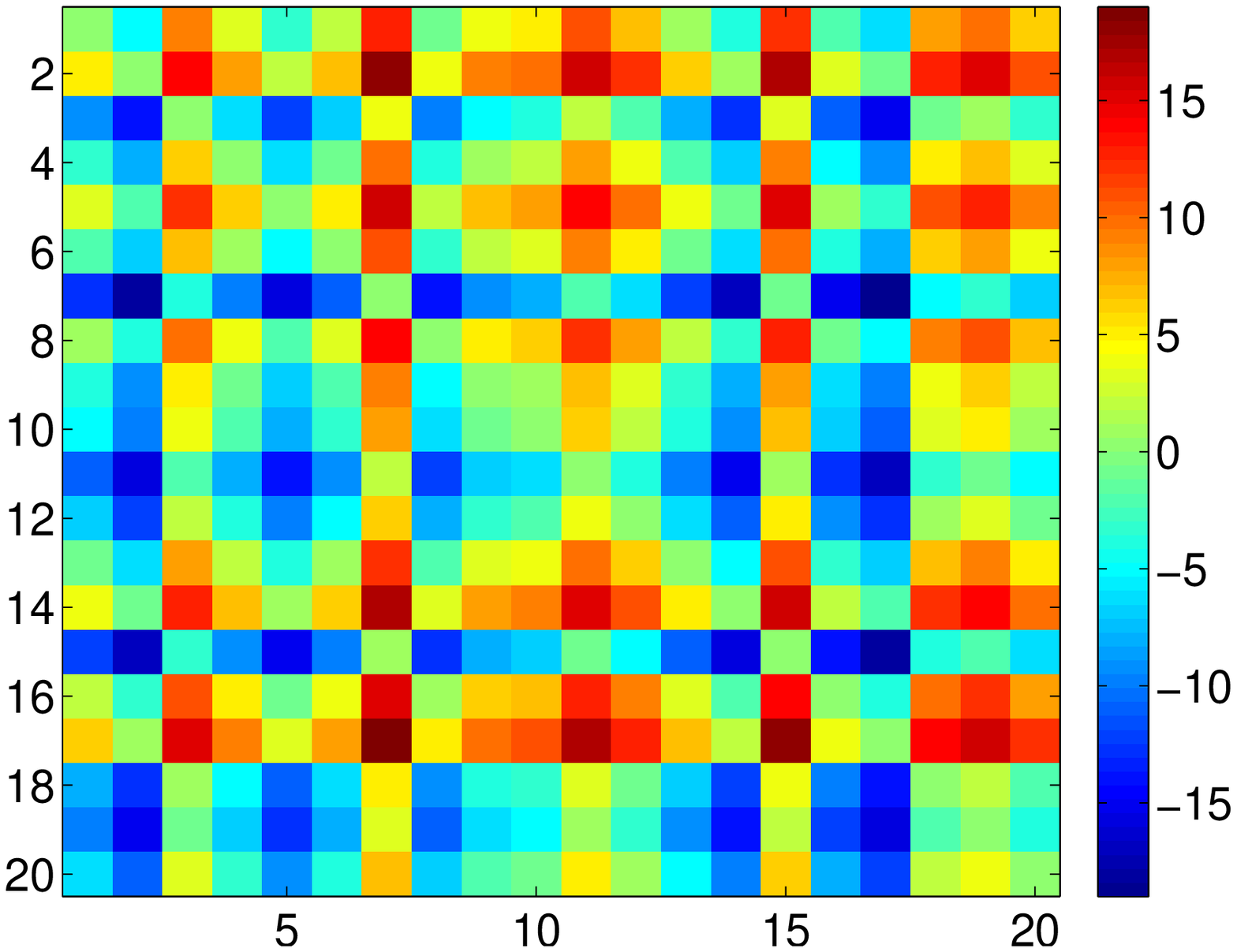}} 
\subfigure[ \textbf{diag}($\theta^{(1)},\ldots,\theta^{(5)}$) ]{\includegraphics[width=0.24\textwidth]{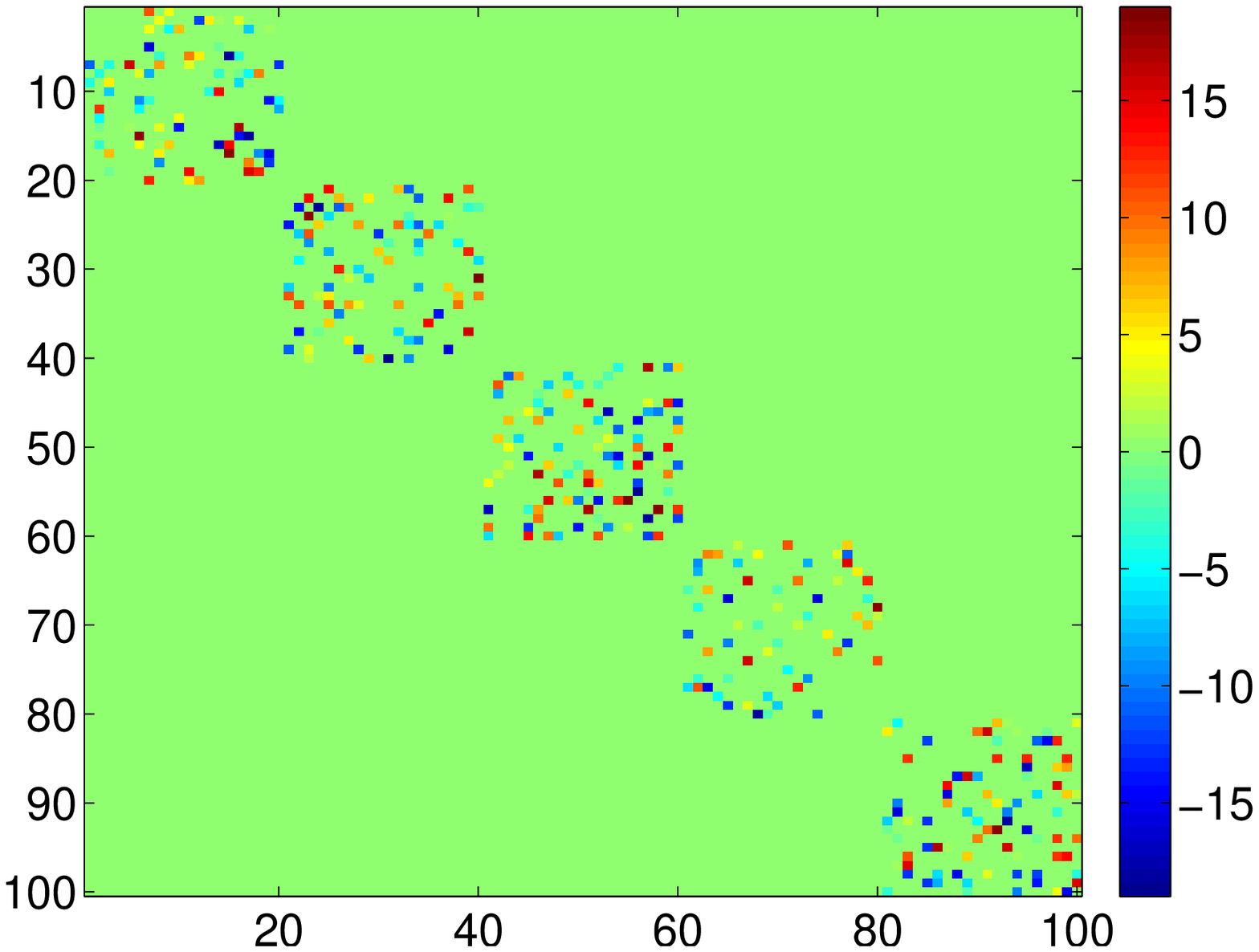}} \\
\subfigure[ $\theta^{(1)}$ ]{\includegraphics[width=0.18\textwidth]{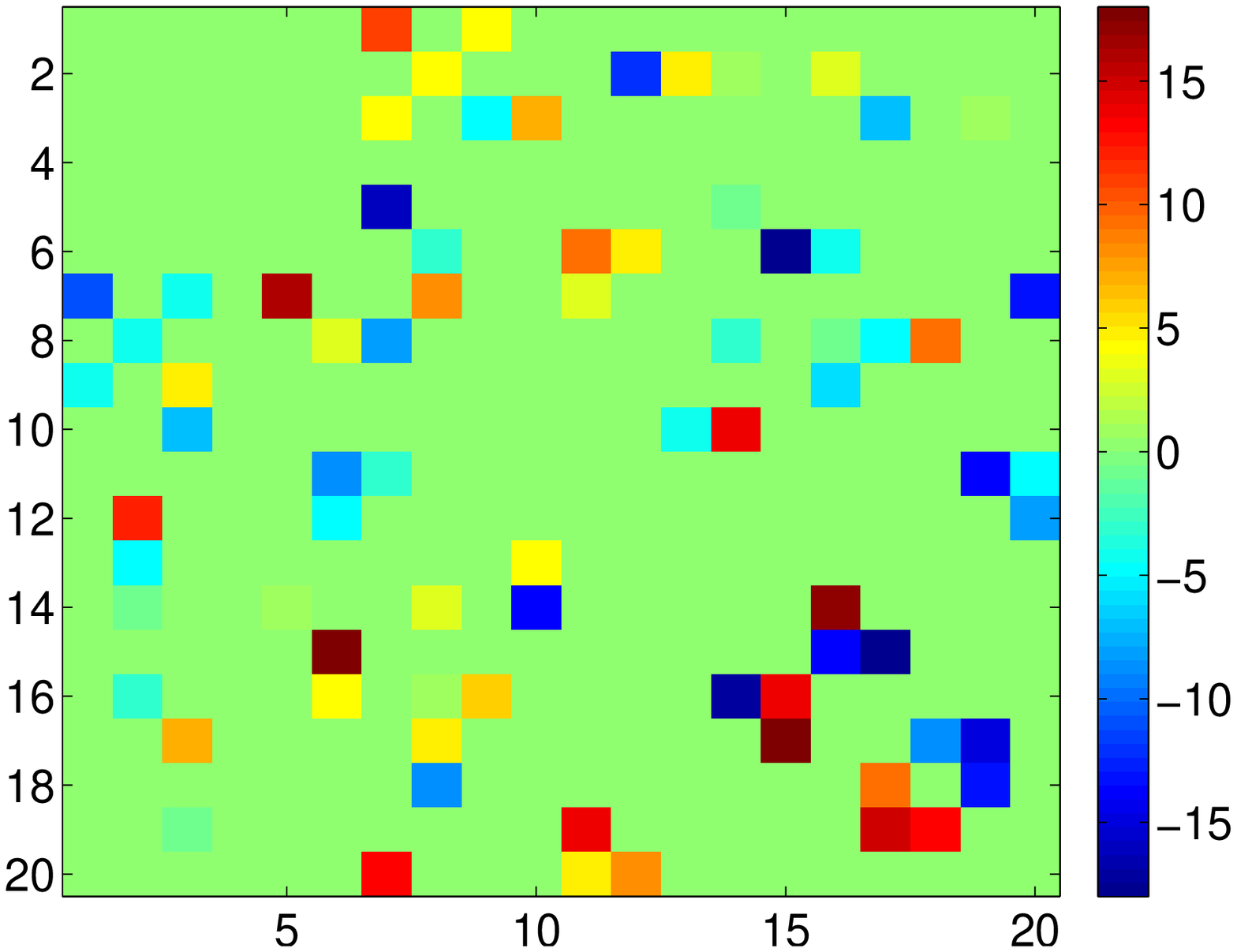}}
\subfigure[ $\theta^{(2)}$ ]{\includegraphics[width=0.18\textwidth]{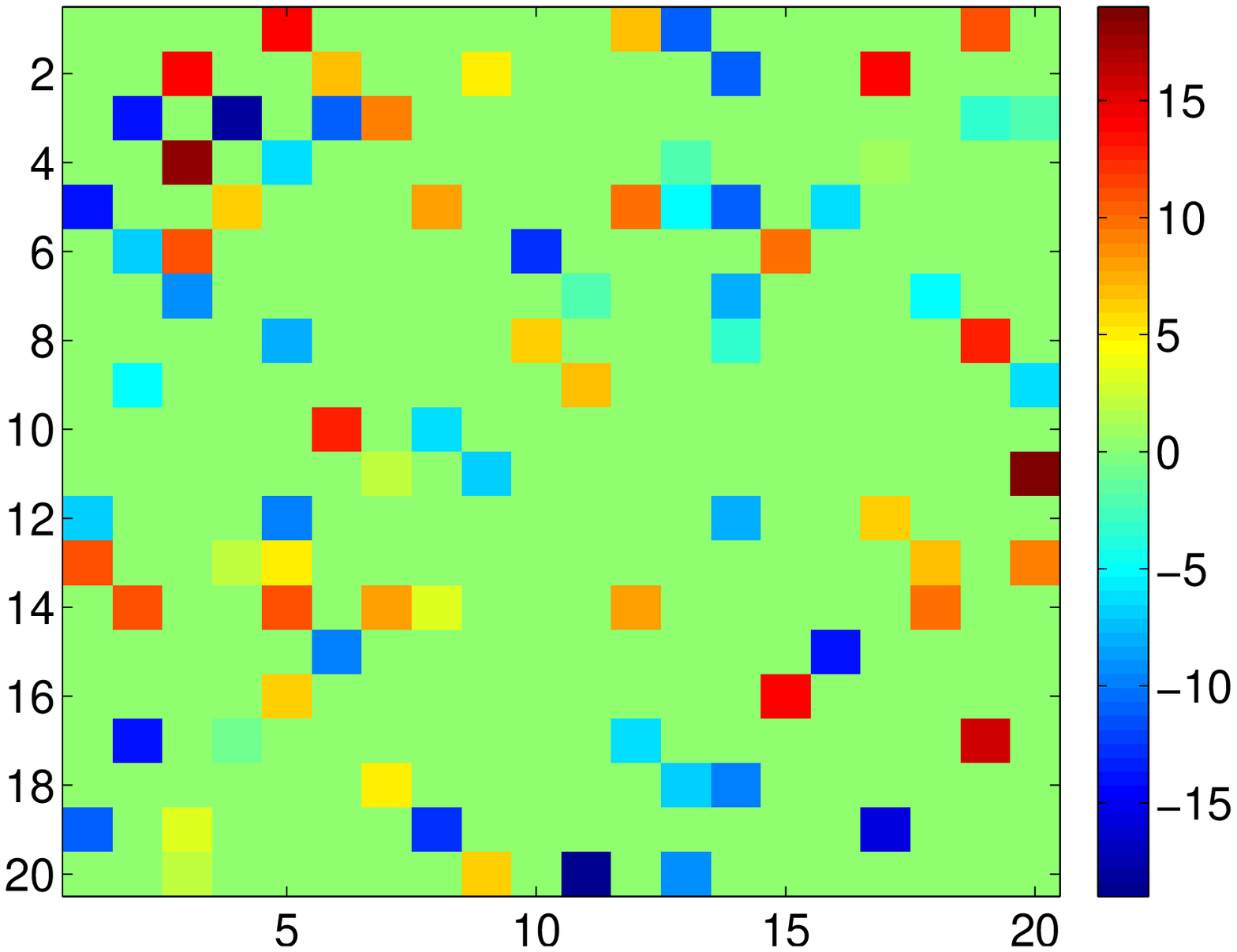}}
\subfigure[ $\theta^{(3)}$ ]{\includegraphics[width=0.18\textwidth]{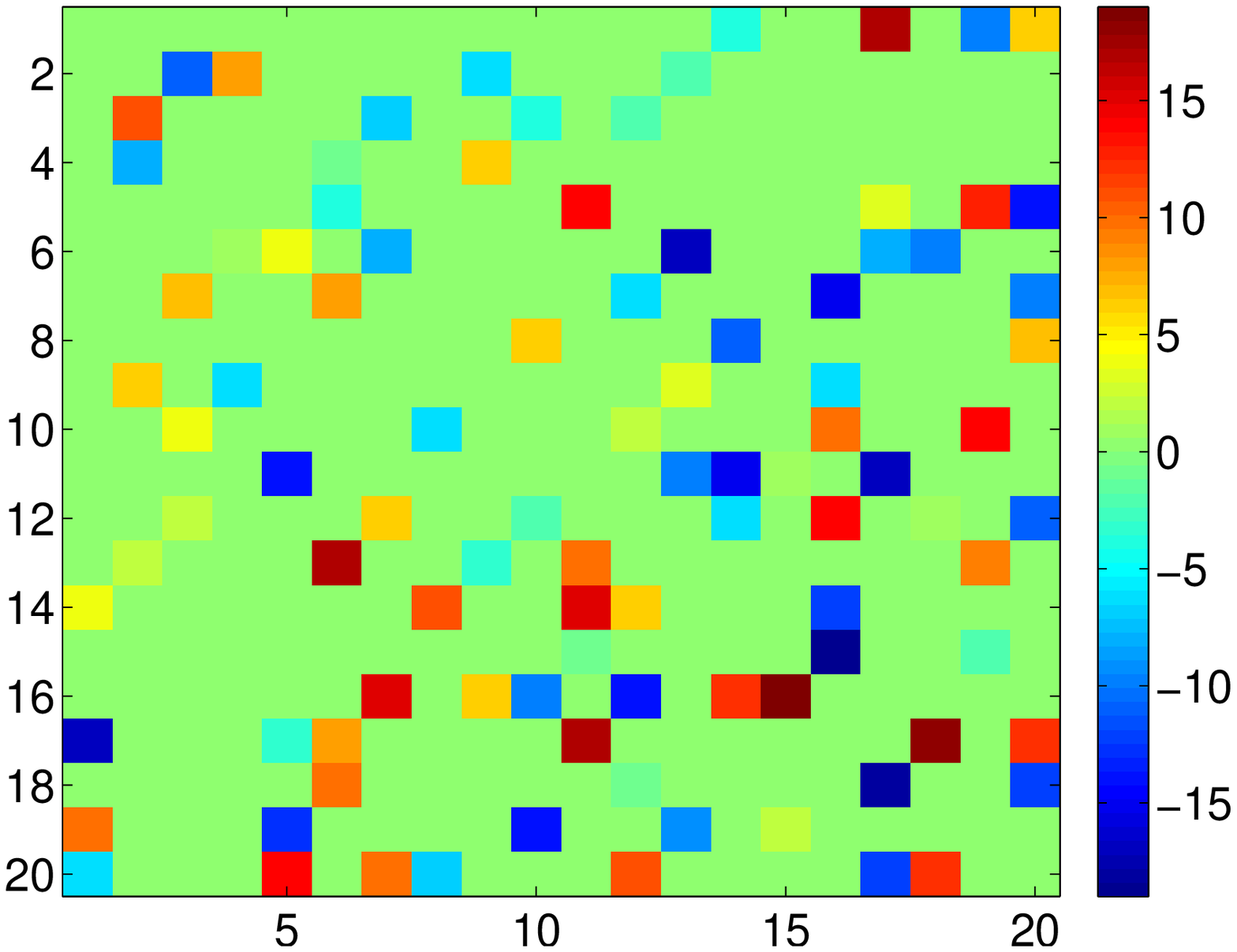}}
\subfigure[ $\theta^{(4)}$ ]{\includegraphics[width=0.18\textwidth]{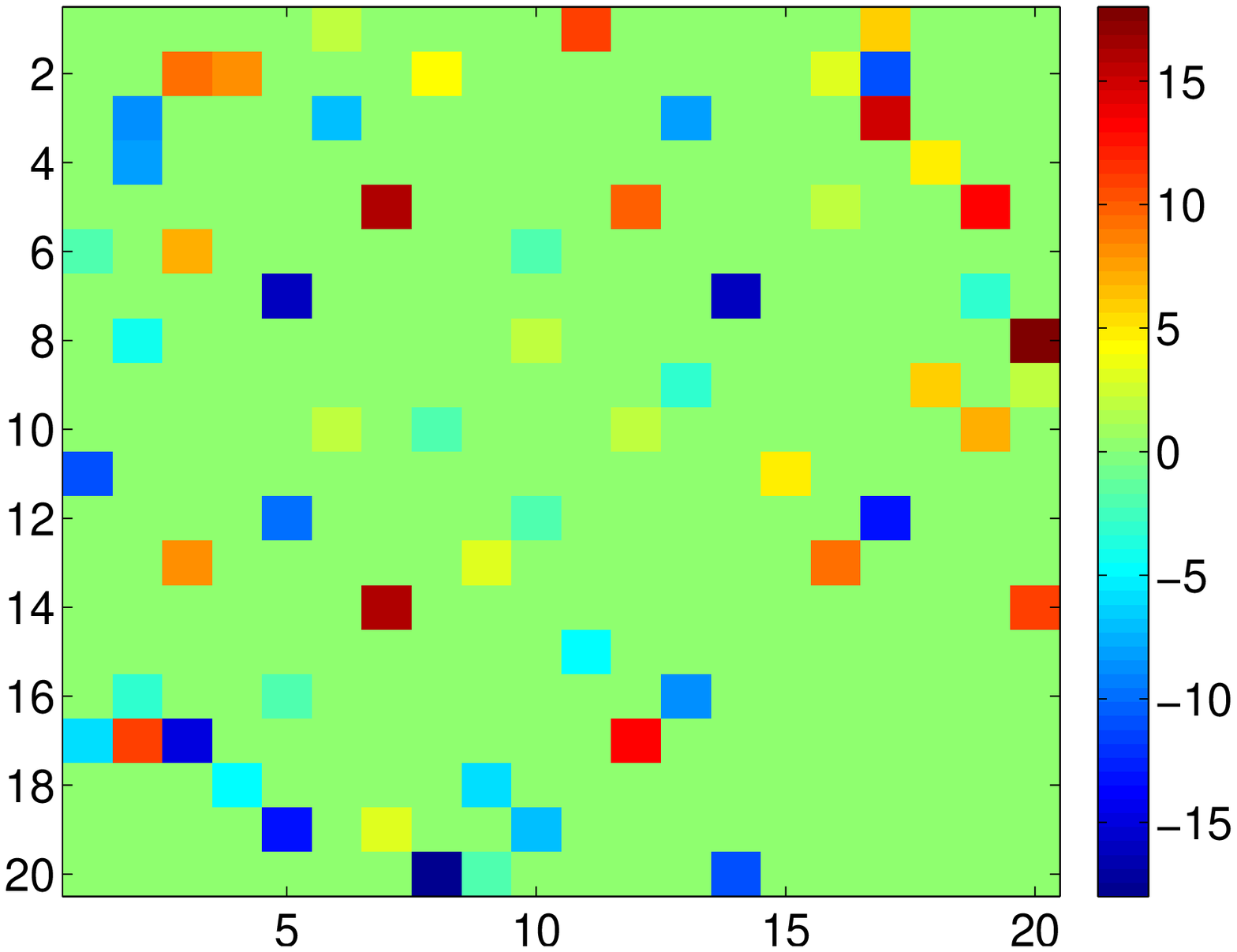}}
\subfigure[ $\theta^{(5)}$ ]{\includegraphics[width=0.18\textwidth]{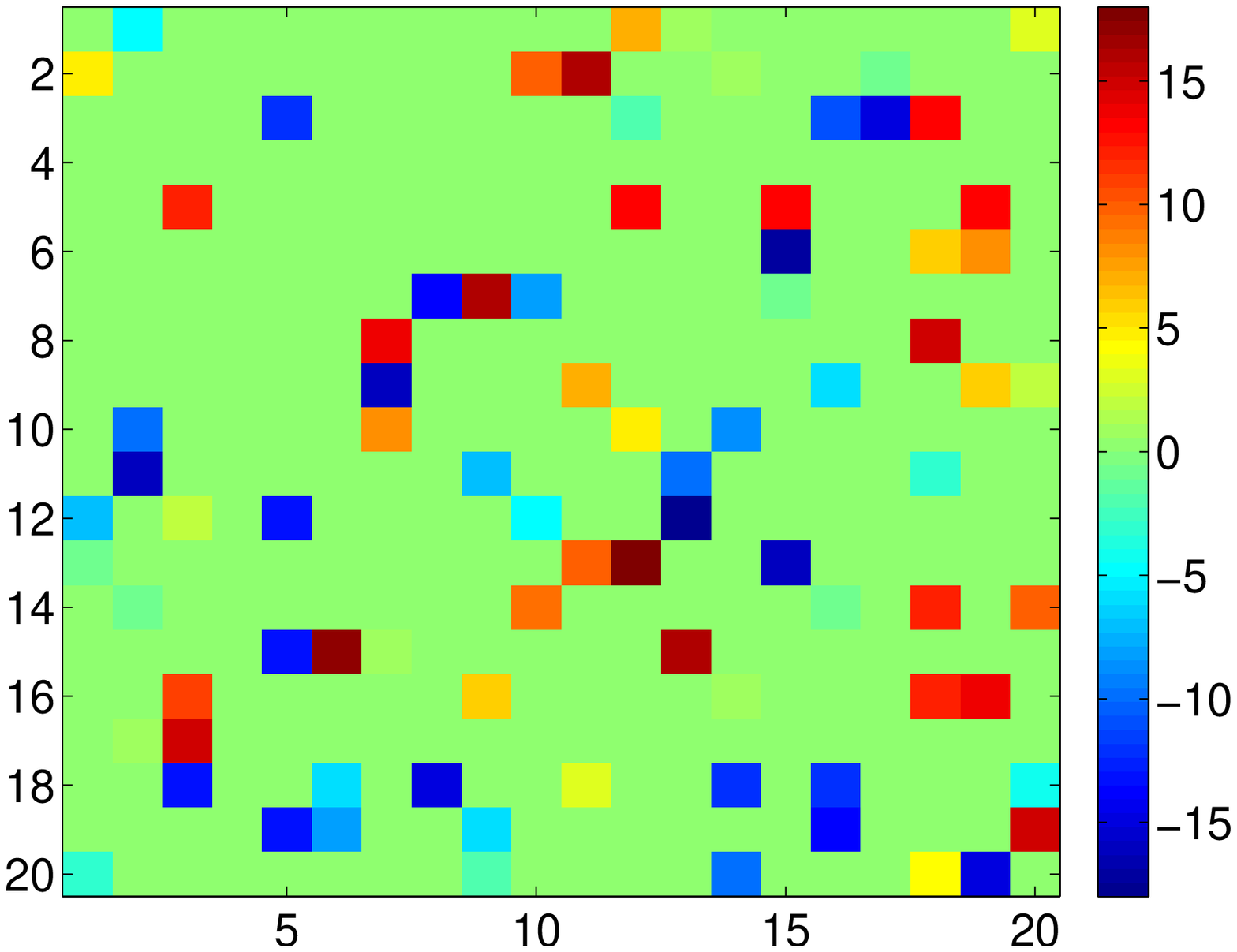}}
\subfigure[ The solution \text{\boldmath$\Theta$}$_{N \times N}$ for the SDP (\ref{SDP_program_rank_agg}) ]{\includegraphics[width=0.25\textwidth]{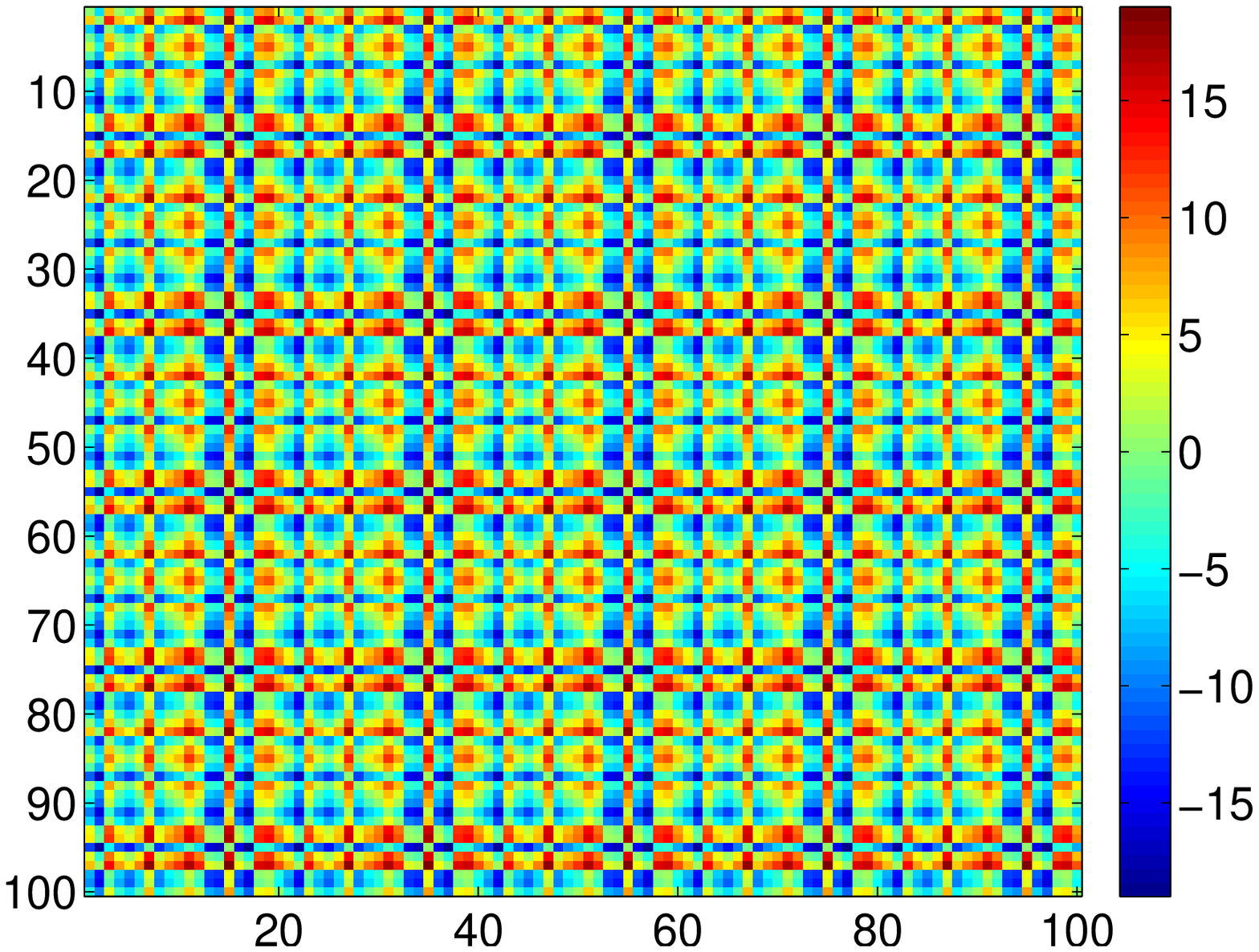}}
\subfigure[ SDP solution ordered by player ID. Note that the $5 \times 5$ blocks on the main diagonal are all zeros.]{\includegraphics[width=0.24\textwidth]{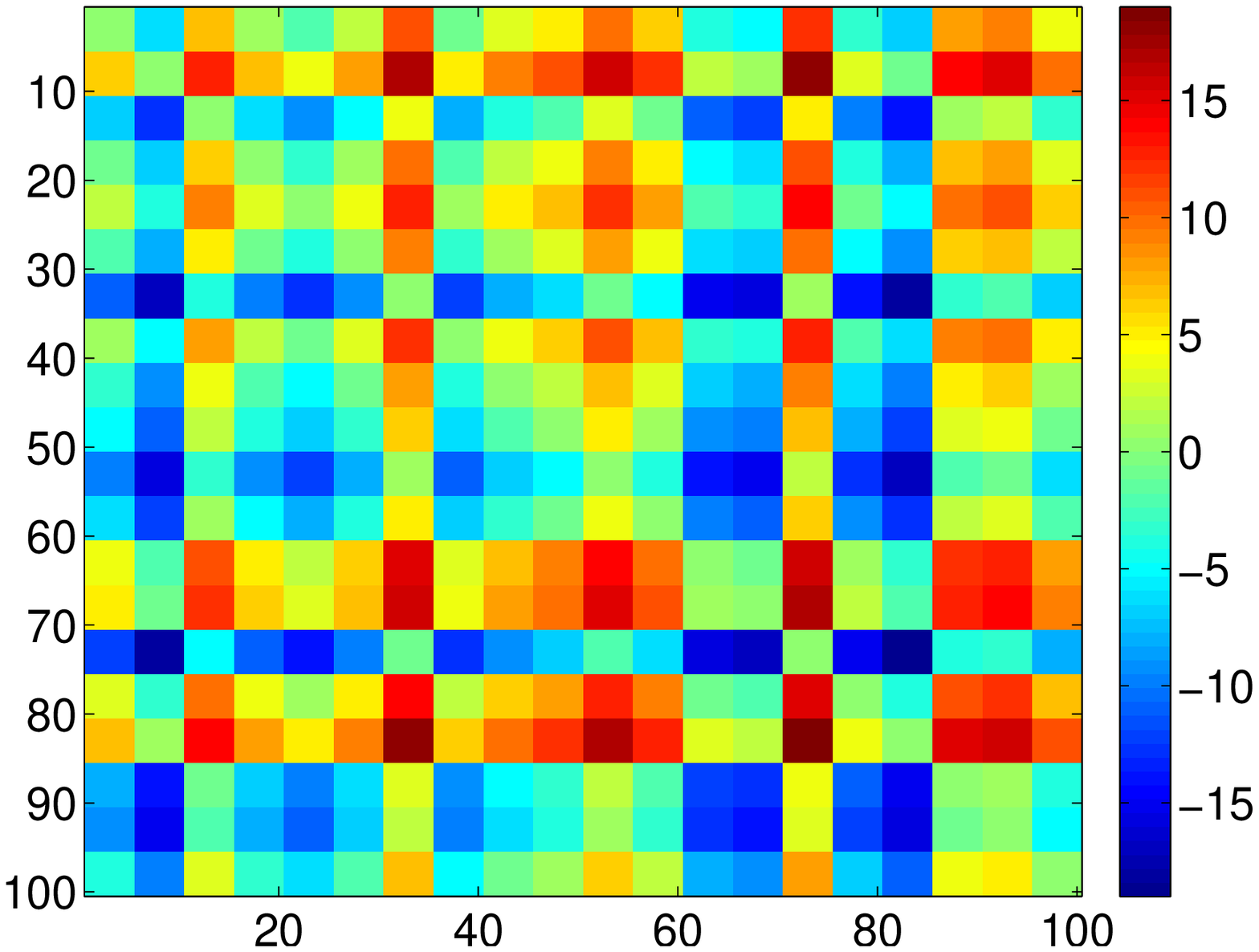}}
\subfigure[ 
The recovered rank offsets (for both SDP-AGG and EIG-AGG)
]{\includegraphics[width=0.24\textwidth]{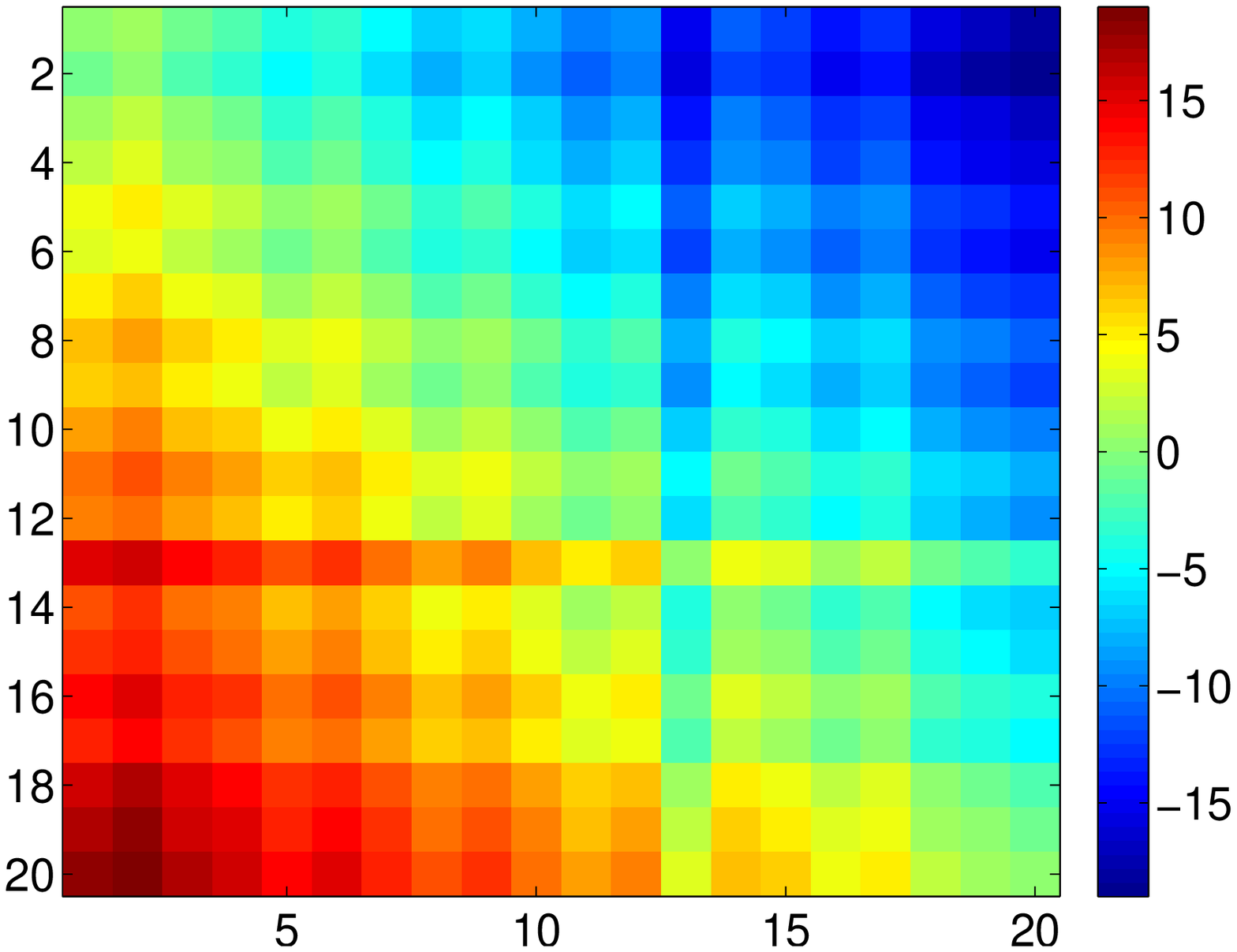}}
\subfigure[ Recovered ranking (for both SDP-AGG and EIG-AGG), $\kappa=0.047$ ]{\includegraphics[width=0.24\textwidth]{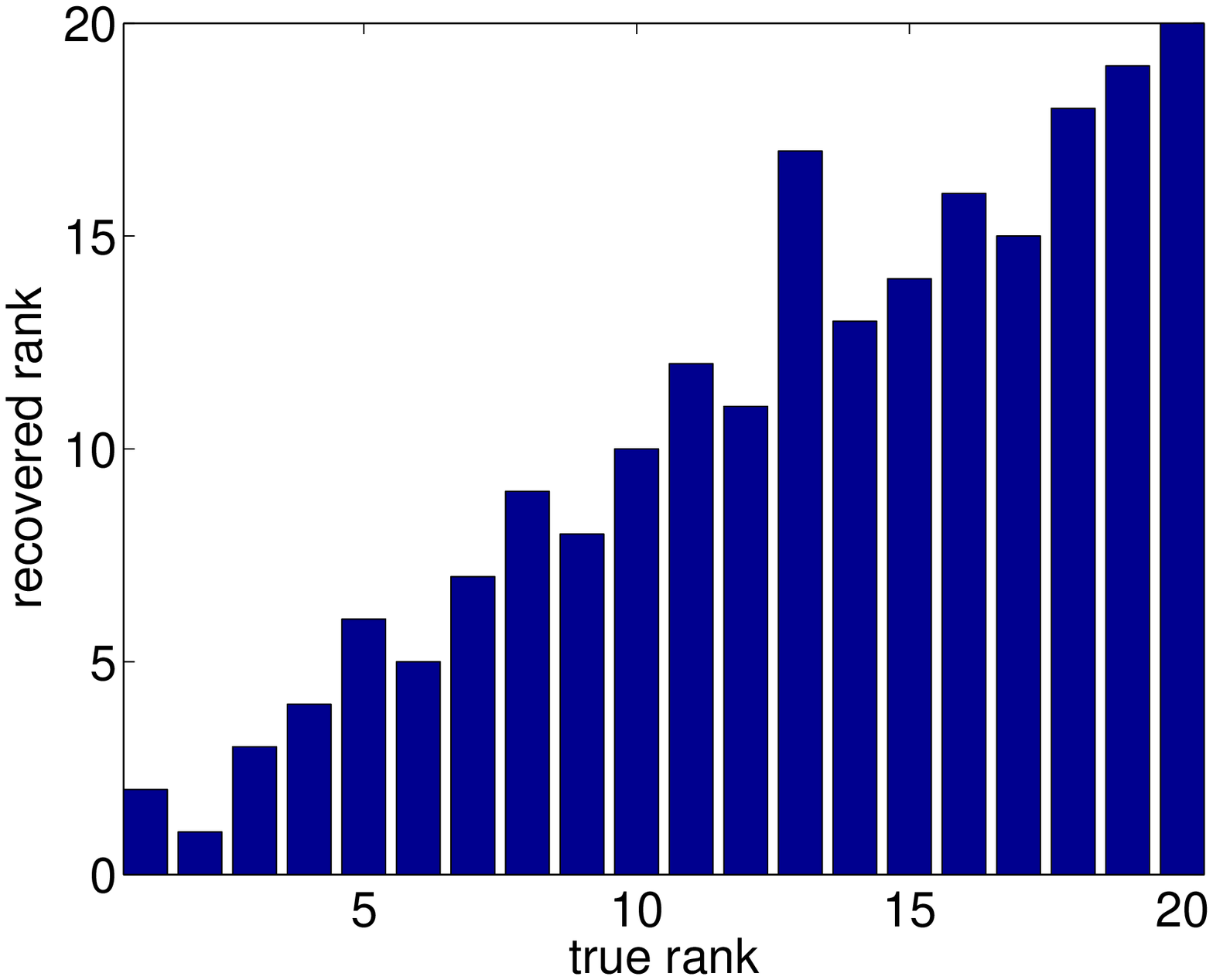}}
%
%
%
\caption{An illustration of the pipeline detailed in Section \ref{secsec:rankAggEnlarged}, for the rank aggregation problem.}
\label{fig:PipelineJuryDemo}
\end{center}
\end{figure}
In Figure \ref{fig:PipelineJuryDemo}, we illustrate our approach on an instance of the rank aggregation problem, as summarized in this Section.
The numerical experiments in Figures \ref{fig:ErrorsJuryMUN} and \ref{fig:ErrorsJuryERO} confirm that, in most scenarios, the SDP-AGG method is significantly more accurate than the other methods of aggregating ranking data from multiple rating systems. (it is surpassed in accuracy only in the case of ordinal comparisons with outliers, and a dense measurement graph, where it is beaten by the naive aggregations based on Serial Rank, shown in Figure \ref{fig:ErrorsJuryERO} (e) and (f).

\subsection{Rank Aggregation via reduced synchronization}
Since the size of the SDP in (\ref{SDP_program_rank_agg}) is $nk \times nk$, one may wonder whether it is possible to reduce its size to $n \times n$, since in some sense, there is a lot of redundant information in the formulation, with many subset of the rows being equal, since those nodes correspond in essence to the same player. Note that one could further impose such constraints to the SDP, that rows $i$ and row $j$ of the inner product matrix $\Upsilon$ are equal whenever $i,j \in \mathcal{A}_u, u=1,\ldots,k$. It is also desirable to be able to solve the rank aggregation problem via the eigenvector method for synchronization, since it is much faster than the SDP   relaxation. Indeed it is the case, and this is based on the observation that the objective function in (\ref{sync_minimization_rankAgg}) (and implicitly in (\ref{syncComplex_minimization_rankAgg})) can be written as
\begin{equation}
\sum_{u=1}^{k} \sum_{ij \in E( \text{\boldmath$G$} ) }  e^{-\iota \theta_i}  \;    \Theta^{(u)}_{ij} \; e^{\iota \theta_j} 
=\sum_{ij \in E( \text{\boldmath$G$} ) }  e^{-\iota \theta_i}    \left( \sum_{u=1}^{k}  e^{ \iota \Theta^{(u)}_{ij}}  \right)   e^{\iota \theta_j}  
= \sum_{ij \in E( \text{\boldmath$G$} ) }  e^{-\iota \theta_i}     \bar{H}_{ij}   e^{\iota \theta_j}
\label{sync_min_sumK}
\end{equation}
where $\bar{H}$ is given by the sum of the $k$ Hermitian matrices
\begin{equation}
\bar{H} = \sum_{u=1}^{k}  H^{(u)}
\label{Havg_CLX}
\end{equation}
Therefore, we may now replace the optimization problem (\ref{sync_minimization_rankAgg}) involving the larger matrix $ \text{\boldmath$H$} $ with a much smaller sized problem based on $\bar{H}$. We denote by EIG-AGG (respectively, SDP-AGG) the method that solves the rank aggregation problem via the eigenvector (respectively, SDP) relaxation for synchronization based on the  matrix $\bar{H}$.

\subsection{Numerical Results for Rank Aggregation}
In this section, we provide numerical experiments comparing the various methods that solve the rank aggregation problem. 
Note that for ordinal measurements, we also compare to the original version of the Rank-Centrality algorithm, considered in \cite{RankCentrality} in the case of multiple rating systems and applicable only to ordinal measurements, whose main steps are summarized in (\ref{aij_RC}),  (\ref{Aij_RC}) and  (\ref{P_RC}). We denote this approach by RCO in the bottom plots of Figures \ref{fig:ErrorsJuryMUN} and \ref{fig:ErrorsJuryERO}.
Furthermore, for both cardinal and ordinal measurements, we also compare to our proposed versions of Rank-Centrality, discussed in Sections \ref{sec:sub_adjRC_ordinal} and \ref{sec:sub_adjRC_cardinal} for the case of a single rating system $k=1$. We adjust these two methods to the setting of multiple rating systems $ k > 1$, by simply averaging out the winning probabilities (given by (\ref{myA_ordinal_RC}) or (\ref{myA_cardinal_RC})) derived from each rating system, as in (\ref{juryVersionMyA}).


In Figure \ref{fig:ErrorsJuryMUN}, we compare the methods in the setting of the Multiplicative Uniform Noise (MUN($n=100,p,\eta$), and average the results over 10 experiments. For cardinal data, we note the the SDP-AGG method yields significantly more accurate results that the rest of the methods, followed by EIG-AGG, and the naive aggregation methods SDP-AVG and SDP-PAR. In the case of the complete graph, the Serial-Rank method SER-GLM comes in second best, together with  SYNC-SDP-PAR. For ordinal data, 
the four methods EIG-AGG, SDP-AGG, SYNC-SDP-PAR and SYNC-SDP−AVG yield very similar results, and are more accurate than the rest of the methods, especially for sparse measurement graphs.

For the  Erd\H{o}s-R\'{e}nyi Outliers model ERO($n=100,p,\eta$) with cardinal measurements, illustrated in Figure \ref{fig:ErrorsJuryERO}, EIG-AGG, SDP-AGG are the best performers, followed very closely by 
SYNC-PAR and SYNC-SDP-PAR. For the case of the complete graph, the SER-GLM comes in next in terms of performance, while all the remaining methods show a rather poor relative performance.

The relative performance of all methods is completely different for the case of ordinal measurents, under the ERO model, where SER-AVG and SER-GLM-AVG (which yield almost identical results) yield the most accurate results, especially at lower levels of noise. Furthermore, the gap between these two methods and all the remaining ones increases as the measurement graphs become denser. For the case of the complete graph with ordinal comparisons with outliers, SER-AVG and SER-GLM-AVG produce results that are 2-3 orders of magnitude more accurate compared to all other methods, at lower levels of noise $\eta \leq 0.2$.
The main observation we would like to point out is that, in the case of  outliers given by the ERO model, our proposed version of Rank-Centrality, denoted by RC, introduced in Section \ref{sec:sub_adjRC_ordinal} and given by (\ref{myA_ordinal_RC}), performs far better than the original RCO method in \cite{RankCentrality}.

\begin{figure}[h!]
\begin{center}
\subfigure[$p=0.2$, cardinal ]{\includegraphics[width=0.30\textwidth]{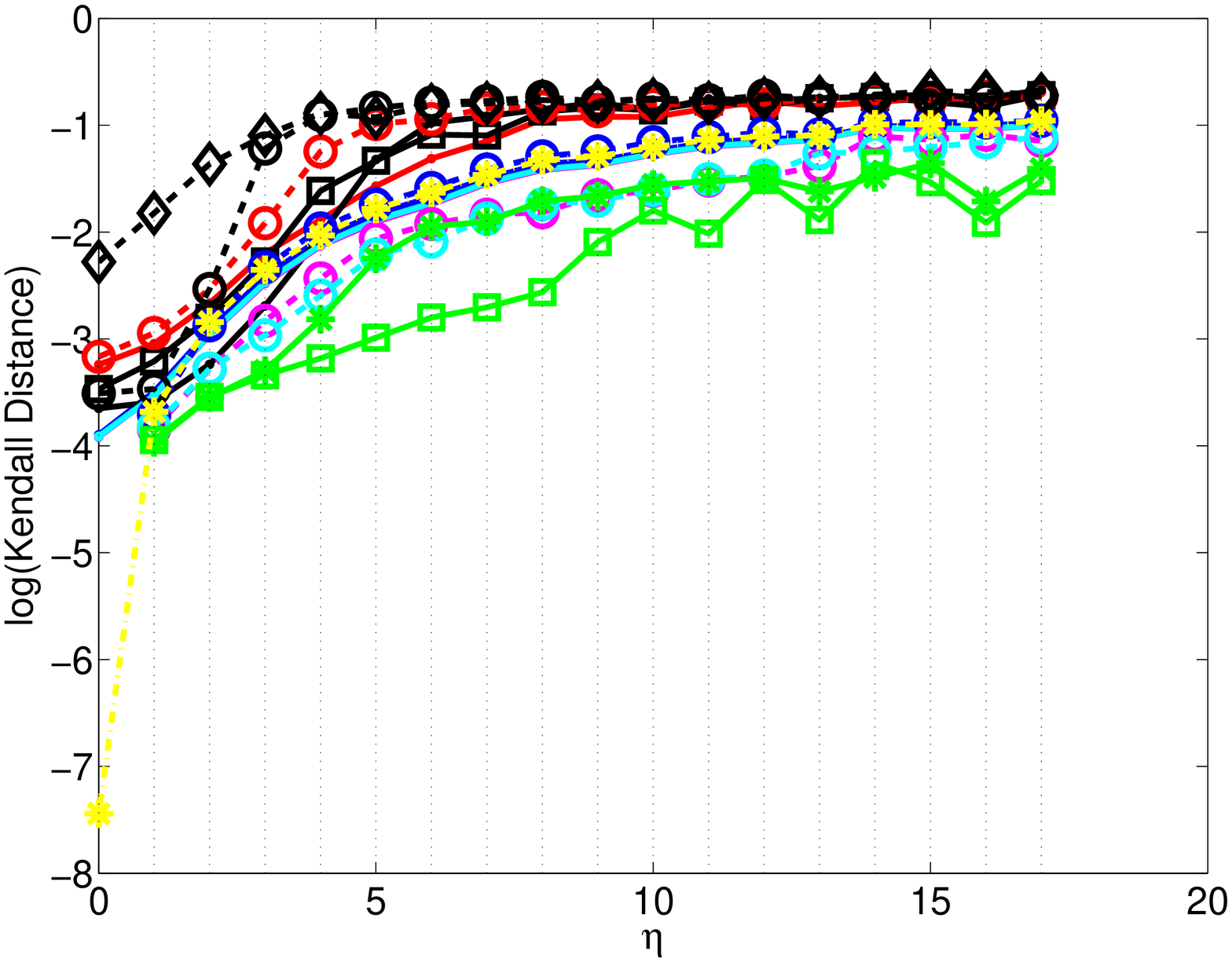}}
\subfigure[$p=0.5$, cardinal]{\includegraphics[width=0.30\textwidth]{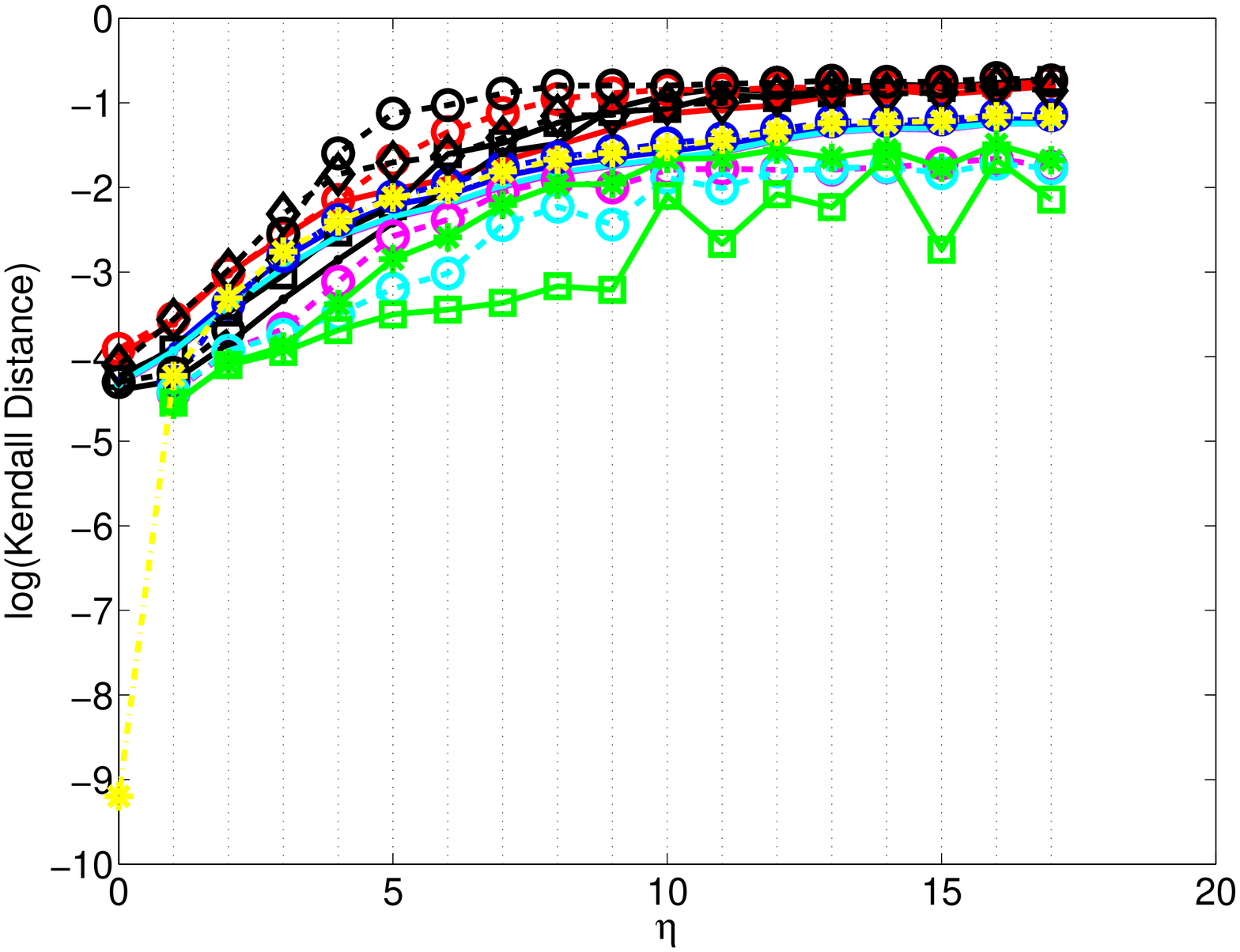}}
\subfigure[$p=1$, cardinal ]{\includegraphics[width=0.30\textwidth]{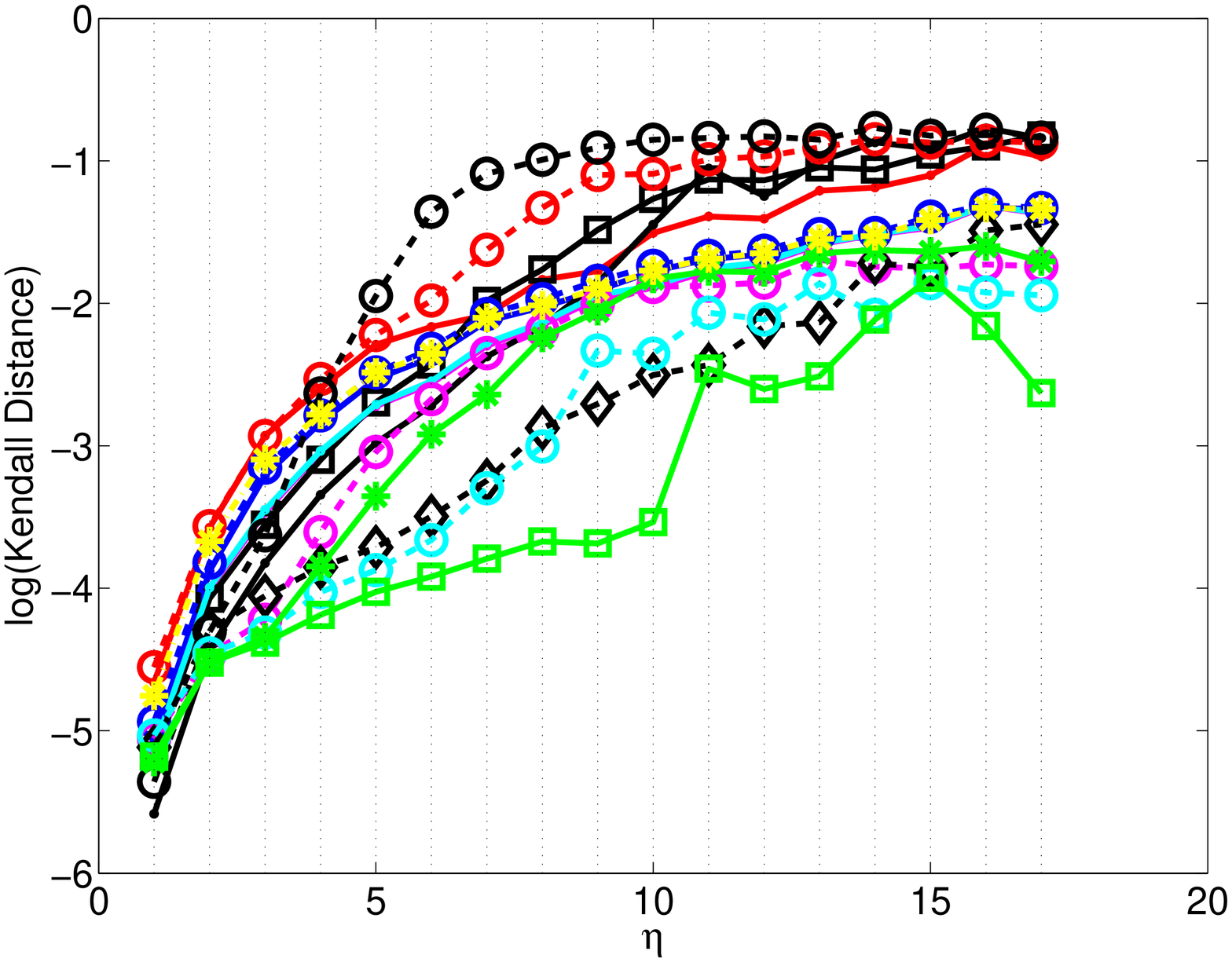}}
\subfigure[$p=0.2$, ordinal ]{\includegraphics[width=0.30\columnwidth]{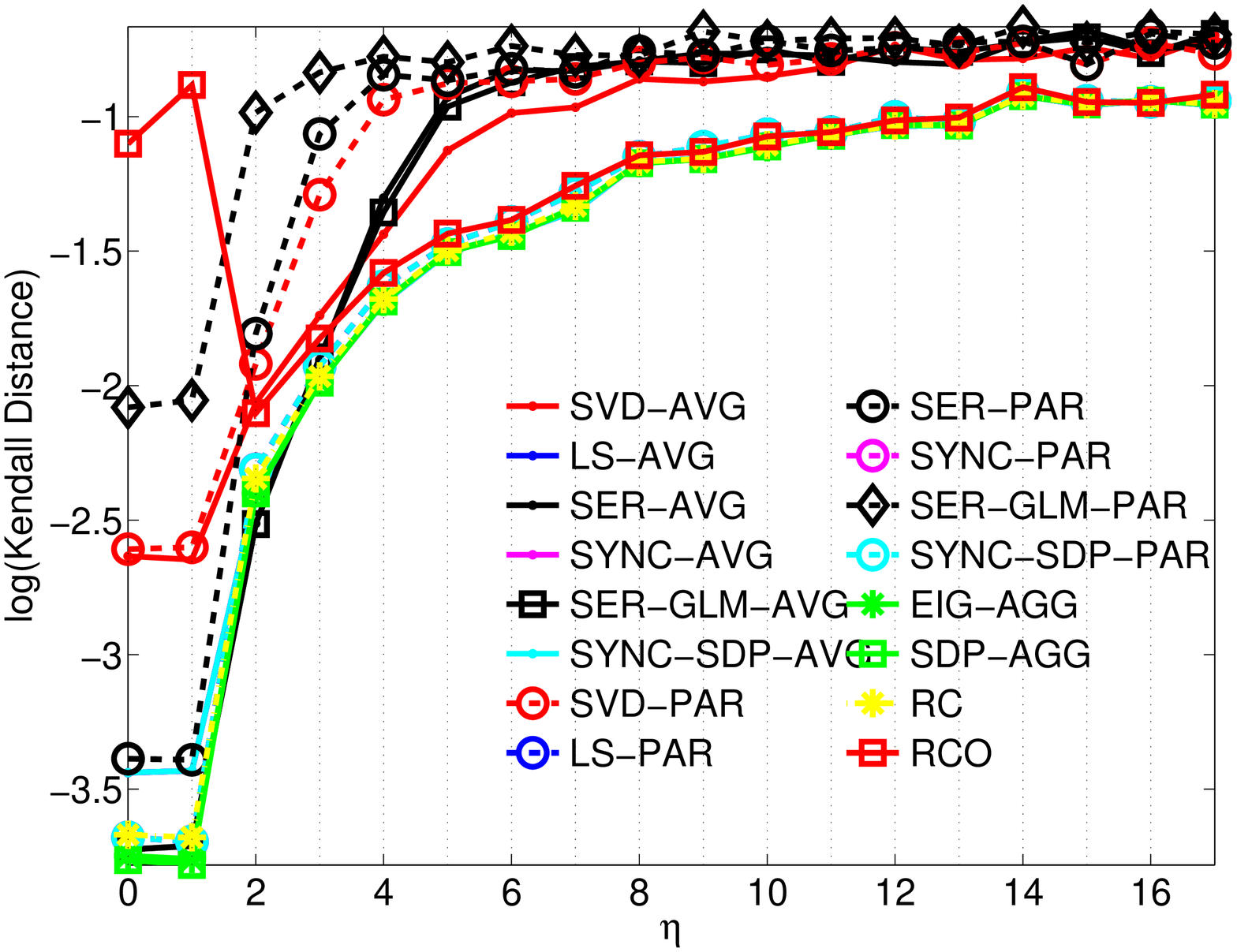}}
\subfigure[$p=0.5$, ordinal ]{\includegraphics[width=0.30\columnwidth]{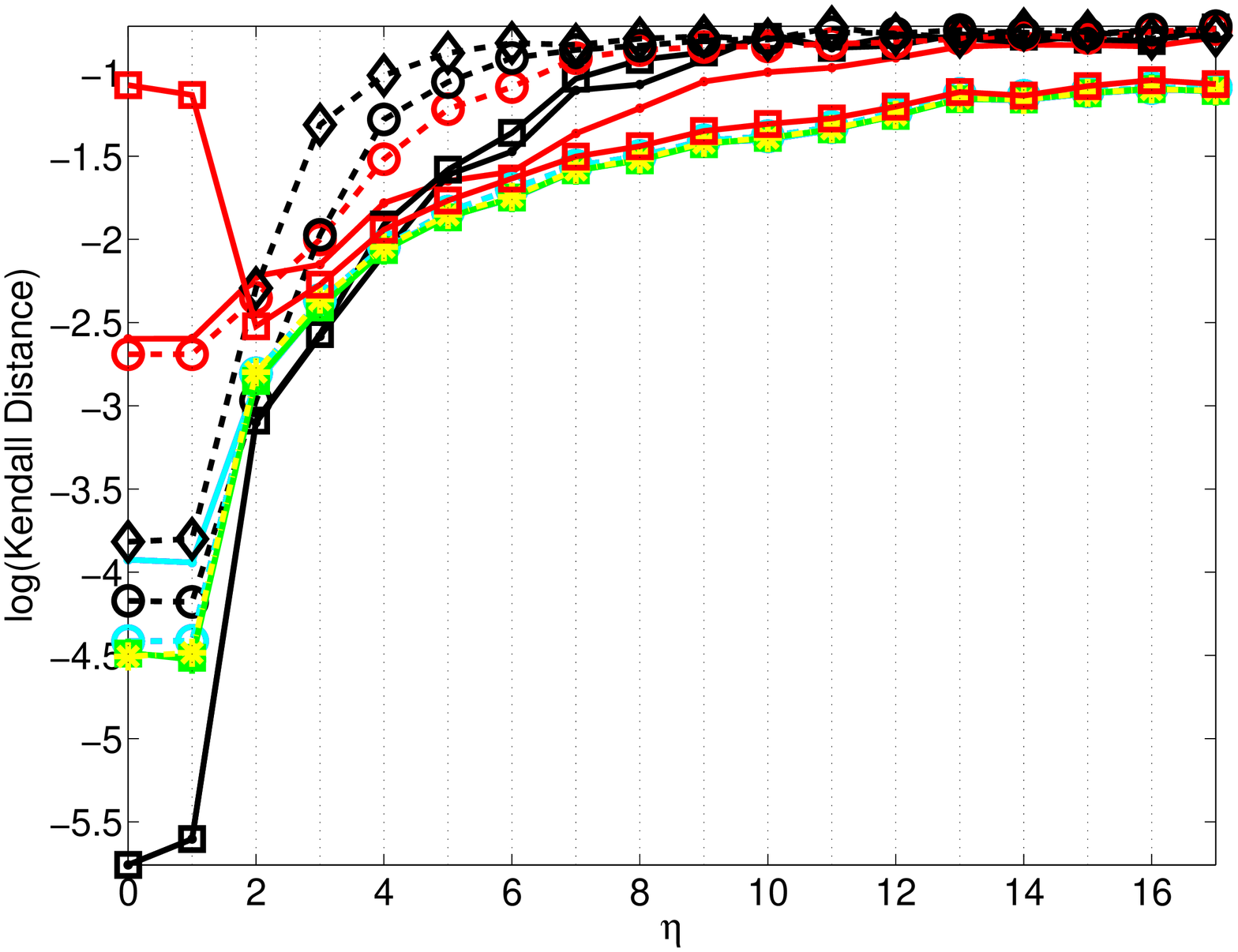}}
\subfigure[$p=1$, ordinal ]{\includegraphics[width=0.30\columnwidth]{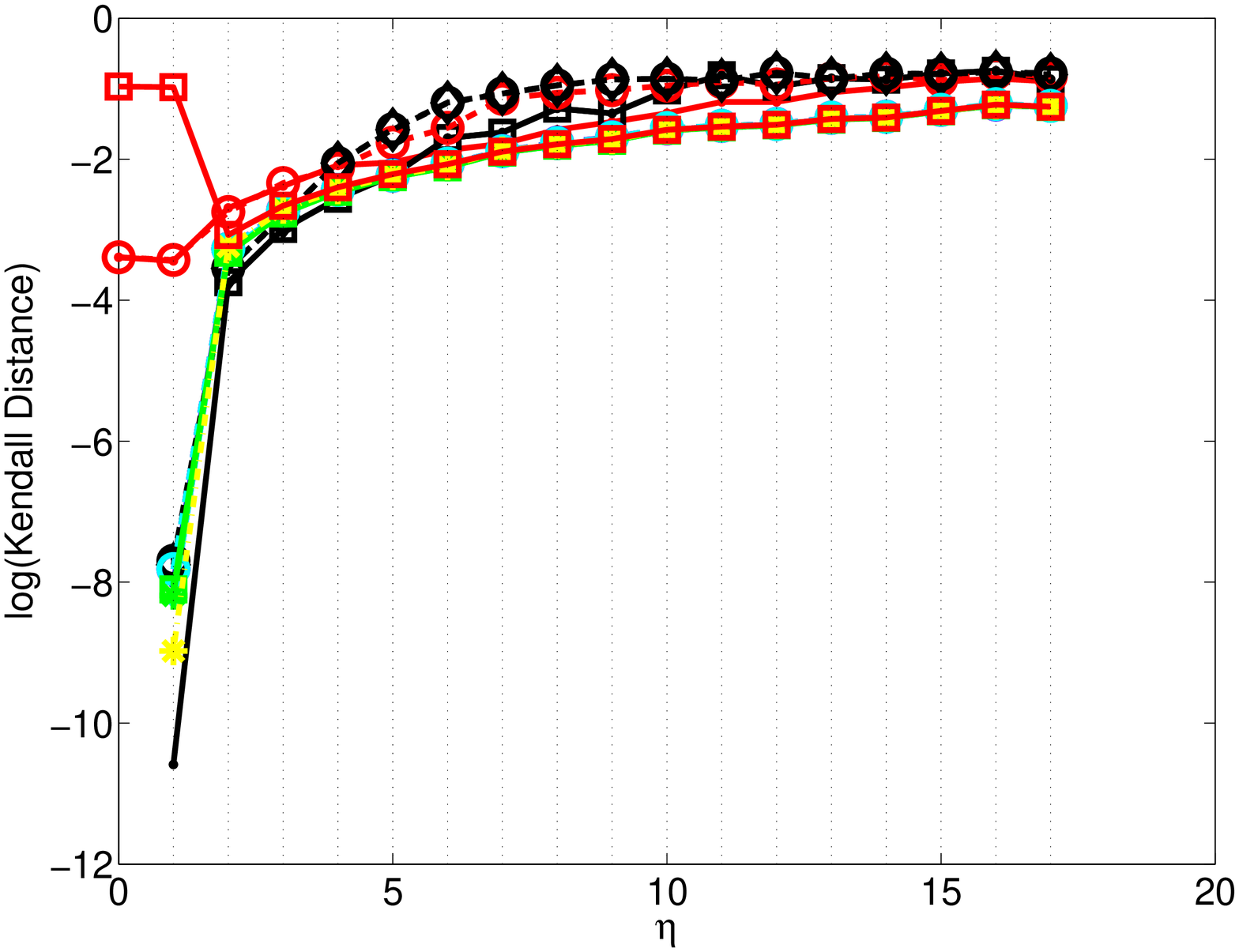}}
\end{center}
\caption{Comparison of results  for the Rank-Aggregation problem with $m=5$ rating systems,  under the Multiplicative Uniform Noise model  MUN($n=100,p,\eta$)  model. We average the results over 10 experiments.}
\label{fig:ErrorsJuryMUN}
\end{figure}

\begin{figure}[h!]
\begin{center}
\subfigure[$p=0.2$, cardinal ]{\includegraphics[width=0.30\textwidth]{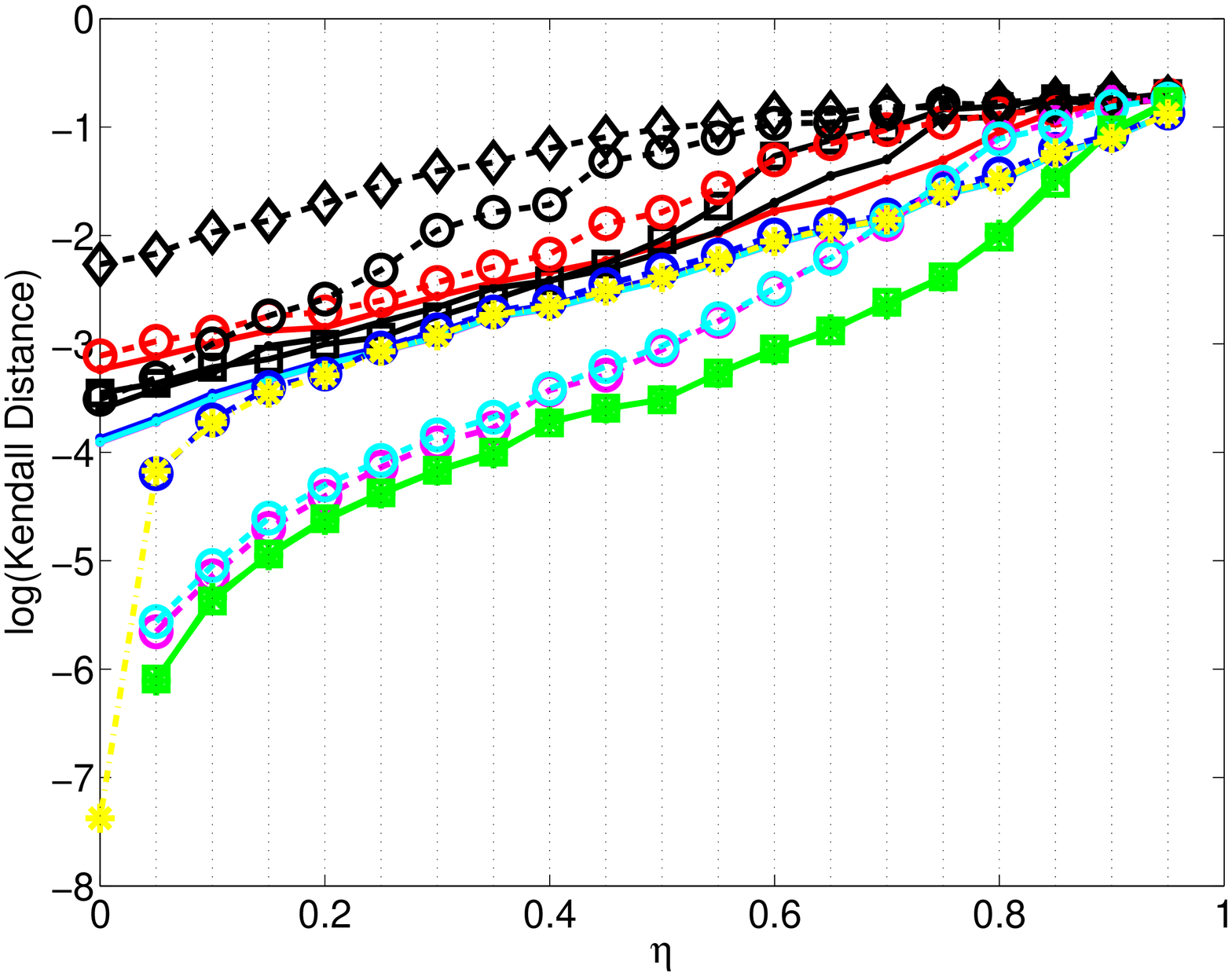}}
\subfigure[$p=0.5$, cardinal]{\includegraphics[width=0.30\textwidth]{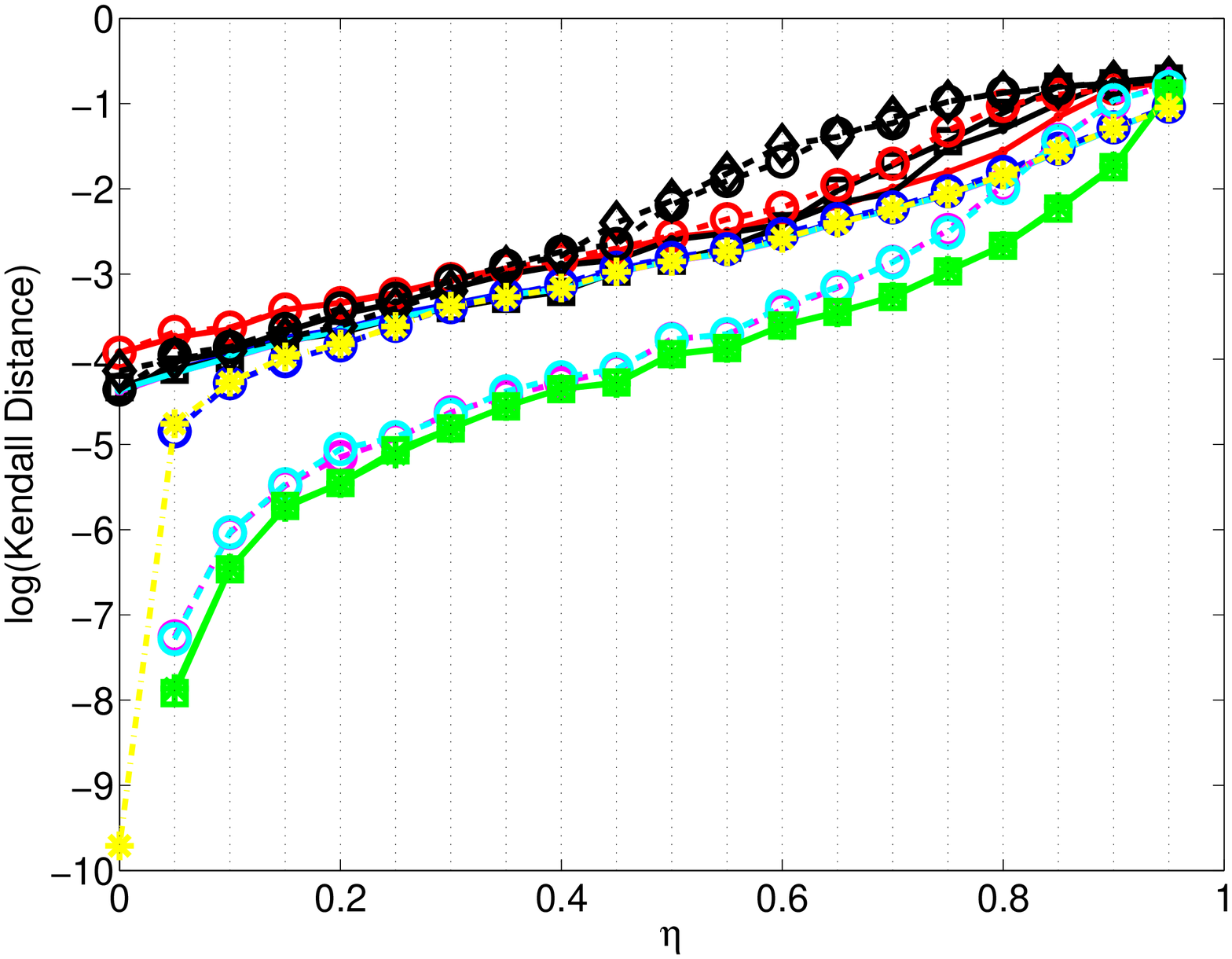}}
\subfigure[$p=1$, cardinal ]{\includegraphics[width=0.30\textwidth]{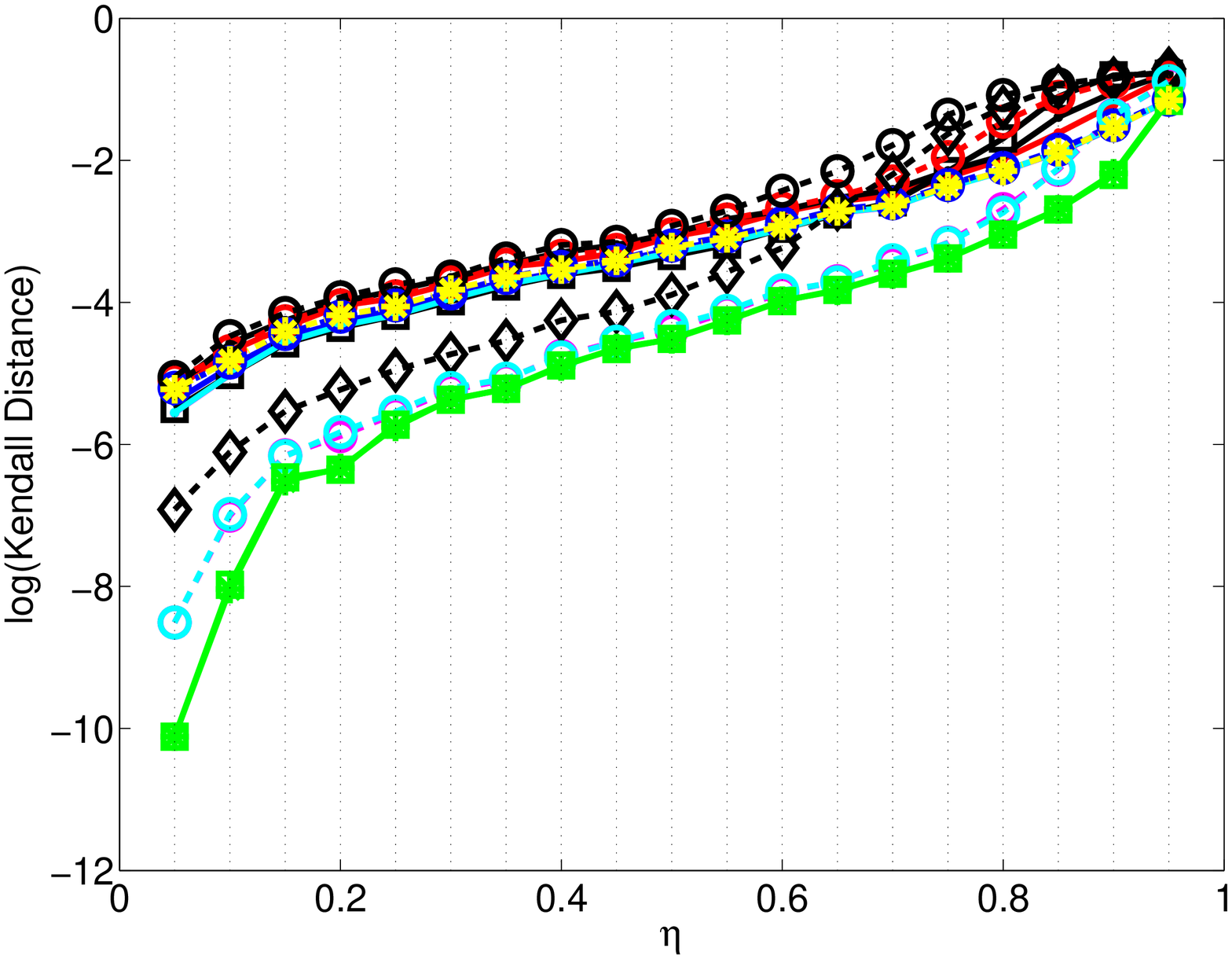}}
\subfigure[$p=0.2$, ordinal ]{\includegraphics[width=0.30\columnwidth]{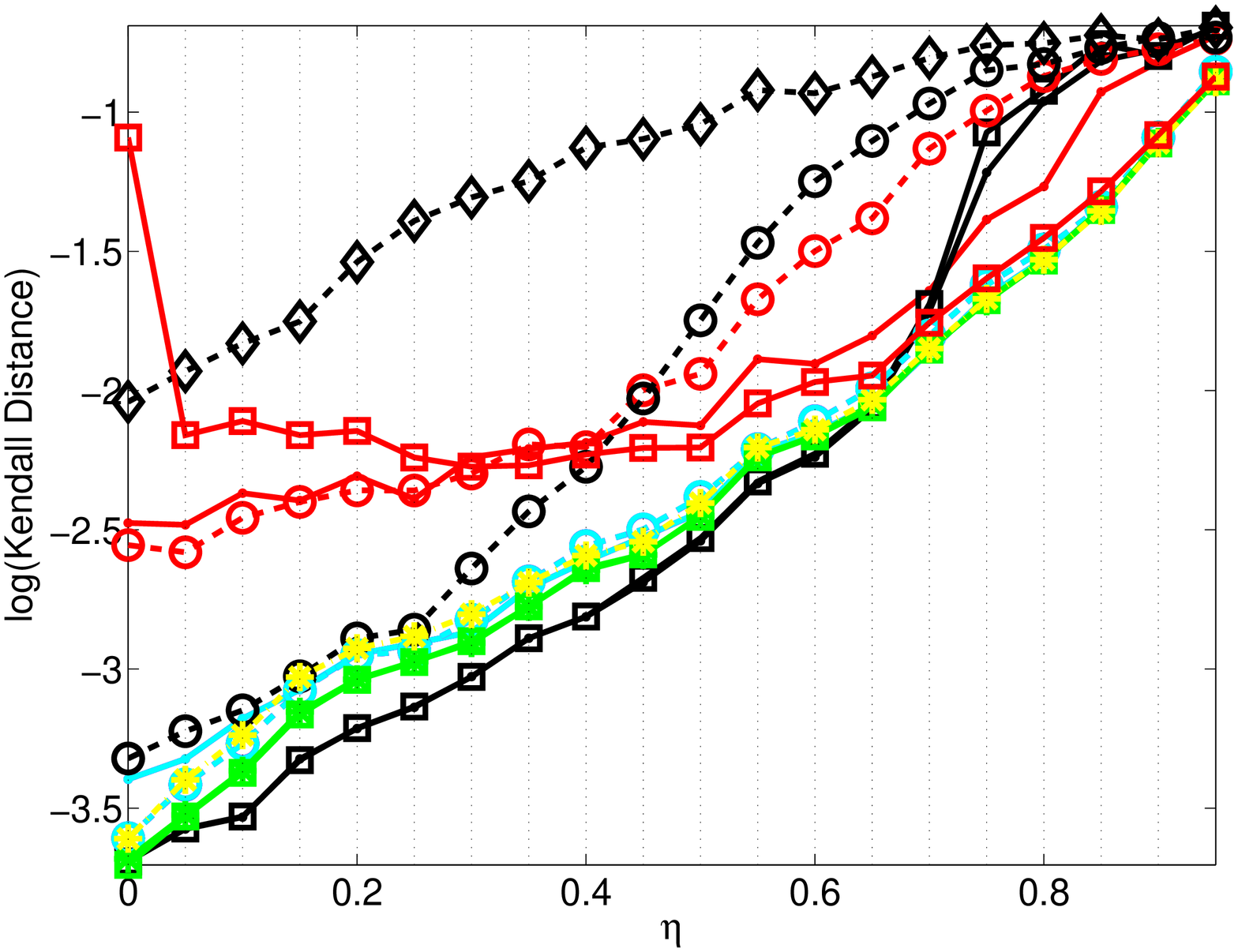}}
\subfigure[$p=0.5$, ordinal ]{\includegraphics[width=0.30\columnwidth]{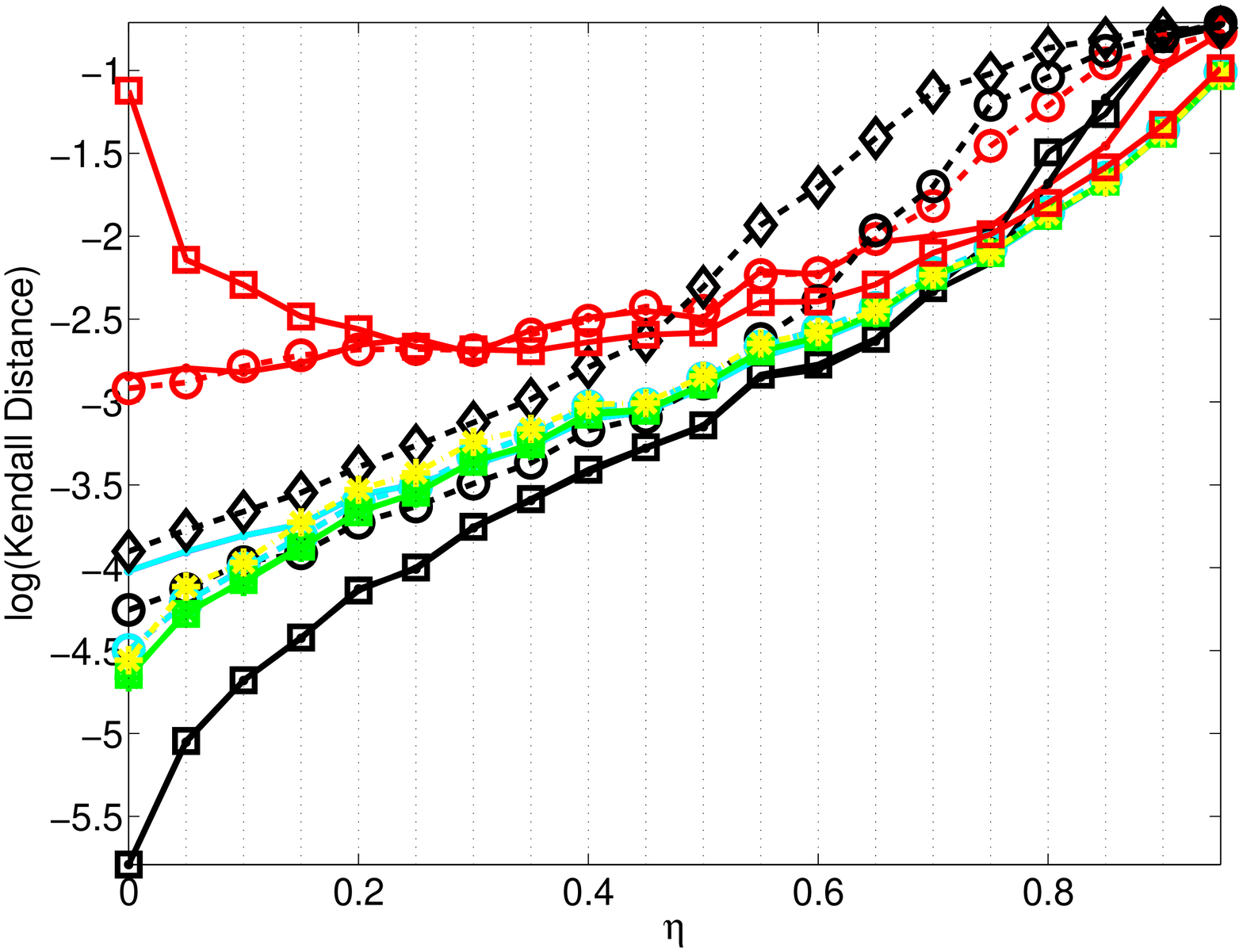}}
\subfigure[$p=1$, ordinal ]{\includegraphics[width=0.30\columnwidth]{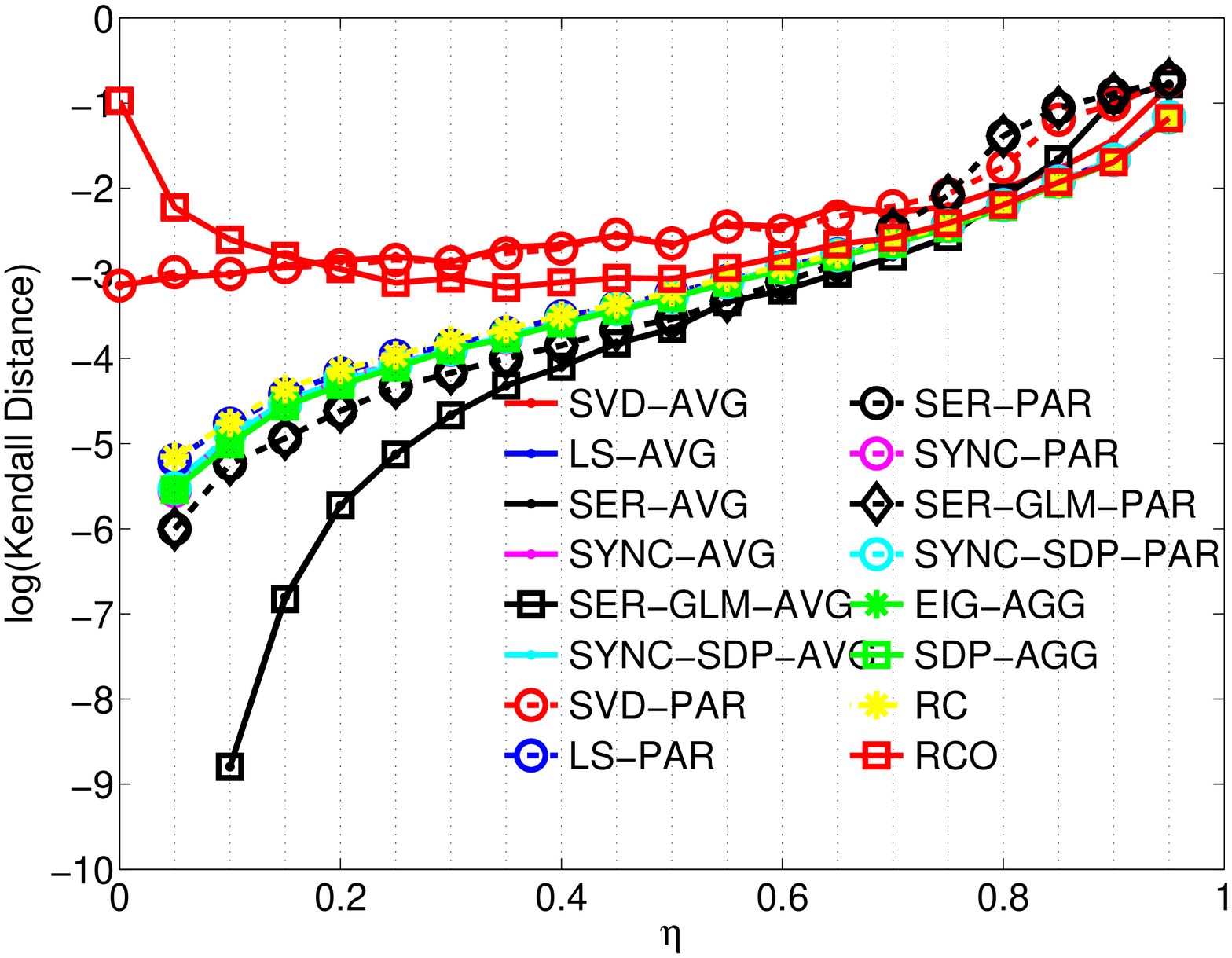}}
\end{center}
\caption{Comparison of results  for the Rank-Aggregation problem with $m=5$ rating systems, for the Erd\H{o}s-R\'{e}nyi Outliers model ERO($n=100,p,\eta$). We average the results over 10 experiments.}
\label{fig:ErrorsJuryERO}
\end{figure}

\FloatBarrier


\section{Ranking with Hard Constraints via Semidefinite Programming}  
\label{sec:constRanking}


In this section, we consider a semi-supervised ranking problem, and illustrate how one can enforce certain types of constraints on the solution. For example, suppose  that the user has readily available information on the true rank of a subset of players, 
and would like to obtain a global solution (a complete ranking of all $n$ players) which obeys the imposed rank constraints. We propose below a modified version of the Sync-Rank algorithm based on the SDP relaxation of the angular synchronization problem followed by a post-processing step, which is able to incorporate such hard constraints on the recovered solution.

The case of ranking with \textit{vertex constraints}, where the user has available information on the true underlying rank of a small subset of players, is, in essence, the problem of \text{angular synchronization with constraints},  analogous to the synchronization over $\mathbb{Z}_2$ with anchor constraints considered previously in \cite{asap3d}, in the context of the three-dimensional graph realization problem and its application to structural biology.

In the spirit of previous work, we shall refer to the subset of players whose rank is already known as \textit{anchors} and denote this set by $\mathcal{A}$, while the set of non-anchor players (referred to as \textit{free} players) shall be denoted  by $ \mathcal{F}$. In the context of the synchronization problem, \textit{anchors} are nodes whose corresponding group element in $SO(2)$ is known a-priori.  Mathematically, the problem of angular synchronization with constraints can be formulated as follows. 
Given our measurement graph $G=(V,E)$ with node set $V$ ($|V|=n$) corresponding to a set of $n$ group elements composed of $h$ anchors $\mathcal{A} = \{a_1,\ldots,a_h\}$ with $a_i \in $ SO(2), and $l$ sensors $\mathcal{F}=\{f_1,\ldots,f_l\}$ with $ f_i \in $ SO(2), with $n = h + l$, and edge set $E$ of size $m$ corresponding to an incomplete set of $m$ (possibly noisy) pairwise angle offsets of the type  $f_i - f_j$ 
or $a_i - f_j$, 
the goal is to provide accurate estimates $\hat{f}_1,\ldots,\hat{f}_l \in $ SO(2) for the unknown non-anchor group elements $f_1,\ldots,f_l$. 

The approach we propose is as follows. Since the ground truth rank of anchor players is known, we compute the ground truth rank offset for all pairs of anchor nodes, use the previously considered transformation (\ref{transfToCircle}) to map such rank offsets to angle offsets  $\Theta_{ij} \in [0, 2 \pi  \delta )$ with $\delta \in [0,1)$ 
\begin{equation}
  \Theta_{ij} :=  2 \pi  \delta  \frac{ r_i - r_j}{n-1}, \forall i,j \in \mathcal{A}
\end{equation}
with $\delta = \frac{1}{2}$, and, finally, enforce the resulting angle offsets as hard constraints in the SDP relaxation given by (\ref{my_SDP_maxAnch}).
We have seen in Section \ref{secsec:syncSDP} that an alternative to the spectral relaxation from Section \ref{secsec:syncEIG} for  angular synchronization relies on semidefinite programming. In light of the optimization problem (\ref{SDP_program_SYNC}), the SDP relaxation of the angular synchronization problem in the presence of anchors is  given by  (\ref{my_SDP_maxAnch}), where the maximization is taken over all semidefinite positive real-valued matrices $\Upsilon \succeq 0$  with
\begin{equation}
\Upsilon_{ij} = \left\{
     \begin{array}{rl}
e^{\iota ( f_i - f_j)} & \;\; \text{ if } i,j \in \mathcal{F} \\
e^{\iota ( f_i - a_j)} & \;\; \text{ if } i \in \mathcal{F}, j \in \mathcal{A} \\
e^{\iota ( a_i - a_j)} & \;\; \text{ if } i,j \in \mathcal{A}. \\
     \end{array}
   \right.
\label{my_upsilondefAnch}
\end{equation}
Note that $\Upsilon$ has ones on its diagonal $ \Upsilon_{ii} =1, \forall i=1,\ldots,n$, and the anchor information gives another layer of hard constraints.
The SDP relaxation for angular synchronization in the presence of anchors is given by
\begin{equation}
	\begin{aligned}
& \underset{\Upsilon\in \mathbb{C}^{n \times n}}{\text{maximize}}
& & Trace(H \Upsilon) \\
& \text{subject to}
  & & \Upsilon_{ii} =1, i=1,\ldots,n \\
  & & &  \Upsilon_{ij} = e^{\iota \Theta_{ij} } , \;\; \text{ if } i,j \in \mathcal{A} \\
  & & &   \Upsilon \succeq 0
	\end{aligned}
\label{my_SDP_maxAnch}
\end{equation}
While in the noiseless case, the SDP in (\ref{my_SDP_maxAnch}) may return a rank-1 one matrix, for noisy data the rank of the optimal solution $\Upsilon$ is higher than 1, thus we compute the solution via the best rank-1 approximation. 
Next, we use the top eigenvector $v_1^{(\Upsilon)}$  of  $\Upsilon$ to recover the estimated angles $ $ (and implicitly the rankings), up to an additive phase (i.e., a global rotation, respectively a circular permutation) since  $ e^{\iota \alpha} v_1^{\Upsilon}$ is also an eigenvector of $\Upsilon$ for any $\alpha \in \mathbb{R}$. 
\begin{figure}[h!]
\begin{center}
\includegraphics[width=0.5\textwidth]{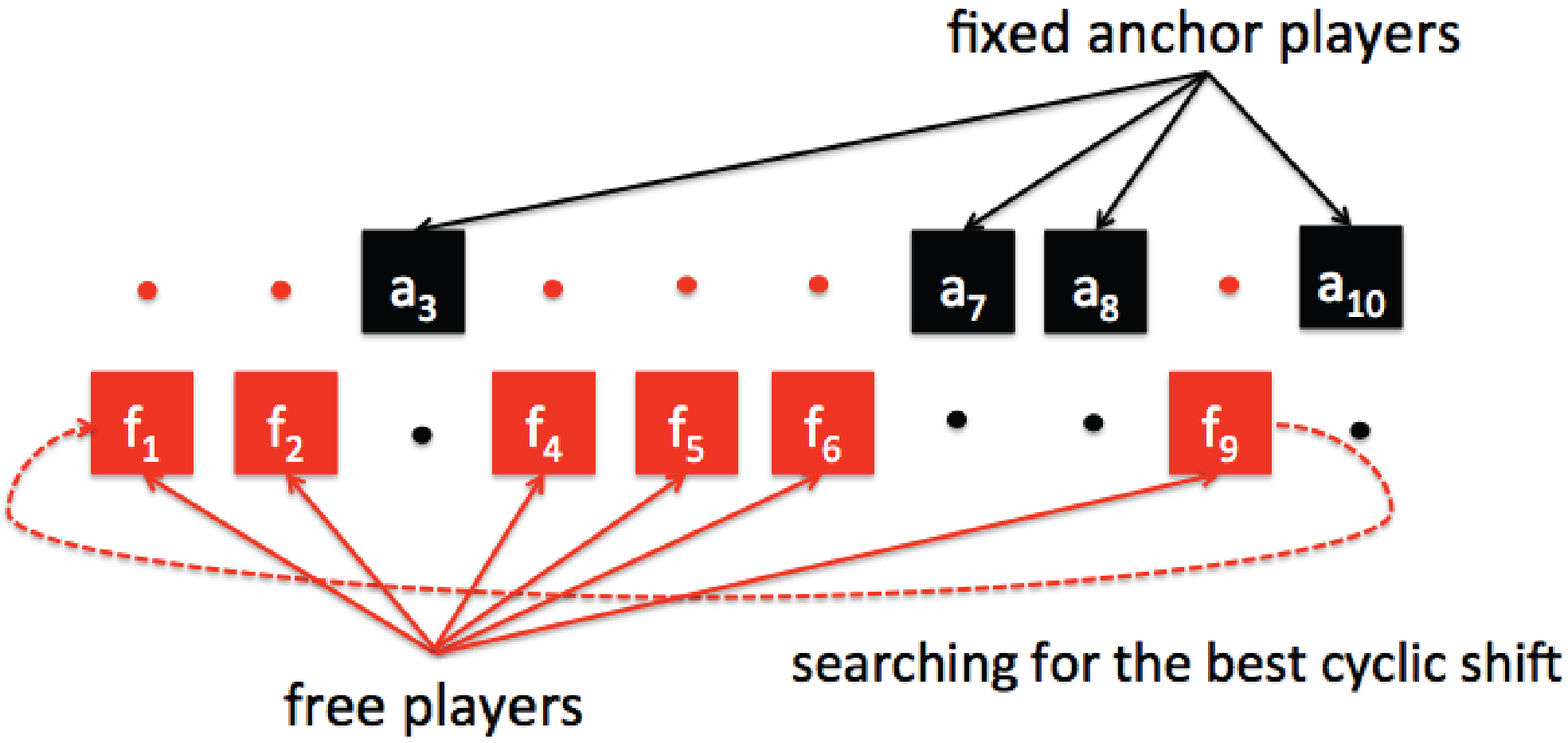}
\hspace{2mm}
\includegraphics[width=0.3\textwidth]{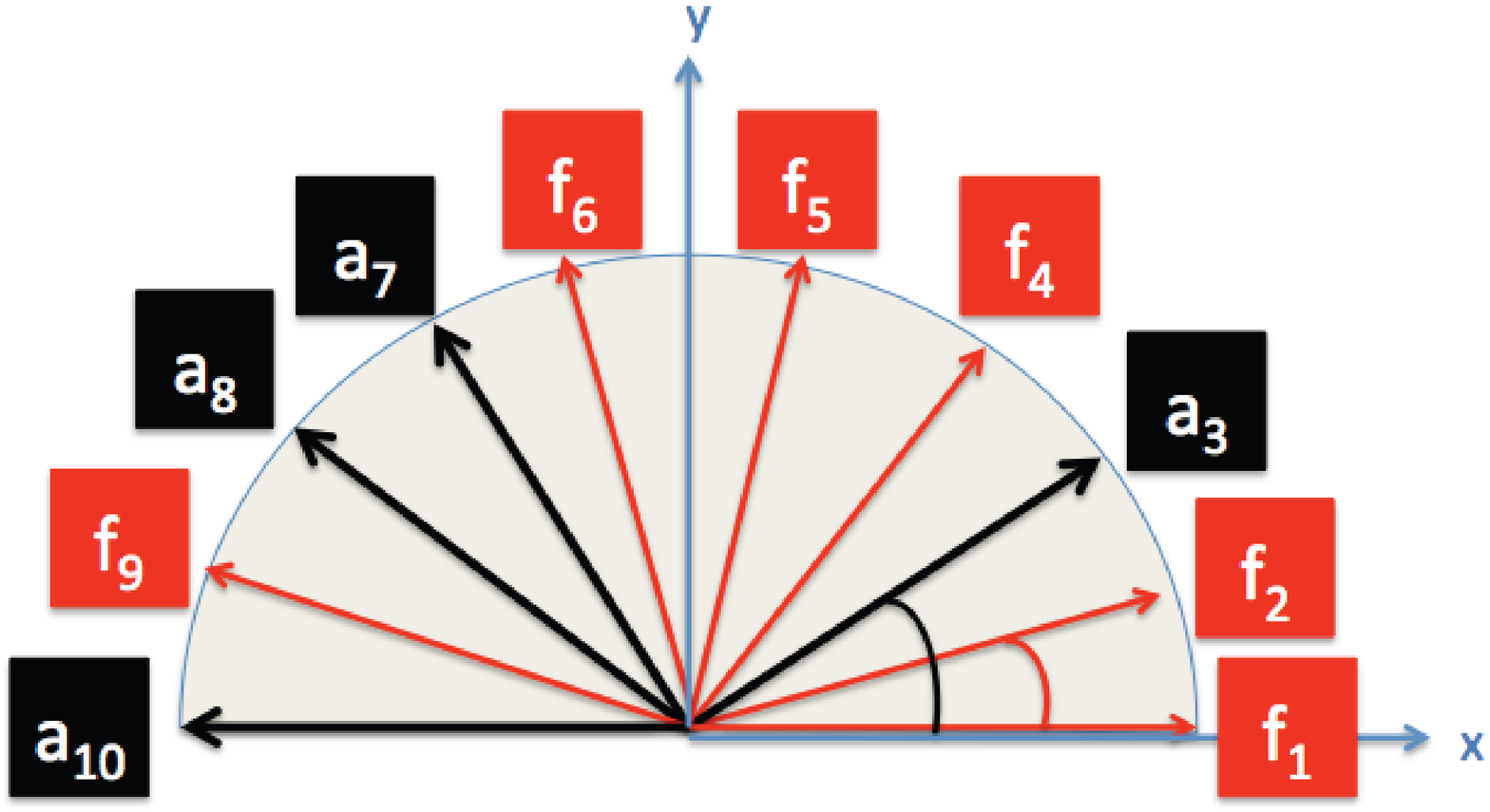}
\end{center}
\caption{ Left: An instance of ranking with anchor constraints, solved by the SDP-based relaxation of the synchronization algorithm. The black nodes denote the anchor players, whose rank is known a-priori and stays fixed throughout the iterations. The red nodes denote the free players, whose relative ranking is recovered (up to a phase shift) from the top eigenvector of the optimal $\Upsilon$ matrix in the SDP program (\ref{my_SDP_maxAnch}). We compute the final combined ranking (of both anchor and free players) by searching for the optimal circular permutation of the free nodes (which \textit{interlace} the anchor nodes), which minimizes the number of upsets in the induced ranking of all players, with respect to the initial pairwise comparison matrix $C$. Right: An illustration of the ground truth ranking, where players correspond to equidistant points placed on the upper half circle, and their rank is given by the magnitude of the corresponding angle; anchor players are shown in black, while free players are shown in red.}
\label{fig:anch_picture_shift}
\end{figure}
With this observation in mind, we propose a post-processing step, depicted in Figure  \ref{fig:anch_picture_shift}, that ensures that the anchor players obey their prescribed rank. We keep the ranks of the anchor players fixed, and sequentially apply a cyclic permutation of the free players on the available rankings. While doing so, the ranks of the free players interlace the ranks of the anchor players, the latter of which stay fixed. We compute the optimal cyclic shift as the one that minimizes the number of upsets in the entire measurement matrix, as previously used in (\ref{argMinSign}).
To measure the accuracy of each candidate total ordering $\text{\boldmath$s$}$ (i.e., a permutation of size $n$), we first compute the  pairwise rank offsets associated to it
$ Z( \text{\boldmath$s$} ) =  \text{\boldmath$s$}   \otimes \mb{1}^T -  \mb{1}  \otimes  \text{\boldmath$s$}^T   $
where $\otimes$ denotes the outer product  $ x \otimes y = x y^T$. 
Next, we choose the circular permutation of the ranks of the set of free players $\mathcal{F}$,  which minimizes the $l_1$ norm of the following residual matrix that counts the total number of upsets
\begin{equation}
\sigma =  \underset{\sigma^{(1)}_{\mathcal{F}},\ldots,\sigma^{(l)}_{\mathcal{F}} }{\text{arg min} }  \;\;\;
 \frac{1}{2} \norm{  \mbox{sign}( Z( \sigma_{\mathcal{A}} \sigma^{(i)}_{\mathcal{F}} ) -  \mbox{sign} (C) }_1,
\end{equation}
where $  \sigma_{\mathcal{A}} $ denotes the permutation associated to the anchor players, which stays fixed in the above minimization. The intuition behind this approach is that the solution obtained from the SDP program preserves as best as possible the relative ordering of the players, in particular of the set of free players, and does so up to a global shift, which we recover via the above procedure.

\begin{figure}[h!]
\begin{center}
\subfigure[$p=0.2$, cardinal, MUN ]{\includegraphics[width=0.283\textwidth]{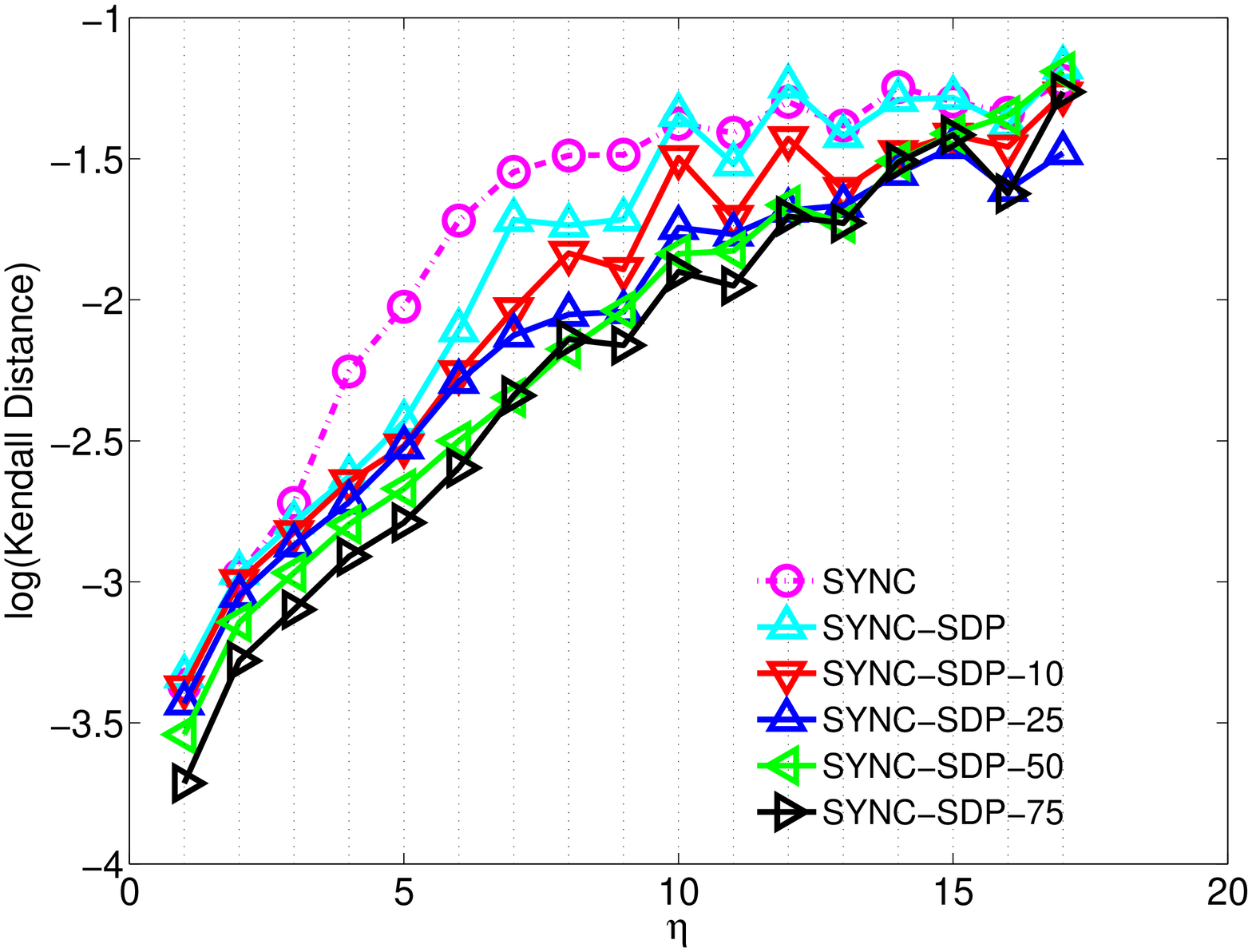}}
\subfigure[$p=0.5$, cardinal, MUN]{\includegraphics[width=0.283\textwidth]{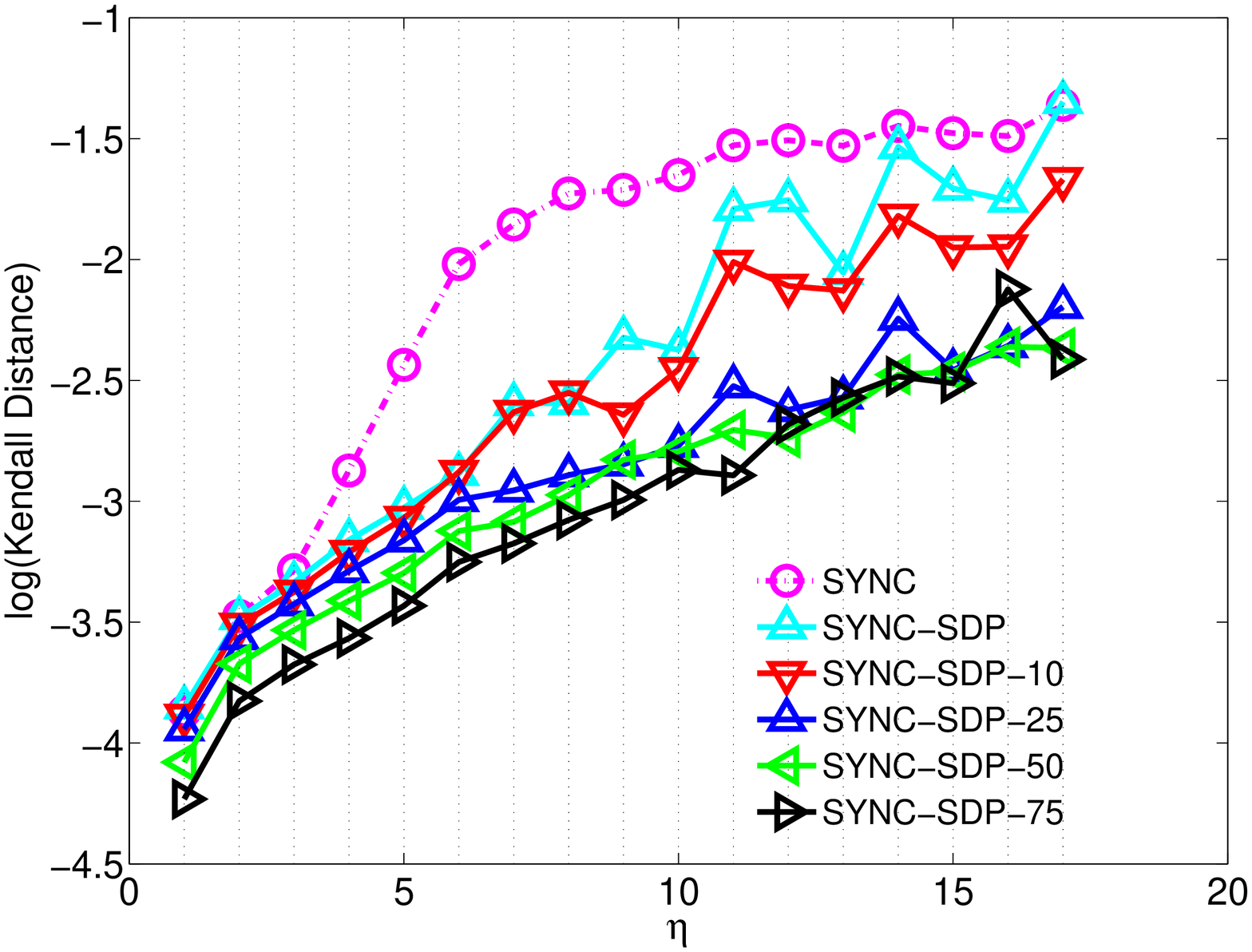}}
\subfigure[$p=1$, cardinal, MUN ]{\includegraphics[width=0.283\textwidth]{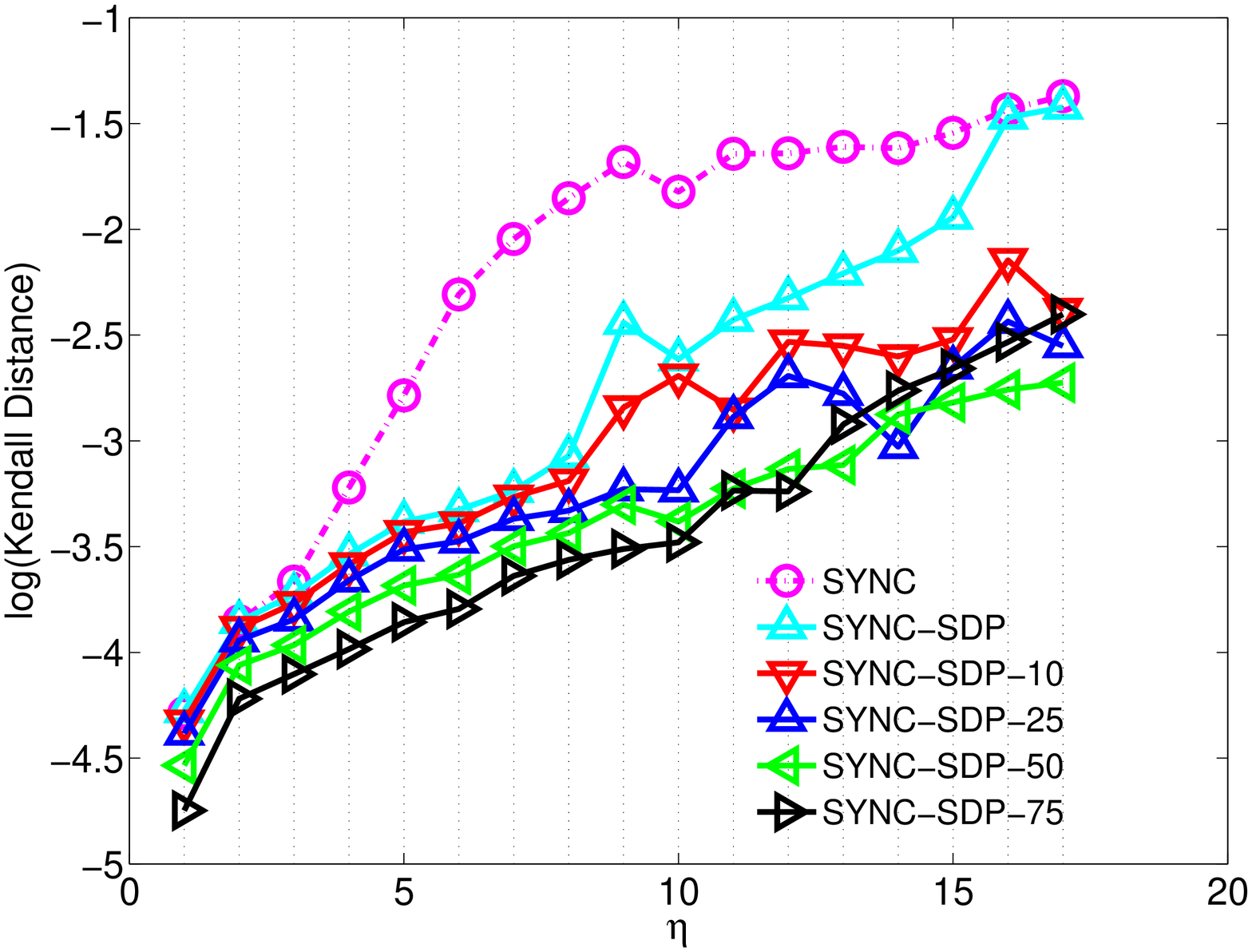}}
\subfigure[ $p=0.2$, ordinal, ERO ]{\includegraphics[width=0.283\textwidth]{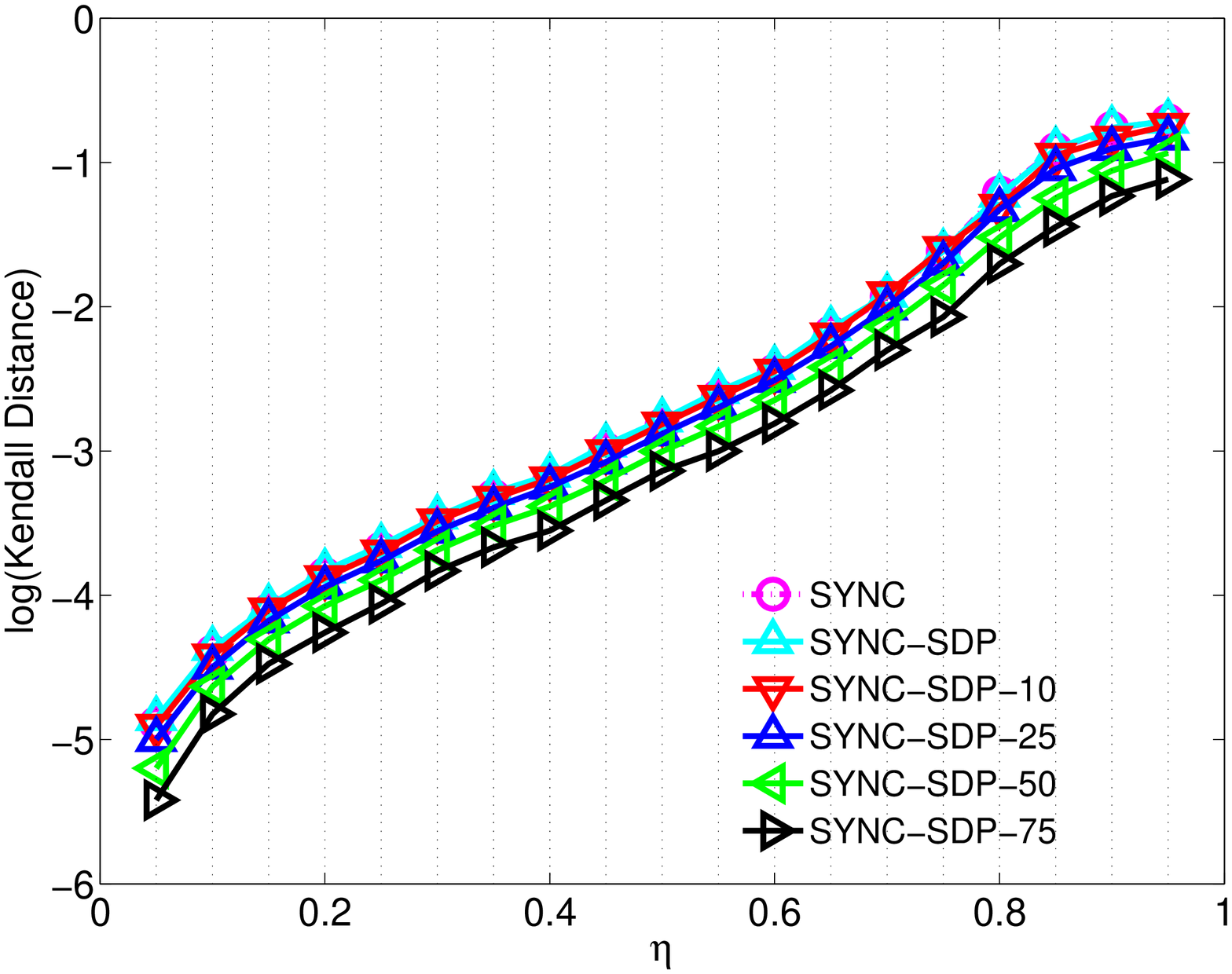}}
\subfigure[ $p=0.5$, ordinal, ERO ]{\includegraphics[width=0.283\textwidth]{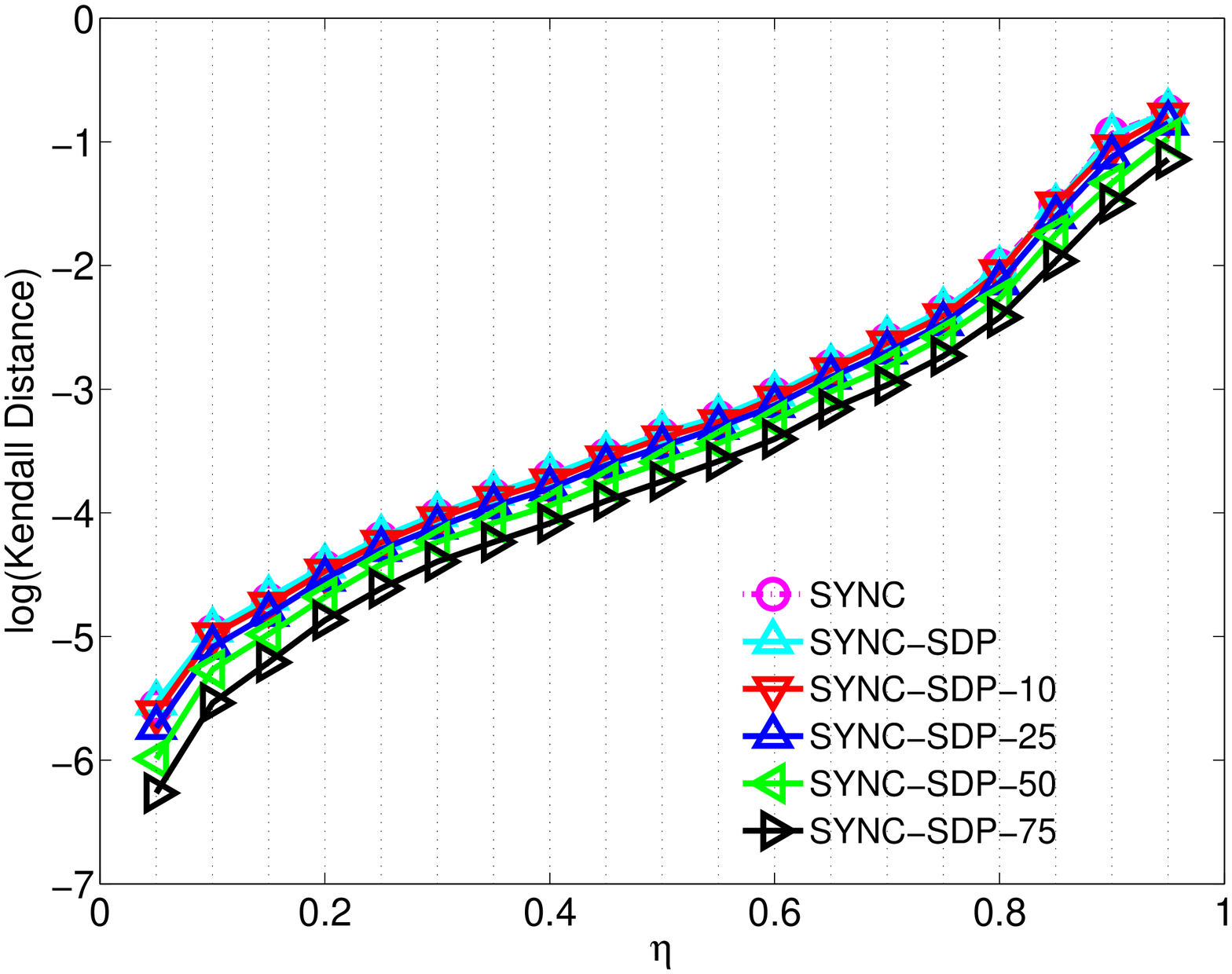}}
\subfigure[ $p=1$, ordinal, ERO  ]{\includegraphics[width=0.283\textwidth]{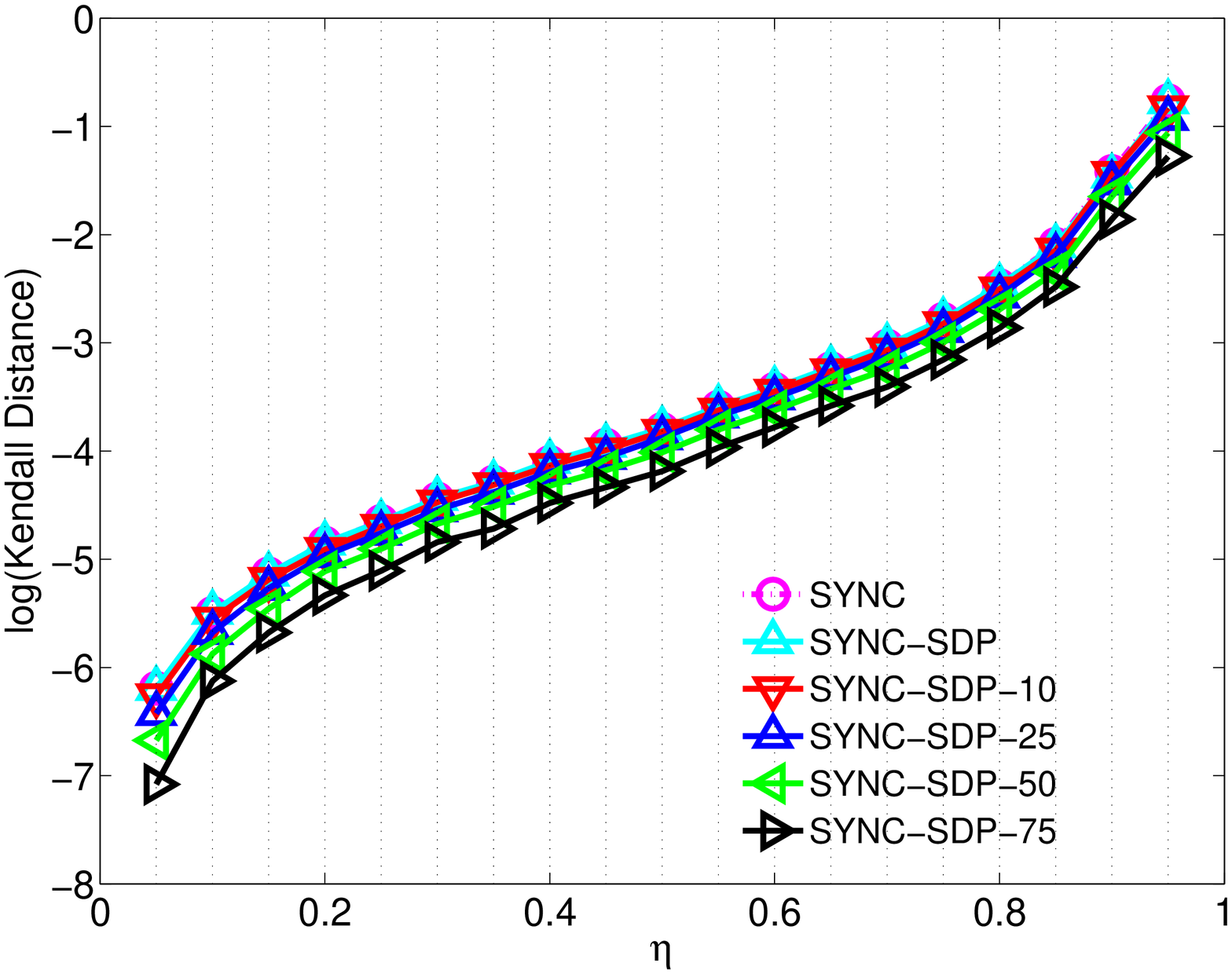}}
\end{center}
\caption{Numerical results for SDP-based synchronization ranking with anchor constraints, for the ERO and MUN noise models with $n=200$ nodes. SYNC-SDP-K denotes the SDP relaxation of the synchronization problem with $K$ anchors added. As expected, the numerical error decreases as more anchor information is available to the user. The effect of the anchor information is emphasized as we increase the amount of noise in the data. We average the results over 15 experiments.}
\label{fig:anchors_SDP}
\end{figure}

Note that, as an alternative, one could think of applying a cyclic permutation to the total ordering (i.e., permutation of size $n$) induced by the top eigenvector $v_1^{(\Upsilon)}$ until encountering the cyclic shift that positions the anchor nodes at their correct ranks, since the rank offsets between the anchors are imposed as hard constraints. However, as a word of  caution, we remark that this is not necessarily guaranteed to happen since the $\Upsilon$ matrix returned by the SDP program is not guaranteed to be of rank 1, and hence the hard constraints get perturbed in the rank-1 projection.

We detail in Figure  \ref{fig:anchors_SDP} the outcome of numerical simulations of Sync-Rank for several instances of the ranking with constraints problem, for the MUN($n=200,p,\eta$) model with cardinal measurements, and the ERO($n=200,p,\eta$) model with ordinal  measurements. We report on experiments where the number of anchors varies in the set $\{10,25,50,75\}$. As expected, the accuracy of the solution increases as the number of anchors increases.   Note that, in the case of cardinal measurements, having as little as 10 or 25 anchors leads  to significantly more accurate results, when compared to the SYNC-SDP, and especially the SYNC-EIG methods that do not account for anchor information.

\section{Future research directions}
\label{sec:varOpen}

We highlight in this section several possible future research directions that we believe are interesting and  worth pursuing further.

\subsection{Ranking with soft (edge) constraints as a generalized eigenvalue problem}

We have seen in the previous Section \ref{sec:constRanking} that adding hard constraints to the ranking problem improves the end results. A natural question to consider is how one may incorporate soft constraints into the problem formulation. Let $H$ denote the Hermitian matrix of available pairwise cardinal rank-offsets after embedding in the angular space, and similarly let $Q$ denote the Hermitian matrix of pairwise cardinal rank-offset \textit{constraints} the user wishes to enforce, also after mapping to the angular embedding space.
One possibility is to incorporate  weights into either the spectral or SDP relaxation of the synchronization problem, 
in which case, the input to the synchronization formulation will be given by
	$$ \tilde{H} = H + \lambda Q $$
where $Q$ is the matrix of rank-offset (soft) constraints. The larger the parameter $\lambda$, the more weight is given to the available pairwise constraints, and thus the more likely they are to be satisfied in the final proposed solution. 
However, in certain instances one may wish to enforce a lower bound on the how well the given constraints are satisfied, 
in which case we adjust the optimization problem as follows.  
We maximize the same quadratic form  $z^* H z $  as in (\ref{FinalRelaxAmitMaxi}), but subject to the condition that the given constraints are satisfied well enough, i.e., $z^* Q z \geq \alpha$, and that the sum of the absolute values of all complex numbers $z_i$ is $n$
\begin{equation}
	\begin{aligned}
	& \underset{\boldsymbol{z} = (z_1, \ldots, z_l) \in \mathbb{C}^n}{\text{max}}
	& & z^* H z  &  \\
	& \text{subject to}
	 & &  z^* Q z \geq \alpha   \\
	& & &  z^*  z = n.
	\end{aligned}
\label{max_zStarz}
\end{equation}Next, we consider the  associated Lagrange Multiplier, given by
\begin{equation}
  \mathcal{L}(z,\lambda, \mu) = z^* H z  + \lambda ( z^* Q z - \alpha ) + \mu (z^* z - n)
\end{equation}
and write down the KKT optimality conditions 
\begin{align}
\text{(Lagrangian Optimality)}  \;\;  &  H z  + \lambda  Q z  + \mu z  = 0\\
\text{(Primal Feasibility)}   \;\;    & z^* Q z = \alpha  \; \text{ and } \\
			                \;\;    &   z^*  z = n   \\
\text{(Dual Feasibility)}     \;\;    &  \lambda \geq 0   \\
\text{(Complementary Slackness)} \;\; &  \lambda ( z^* Q z - \alpha ) = 0  
\end{align}
We remark that this approach is extremely similar to the one considered in the constrained clustering algorithm proposed in \cite{WangQD14}, where the goal is to cluster a given weighted graph with adjacency matrix $H$, subject to soft  constraints captured in a  sparse matrix $Q$, of the form $Q_{ij}=1$ (respectively, $Q_{ij}=-1$) if nodes $i$ and $j$ should be (respectively, should not be) in the same cluster, and $Q_{ij}=0$ if no information is available. 
As shown in \cite{WangQD14}, one also arrives at a generalized eigenvalue problem 
\begin{equation}
   Hz = (-\lambda)  (Q + \frac{\beta}{n} I) z
\end{equation}
where $\beta = \frac{\mu}{\lambda}n$. 
Furthermore, instead of working with the above $H$ and $Q$ matrices (which in our setting (\ref{max_zStarz}) are Hermitian matrices) one could instead consider their associated 
graph Connection Laplacians, and solve the resulting generalized eigenvalue problem where both matrices are now diagonally dominant. 
We point out our recent work \cite{consClust}, where we present a principled spectral approach to the above well-studied constrained clustering problem, and reduce it to a generalized eigenvalue problem  involving Laplacian matrices that can be solved in nearly-linear time due to recent progress in the area of Laplacian linear systems solvers. In practice, this translates to a very fast implementation that consistently outperforms the approach in \cite{WangQD14}, both in terms of accuracy and running time. It would be interesting to explore whether our approach in \cite{consClust} can be extended to the setting of ranking with soft constraints.

\subsection{Semi-supervised ranking}  \label{sec:semiSupervised}
In the semi-supervised setting, one is allowed to enforce constraints on the retrieved ranking. The scenario when one wishes to enforce that the rank of a given player takes a certain value is an instance of \textbf{synchronization over $SO(2)$ with anchors}, and solved either via SDP, as shown in Section \ref{sec:constRanking}.
Furthermore, one could shrink the size of the SDP from $n \times n$ to $(n-k) \times (n-k)$ where $k$ is the number of anchors, and add to the objection function the pairwise comparisons between the anchors and non-anchor players. 
Alternatively, one could   also explore a method based on QCQP (Quadratically Constrained Quadratic Programming), similar to an approach we have previously investigated in \cite{asap3d} in the context of synchronization over $ \mathbb{Z}_2$ with anchors and its application to structural biology. 
An interesting variation one could further explore concerns the scenario when  one wishes to enforce \textbf{ordinal hard  constraints} in the form of a certain player $i$ being ranked higher than another player $j$, i.e., $ \hat{r}_i > \hat{r}_j$.


\subsection{(Signless) Ranking with Unsigned Cardinal Comparisons}  \label{sec:SignlessRanking}

Consider for example the scenario where one is given the score sheet of all soccer games in the England Football Premier League, recording the goal difference for each game, but without disclosing who won and who lost the game. In other words, one wishes to reconstruct the underlying  rankings (up to a global ordering) only using information on the magnitude of the rank offsets
\begin{equation}
	C_{ij} = | r_i - r_j |
\end{equation}
We remark that the resulting problem is nothing but an instance of the \textit{graph realization problem} on the line, with noisy distances \cite{asap2d}. In practice, an  instance of this problem arises in shotgun genome sequencing experiments, where the goal is to reorder substrings of cloned DNA strands (called \textit{reads}) using assembly algorithms that exploit the overlap between the reads. We refer the reader to the recent work of Fogel et al. on convex relaxations for the \textit{seriation problem} \cite{AspremontRelaxPermutation}, including an application to the above shotgun gene sequencing task.  
We remark that preliminary numerical results indicate that SVD-based ranking performs remarkably well in this setting of recovering the ordering (up to a global sign) using information only on the magnitude of the rank offsets but not their sign, and defer this investigation to future work. 

\subsection{Ranking both items and raters}
Another possible direction concerns the detection of outlier voters, in the setup when there are multiple voting systems providing pairwise  comparisons on the same set of players. For example, in Section \ref{sec:RankAggregation} 
we considered the problem of rank aggregation, of combining information from multiple voters or rating systems, which provide incomplete and inconsistent rankings or pairwise comparisons between the same set of players or items. A natural question asks whether it is possible to derive a reputation for both raters and items, and identify outlier voters of low credibility. 
Furthermore, one can incorporate the credibility of the voters when estimating the reputation of the objects, as considered in the recent work of  de Kerchove and Van Dooren \cite{deKerchovePaul}, who introduce a class of iterative voting systems that assign a reputation to both voters and items, by applying a filter to the votes.

\subsection{Ranking via $l_1$ Angular Synchronization}

Singer \cite{sync} and Yu \cite{StellaAE}   observe that embedding in the angular space is significantly more robust to outliers when compared to $l_1$ or $l_2$ based  embeddings in the linear space.
Since the existing spectral and SDP relaxations for the angular synchronization problem \cite{sync}  approximate a least squares solution in the angular space, a natural question to ask is whether an $l_1$-based formulation of the angular synchronization problem could further improve the robustness to noise or sparsity of the measurement graph.
A preliminary investigation of such an $l_1$ based approach for angular synchronization (which we did not further apply to the ranking problem) using the splitting orthogonality constraints approach of Lai and Osher \cite{LaiStanSOC} based on Bregman iterations, suggests that it often yields more accurate results than the existing spectral relaxation of its least-squares solution only for favorable noise regimes. At high levels of noise, the spectral relaxation performs better, and does so at a much lower computational cost. 






\subsection{Ranking over the Sphere and Synchronization over SO(3)}   
Another research direction we find very interesting but perhaps rather esoteric, concerns the possibility of casting the ranking problem as a synchronization problem over the group SO(3), as opposed to SO(2) which we considered in this paper. 
Consider for example a sphere centered at the origin, with points on the sphere corresponding to the players and the $z$-axis to their ranks, such that the closer a player is to the South Pole, the higher her or his ranking is.
The extra \textit{dimension} would  facilitate the extraction of  partial rankings, which one can collect by traversing along longitudinal lines and their immediate neighborhoods. In addition, one can also extract sets of players which are about the same rank (roughly speaking, they have about the same latitude), but cannot be positioned accurately with respect to each other (for example, a set of $10$ players out of $100$, is ranked in positions $81-90$, but based on the data, their relative ordering cannot be established).
In other words, if one considers a slice of the sphere (parallel to the $xy$-plane) all the players on or nearby it would have about the same rank. 
Perhaps most important, it would also be interesting to investigate the extent to which the extra dimension increases the overall robustness to noise of the algorithm. 
Note that, while for SO(2), finding an optimal circular permutation was required to eliminate the additional degree of freedom, for the case of SO(3) one could search for the best rotation around the origin of the sphere, such that the resulting rank offsets agree with the initial data as best as possible.

\subsection{Tighter bounds in terms of robustness to noise}  
Another future direction would be to aim for tighter guarantees in terms of robustness to noise. We remark that  existing theoretical guarantees from the recent literature on the group synchronization problem by various works of Bandeira, Boumal,  Charikar, Kennedy, Singer and Spielman \cite{sync,littleGrothendieck,tightSDPMLEsync,afonso}, 
trivially translate to lower bounds for the largest amount of noise permissible in the measurements, while still achieving exact or almost exact recovery of the underlying ground truth ranking. However, note that a perfect recovery of the angles in the angular synchronization problem is not a necessary condition for a perfect recovery of the underlying ground truth ranking, since it is enough that only the relative ordering of the angles is preserved.


\subsection{The Minimum Linear Arrangement Problem}
One could also investigate any potential connections between synchronization and SVD-based ranking using only magnitude information (as detailed in Section \ref{sec:SignlessRanking}) and the Minimum Linear Arrangement (MLA) problem, which is defined as follows. Given a graph $G=(V,E)$ and positive edge weights $w:E \mapsto \mathbb{R}_{+}$, a linear arrangement is a permutation $\pi : V \mapsto \{1,2,\ldots,n\}$. The cost of the arrangement is given by 
\begin{equation}
 \sum_{ij \in E} w(u,v) | \pi(u) - \pi(v) | 
\end{equation}
In the MLA problem, the goal is to find a linear arrangement of minimum cost a task known to be NP-complete. In the approximation theory literature, the work of Feige and Lee \cite{Feige_JamesLee2007} provides an $O(\sqrt{\log n} \log \log n )$-approximation SDP-based algorithm for the MLA problem. 

\section{Summary and Discussion} \label{sec:summaryDisc}

In this paper, we considered the problem of ranking with noisy incomplete information  and made an explicit connection with the angular synchronization problem, for which spectral and SDP relaxations already exist in the literature with provable guarantees. This approach leads to a computationally efficient (as is the case for the spectral relaxation), non-iterative algorithm that is model independent and relies exclusively on the available data.
We provided extensive numerical simulations, on both synthetic and real data sets (English Premier League soccer matches, a Microsoft Halo 2 tournament, and NCAA College Basketball matches), that show that our proposed procedure compares favorably with state-of-the-art methods from the literature, across a variety of measurement graphs and noise models, both for cardinal and ordinal pairwise  measurements.

We took advantage of the spectral, and especially SDP, relaxations and proposed methods for ranking in the semi-supervised setting where a subset of the players have a prescribed rank to be enforced as a hard constraint. In addition, we considered the popular rank aggregation problem of combining ranking information from multiple rating systems that provide independent incomplete and inconsistent pairwise measurements on the same set of items, with the goal of producing a single global ranking.

Aside from comparing our method to two recently proposed state-of-the-art algorithms, \textit{Serial-Rank} \cite{serialRank}, and \textit{Rank-Centrality} \cite{RankCentrality}, we have also compared to the more traditional Least-Squares method, and have also proposed a simple (Singular Value Decomposition) SVD-based ranking algorithm which we expect to be amenable to a theoretical analysis using tools from random matrix theory, and constitutes ongoing work. 
%
%
Finally, in Appendix A, we considered the problem of planted consistent partial rankings, which arises in the setting when a small subset of players have their pairwise measurements significantly more accurate than the rest of the network. We proposed an algorithm based on the first non-trivial eigenvector of the random-walk Laplacian associated to the residual matrix of rank offsets, which is able to extract the set of nodes in the hidden partial ranking, exploiting the fact that, unlike other methods, synchronization-based ranking is  able to preserve the ordering of the players associated to the planted partial ranking.

\section*{Acknowledgements}
The author would like to thank the Simons Institute for their warm hospitality and fellowship support during his stay in Berkeley throughout the Fall 2014 semester, when most of this research was carried out. He is very grateful to Amit Singer for suggesting, a few years back, the possibility of applying  angular synchronization to the ranking problem.  He would like to thank Andrea Bertozzi for her support via AFOSR MURI grant FA9550-10-1-0569, Alex d'Aspremont and Fajwel Fogel for useful discussions on their Serial-Rank algorithm and pointers to the relevant data sets,  
and Xiuyuan Cheng for discussions and references to the random matrix theory literature. 
Credits for the use of the Halo 2 Beta Dataset are given to Microsoft Research Ltd. and Bungie.

\bibliographystyle{siam}
\bibliography{main_SyncRank}

\begin{thebibliography}{10}

\bibitem{Alon98findinga}
{\sc N.~Alon, M.~Krivelevich, and B.~Sudakov}, {\em Finding a large hidden
  clique in a random graph.}, in SODA, H.~J. Karloff, ed., ACM/SIAM, 1998,
  pp.~594--598.

\bibitem{AmmarShah}
{\sc A.~Ammar and D.~Shah}, {\em Efficient rank aggregation using partial
  data}, in Proceedings of the 12th ACM SIGMETRICS/PERFORMANCE Joint
  International Conference on Measurement and Modeling of Computer Systems,
  SIGMETRICS '12, New York, NY, USA, 2012, ACM, pp.~355--366.

\bibitem{AndersonZeitouni}
{\sc G.~Anderson and O.~Zeitouni}, {\em A law of large numbers for finite-range
  dependent random matrices}, Communications on Pure and Applied Mathematics,
  61 (2008), pp.~1118--1154.

\bibitem{AtkinsSeriation}
{\sc J.~E. Atkins, E.~G. Boman, and B.~Hendrickson}, {\em A spectral algorithm
  for seriation and the consecutive ones problem}, SIAM Journal on Computing,
  28 (1998), pp.~297--310.

\bibitem{AzariSoufiani_nips13_b}
{\sc H.~{{Azari Soufiani}}, W.~Chen, D.~C. Parkes, and L.~Xia}, {\em
  {Generalized Method-of-Moments for Rank Aggregation}}, in Proceedings of the
  Annual Conference on Neural Information Processing Systems (NIPS 2013), 2013.

\bibitem{AzariSoufiani_icml14}
{\sc H.~{{Azari Soufiani}}, D.~Parkes, and L.~Xia}, {\em {Computing Parametric
  Ranking Models via Rank-Breaking}}, in Proceedings of the International
  Conference on Machine Learning (ICML 2014), 2014.

\bibitem{bandeira2014tightness}
{\sc A.~Bandeira, N.~Boumal, and A.~Singer}, {\em Tightness of the maximum
  likelihood semidefinite relaxation for angular synchronization}, arXiv
  preprint arXiv:1411.3272,  (2014).

\bibitem{littleGrothendieck}
{\sc A.~S. Bandeira, N.~Boumal, and A.~Singer}, {\em Approximating the little
  grothendieck problem over the orthogonal and unitary groups},
  arXiv:1308.5207.

\bibitem{tightSDPMLEsync}
{\sc A.~S. Bandeira, C.~Kennedy, and A.~Singer}, {\em Tightness of the maximum
  likelihood semidefinite relaxation for angular synchronization},
  arXiv:1411.3272.

\bibitem{afonso}
{\sc A.~S. Bandeira, A.~Singer, and D.~A. Spielman}, {\em A {C}heeger
  inequality for the graph {C}onnection {L}aplacian}, SIAM Journal on Matrix
  Analysis and Applications, 34 (2013), pp.~1611--1630.

\bibitem{Benaych_georges_thesingular}
{\sc F.~Benaych-Georges and R.~R. Nadakuditi}, {\em The singular values and
  vectors of low rank perturbations of large rectangular random matrices},
  Journal of Multivariate Analysis, 111 (2012), pp.~120--135.

\bibitem{Boyd_ADMM}
{\sc S.~Boyd, N.~Parikh, E.~Chu, B.~Peleato, and J.~Eckstein}, {\em Distributed
  optimization and statistical learning via the alternating direction method of
  multipliers}, Found. Trends Mach. Learn., 3 (2011), pp.~1--122.

\bibitem{BradleyTerry1952}
{\sc R.~A. Bradley and M.~E. Terry}, {\em Rank analysis of incomplete block
  designs: I. the method of paired comparisons}, Biometrika,  (1952),
  p.~324–345.

\bibitem{BravermanMossel}
{\sc M.~Braverman and E.~Mossel}, {\em Noisy sorting without resampling}, in
  Proceedings of the Nineteenth Annual ACM-SIAM Symposium on Discrete
  Algorithms, SODA '08, Philadelphia, PA, USA, 2008, Society for Industrial and
  Applied Mathematics, pp.~268--276.

\bibitem{Condorcet}
{\sc M.~Condorcet}, {\em {Essai sur l'application de l'analyse \`{a} la
  probabilit\'{e} des d\'{e}cisions rendues \`{a} la pluralit\'{e} des voix}},
  Imprimerie Royale, Paris, 1785.

\bibitem{cremonesi2010performance}
{\sc P.~Cremonesi, Y.~Koren, and R.~Turrin}, {\em Performance of recommender
  algorithms on top-n recommendation tasks}, in Proceedings of the fourth ACM
  conference on Recommender systems, ACM, 2010, pp.~39--46.

\bibitem{consClust}
{\sc M.~Cucuringu, I.~Koutis, and S.~Chawla}, {\em {Constrained Spectral
  Clustering on Massive Data Sets: A Generalized Laplacian Eigenproblem
  Perspective}}, submitted,  (2014).

\bibitem{asap2d}
{\sc M.~Cucuringu, Y.~Lipman, and A.~Singer}, {\em Sensor network localization
  by eigenvector synchronization over the {E}uclidean group}, ACM Trans. Sen.
  Netw., 8 (2012), pp.~19:1--19:42.

\bibitem{asap3d}
{\sc M.~Cucuringu, A.~Singer, and D.~Cowburn}, {\em Eigenvector
  synchronization, graph rigidity and the molecule problem}, Information and
  Inference, 1 (2012), pp.~21--67.

\bibitem{deKerchovePaul}
{\sc C.~de~Kerchove and P.~Van~Dooren}, {\em Iterative filtering in reputation
  systems}, SIAM J. Matrix Anal. Appl., 31 (2010), pp.~1812--1834.

\bibitem{Elo1978}
{\sc A.~Elo}, {\em The rating of Chess Players, Past and present}, Arco Pub,
  1978.

\bibitem{EstradaCookCLX}
{\sc E.~Estrada}, {\em {The Structure of Complex Networks : Theory and
  Applications}}, Oxford University Press, New York, 2012.

\bibitem{Feige_JamesLee2007}
{\sc U.~Feige and J.~R. Lee}, {\em An improved approximation ratio for the
  minimum linear arrangement problem}, Inf. Process. Lett., 101 (2007),
  pp.~26--29.

\bibitem{densestKsubgraphFeigeSeltzer}
{\sc U.~Feige and M.~Seltzer}, {\em On the densest k-subgraph problem},
  Technical report CS97-16, Tel-Aviv University, Dept. of Computer Science,
  (1997).

\bibitem{FeralPeche}
{\sc D.~F\'{e}ral and S.~P\'{e}ch\'{e}}, {\em The largest eigenvalue of rank
  one deformation of large {W}igner matrices}, Communications in Mathematical
  Physics, 272 (2007), pp.~185--228.

\bibitem{serialRank}
{\sc F.~Fogel, A.~d'Aspremont, and M.~Vojnovic}, {\em Serialrank: Spectral
  ranking using seriation}, in Advances in Neural Information Processing
  Systems 27, 2014, pp.~900--908.

\bibitem{AspremontRelaxPermutation}
{\sc F.~Fogel, R.~Jenatton, F.~Bach, and A.~d'Aspremont}, {\em Convex
  relaxations for permutation problems.}, in NIPS, C.~J.~C. Burges, L.~Bottou,
  Z.~Ghahramani, and K.~Q. Weinberger, eds., 2013, pp.~1016--1024.

\bibitem{FreundSchapireSinger}
{\sc Y.~Freund, R.~Iyer, R.~E. Schapire, and Y.~Singer}, {\em An efficient
  boosting algorithm for combining preferences}, J. Mach. Learn. Res., 4
  (2003), pp.~933--969.

\bibitem{RankingDynamicNetworks}
{\sc R.~Ghosh, T.-T. Kuo, C.-N. Hsu, S.-D. Lin, and K.~Lerman}, {\em Time-aware
  ranking in dynamic citation networks}, in Data Mining Workshops (ICDMW), 2011
  IEEE 11th International Conference on, Dec 2011, pp.~373--380.

\bibitem{giles1998citeseer}
{\sc C.~L. Giles, K.~D. Bollacker, and S.~Lawrence}, {\em Citeseer: An
  automatic citation indexing system}, in Proceedings of the third ACM
  conference on Digital libraries, ACM, 1998, pp.~89--98.

\bibitem{Gleich04svdbased}
{\sc D.~Gleich}, {\em {SVD} based term suggestion and ranking system}, in In:
  IEEE International Conference on Data Mining, 2004, pp.~391--394.

\bibitem{harville1977use}
{\sc D.~Harville}, {\em The use of linear-model methodology to rate high school
  or college football teams}, Journal of the American Statistical Association,
  72 (1977), pp.~278--289.

\bibitem{hirani2010least}
{\sc A.~N. Hirani, K.~Kalyanaraman, and S.~Watts}, {\em Least squares ranking
  on graphs}, arXiv preprint arXiv:1011.1716,  (2010).

\bibitem{horton2010labor}
{\sc J.~J. Horton and L.~B. Chilton}, {\em The labor economics of paid
  crowdsourcing}, in Proceedings of the 11th ACM conference on Electronic
  commerce, ACM, 2010, pp.~209--218.

\bibitem{HuberRowSum1963}
{\sc P.~J. Huber}, {\em Pairwise comparison and ranking: Optimum properties of
  the row sum procedure}, The Annals of Mathematical Statistics, 34 (1963),
  pp.~511--520.

\bibitem{JamiesonNowak2011}
{\sc K.~G. Jamieson and R.~D. Nowak}, {\em Active ranking using pairwise
  comparisons}, in NIPS, J.~Shawe-Taylor, R.~S. Zemel, P.~L. Bartlett, F.~C.~N.
  Pereira, and K.~Q. Weinberger, eds., 2011, pp.~2240--2248.

\bibitem{HodgeRanking}
{\sc X.~Jiang, L.-H. Lim, Y.~Yao, and Y.~Ye}, {\em Statistical ranking and
  combinatorial hodge theory}, Mathematical Programming, 127 (2011),
  pp.~203--244.

\bibitem{KendallSmith1940}
{\sc M.~G. Kendall and B.~B. Smith}, {\em On the method of paired comparisons},
  Biometrika, 31 (1940), pp.~324--345.

\bibitem{KenyonMathieu}
{\sc C.~Kenyon-Mathieu and W.~Schudy}, {\em How to rank with few errors}, in
  Proceedings of the Thirty-ninth Annual ACM Symposium on Theory of Computing,
  STOC '07, New York, NY, USA, 2007, ACM, pp.~95--103.

\bibitem{KleinberHITS}
{\sc J.~M. Kleinberg}, {\em Authoritative sources in a hyperlinked
  environment}, J. ACM, 46 (1999), pp.~604--632.

\bibitem{koehler1982application}
{\sc K.~J. Koehler and H.~Ridpath}, {\em An application of a biased version of
  the bradley-terry-luce model to professional basketball results}, Journal of
  Mathematical Psychology, 25 (1982), pp.~187--205.

\bibitem{LaiStanSOC}
{\sc R.~Lai and S.~Osher}, {\em A splitting method for orthogonality
  constrained problems}, J. Sci. Comput., 58 (2014), pp.~431--449.

\bibitem{langville2011google}
{\sc A.~N. Langville and C.~D. Meyer}, {\em Google's PageRank and beyond: The
  science of search engine rankings}, Princeton University Press, 2011.

\bibitem{LeeMultiWay}
{\sc J.~R. Lee, S.~Oveis~Gharan, and L.~Trevisan}, {\em Multi-way spectral
  partitioning and higher-order cheeger inequalities}, in Proceedings of the
  Forty-fourth Annual ACM Symposium on Theory of Computing, STOC '12, New York,
  NY, USA, 2012, ACM, pp.~1117--1130.

\bibitem{LiuLearningToRank}
{\sc T.-Y. Liu}, {\em Learning to rank for information retrieval}, Found.
  Trends Inf. Retr., 3 (2009), pp.~225--331.

\bibitem{bookLiuLearningToRank}
\leavevmode\vrule height 2pt depth -1.6pt width 23pt, {\em Learning to Rank for
  Information Retrieval}, Springer, 2011.

\bibitem{Luce1959individual}
{\sc R.~D. Luce}, {\em Individual Choice Behavior a Theoretical Analysis}, john
  Wiley and Sons, 1959.

\bibitem{MaReciprocal}
{\sc M.~M.}, {\em A matrix approach to asset pricing in foreign exchange
  market'}, preprint,  (2008).

\bibitem{Mallows1957}
{\sc C.~Mallows}, {\em Non null ranking models {I}}, Biometrika, 44 (1957),
  pp.~114--130.

\bibitem{BrainMasonJCN}
{\sc A.~V. Mantzaris, D.~S. Bassett, N.~F. Wymbs, E.~Estrada, M.~A. Porter,
  P.~J. Mucha, S.~T. Grafton, and D.~J. Higham}, {\em Dynamic network
  centrality summarizes learning in the human brain}, Journal of Complex
  Networks, 1 (2013), pp.~83--92.

\bibitem{RankCentrality}
{\sc S.~Negahban, S.~Oh, and D.~Shah}, {\em Iterative ranking from pair-wise
  comparisons}, in Advances in Neural Information Processing Systems 25, 2012,
  pp.~2474--2482.

\bibitem{booknewman}
{\sc M.~E.~J. Newman}, {\em Networks: An Introduction}, Oxford University
  Press, USA, 2010.

\bibitem{osting2012statistical}
{\sc B.~Osting, J.~Darbon, and S.~Osher}, {\em Statistical ranking using the
  l1-norm on graphs}, vol.~7, 2013, pp.~907--926.

\bibitem{Pageetal98}
{\sc L.~Page, S.~Brin, R.~Motwani, and T.~Winograd}, {\em The pagerank citation
  ranking: Bringing order to the web}, in Proceedings of the 7th International
  World Wide Web Conference, 1998, pp.~161--172.

\bibitem{raykar2011ranking}
{\sc V.~C. Raykar and S.~Yu}, {\em Ranking annotators for crowdsourced labeling
  tasks}, in Advances in neural information processing systems, 2011,
  pp.~1809--1817.

\bibitem{SaatyReciprocal}
{\sc T.~L. Saaty}, {\em {A scaling method for priorities in hierarchical
  structures}}, Journal of Mathematical Psychology, 15 (1977), pp.~234--281.

\bibitem{sync}
{\sc A.~Singer}, {\em Angular synchronization by eigenvectors and semidefinite
  programming}, Appl. Comput. Harmon. Anal., 30 (2011), pp.~20--36.

\bibitem{AmitVDM}
{\sc A.~Singer and H.-T. Wu}, {\em Vector diffusion maps and the connection
  laplacian}, Communications on Pure and Applied Mathematics, 65 (2012),
  pp.~1067--1144.

\bibitem{talluri2006theory}
{\sc K.~T. Talluri and G.~J. Van~Ryzin}, {\em The theory and practice of
  revenue management}, vol.~68, springer, 2006.

\bibitem{RUMmodel}
{\sc L.~Thurstone}, {\em A law of comparative judgement}, Psychological Review,
  34 (1927), pp.~278--286.

\bibitem{Vandenberghe94sdp}
{\sc L.~Vandenberghe and S.~Boyd}, {\em Semidefinite programming}, SIAM Review,
  38 (1994), pp.~49--95.

\bibitem{WangQD14}
{\sc X.~Wang, B.~Qian, and I.~Davidson}, {\em On constrained spectral
  clustering and its applications}, Data Min. Knowl. Discov., 28 (2014),
  pp.~1--30.

\bibitem{WassermanSocialBook}
{\sc S.~Wasserman and K.~Faust}, {\em Social Network Analysis: Methods and
  Applications}, Cambridge University Press, 1994.

\bibitem{RankingWauthierJordan}
{\sc F.~L. Wauthier, M.~I. Jordan, and N.~Jojic}, {\em Efficient ranking from
  pairwise comparisons.}, in ICML (3), vol.~28 of JMLR Proceedings, JMLR.org,
  2013, pp.~109--117.

\bibitem{xiong2003reputation}
{\sc L.~Xiong and L.~Liu}, {\em A reputation-based trust model for peer-to-peer
  e-commerce communities}, in E-Commerce, 2003. CEC 2003. IEEE International
  Conference on, IEEE, 2003, pp.~275--284.

\bibitem{StellaAE}
{\sc S.~Yu}, {\em Angular embedding: A robust quadratic criterion.}, IEEE
  Trans. Pattern Anal. Mach. Intell., 34 (2012), pp.~158--173.

\bibitem{ADMM_Sparse_Low_Rank}
{\sc K.~Zhou, H.~Zha, and L.~Song}, {\em Learning social infectivity in sparse
  low-rank networks using multi-dimensional hawkes processes.}, in AISTATS,
  vol.~31 of JMLR Proceedings, JMLR.org, 2013, pp.~641--649.

\end{thebibliography}

\clearpage

\section{ Appendix A: Recovering Planted Locally-Consistent Partial Rankings} 
\label{sec:PartialRank}

Aside from its increased robustness to noise, we point out another appealing feature of the Sync-Rank algorithm. In many real-life scenarios, one is not necessarily interested in recovering the entire ranking, but only the ranking of a subset of the players.
Often, the noise in the available measurements may not be distributed uniformly throughout the network, with part of the network containing pairwise comparisons that are a lot less noisy than those in the rest of the network. This asymmetry provides us with an opportunity and raises the question of whether one can recover partial rankings which are locally consistent with the given data. In this Appendix we demonstrate empirically that Sync-Rank is able to preserve such partial rankings. In addition, we  extract such hidden locally consistent partial rankings, relying on a spectral algorithm in the spirit of existing algorithms for the planted clique and densest subgraph problems.

We remark that this approach can be extended to extracting multiple partial rankings by relying on recent work of Lee et al. on multi-way spectral partitioning and higher-order Cheeger inequalities \cite{LeeMultiWay}. Note that in the case of $k$ partial rankings, one does not necessarily seek a partition of the residual graph into $k$ clusters, but rather seeks $k$ disjoint clusters, whose union is not necessarily the entire graph.  The \textit{Cheeger Sweep} final step of their algorithm allows one to extract $k$ such sets of smallest expansion. For simplicity, we only consider the case of a single planted locally consistent partial ranking. We denote by $ \Lambda$ the set of players whose pairwise ranking measurements contain  less noise than the rest of the available data, and we shall refer to such players as 
\textit{$ \Lambda$-players}, and to the remaining ones as \textit{$ \Lambda^C$-players}.

We evaluate our proposed procedure on graphs generated from the following ensemble $\mathcal{G}(n,p, \beta,\eta_1, \eta_2)$, where the measurement graph $G$ is an Erd\H{o}s-R\'{e}nyi random graph $G(n,p)$, $| \Lambda | = \beta n$, and the noise model is given by the following mixture
\begin{equation}
C_{ij} = \left\{
 \begin{array}{rll}
 r_i - r_j & \;\; \{i,j\} \in \Lambda, \text{ for a correct edge} & \text{ with probability } (1-\eta_1) \alpha \\
 \sim \text{Uniform}[-(n-1), n-1]  & \;\; \{i,j\} \in \Lambda  \text{ for an incorrect edge} & \text{ with probability } \eta_1 \alpha	\\
 r_i - r_j & \;\; \{i,j\} \notin \Lambda, \text{ for a correct edge} & \text{ with probability } (1-\eta_2) \alpha \\
 \sim \text{Uniform}[-(n-1), n-1]  & \;\; \{i,j\} \notin \Lambda & \text{ with probability }  \eta_2 \alpha	\\
  0      & \;\; \text{ for a missing edge},   & \text{ with probability } 1-\alpha.  	\\
     \end{array}
   \right.
\label{plantedERoutliers}
\end{equation}

The approach we propose for recovering such a locally consistent partial ranking  starts as follows. For a given ranking method of choice (such as the ones we have considered throughout the paper),   
we estimate the complete ranking $\hat{r}_1, \ldots, \hat{r}_n$, and consider the resulting matrix of rank offsets
\begin{equation}
\hat{C} =  \left( \hat{r}  \otimes \mb{1}^T -  \mb{1}  \otimes \hat{r}^T  \right) \circ A
\end{equation}
where $\otimes$ denotes the outer product of two vectors $ x \otimes y = x y^T$, 
$\circ$ denotes the Hadamard product of two matrices (entrywise product), and $A$ is the adjacency matrix of the graph $G$; in other words $	\hat{C}_{ij} = \hat{r}_i - \hat{r}_j , (i,j) \in E(G)$.  Next, we consider the \textit{residual matrix} of pairwise rank offsets
\begin{equation}
	R = | C - \hat{C} |
	\label{def:ResidMtxCard}
\end{equation}
for the case of cardinal measurements, and 
\begin{equation}
	R = | C - \mbox{ sign} (\hat{C}) |
\end{equation}
for the case of ordinal measurements. 
\begin{figure}[h!]
\begin{center}
\subfigure[ SVD ]{\includegraphics[width=0.161\textwidth]{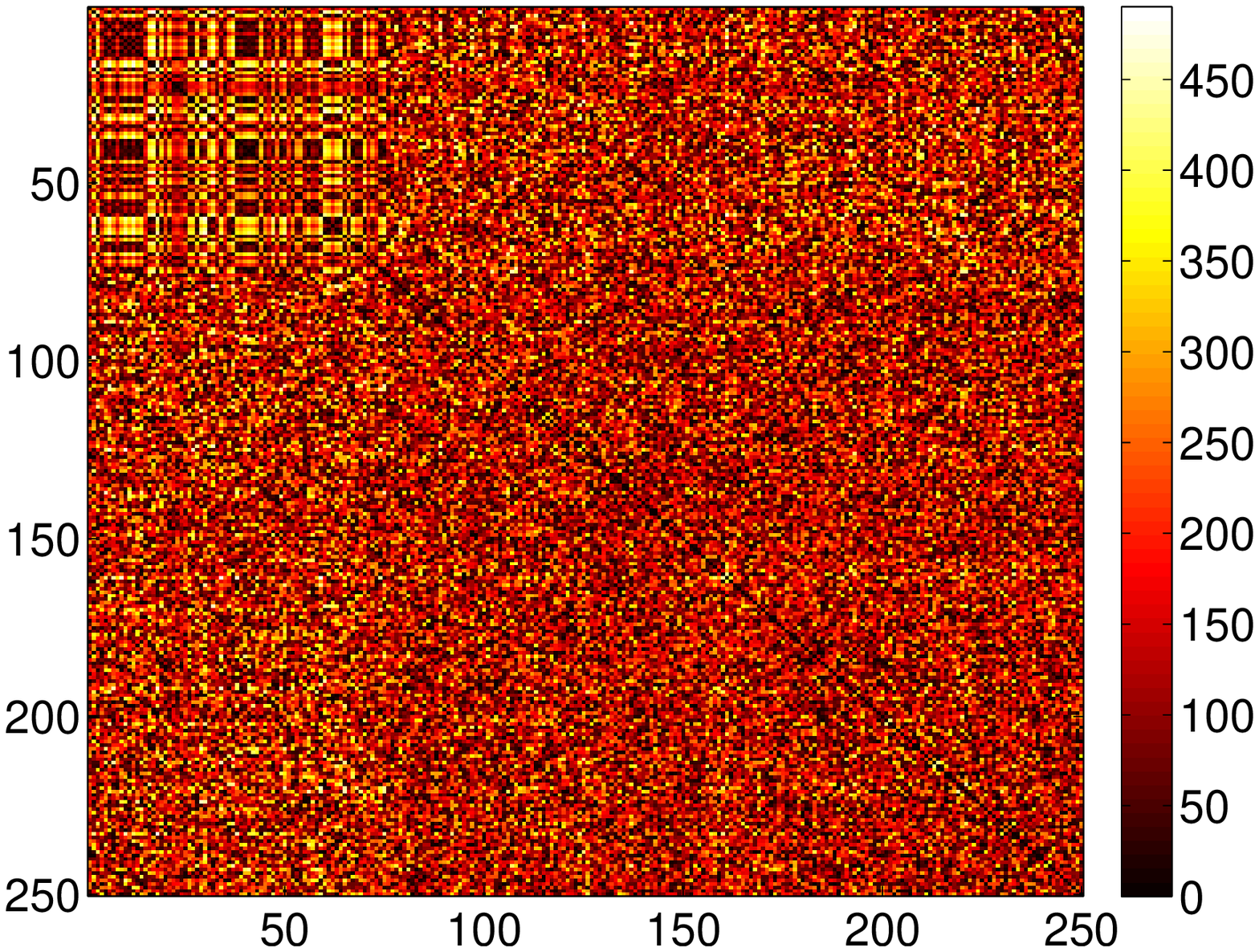}}
\subfigure[ LS ]{\includegraphics[width=0.161\textwidth]{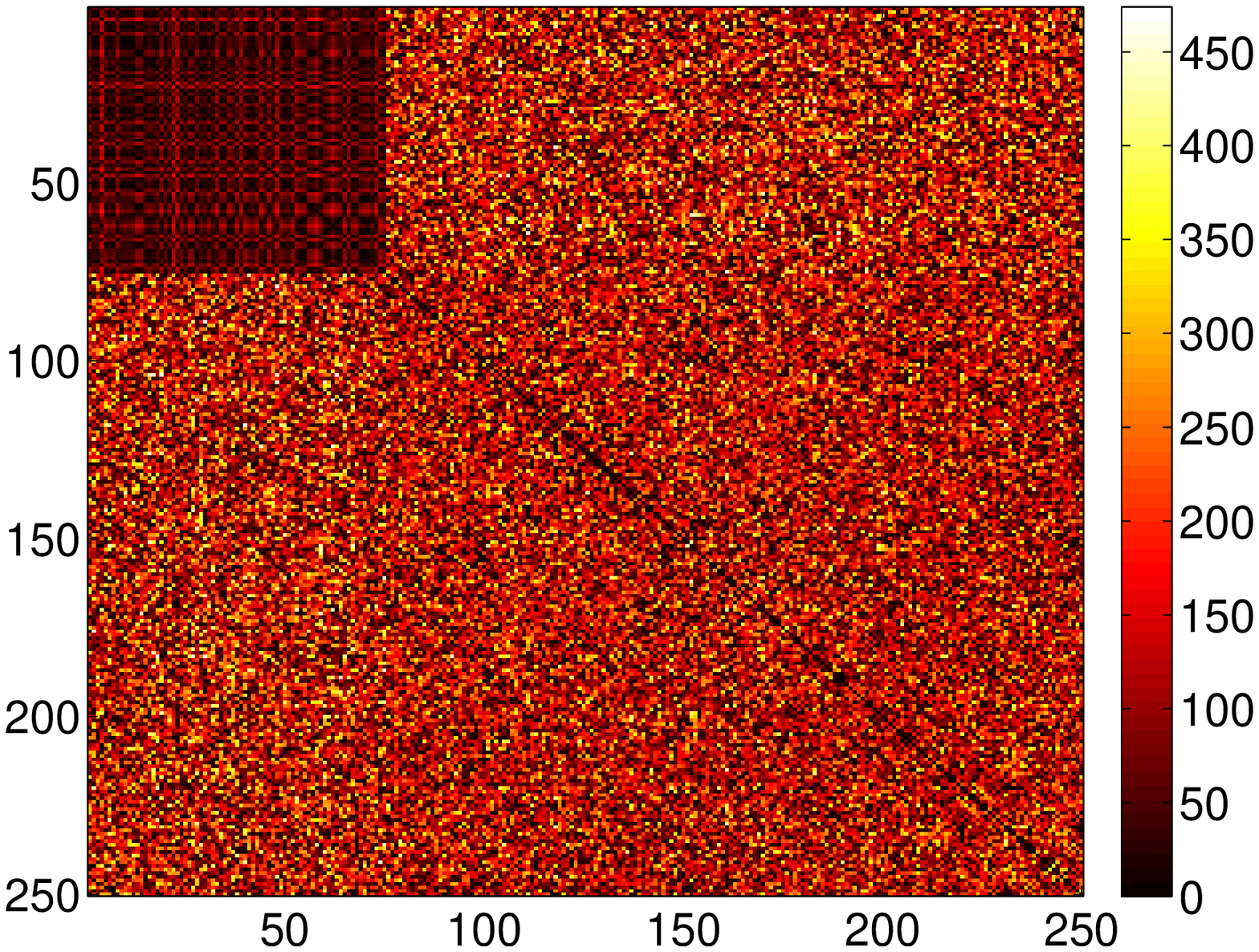}}
\subfigure[ SER ]{\includegraphics[width=0.161\textwidth]{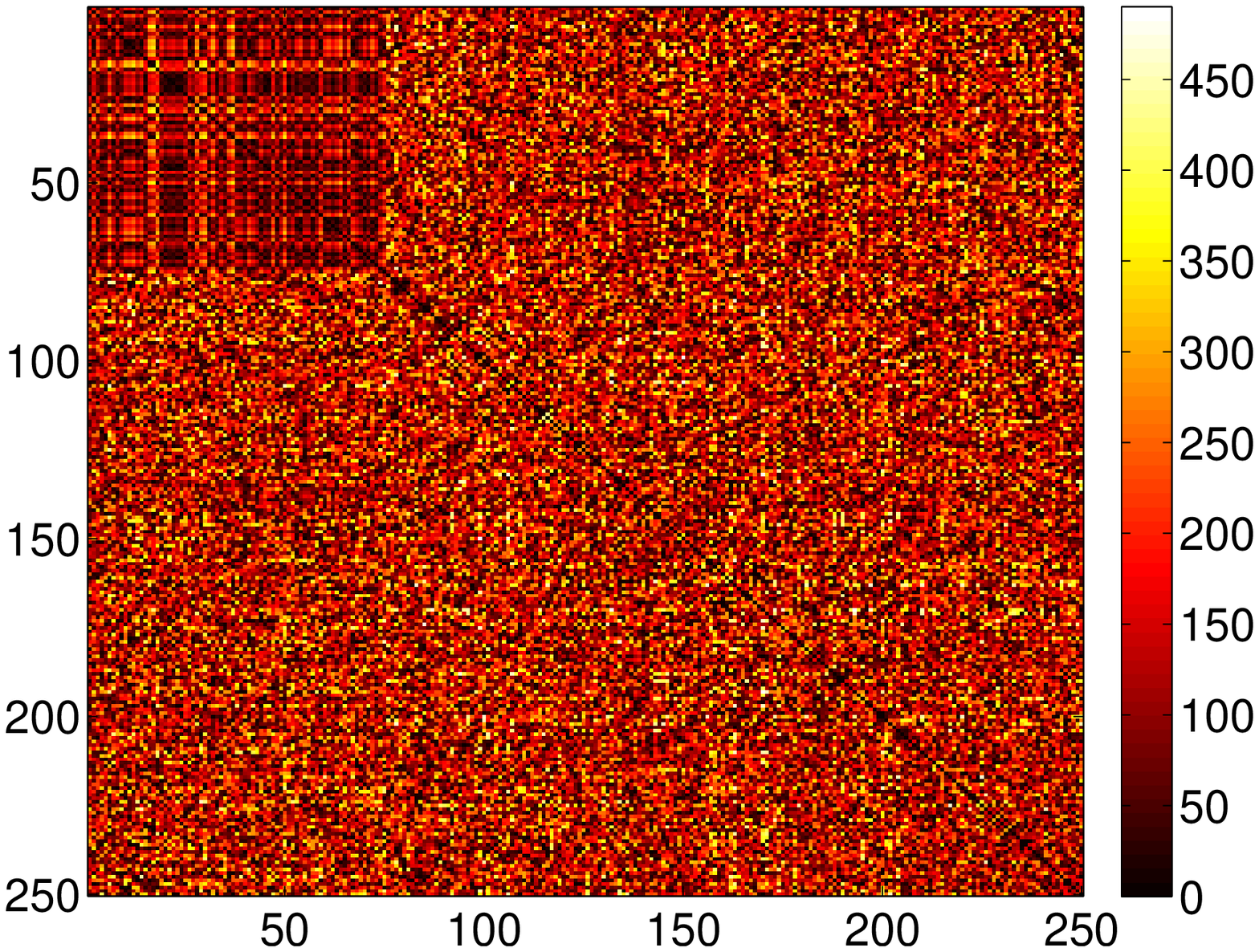}}
\subfigure[ RC ]{\includegraphics[width=0.161\textwidth]{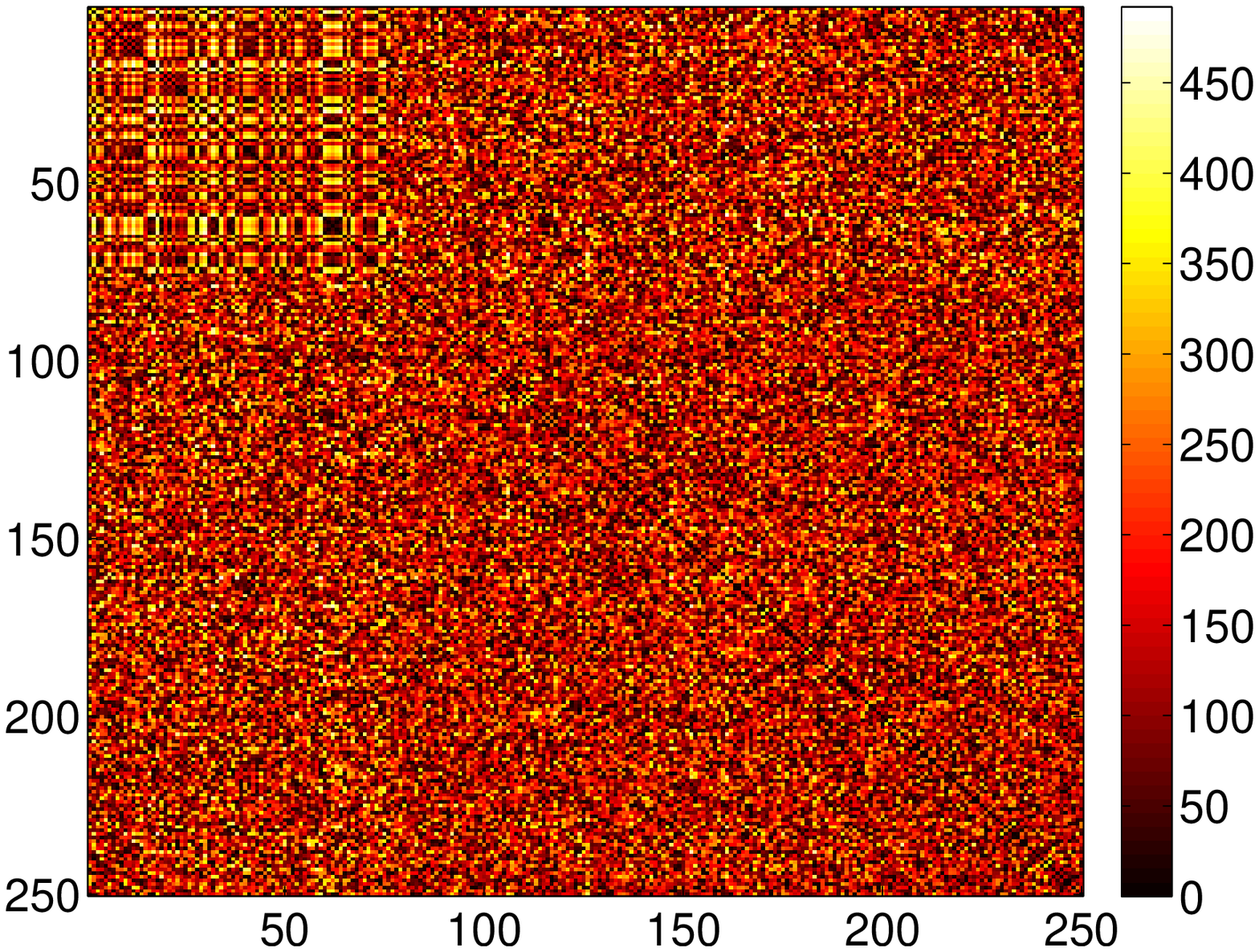}}
\subfigure[SYNC-EIG]{\includegraphics[width=0.161\textwidth]{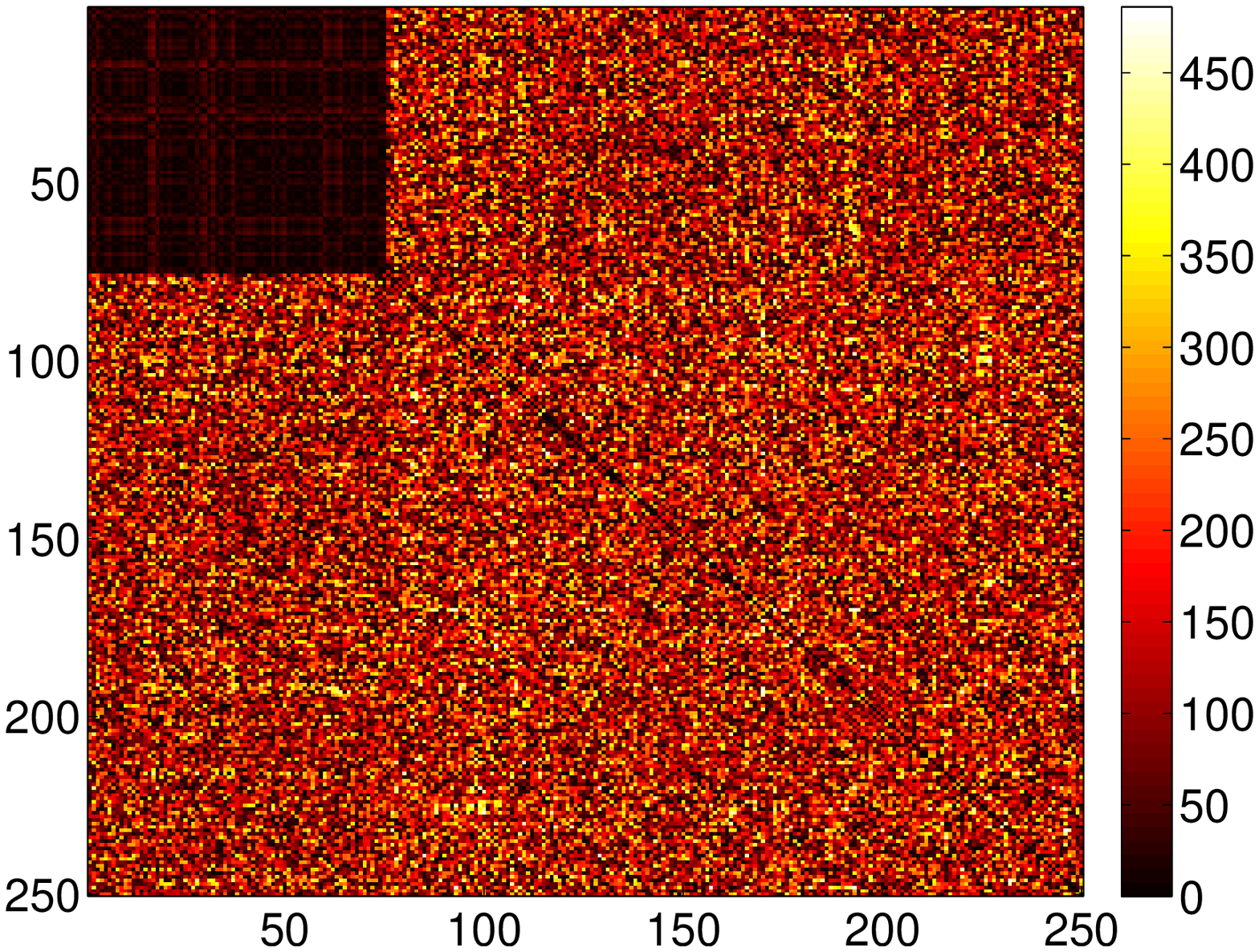}}
\subfigure[SYNC-SDP]{\includegraphics[width=0.161\textwidth]{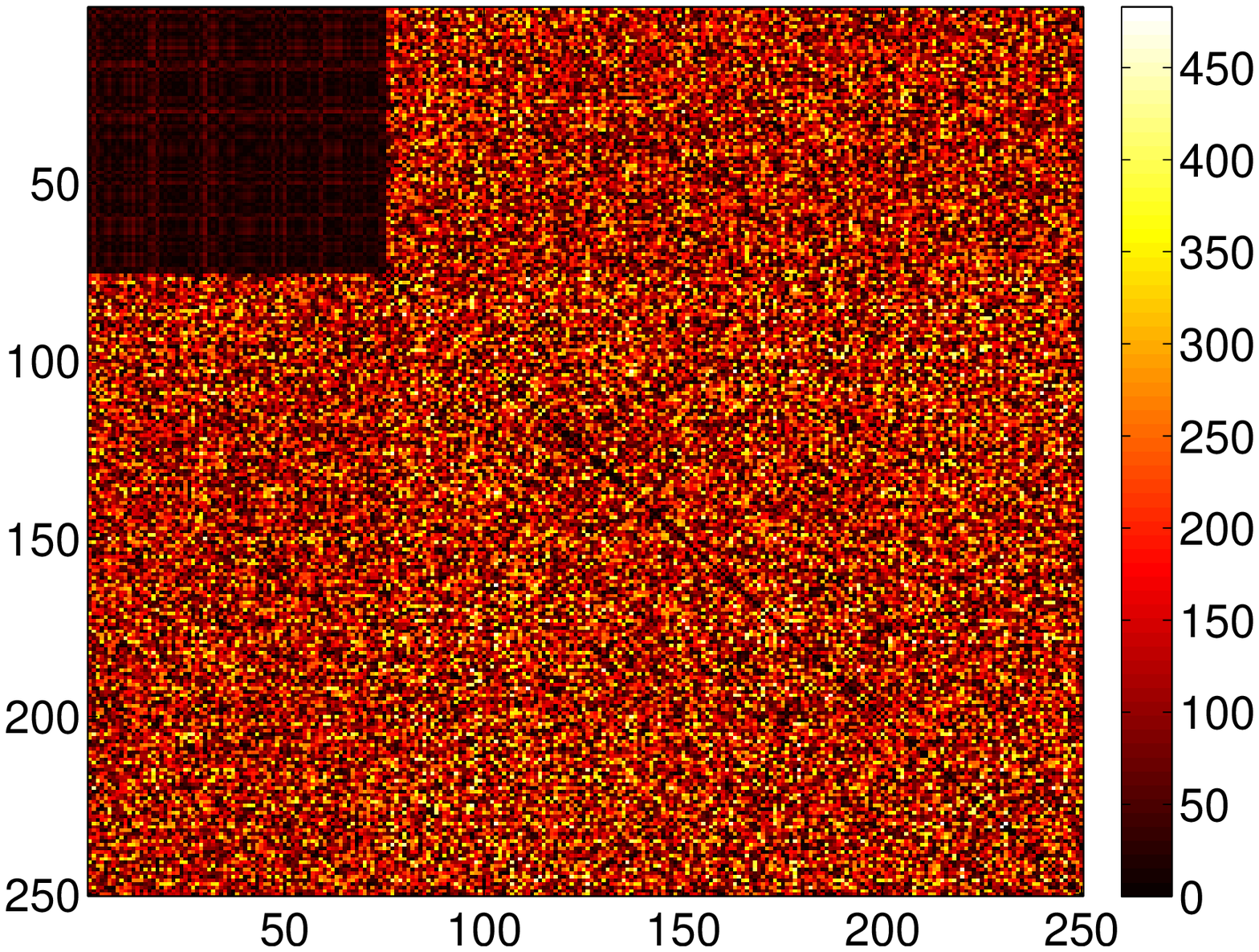}}
\caption{The residual matrices given by (\ref{def:ResidMtxCard}), for each of the six methods, for a random instance of the ensemble $\mathcal{G}(n=250, \beta=0.3, \eta_1=0, \eta_2=1)$. The $\Lambda$-nodes correspond to the first $ \beta n = 75 $ positions.}
\end{center}
\label{fig:ResidualMatrices}
\end{figure}
We illustrate in Figure \ref{fig:ResidualMatrices} the   residual matrices obtained by each of the methods, and remark that, for ease of visualization,  
the set $\Lambda$ consists of the first $ \beta n = 75 $ nodes, corresponding to the top left corner of each of the residual matrices shown in the heatmap.
Whenever an estimated rank-offset $\hat{C}_{ij}$ (induced by the recovered ranking)  matches very well with the initial measurement $C_{ij}$, we expect the residual $R_{ij}$ to have small magnitude, and conversely, whenever the recovered offset $\hat{C}_{ij}$ is far from the initial measurement $C_{ij}$ 
then we expect $R_{ij}$ to have large magnitude. Furthermore, we show in Figure  \ref{fig:BarplotRankingsExample}, the recovered rankings by each of the methods, where we separate the rankings of the   $\Lambda^C$-players (shown in the top subplot, in blue), from those of the $\Lambda$-players (shown in the bottom subplot, in red). Note that the set $\Lambda$ is not available to the user, and at this point  we do not  have yet an estimate $\hat{\Lambda}$ for $\Lambda$; however this plot already highlights the fact that synchronization-based ranking preserves almost perfectly the relative ranking of the $\Lambda$-players, while all the other methods fail to do so.  Note that the title in the bottom plots of each sub-Figure in \ref{fig:BarplotRankingsExample}, compute the Jaccard Index between   $\Lambda$ and $\hat{\Lambda}$ (which we show next how to estimate), and the Kendall distance between the ground truth rankings of the $\hat{\Lambda}$-players, and the estimated rankings of the $\hat{\Lambda}$-players.


In practice, when the measurements between pairs of $\Lambda$-nodes also contain noise, the corresponding sub-matrix has no longer zero or close to zero residuals, and it is harder to identify such a set of nodes. The resulting task at hand is to identify a subset of nodes for which the average inter-edge weights (the residuals shown in Figure \ref{fig:ResidualMatrices}) is as small as possible. This problem is equivalent\footnote{Perhaps after hard thresholding the entries of the residual matrix to 0 or 1.} to the well-known \textit{densest subgraph} problem, investigated in the theoretical computer science literature, for the case when the graph $G$ is unweighted. 
\textit{Densest-k-Subgraph} (DkS) on an undirected unweighted graph $G$ concerns finding a subset of nodes $U \in V(G)$ of size $|U|=k$ with the maximum induced average degree. Using a reduction from the Max-Clique problem, the \textit{Densest-k-Subgraph}  can be shown to be NP-hard to solve exactly. Feige and Seltzer   \cite{densestKsubgraphFeigeSeltzer} showed that the \textit{Densest-k-Subgraph} problem is NP-complete even when restricted to bipartite graphs of maximum degree $3$.
If the parameter $k$ is not known apriori, the problem becomes the well known \textit{Densest Subgraph} problem, where one seeks to find a subset $U$ (regardless of its size) in order to maximize the average degree induced by $U$. We note that the \textit{Densest Subgraph} problem can be solved in polynomial time using either linear programming 
or flow techniques.

In a seminal paper \cite{Alon98findinga}, Alon, Krivelevich, and Sudakov considered the graph ensemble $G(n,1/2,k)$, where one starts with an Erd\H{o}s-R\'{e}nyi $G(n,p=1/2)$ and randomly places  a clique of size $k$ in $G$. They proposed an efficient spectral algorithm that relies on the second eigenvector of the adjacency matrix of $G$, which
almost surely finds the clique of size $k$, as long as $k > c \sqrt{n}$. 
We remark that in our case the residual graph is weighted and thus a different procedure is required, though one can perhaps think of various heuristics for thresholding the entries of W, setting to 1 all remaining weights (thus rendering the graph unweighted), and perhaps further adding edges to the graph between r-hop neighbors away pairs of nodes, with the hope of producing a clique, which we can then detect using the algorithm proposed in \cite{Alon98findinga}.


However, in this work, we take none of the above approaches, and rely on another spectral algorithm to uncover the planted dense subgraph, and hence the partial ranking.  We use the following approach to uncover the planted densest subgraph of given  size $k$. We first compute a similarity between the adjacent nodes of the graph using the Gaussian Kernel
$$ W_{ij} = e^{ - \frac{R_{ij}}{\epsilon^2}}, (i,j)  \in E(G) $$
for the parameter choice $ \epsilon^2 = 2n/10 $, and consider the random-walk Laplacian  $L = D^{-1} W$, whose trivial eigenvalue-eigenvector pair is given by $\lambda_1=1$ and the all-ones vector $v_1=\mb{1}$. To extract a cluster of given size $k$ (our proposed estimate for the set $\Lambda$), we compute the first non-trivial eigenvector $v_2$ of $L$, and consider its top $k$ largest entries to produce an  estimate $\hat{\Lambda}$ for the set $\Lambda$. We detail these steps in Algorithm \ref{Algo:densestSubgrph}.
\begin{algorithm}[h!]
\begin{algorithmic}[1]
\REQUIRE A weighted graph $H=(V,E)$, and its symmetric adjacency matrix $W$, and $k < n$
\STATE Compute the random-walk Laplacian $L = D^{-1} W$, where $D$ is a diagonal matrix with $D_{ii} = \sum_{j=1}^{n} W_{ij} $ 
\STATE Find the second eigenvector $v_2$ of $L$, corresponding to the second largest eigenvalue $\lambda_2 < \lambda_1=1 $ 
\STATE Sort the entries of $v_2$ in decreasing order, consider the top $k$  largest entries in $v_2$, let $\Lambda^d$ denote the set of corresponding vertices, and compute the residual error associated to $\Lambda^d$
	$$ ERR^{\Lambda_d} = \norm{C_{\Lambda^d,\Lambda^d} - \hat{C}_{\Lambda^d,\Lambda^d} }_{1}$$
\STATE Sort the entries of $v_2$ in increasing order, consider the top $k$  largest entries in $v_2$, let $\Lambda^i$ denote the set of corresponding vertices, and compute the residual error associated to $\Lambda^i$
	$$ ERR^{\Lambda_i} = \norm{ C_{\Lambda^i,\Lambda^i} - \hat{C}_{\Lambda^i,\Lambda^i} }_{1} $$
\STATE  Output the final estimate $\hat{\Lambda} $ for  $\Lambda$ as
\begin{equation}
\hat{\Lambda} =  \underset{ \Lambda^d, \Lambda^i }{\text{arg min} }  \;\;\;  \{ ERR^{\Lambda_d}, ERR^{\Lambda_i} \} 
\label{argMinQdensest}
\end{equation}
\end{algorithmic}
\caption{Algorithm for recovering the densest (weighted) subgraph of size $k$ of the residual graph whose adjacency matrix is denoted by $W$.}
\label{Algo:densestSubgrph}
\end{algorithm}
Figure \ref{fig:New_JI_LocalRankComp} shows a comparison of the methods (results averaged across 15 experiments) in terms of the Jaccard Similarity Index between the ground truth set $\Lambda$ and our proposed estimate $\hat{\Lambda}$, 
 defined as 
\begin{equation}
	\mathcal{J}(\Lambda,\hat{\Lambda}) = \frac{\Lambda \cap \hat{\Lambda}}{\Lambda \cup \hat{\Lambda}},
\end{equation}
meaning that for values equal to $1$ the algorithm would perfectly recover the planted nodes in $\Lambda$.
Furthermore, we plot in Figure \ref{fig:New_KD_LocalRankComp} the corresponding 
Kendall distance between the recovered ranking of the $\hat{\Lambda}$-nodes (from Figure \ref{fig:New_JI_LocalRankComp}) and their ground truth values, 
for the ensemble given by $\mathcal{G}(n=250,p,\beta,\eta_1, \eta_2=1)$, for varying $p = \{0.2, 0.5, 1 \}$ and  $ \eta_1 = \{0, 0.2, 0.4\}$. Note that across all experiments, the SYNC-SDP algorithm is by far the best performer, both in terms of the Jaccard Index (thus being able to recover the set $\Lambda$) and the Kendall distance (thus being able to recover accurately the ranking\footnote{This partial ranking is simply extracted from the complete ranking solution, as computed using the method of choice.} of the estimated $\hat{\Lambda}$-nodes).  
Note that an alternative approach to estimating the ranking of the  $\hat{\Lambda}$-nodes (once $\hat{\Lambda}$ has already been estimated) would be to recompute it by re-running the method of choice on the original measurement matrix $C$, restricted to the sub-matrix corresponding to nodes $\hat{\Lambda}$. We expect this approach to yield more accurate results, especially in the case when the measurements between a pair of $(\Lambda^C,\Lambda^C)$-nodes, or a pair of $(\Lambda,\Lambda^C)$-nodes are very noisy (i.e., $\eta_2$ is large in the ensemble $\mathcal{G}(n,p,\beta,\eta_1, \eta_2)$ defined by (\ref{plantedERoutliers})), as was the case for the numerical experiments detailed here, for which we have chosen $\eta_2 = 1$, meaning such measurements are pure noise and chosen uniformly at random from $[-(n-1), (n-1)]$.

\begin{figure}[h!]
\begin{center}
\subfigure[ SVD ]{\includegraphics[width=0.32324\textwidth]{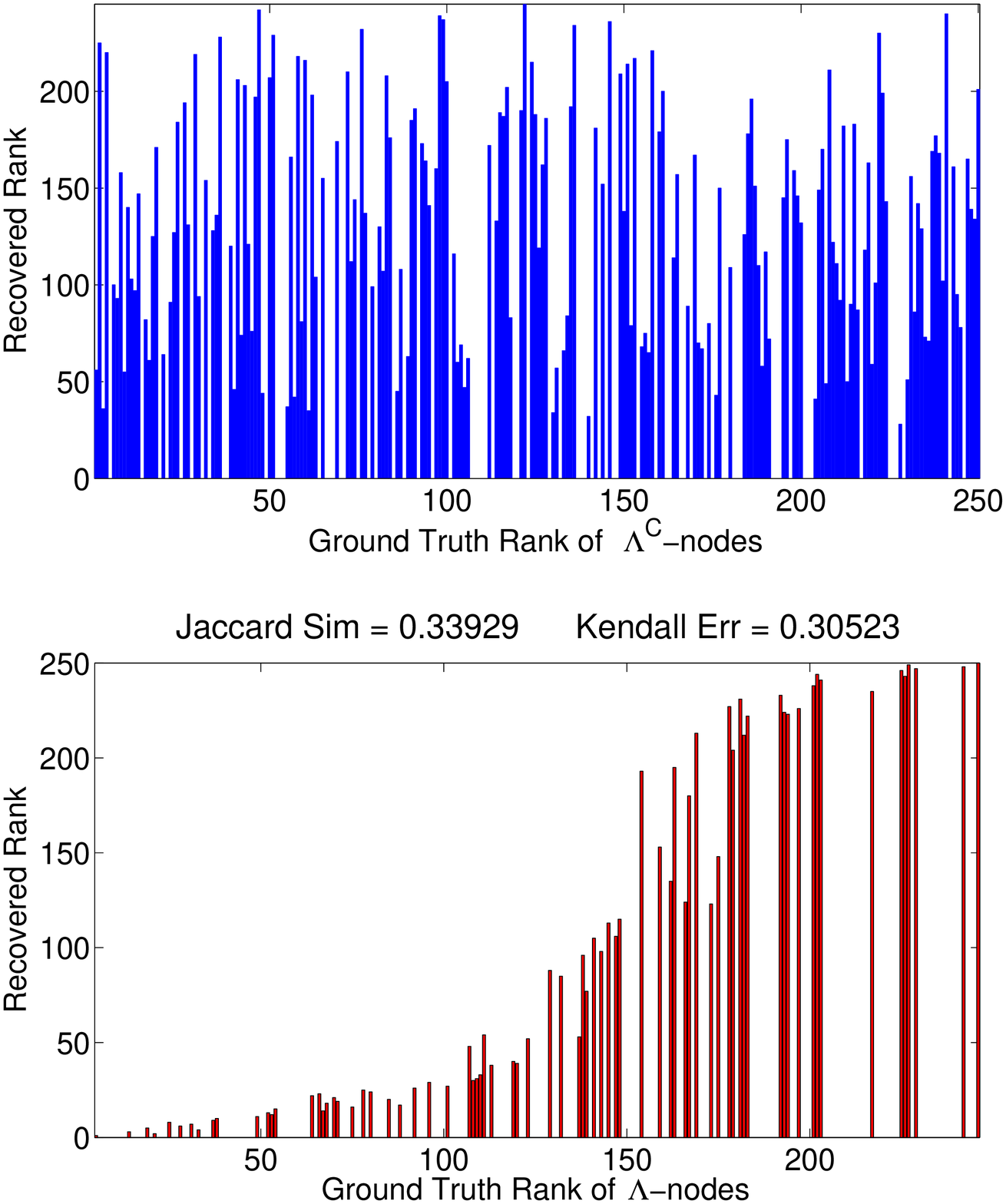}}
\subfigure[ LS ]{\includegraphics[width=0.324\textwidth]{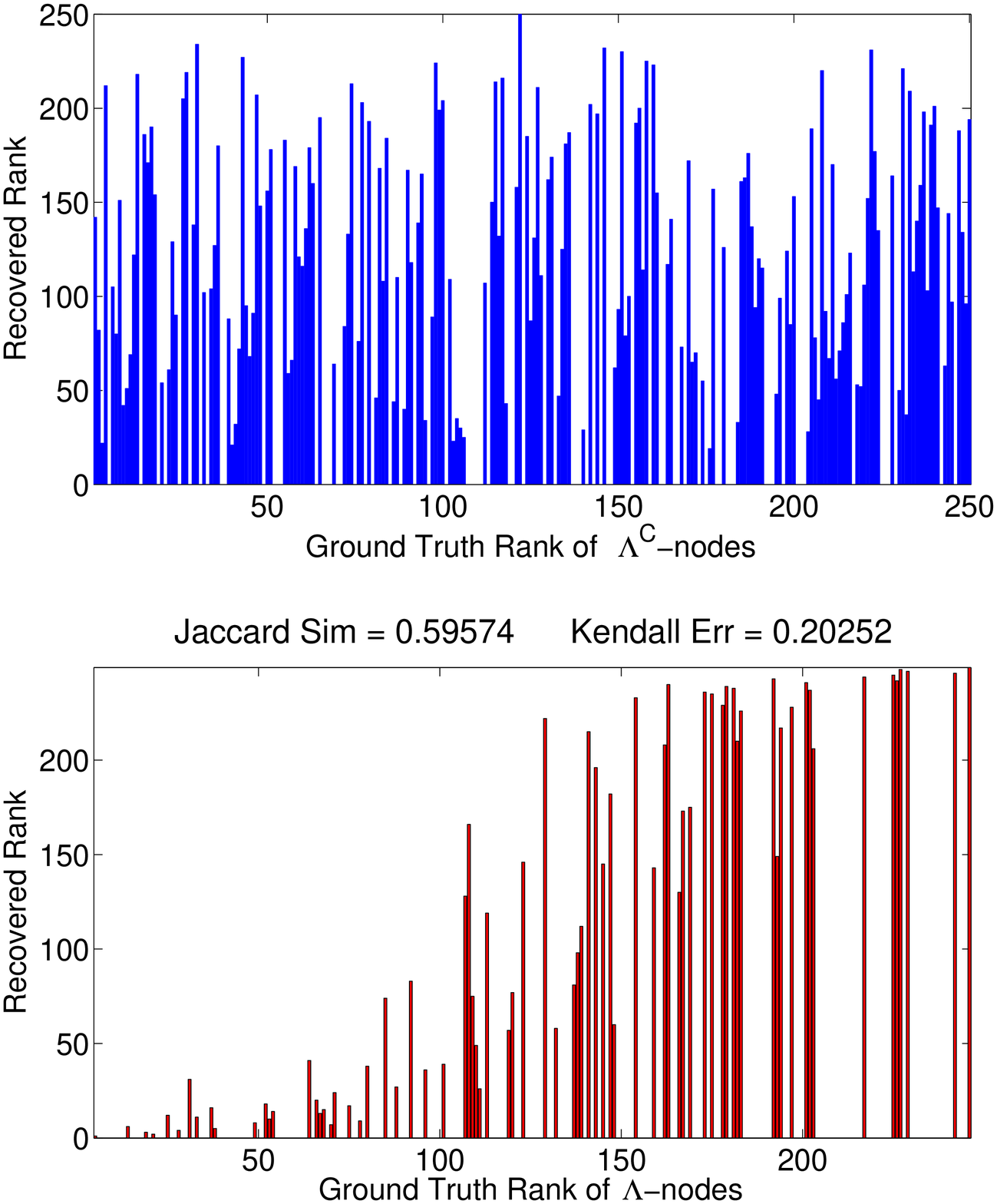}}
\subfigure[ SER ]{\includegraphics[width=0.324\textwidth]{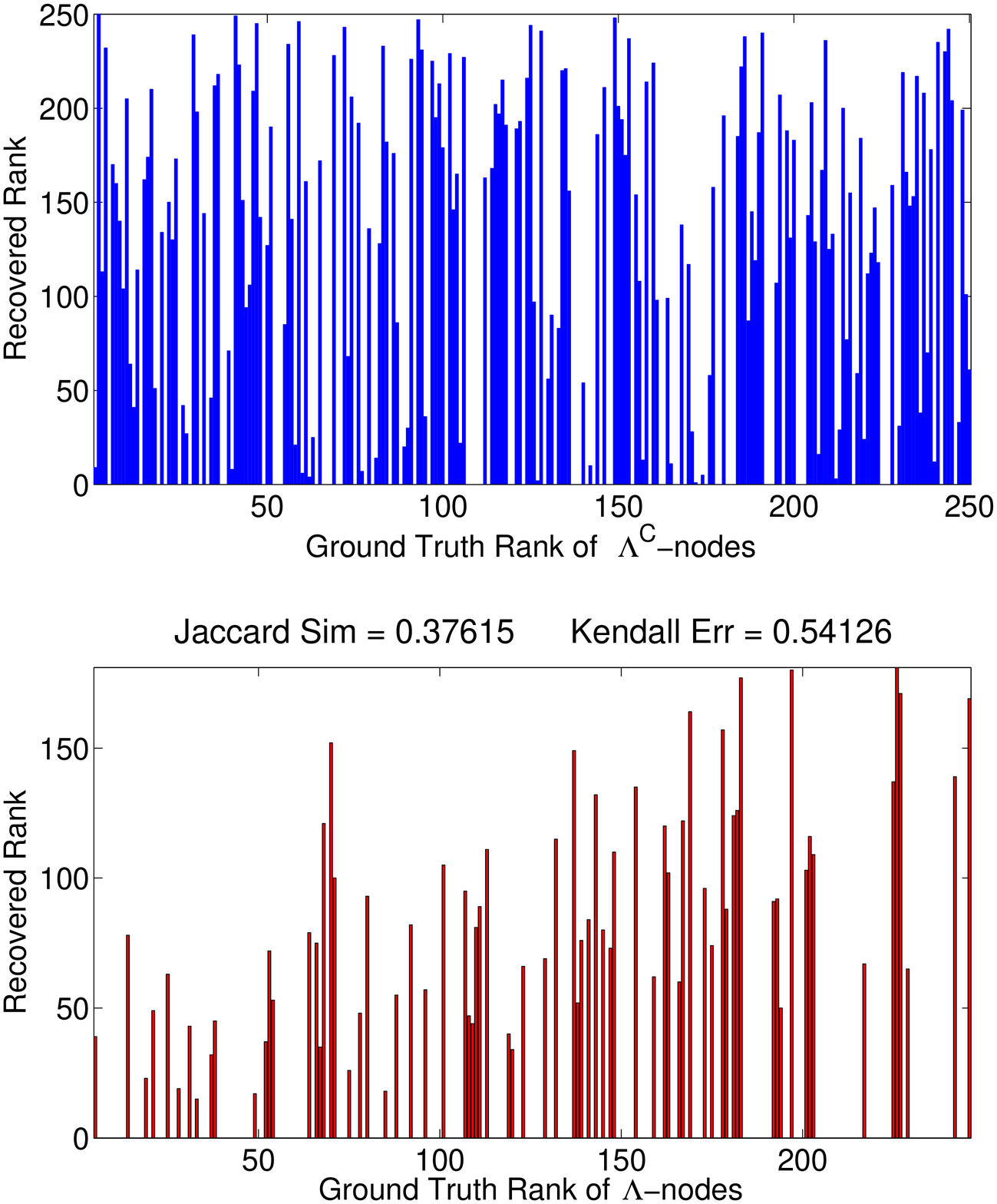}}
\subfigure[ RC ]{\includegraphics[width=0.324\textwidth]{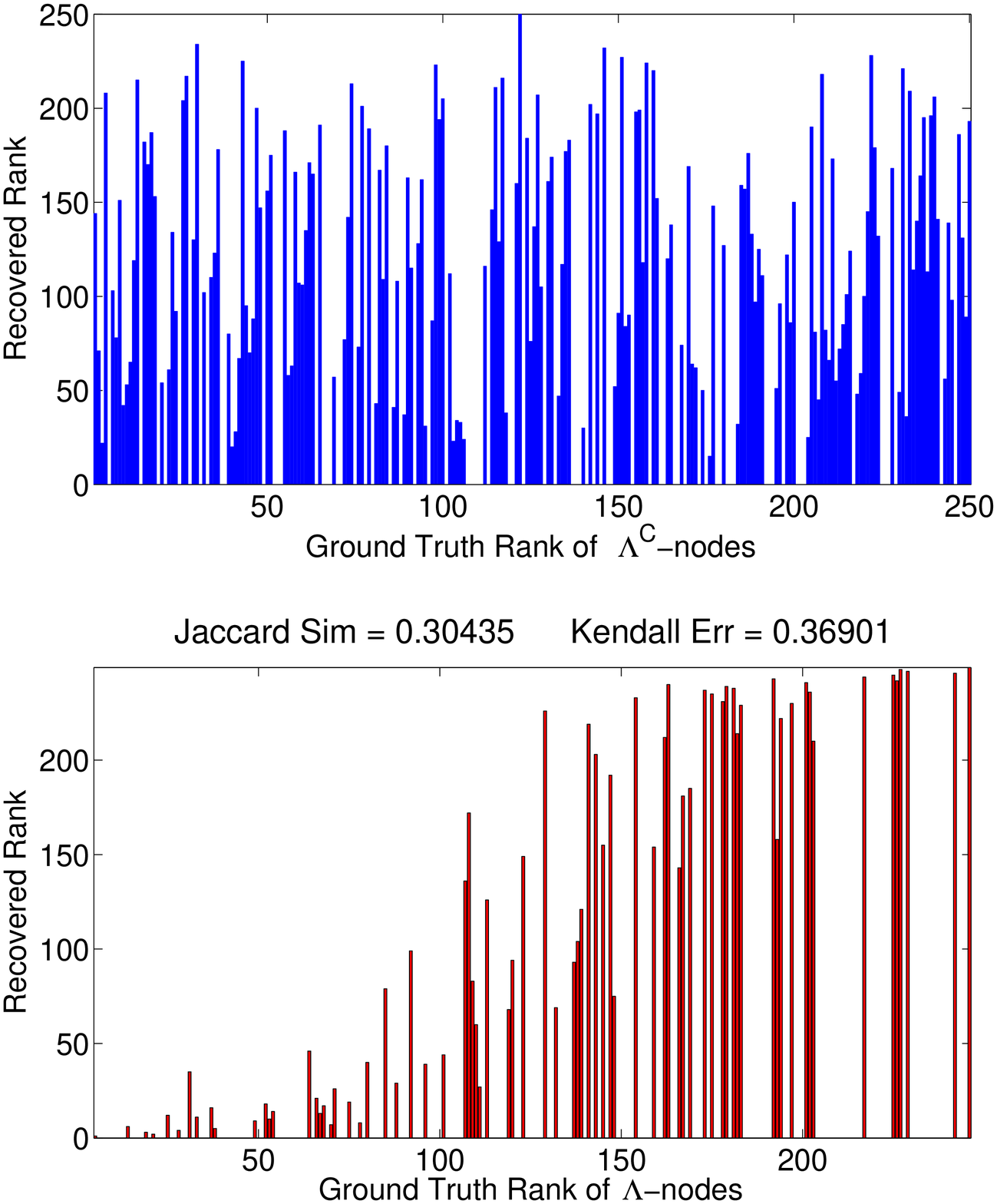}}
\subfigure[ SYNC-EIG  ]{\includegraphics[width=0.324\textwidth]{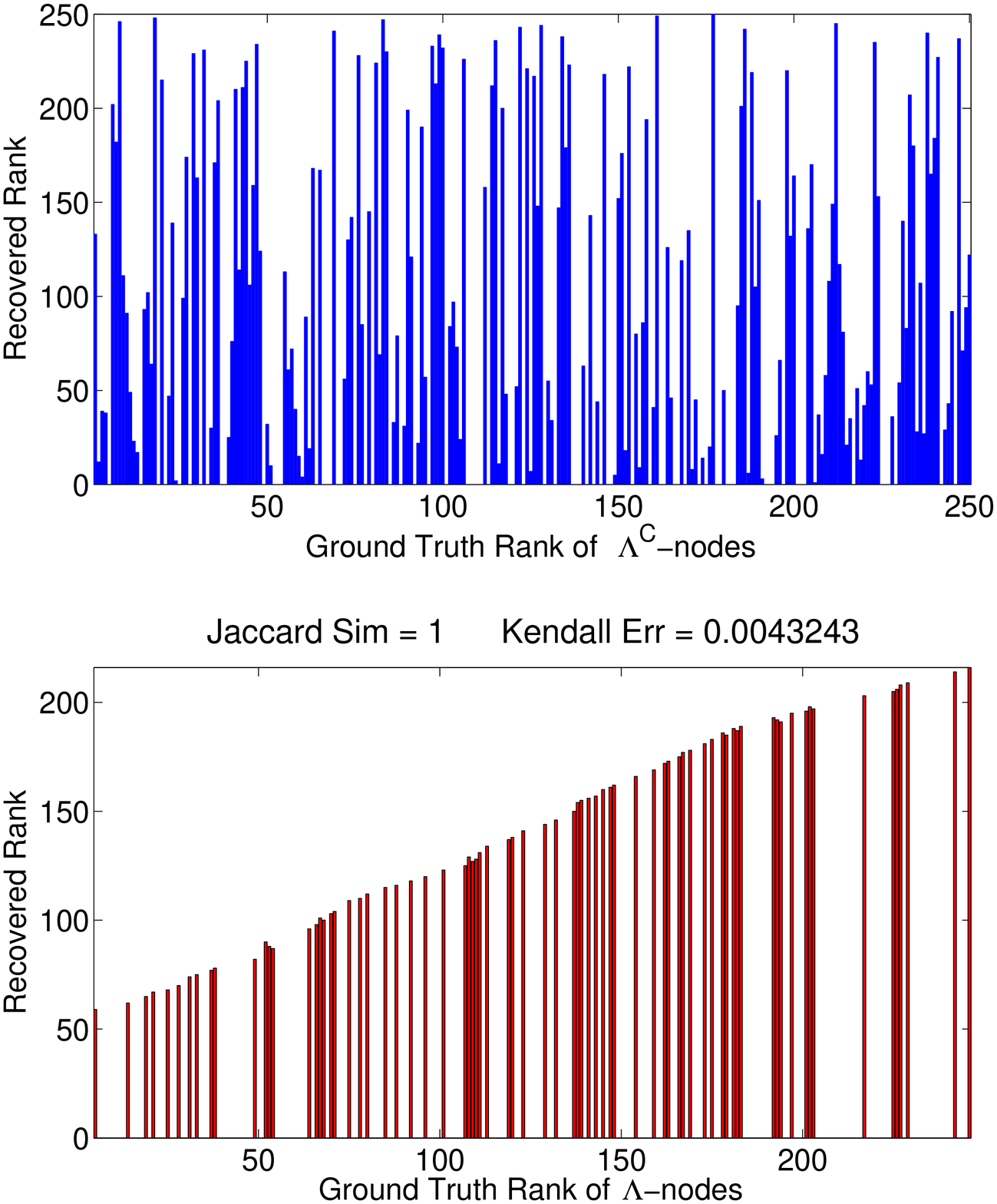}}
\subfigure[ SYNC-SDP  ]{\includegraphics[width=0.324\textwidth]{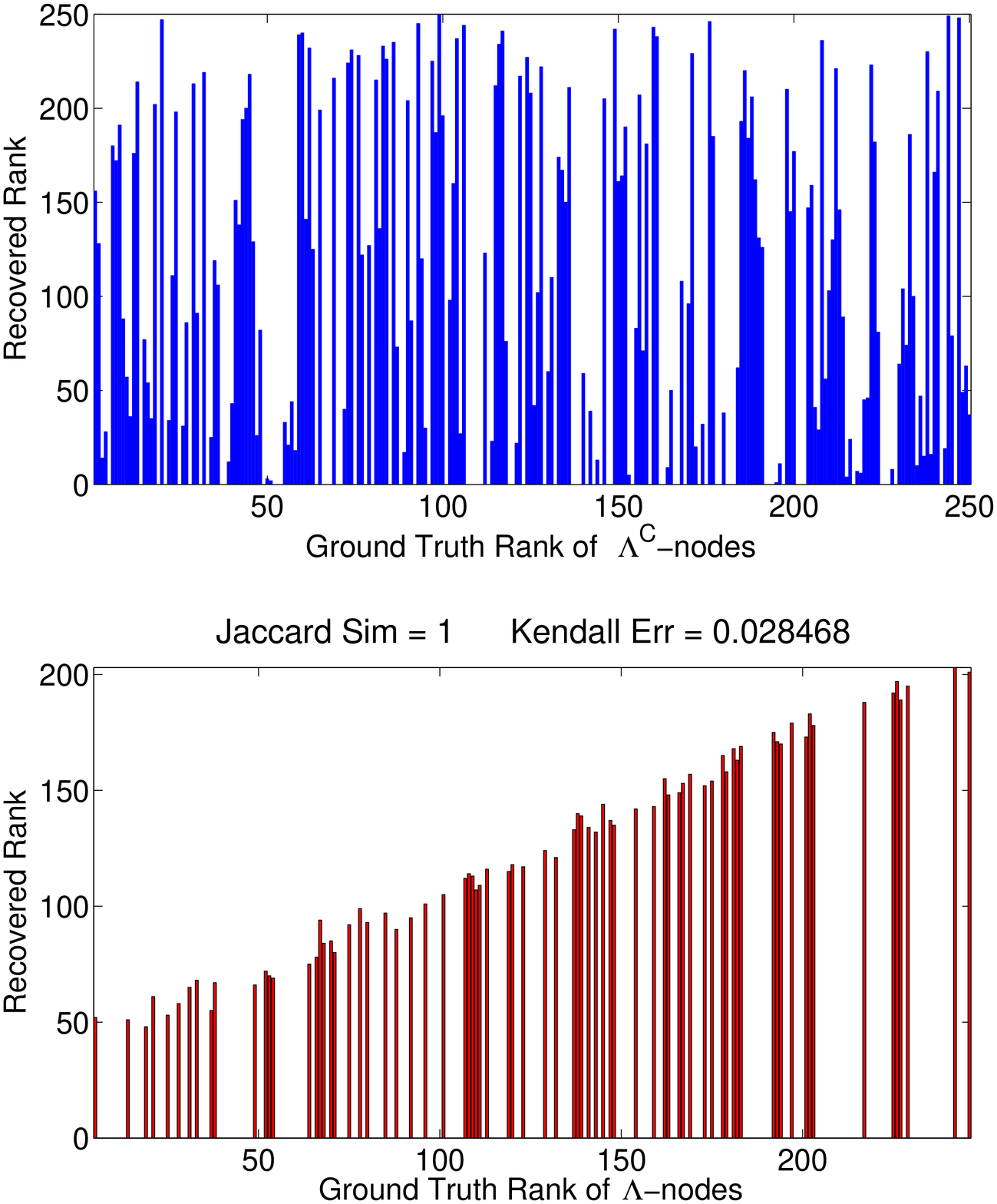}}
\end{center}
\caption{The recovered rankings by each of the six methods, for a random instance of the ensemble $\mathcal{G}(n=250, \beta=0.3, \eta_1=0, \eta_2=1)$, for which $|\Lambda| = \beta n = 75 $. For each plot, the top subplot, respectively bottom subplot, corresponds to the recovered rankings of the (ground truth) $\Lambda^C$-nodes, respectively the (ground truth) $\Lambda$-nodes. Note that, in practice, the set $\Lambda$ is not known a-priori, but we use this information here only for the purpose of making the point that the synchronization-based method is able to perfectly preserve the relative ranking of the $\Lambda$-players. We exploit this phenomenon, and are able to recover the planted set $\Lambda$ in a totally unsupervised manner.}
\label{fig:BarplotRankingsExample}
\end{figure}

\begin{figure}[h!]
\begin{center}
\subfigure[ $p=0.2, \eta_1=0$  ]{\includegraphics[width=0.30\textwidth]{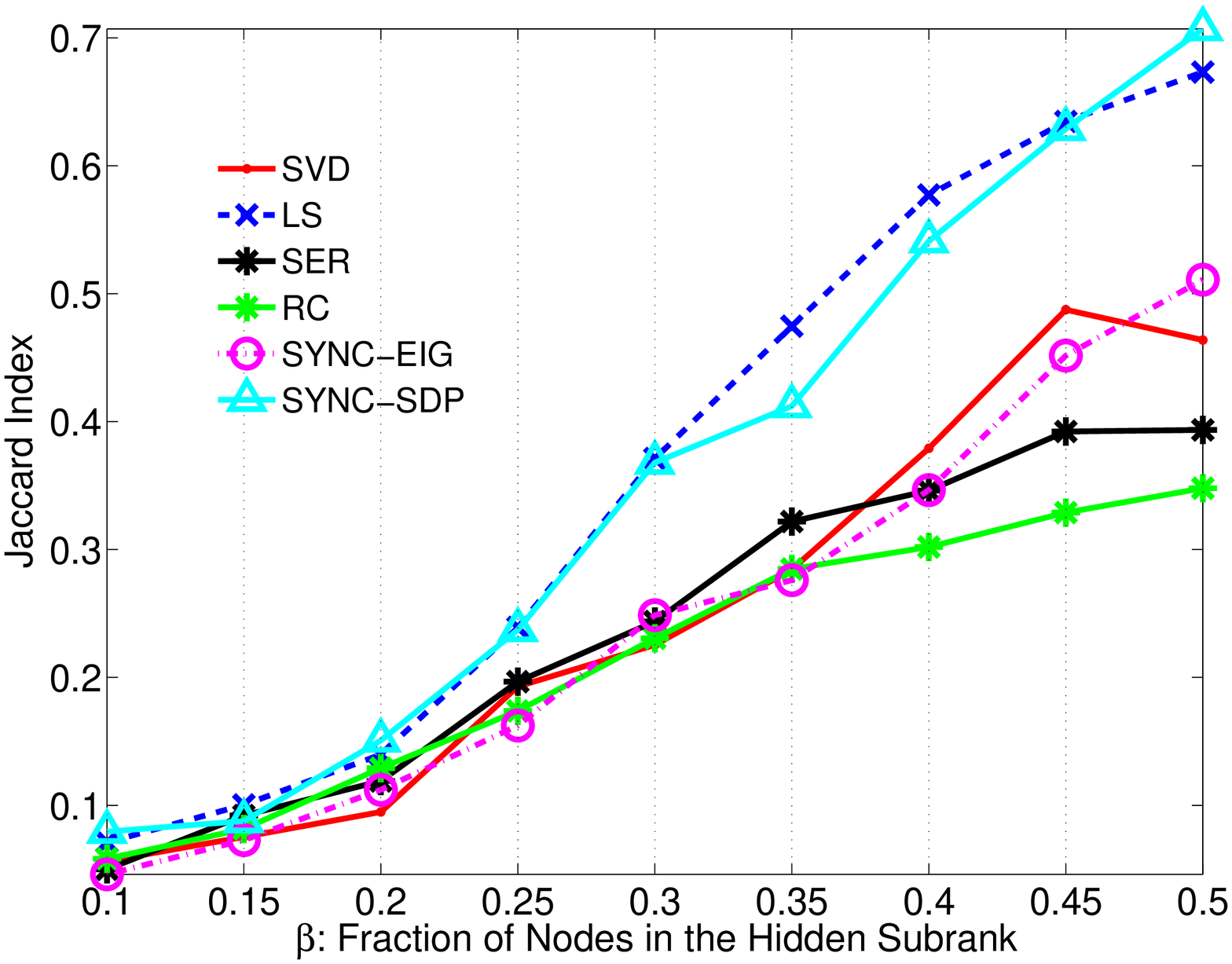}}
\subfigure[ $p=0.2, \eta_1=0.2$  ]{\includegraphics[width=0.30\textwidth]{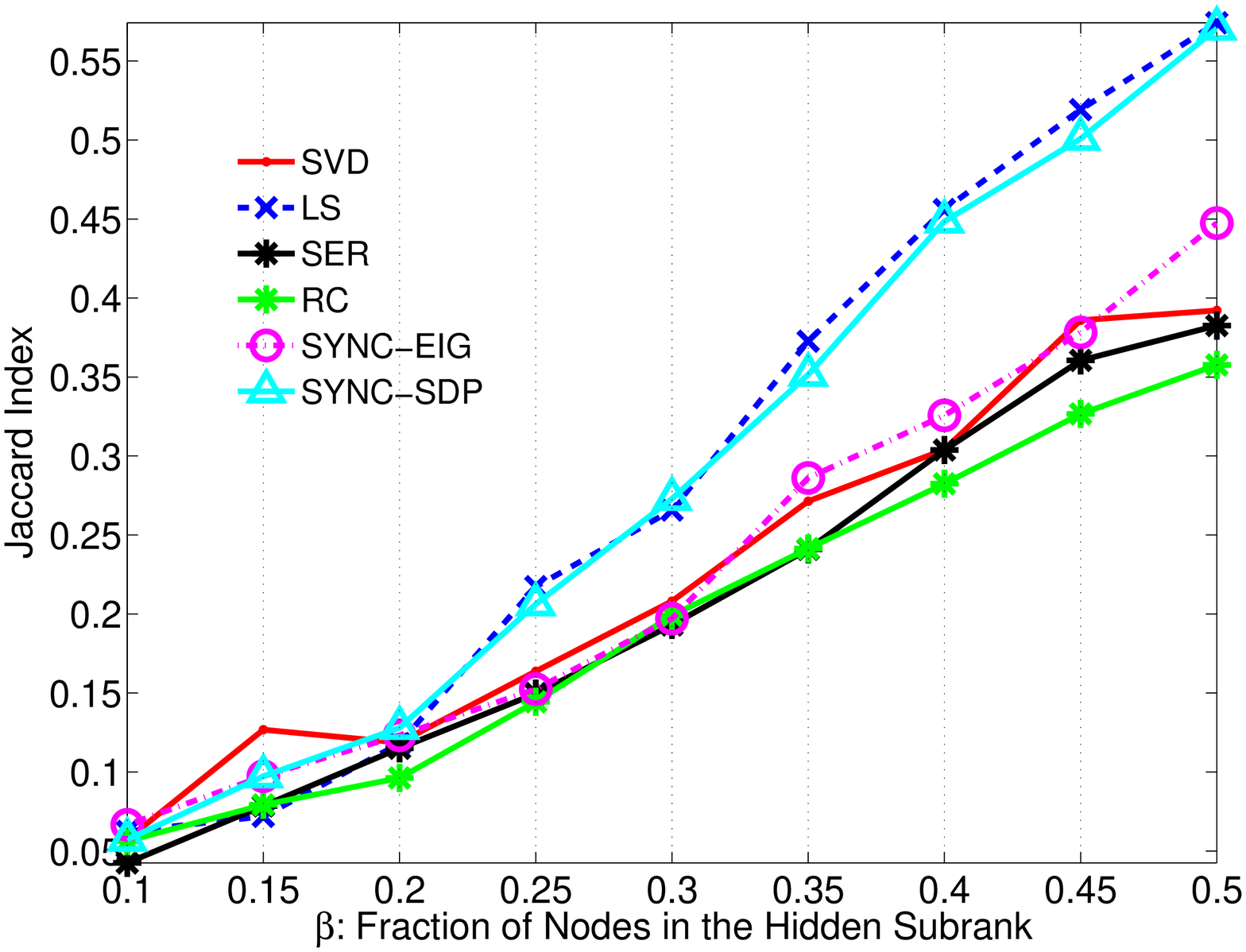}}
\subfigure[ $p=0.2, \eta_1=0.4$  ]{\includegraphics[width=0.30\textwidth]{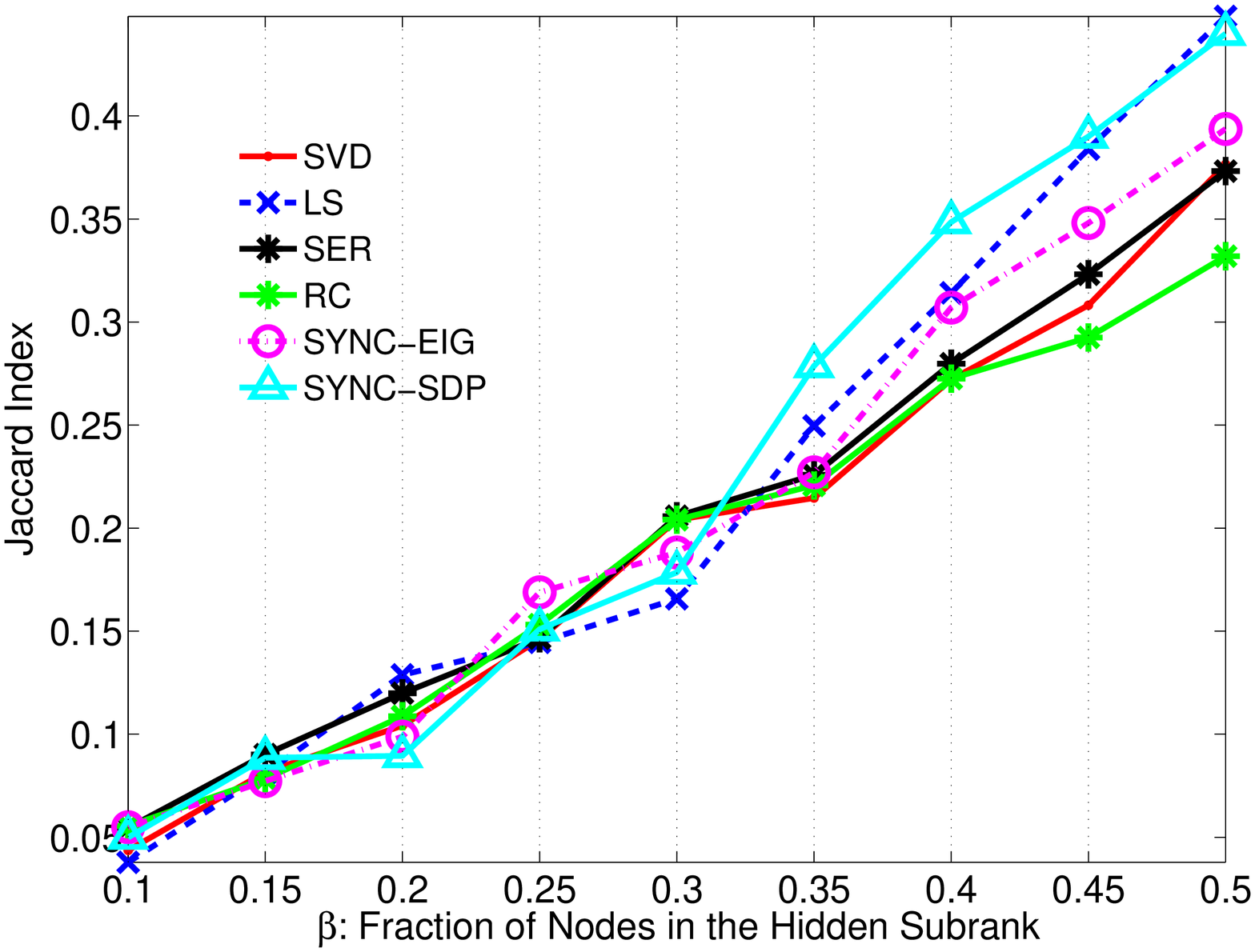}}
\subfigure[ $p=0.5, \eta_1=0$  ]{\includegraphics[width=0.30\textwidth]{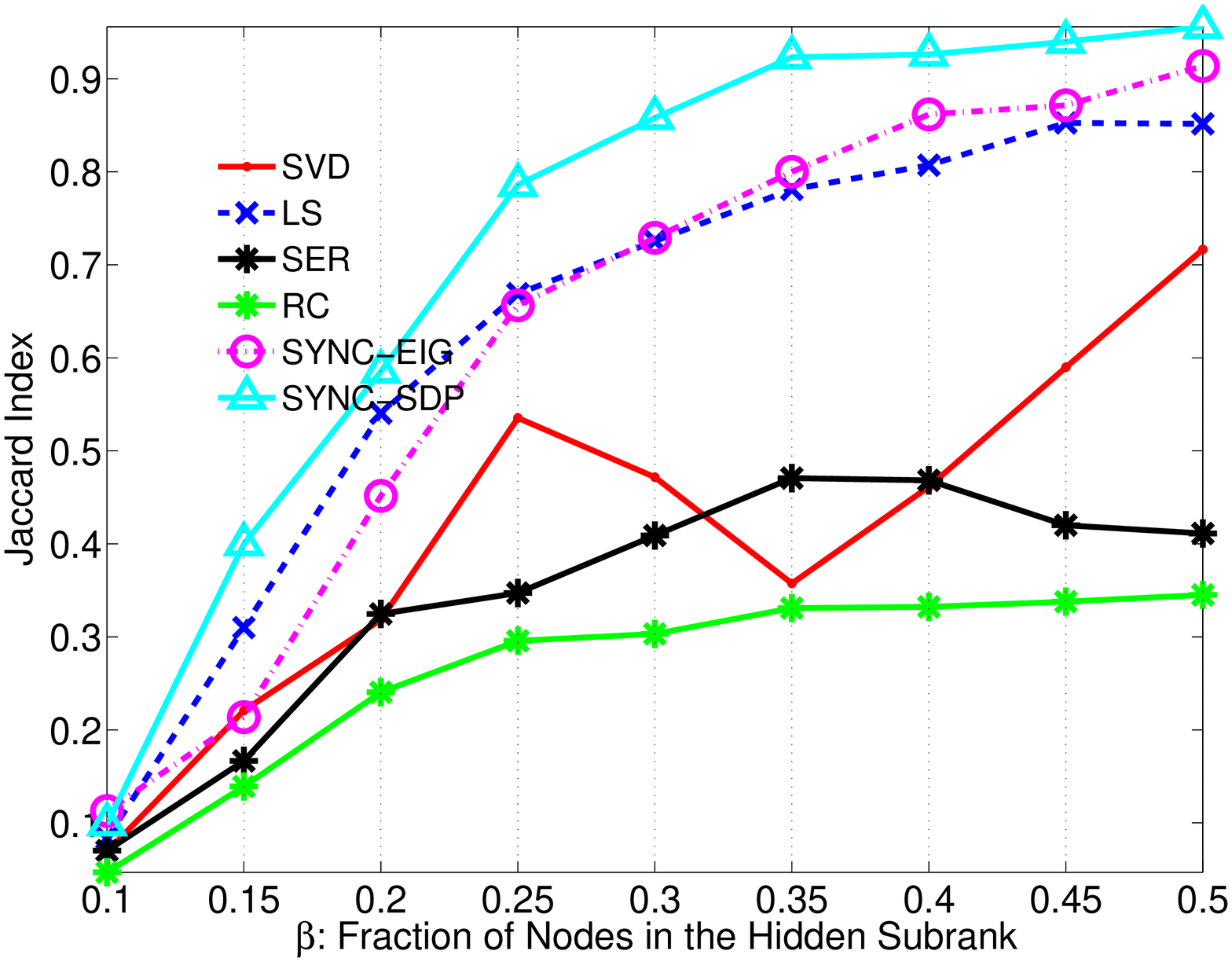}}
\subfigure[ $p=0.5, \eta_1=0.2$  ]{\includegraphics[width=0.30\textwidth]{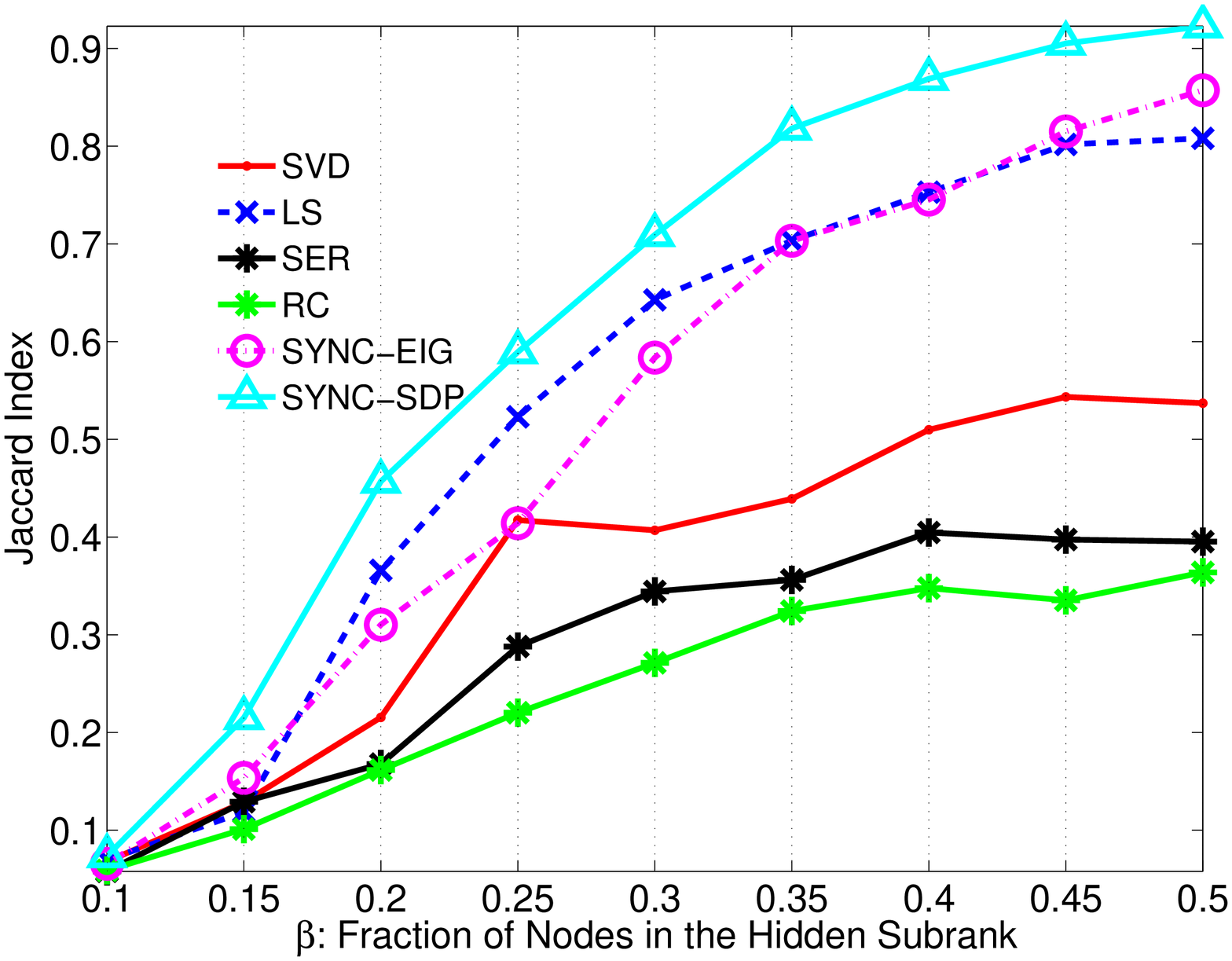}}
\subfigure[ $p=0.5, \eta_1=0.4$  ]{\includegraphics[width=0.30\textwidth]{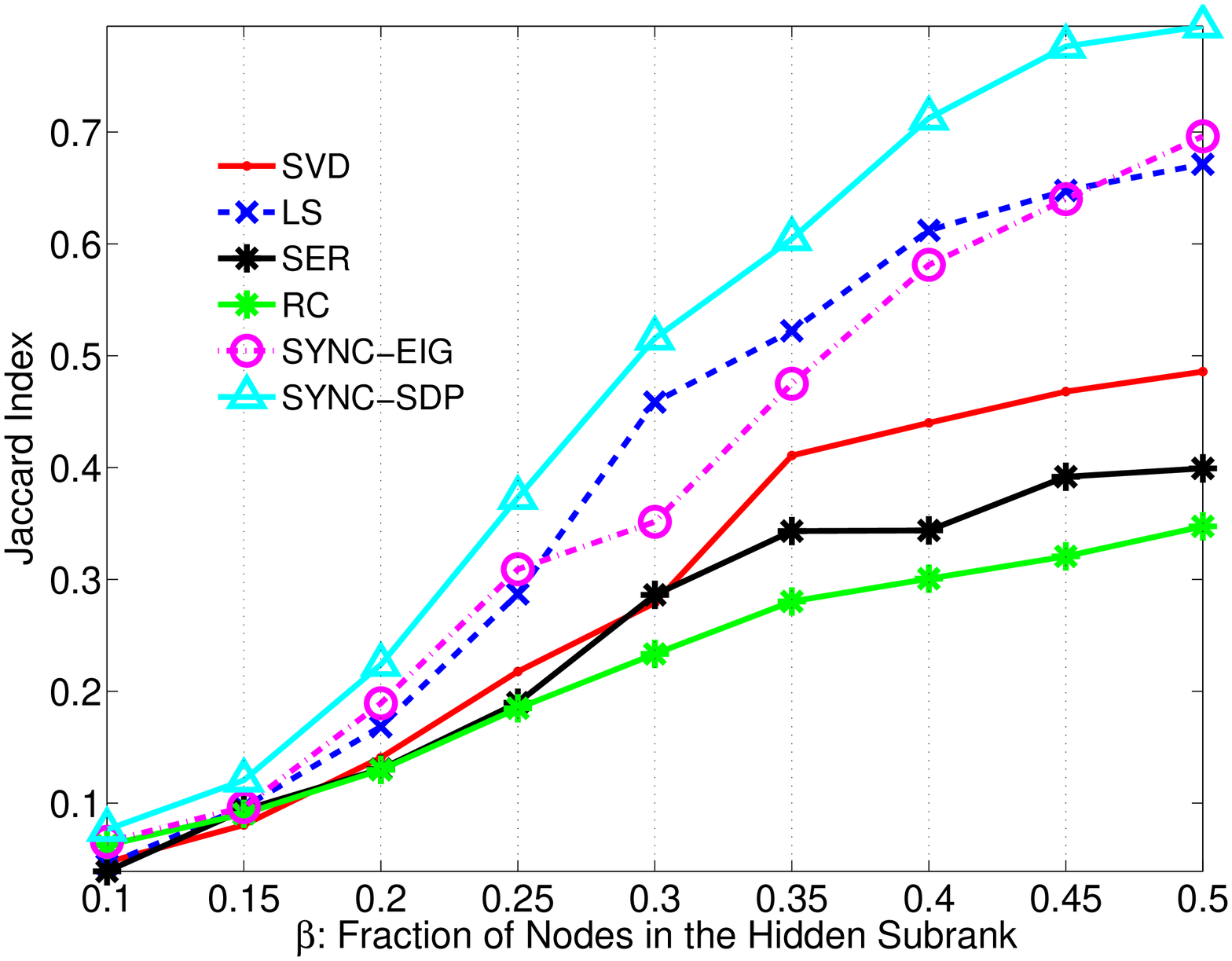}}
\subfigure[ $p=1, \eta_1=0$  ]{\includegraphics[width=0.30\textwidth]{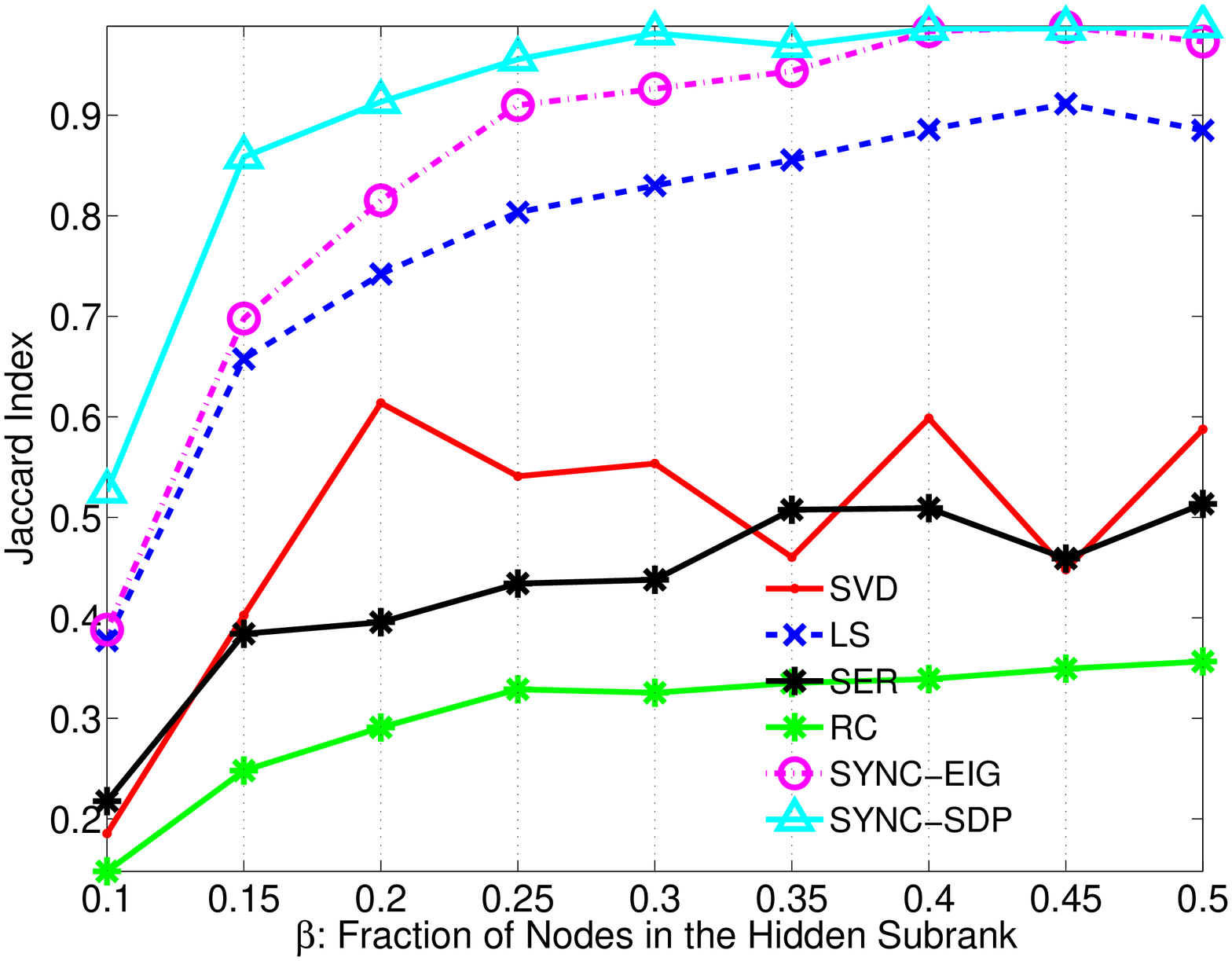}}
\subfigure[ $p=1, \eta_1=0.2$  ]{\includegraphics[width=0.30\textwidth]{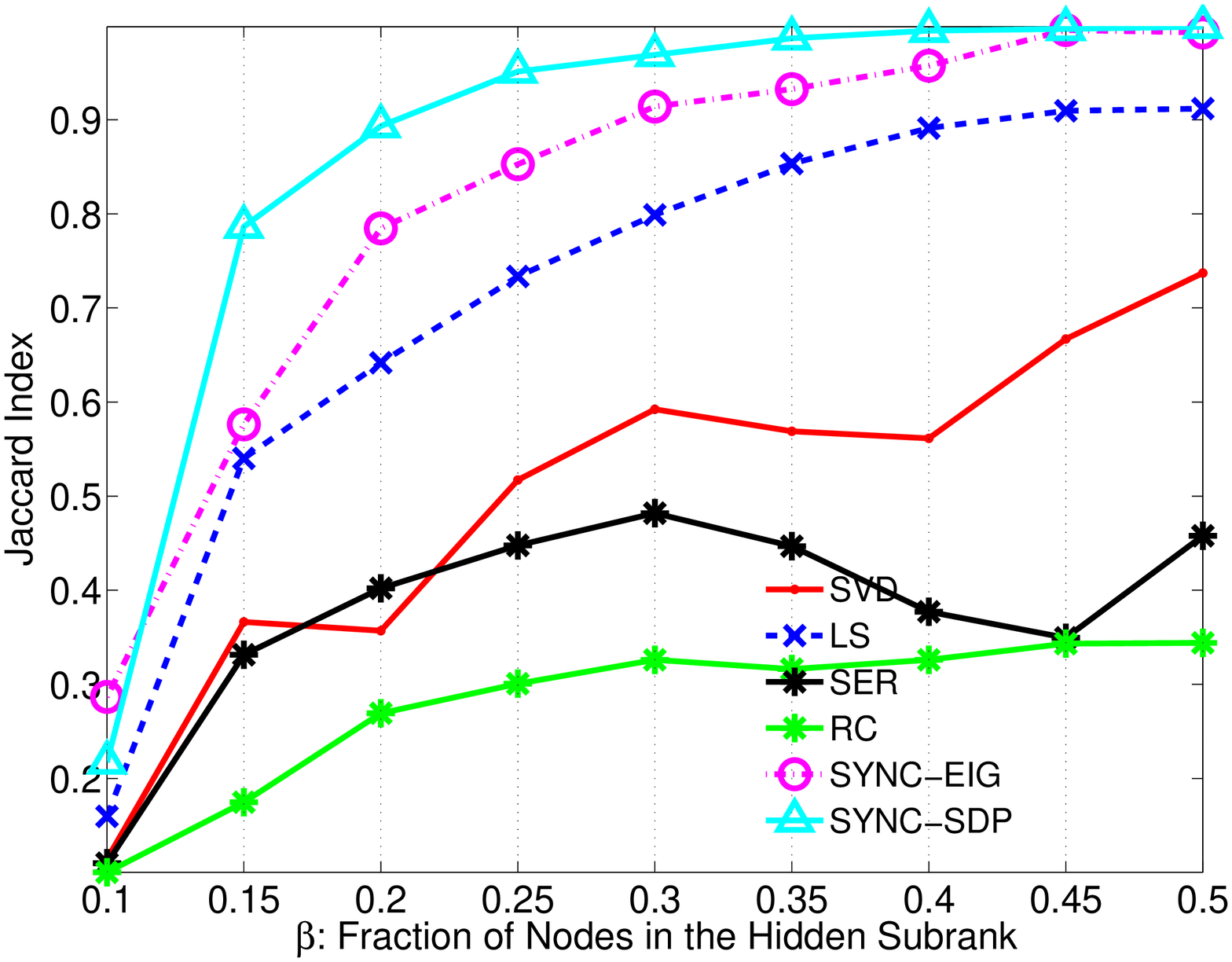}}
\subfigure[ $p=1, \eta_1=0.4$  ]{\includegraphics[width=0.30\textwidth]{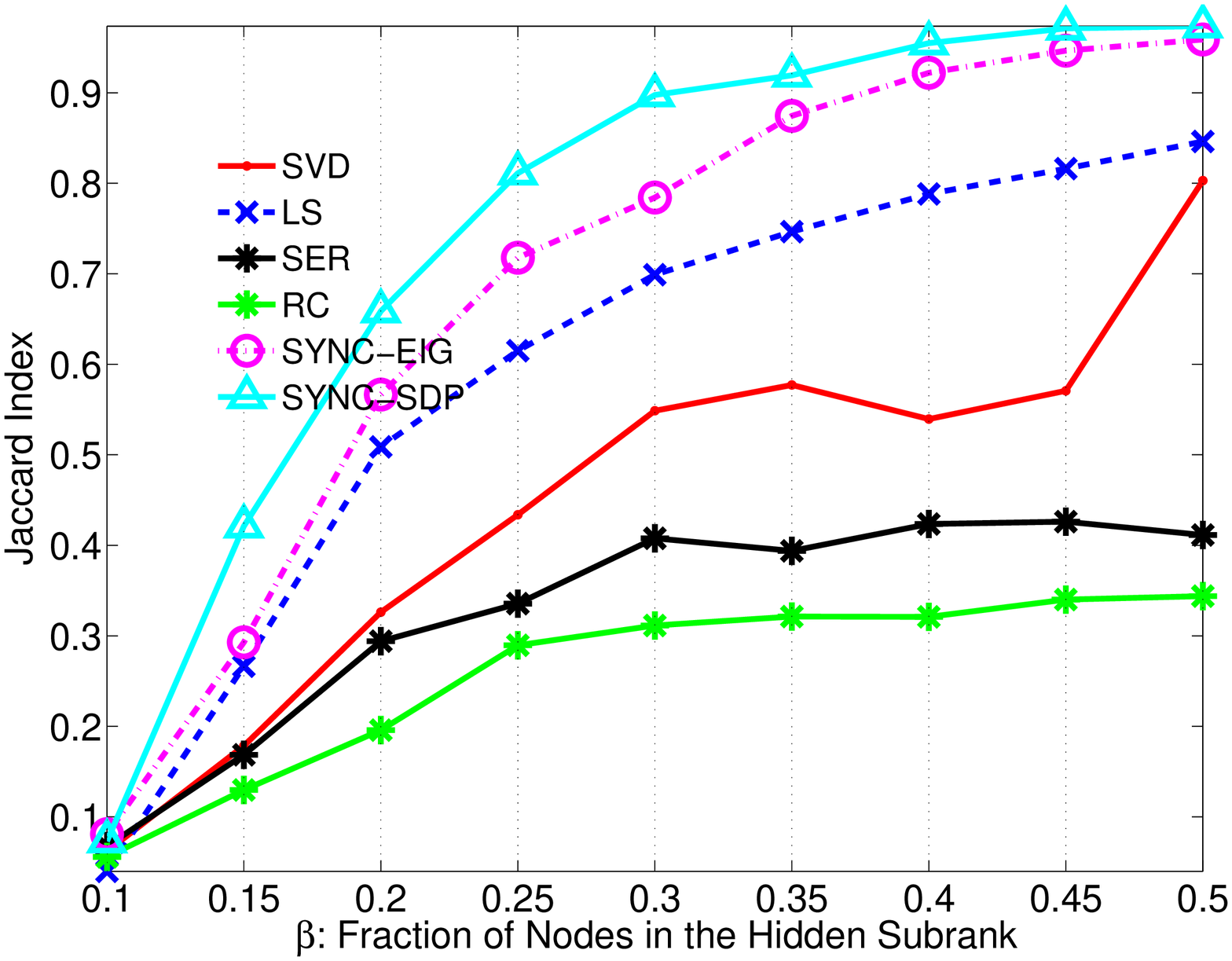}}
\end{center}
\caption{Comparison of the methods in terms of the Jaccard Similarity Index (higher is better) between the recovered $\hat{\Lambda}$ and the ground truth $\Lambda$, from the ensembles given by $\mathcal{G}(n=250,p,\beta,\eta_1, \eta_2=1)$, for varying $p = \{0.2, 0.5, 1 \}$ and  $ \eta_1 = \{0, 0.2, 0.4\}$. 
Experiments are averaged over 15 runs.}
\label{fig:New_JI_LocalRankComp}
\end{figure}

\begin{figure}[h!]
\begin{center}
\subfigure[ $p=0.2, \eta_1=0$  ]{\includegraphics[width=0.32\textwidth]{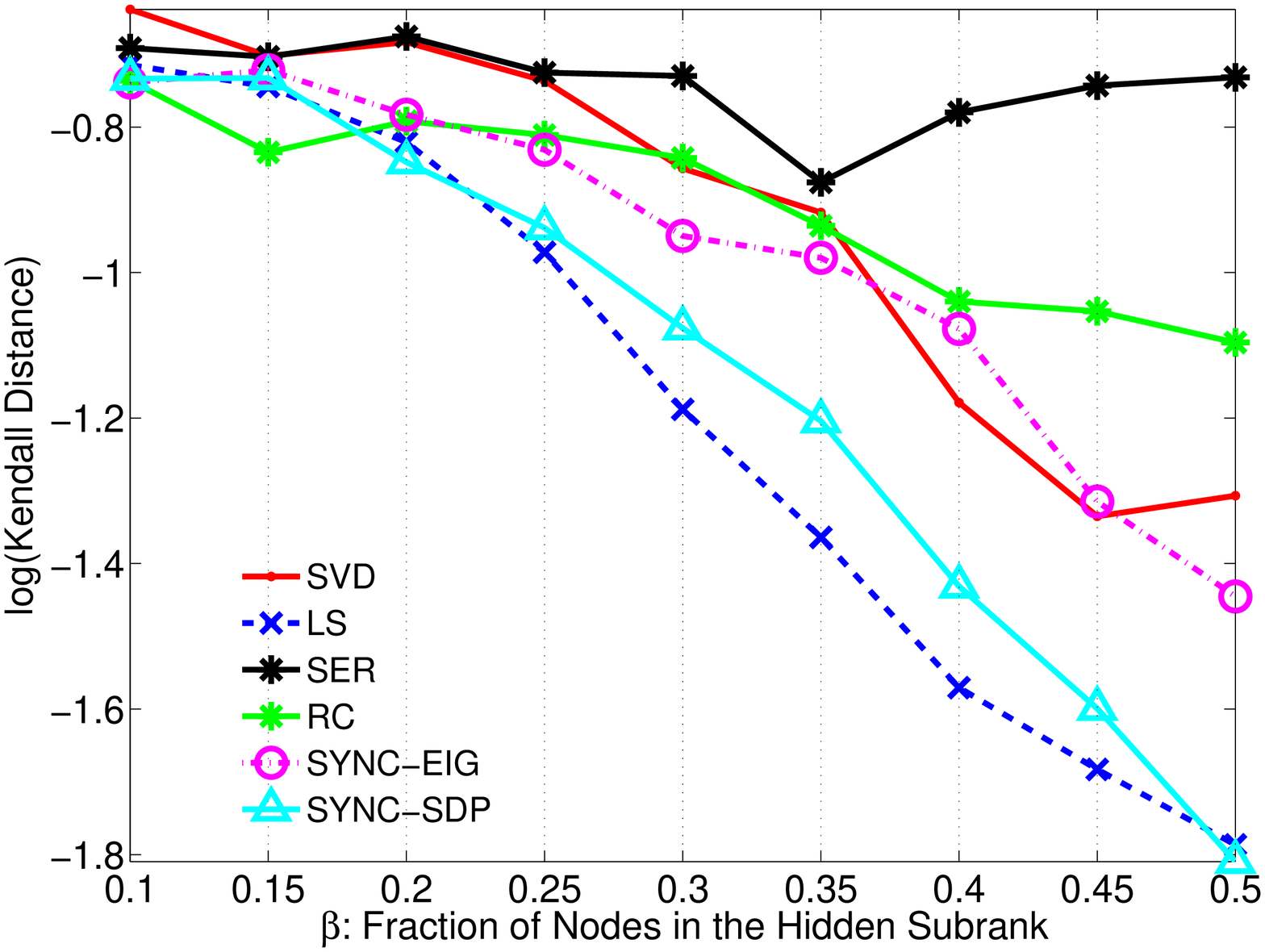}}
\subfigure[ $p=0.2, \eta_1=0.2$  ]{\includegraphics[width=0.32\textwidth]{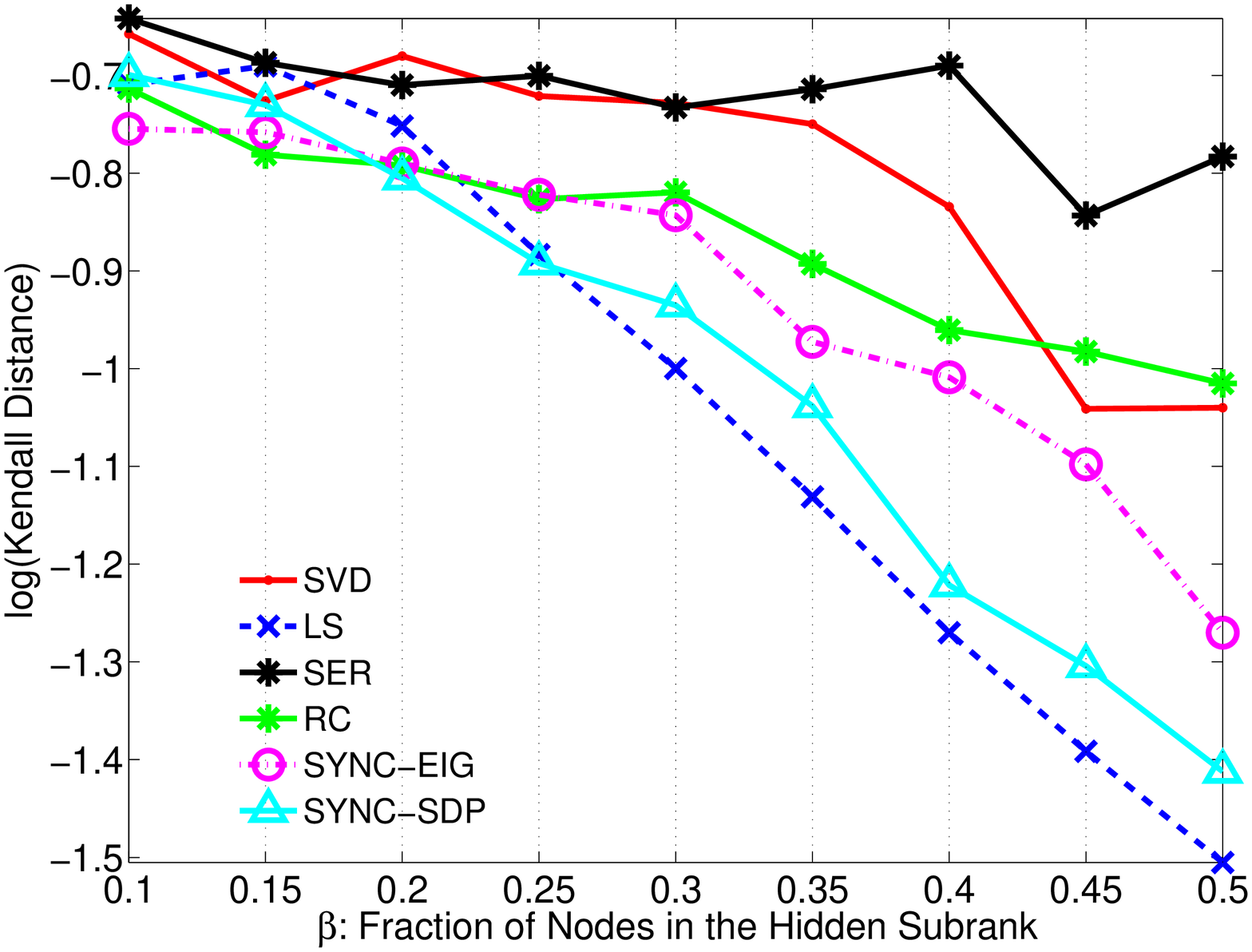}}
\subfigure[ $p=0.2, \eta_1=0.4$  ]{\includegraphics[width=0.32\textwidth]{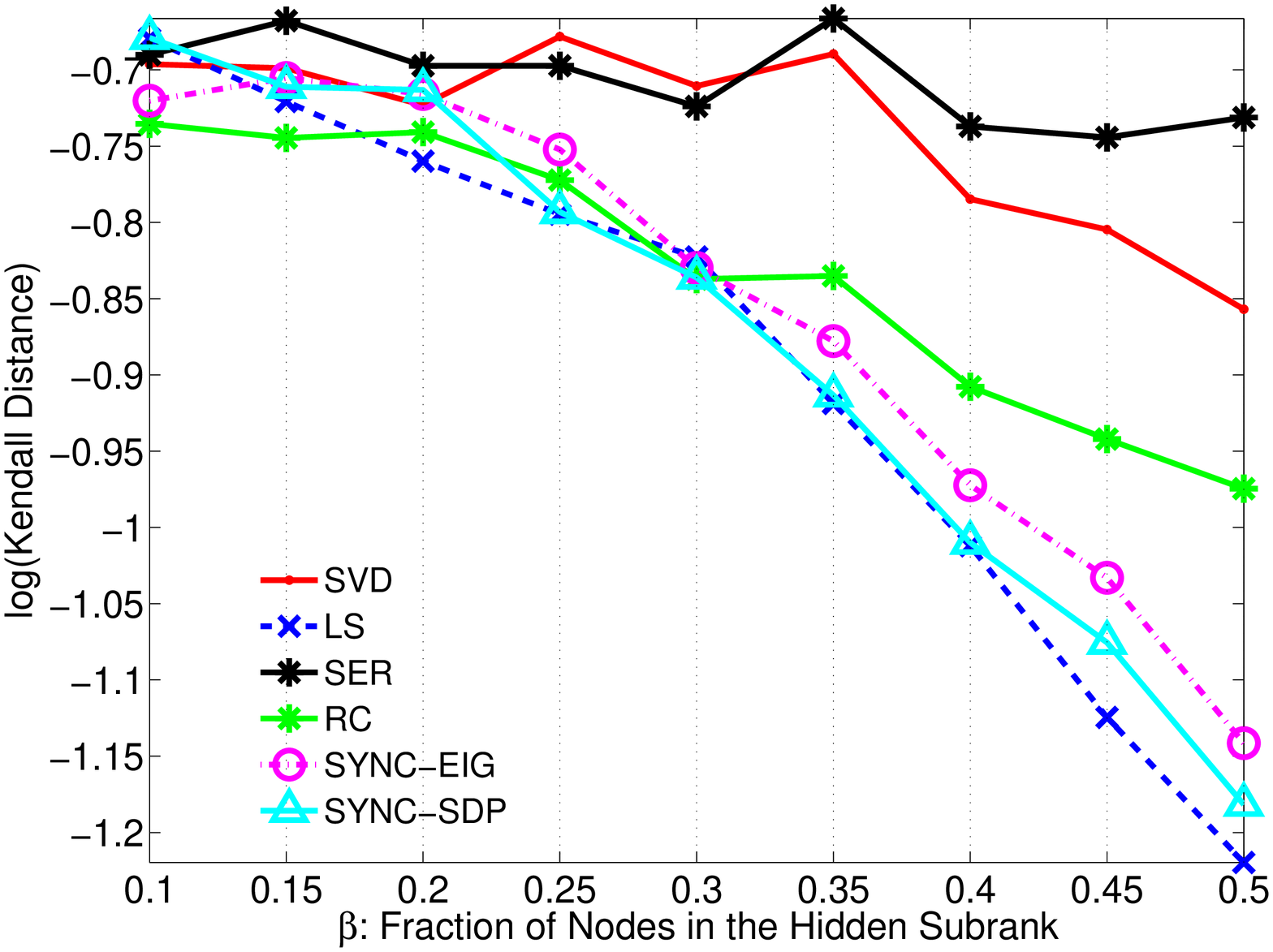}}
\subfigure[ $p=0.5, \eta_1=0$  ]{\includegraphics[width=0.32\textwidth]{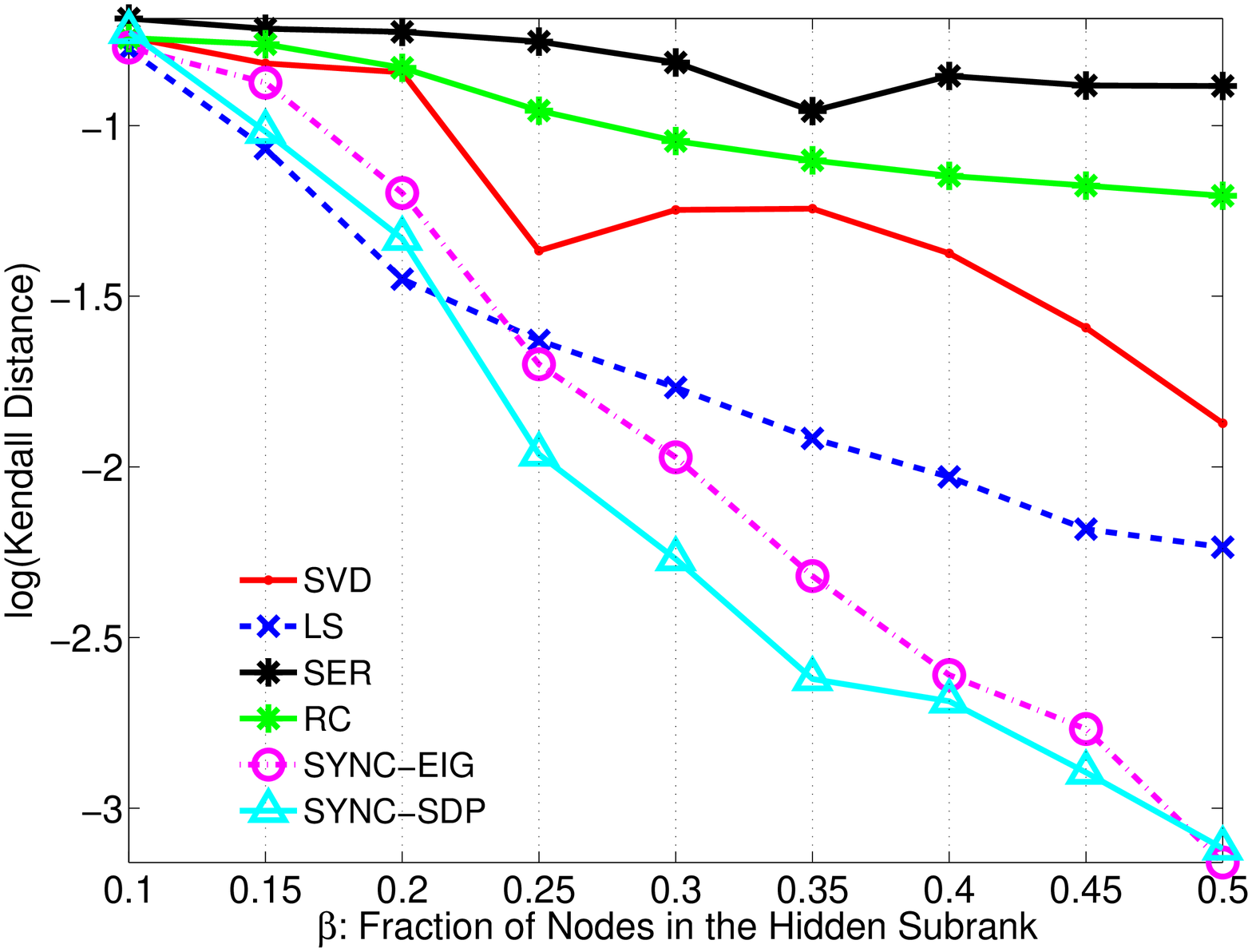}}
\subfigure[ $p=0.5, \eta_1=0.2$  ]{\includegraphics[width=0.32\textwidth]{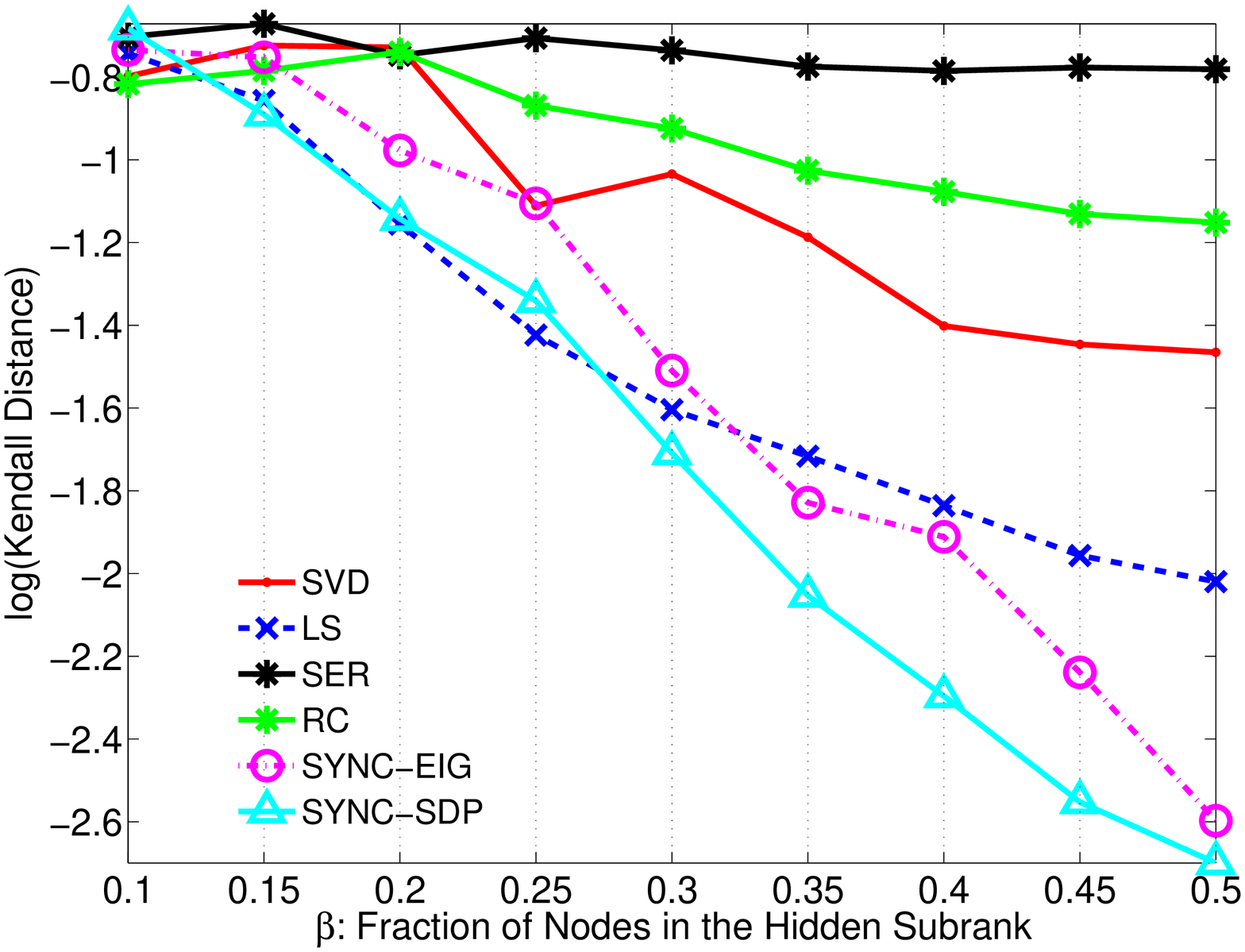}}
\subfigure[ $p=0.5, \eta_1=0.4$  ]{\includegraphics[width=0.32\textwidth]{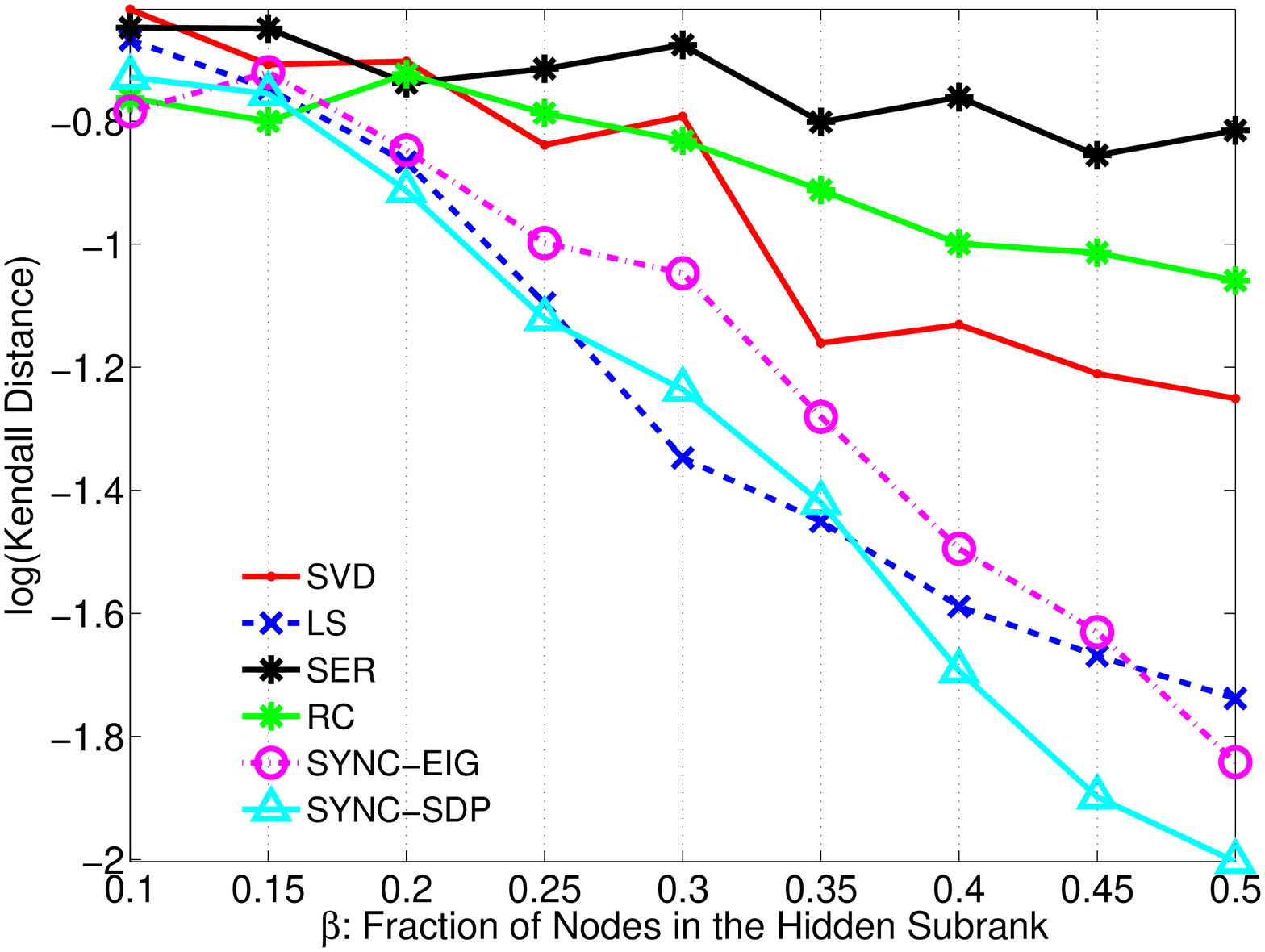}}
\subfigure[ $p=1, \eta_1=0$  ]{\includegraphics[width=0.32\textwidth]{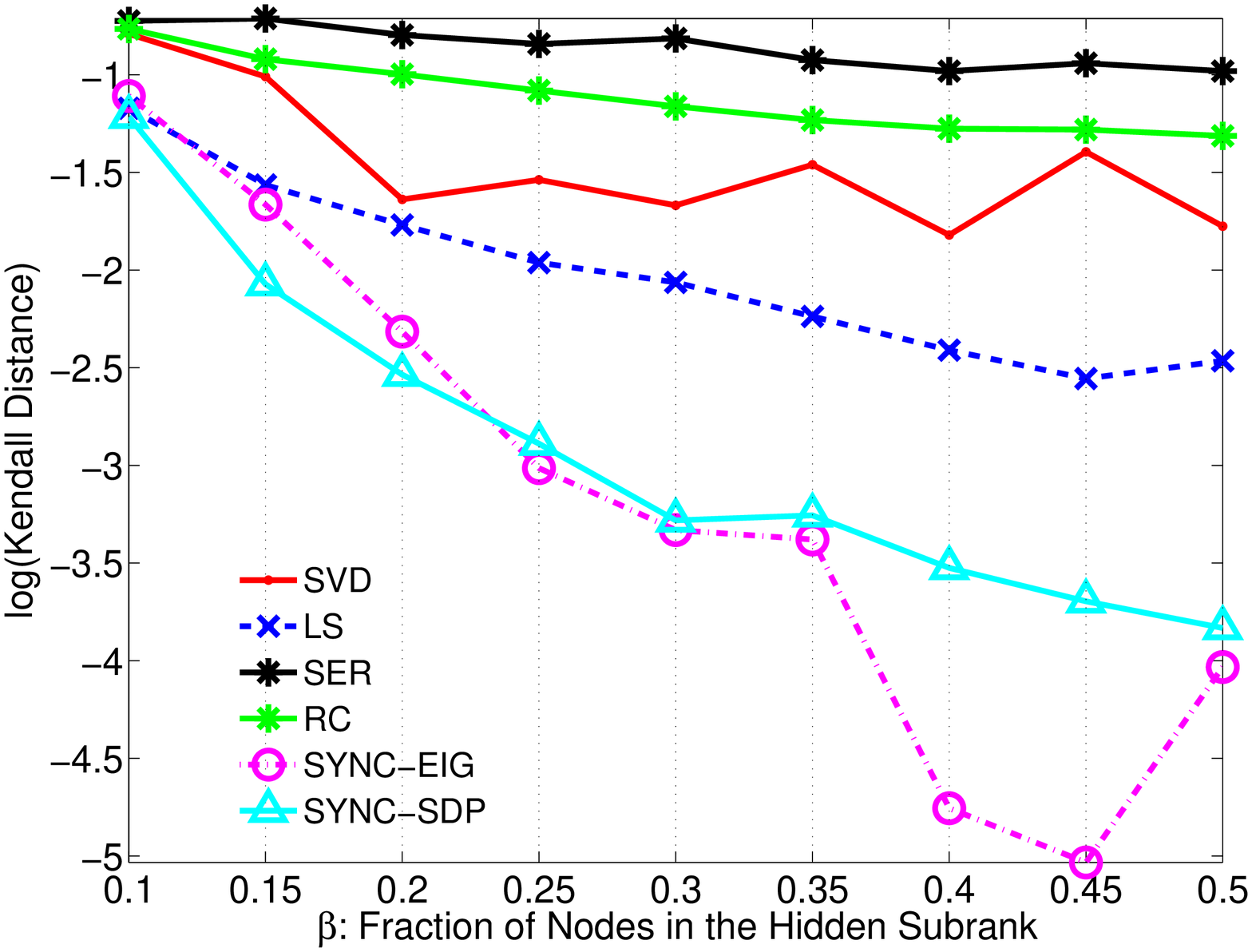}}
\subfigure[ $p=1, \eta_1=0.2$  ]{\includegraphics[width=0.32\textwidth]{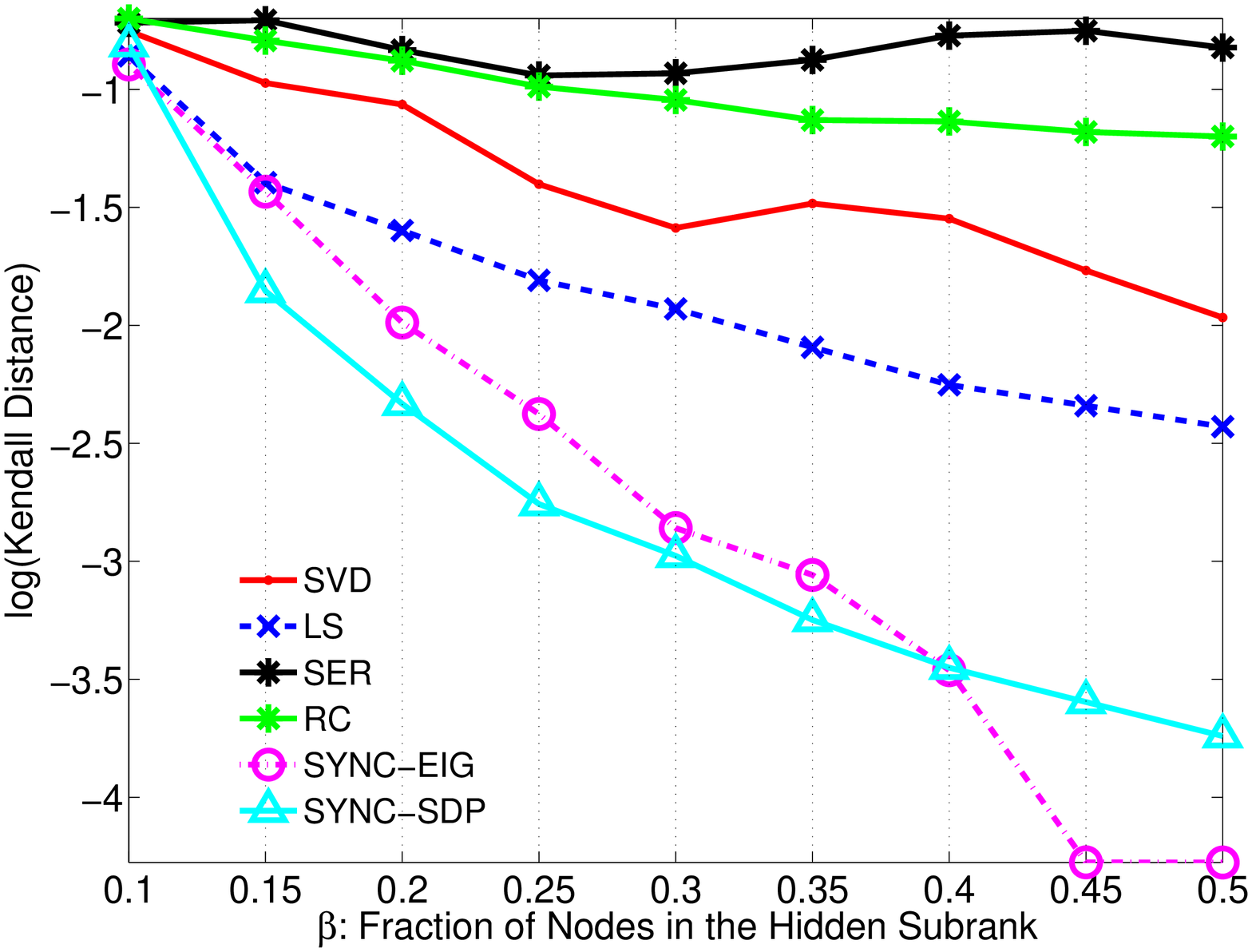}}
\subfigure[ $p=1, \eta_1=0.4$  ]{\includegraphics[width=0.32\textwidth]{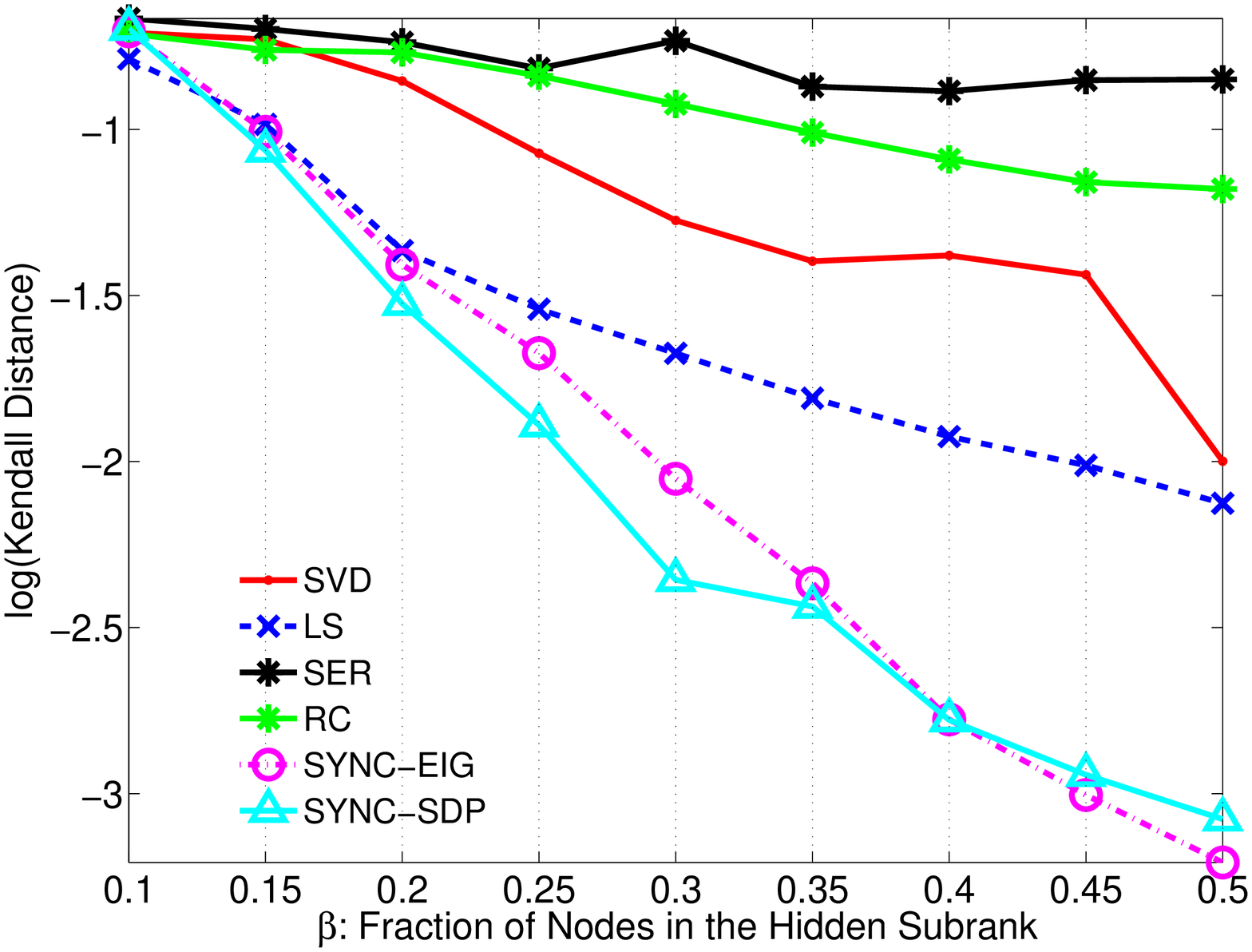}}
\end{center}
\caption{The Kendall Distance (lower is better) between the recovered ranking of the $\hat{\Lambda}$-nodes by each of the methods (from Figure \ref{fig:New_JI_LocalRankComp}) and their ground truth values, 
for the ensemble given by $\mathcal{G}(n=250,p,\beta,\eta_1, \eta_2=1)$, for varying $p = \{0.2, 0.5, 1 \}$ and  $ \eta_1 = \{0, 0.2, 0.4\}$.   Experiments are averaged over 15 runs.}
\label{fig:New_KD_LocalRankComp}
\end{figure}

\clearpage

\section{Appendix B: Serial-Rank}  \label{sec:appSER}

A spectral algorithm that exactly solves the noiseless seriation problem (and the related \textit{continuous ones problem}) was proposed by Atkins et al. \cite{AtkinsSeriation}, based on the observation that given similarity matrices computed from serial variables, the ordering induced by the second eigenvector of the associated Laplacian matrix (i.e., the \textit{Fiedler} vector) matches that of the variables. In other words, this approach (reminiscent of spectral clustering) exactly reconstructs the correct ordering if the items in question lie on a one dimensional chain. In \cite{serialRank}, Fogel et al. adapt the above seriation procedure to the ranking problem, and propose an efficient polynomial-time algorithm with provable recovery and robustness guarantees. Under certain conditions on the pattern of noisy entries, Serial-Rank is able to perfectly recover the underlying true ranking, even when a fraction of the comparisons are either corrupted by noise or completely missing.
In this noisy setting, when the underlying measurement graph is dense, in other words, a high fraction of all the pairwise comparisons are observed, the spectral solution obtained by Serial-Rank is more robust to noise that other classical scoring-based methods from the literature.

The authors of \cite{serialRank} propose two approaches for computing a pairwise similarity matrix from both ordinal  and cardinal comparisons between the players. In the case of ordinal measurements, the similarity measure counts the number of \textit{matching comparisons}. More precisely, given as input a skew symmetric matrix $C$ of size $n \times n$ of pairwise comparisons $C_{ij} = \{ -1, 0, 1 \}$, with $C_{ij} = -C_{ji} $, 
 given by the following model
\begin{equation}
C_{ij} = \left\{
 \begin{array}{rl}
1 & \;\; \text{ if i is ranked higher than j} \\
0 & \;\; \text{ if i and j are tied, or comparison is not available} \\
-1  & \;\; \text{ if j is ranked higher than i} \\
     \end{array}
   \right.
\label{binaryCmodel}
\end{equation}
For convenience, the diagonal of $C$ is set to $C_{ii}=1, \forall i= 1,2,\ldots,n$. Finally, the pairwise similarity matrix is given by 
\begin{equation}
  S_{ij}^{match} = \sum_{k=1}^{n} \left( \frac{1 + C_{ik}  C_{jk} }{ 2 } \right).
  \label{SijMatch}
\end{equation}
Note that $C_{ik} C_{jk}=1$ whenever $i$ and $j$ have the same signs, and $C_{ik} C_{jk} = -1$ whenever they have opposite signs, thus $S_{ij}^{match}$ counts the number of matching comparisons between $i$ and $j$ with a third  reference item $k$. The lack of a comparison of item $k$ with either $i$ or $j$ contributes with a $ \frac{1}{2}$  to the summation in (\ref{SijMatch}).
The intuition behind this similarity measure proposed by Fogel et al. is that players that beat the same players and are beaten by the same players should have a similar ranking in the final solution. Written in a compact form, the final similarity matrix is given by 
\begin{equation}
 	S^{match} = \frac{1}{2} \left( n \mb{1} \mb{1}^T + C C^T \right) 
\label{SMatch}
\end{equation}
We summarize in Algorithm \ref{Algo:SerialRank} the main steps of the Serial-Rank method of Fogel et al. for ranking via seriation.

\begin{algorithm}[h!]
\begin{algorithmic}[1]   
\REQUIRE A set of pairwise comparisons $C_{ij} \in \{-1,0,1 \}$ or [-1,1]
\STATE Compute a similarity matrix as shown in (\ref{SijMatch})
\STATE Compute the associated graph Laplacian matrix 		
\begin{equation}
		L_S = D - S
\end{equation}
where D is a diagonal matrix $D =$ \textbf{diag} $(S \mb{1})$, i.e., $D_{ii} = \sum_{j=1}^{n}  G_{i,j} $ is the degree of node $i$ in the measurement graph $G$. 
\STATE Compute the Fiedler vector of S (eigenvector corresponding to the smallest nonzero eigenvalue of $L_S$).
\STATE Output the ranking induced by sorting the Fiedler vector of $S$, with the global ordering (increasing or decreasing order) chosen such that the number of upsets is minimized.
\end{algorithmic}
\caption{ \textbf{Serial-Rank}: an algorithm for spectral ranking using seriation \cite{serialRank}. }
\label{Algo:SerialRank}
\end{algorithm}


In the generalized linear model setting, one assumes that the paired comparisons are generated according to a generalized linear model, where paired comparisons are independent, and item $i$ is preferred to item $j$ with probability $$ P_{ij} = H(\nu_i - \nu_j) $$
where $\nu \in \mathbb{R}^n$ is a vector denoting the strength, rank, or skill level of the $n$ players.
In the context of the GLM model, Fogel et al. propose the following similarity matrix 
\begin{equation}
	S_{i,j}^{glm} = \sum_{k=1}^{n}  \mb{1}_{ \{ m_{i,k}  m_{j,k} > 0 \}  } \left( 1 - \frac{ |C_{i,k} -  C_{j,k}|}{2}   \right) + 	\frac{ \mb{1}_{ \{ m_{i,k}  m_{j,k} = 0 \} } } {2}
	\label{def:SimGLM}
\end{equation}
where $m_{i,k}=1$  if $i$ and $j$ played in a match, and 0 otherwise. 
We denote by SER-GLM the Serial-Rank algorithm based on the above GLM model given by (\ref{def:SimGLM}). For the cardinal case, SER-GLM is roughly similar to SER, except in the complete graph case, when SER-GLM consistently outperforms SER under both the MUN and ERO noise models. We refer the reader to  Figures
\ref{fig:Meth6_n200_num} and \ref{fig:Meth6_n200_ord} for the numerical results showing how SER and SER-GLM compare to the existing  and newly proposed methods.

\section{Appendix C: Additional Numerical Results}  \label{sec:appEngland}


We present in this Section additional experiments for the English Premier League data set, when the input measurements are given by 
\begin{itemize}
\item $C^{tpd}$ (\textit{Total-Point-Difference}), shown in Table \ref{tab:EnglandStandings_sumGoalDif}
\item $C^{stpd}$ (\textit{Sign-Point-Goal-Difference}), shown in Table  \ref{tab:EnglandStandings_signsumGoalDif}
\item $C^{snw}$ (\textit{Sign(Net Wins)}), shown in Table \ref{tab:EnglandStandings_signNrWonLost}
\end{itemize}

Figure \ref{fig:HaloDegDist} shows a histogram of the degrees of the nodes in the comparison graph for  the Halo data set.

\begin{figure}[h!]
\begin{center}
\includegraphics[width=0.34\textwidth]{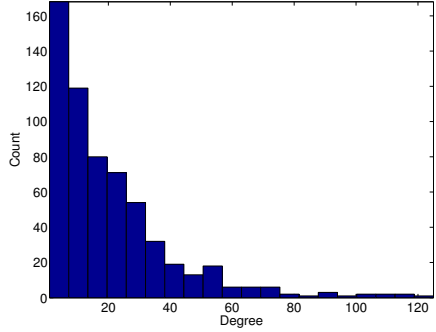}
\caption{Histogram of degrees in the comparison graph for the Halo data set, after discard the nodes with degree less than three.}
\end{center}
\label{fig:HaloDegDist}
\end{figure}

\begin{table}[tpb]
\begin{minipage}[b]{0.95\linewidth}
\begin{center}
\begin{tabular}{|c| C{1cm} |C{1cm}|C{1cm}|C{1cm}|C{1cm}|C{1cm}|C{1cm}|C{1cm}|C{1cm}|}
\hline
 Team   & SVD & LS & SER & SYNC &  SER-GLM & SYNC SUP &  SYNC SDP & RC & GT  \\
\hline
Arsenal   & 5  & 4  & 5  & 4  & 4  & 5  & 4  & 4  & 4   \\
 Aston Villa   & 14  & 16  & 18  & 16  & 16  & 14  & 16  & 16  & 15   \\
 Cardiff   & 17  & 19  & 20  & 19  & 20  & 20  & 19  & 19  & 20   \\
 Chelsea   & 2  & 2  & 1  & 2  & 3  & 1  & 2  & 2  & 3   \\
 Crystal Palace   & 16  & 14  & 15  & 14  & 15  & 13  & 14  & 13  & 11   \\
 Everton   & 4  & 6  & 4  & 6  & 6  & 7  & 6  & 5  & 5   \\
 Fulham   & 20  & 20  & 17  & 20  & 19  & 19  & 20  & 20  & 19   \\
 Hull   & 15  & 13  & 12  & 13  & 13  & 15  & 13  & 14  & 16   \\
 Liverpool   & 3  & 3  & 3  & 3  & 2  & 2  & 3  & 3  & 2   \\
 Man City   & 1  & 1  & 2  & 1  & 1  & 3  & 1  & 1  & 1   \\
 Man United   & 6  & 5  & 6  & 5  & 5  & 6  & 5  & 6  & 7   \\
 Newcastle   & 18  & 12  & 13  & 12  & 14  & 12  & 12  & 12  & 10   \\
 Norwich   & 19  & 18  & 16  & 18  & 17  & 17  & 18  & 18  & 18   \\
 Southampton   & 7  & 7  & 7  & 7  & 8  & 8  & 7  & 7  & 8   \\
 Stoke   & 9  & 10  & 11  & 10  & 10  & 9  & 10  & 10  & 9   \\
 Sunderland   & 11  & 17  & 19  & 17  & 18  & 16  & 17  & 17  & 14   \\
 Swansea   & 13  & 9  & 10  & 9  & 9  & 11  & 9  & 9  & 12   \\
 Tottenham   & 12  & 8  & 8  & 8  & 7  & 4  & 8  & 8  & 6   \\
 West Brom   & 8  & 15  & 14  & 15  & 12  & 18  & 15  & 15  & 17   \\
 West Ham   & 10  & 11  & 9  & 11  & 11  & 10  & 11  & 11  & 13   \\ 
\hline
Nr. upsets & 78 & 66 & 64 & 66 & 70 & 54 & 66 & 62 & 64   \\ 
Score/100 &   27.4 &   29.1 &   27.9 &   29.1 &   29.0 &   27.8 &   29.1 &   29.1 &   28.4   \\ 
W-Score/1000 &   28.1 &   29.2 &   27.8 &   29.2 &   29.3 &   27.6 &   29.2 &   29.3 &   28.8   \\ 
Corr w. GT &   0.66 &   0.83 &   0.75 &   0.83 &   0.81 &   0.85 &   0.83 &   0.85 &   1.00   \\ 
\hline
\end{tabular}
\end{center}
\end{minipage} 
\caption{English Premier League Standings for the 2013-2014 Season, based on the input matrix  $C^{tpd}$ given by (\ref{Ctpd}) (\textit{Total-Goal-Difference}).
}
\label{tab:EnglandStandings_sumGoalDif}
\end{table}

\begin{table}[tpb]
\begin{minipage}[b]{0.95\linewidth}
\begin{center}
\begin{tabular}{|c| C{1cm} |C{1cm}|C{1cm}|C{1cm}|C{1cm}|C{1cm}|C{1cm}|C{1cm}|C{1cm}|}
\hline
 Team   & SVD & LS & SER & SYNC &  SER-GLM & SYNC SUP &  SYNC SDP & RC & GT  \\
\hline
Arsenal   & 2  & 7  & 6  & 7  & 6  & 5  & 7  & 6  & 4   \\
 Aston Villa   & 14  & 14  & 13  & 14  & 14  & 14  & 14  & 14  & 15   \\
 Cardiff   & 16  & 19  & 16  & 19  & 16  & 20  & 19  & 19  & 20   \\
 Chelsea   & 11  & 1  & 1  & 1  & 1  & 1  & 1  & 1  & 3   \\
 Crystal Palace   & 15  & 13  & 15  & 13  & 15  & 13  & 13  & 13  & 11   \\
 Everton   & 7  & 6  & 4  & 6  & 5  & 7  & 6  & 7  & 5   \\
 Fulham   & 18  & 18  & 18  & 18  & 17  & 19  & 18  & 18  & 19   \\
 Hull   & 13  & 15  & 14  & 15  & 13  & 15  & 15  & 15  & 16   \\
 Liverpool   & 5  & 2  & 3  & 2  & 3  & 2  & 2  & 2  & 2   \\
 Man City   & 4  & 3  & 2  & 3  & 2  & 3  & 3  & 3  & 1   \\
 Man United   & 1  & 5  & 7  & 5  & 7  & 6  & 5  & 5  & 7   \\
 Newcastle   & 9  & 9  & 9  & 9  & 9  & 12  & 9  & 10  & 10   \\
 Norwich   & 17  & 16  & 17  & 16  & 18  & 17  & 16  & 17  & 18   \\
 Southampton   & 6  & 8  & 8  & 8  & 8  & 8  & 8  & 8  & 8   \\
 Stoke   & 10  & 11  & 11  & 11  & 11  & 9  & 11  & 12  & 9   \\
 Sunderland   & 20  & 17  & 19  & 17  & 19  & 16  & 17  & 16  & 14   \\
 Swansea   & 8  & 10  & 12  & 10  & 12  & 11  & 10  & 11  & 12   \\
 Tottenham   & 3  & 4  & 5  & 4  & 4  & 4  & 4  & 4  & 6   \\
 West Brom   & 19  & 20  & 20  & 20  & 20  & 18  & 20  & 20  & 17   \\
 West Ham   & 12  & 12  & 10  & 12  & 10  & 10  & 12  & 9  & 13   \\ 
\hline
Nr. upsets & 78 & 46 & 52 & 46 & 50 & 54 & 46 & 52 & 64   \\ 
Score/100 &    6.6 &    8.1 &    7.9 &    8.1 &    7.9 &    8.1 &    8.1 &    8.1 &    8.1   \\ 
W-Score/1000 &    6.6 &    7.9 &    7.6 &    7.9 &    7.6 &    8.1 &    7.9 &    7.9 &    8.1   \\ 
Corr w. GT &   0.60 &   0.81 &   0.78 &   0.81 &   0.75 &   0.85 &   0.81 &   0.80 &   1.00   \\ 
\hline
\end{tabular}
\end{center}
\end{minipage} 
\caption{English Premier League Standings
for the 2013-2014 Season, based on the input matrix  $C^{stpd}$ (\textit{Sign-Total-Goal-Difference}) given by (\ref{Cstpd}).}
\label{tab:EnglandStandings_signsumGoalDif}
\end{table}

\begin{table}[tpb]
\begin{minipage}[b]{0.95\linewidth}
\begin{center}
\begin{tabular}{|c| C{1cm} |C{1cm}|C{1cm}|C{1cm}|C{1cm}|C{1cm}|C{1cm}|C{1cm}|C{1cm}|}
\hline
 Team   & SVD & LS & SER & SYNC &  SER-GLM & SYNC SUP &  SYNC SDP & RC & GT  \\
\hline
Arsenal   & 6  & 4  & 5  & 4  & 4  & 5  & 4  & 4  & 4   \\
 Aston Villa   & 15  & 15  & 14  & 15  & 19  & 19  & 15  & 16  & 15   \\
 Cardiff   & 17  & 19  & 15  & 19  & 16  & 20  & 19  & 19  & 20   \\
 Chelsea   & 1  & 1  & 1  & 1  & 2  & 1  & 1  & 1  & 3   \\
 Crystal Palace   & 14  & 13  & 16  & 13  & 17  & 13  & 13  & 13  & 11   \\
 Everton   & 4  & 5  & 4  & 5  & 5  & 6  & 5  & 5  & 5   \\
 Fulham   & 20  & 16  & 17  & 16  & 14  & 16  & 16  & 17  & 19   \\
 Hull   & 16  & 14  & 12  & 14  & 13  & 14  & 14  & 14  & 16   \\
 Liverpool   & 3  & 2  & 3  & 2  & 3  & 3  & 2  & 2  & 2   \\
 Man City   & 2  & 3  & 2  & 3  & 1  & 2  & 3  & 3  & 1   \\
 Man United   & 7  & 7  & 7  & 7  & 7  & 7  & 7  & 7  & 7   \\
 Newcastle   & 12  & 12  & 10  & 12  & 10  & 10  & 12  & 12  & 10   \\
 Norwich   & 18  & 17  & 19  & 17  & 18  & 18  & 17  & 18  & 18   \\
 Southampton   & 8  & 8  & 8  & 8  & 8  & 8  & 8  & 8  & 8   \\
 Stoke   & 9  & 9  & 9  & 9  & 9  & 9  & 9  & 9  & 9   \\
 Sunderland   & 10  & 18  & 20  & 18  & 20  & 17  & 18  & 15  & 14   \\
 Swansea   & 19  & 10  & 13  & 10  & 11  & 12  & 10  & 11  & 12   \\
 Tottenham   & 5  & 6  & 6  & 6  & 6  & 4  & 6  & 6  & 6   \\
 West Brom   & 11  & 20  & 18  & 20  & 15  & 15  & 20  & 20  & 17   \\
 West Ham   & 13  & 11  & 11  & 11  & 12  & 11  & 11  & 10  & 13   \\ 
\hline
Nr. upsets & 54 & 42 & 40 & 42 & 52 & 48 & 42 & 44 & 54   \\ 
Score/100 &    5.6 &    5.8 &    5.8 &    5.8 &    5.6 &    5.8 &    5.8 &    5.8 &    5.9   \\ 
W-Score/1000 &    5.9 &    5.7 &    5.7 &    5.7 &    5.6 &    5.7 &    5.7 &    5.8 &    6.0   \\ 
Corr w. GT &   0.77 &   0.83 &   0.77 &   0.83 &   0.78 &   0.84 &   0.83 &   0.85 &   1.00   \\ 
\hline
\end{tabular}
\end{center}
\end{minipage} 
\caption{English Premier League Standings for the 2013-2014 Season, based on the input matrix $C^{snw}$ (\textit{Sign-Net-Wins}) given by (\ref{Csnw}).}
\label{tab:EnglandStandings_signNrWonLost}
\end{table}


\end{document}